\documentclass{sig-alternate}

\usepackage[boxed]{algorithm2e}
\usepackage{bm}
\usepackage{url}
\usepackage{comment}
\usepackage{url}
\usepackage{multirow}
\usepackage{graphicx}
\usepackage{color}

\newtheorem{definition}{Definition}[section]
\newtheorem{theorem}{Theorem}[section]
\newtheorem{lemma}{Lemma}[section]

\begin{document}
%

\title{Differentially- and non-differentially-private random decision trees}

%
%
%
%
%

\numberofauthors{4} 
%
\author{
%
%
\alignauthor
Mariusz Bojarski\thanks{equal contribution}\\
       \affaddr{NYU Polytechnic School of Engineering}\\
       \affaddr{Brooklyn, NY}\\
       \email{mb4496@nyu.edu}
\alignauthor
Anna Choromanska$^{*}$\\
       \affaddr{Courant Institute of Mathematical Sciences}\\
       \affaddr{New York, NY}\\
       \email{achoroma@cims.nyu.edu}
\alignauthor 
Krzysztof Choromanski$^{*}$\\
       \affaddr{Google Research}\\
       \affaddr{New York, NY}\\
       \email{kchoro@google.com}
\and
\alignauthor 
Yann LeCun\\
       \affaddr{Courant Institute of Mathematical Sciences and Facebook}\\
       \affaddr{New York, NY}\\
       \email{yann@cs.nyu.edu}
}

\maketitle

\begin{abstract}
We consider  supervised learning with random decision trees, where the tree construction is completely random. The method is popularly used and works well in practice despite the simplicity of the setting, but its statistical mechanism is not yet well-understood. In this paper we provide strong  theoretical guarantees regarding learning with random decision trees.  We analyze and compare three different variants of the algorithm that have minimal memory requirements: majority voting, threshold averaging and probabilistic averaging. The random structure of the tree enables us to adapt these methods to a differentially-private setting thus we also propose differentially-private versions of all  three schemes. We give upper-bounds on the generalization error and mathematically explain how the accuracy depends on the number of random decision trees. Furthermore, we prove that only logarithmic (in the size of the dataset) number of independently selected random decision trees suffice to correctly classify most of the data, even when differential-privacy guarantees must be maintained. We empirically show that majority voting and threshold averaging give the best accuracy, also for conservative users requiring high privacy guarantees. Furthermore, we demonstrate that a simple majority voting rule is an especially good candidate for the differentially-private classifier since it is much less sensitive to the choice of forest parameters than other methods.
\end{abstract}

\category{}{Security and privacy}{Security services}[Privacy-\\preserving protocols]
\category{}{Security and privacy}{Database and storage security}[Data anonymization and sanitization]
\category{}{Machine learning}{Machine learning approaches}[Classification and regression trees]




\terms{
Theory, Algorithms, Security
}


\section{Introduction}

Decision tree is one of the most fundamental structures used in machine learning. Constructing a tree of good quality is a hard computational problem though. Needless to say, the choice of the optimal attribute according to which the data partitioning should be performed in any given node of the tree requires nontrivial calculations involving data points located in that node. Nowadays, with an increasing importance of the mechanisms preserving
privacy of the data handled by machine learning algorithms, the need arises to construct  these algorithms with strong privacy 
guarantees (see e.g.~\cite{srikant},~\cite{du},~\cite{choro2},~\cite{chaudhuri},~\cite{geetha2}). One of the strongest currently used notions of privacy is the so-called \textit{differential privacy} that was introduced~\cite{nissim} in a quest to achieve the dual goal of maximizing  data utility and preserving data confidentiality.  A differentially-private database access mechanism preserves the privacy of any individual in the database, irrespectively of the amount of auxiliary information available to an adversarial database client.
Differential-privacy techniques add noise to perturb data (such as Laplacian noise). Its magnitude depends on the sensitivity of the statistics that are being output.  Even though the overall scheme looks simple, in practice it is usually very difficult to obtain a reasonable level of differential privacy and at the same time maintain good accuracy. This is the case since usually too big perturbation error needs to be added. In particular, this happens when
machine learning computations access data frequently during the entire execution of the algorithm and output structures that are very sensitive to the data. This is also an obstacle for proposing a scheme that computes an optimal decision tree in a differentially-private way. In such a scenario the attribute chosen in every node and any additional information stored there depends on the data and that is why it must be perturbed in order to keep the desired level of differential privacy. Big perturbation added in this setting leads to the substantially smaller quality of the constructed tree. 

Instead of constructing one differentially-private decision tree, in this paper we consider constructing a random forest. Random forests~\cite{breiman} constitute an important member of the family of the decision tree-based algorithms due to their effectiveness and excellent performance. They are also the most accurate general-purpose classifiers available~\cite{Biau:2008:CRF:1390681.1442799, breiman}. In this paper we construct a forest consisting of $O(\log(n))$ random decision trees ($n$ is the size of the dataset, e.g. number of data samples). An attribute according to which the selection is performed in any given node is chosen uniformly at random from all the attributes, independently from the dataset in that node. In the continuous case,
the threshold value for the chosen attribute is then also
chosen uniformly at random from the range of all possible values. That simple rule 
enables us to construct each decision tree very fast since the choice of nodes' attributes does not depend on the data at all. The obtained algorithm is therefore fast and scalable with minimal memory requirements. It also takes only one pass over the data to construct the classifier. Since most of the structure of each random decision tree is constructed without examining the data, the algorithm suits the differentially-private scenario very well. After a sufficient number of random decision trees is constructed, the classification of every point from a dataset takes place. Classification is done according to one of the three schemes: majority voting (\cite{breiman}), threshold averaging or probabilistic averaging (\cite{weifan2}).
In the differentially-private setting  we add perturbation error to the counters in leaves, but 
no perturbation error is added to the inner nodes. This leads to a much more accurate learning mechanism. 
Performing voting/averaging (see: \cite{schapire} for applications of the voting methods) instead of just taking the best tree for a given dataset is important since it enables us to add smaller perturbation error to obtain the same level of differential privacy. 

In this paper we analyze both non-differentially-private and differentially-private setting in all three variants: majority voting, threshold averaging, and probabilistic averaging. To the best of our knowledge, we are the first to give a comprehensive and unified theoretical analysis of all three models in both settings, where in case of differentially-private setting no theoretical analysis was ever provided in the context of random decision trees. The differentially-private setting is especially difficult to analyze since increasing the number of trees does not necessarily decrease the training (and test) error in this setting. Having more random decision trees require adding bigger perturbation error that may decrease the efficiency of the learning algorithm. In this paper we thoroughly investigate this phenomenon. The dependence of the quality of the random decision tree methods on the chosen level of differential privacy, the height of the tree and the number of trees in the forest is in the central focus of our theoretical and empirical analysis. Understanding these dependencies is crucial while applying these methods in practice. Our theoretical analysis relate the empirical error and the generalization error of the classifier to the average tree accuracy and explain quantitatively how the quality of the system depends on the number of chosen trees. Furthermore, we show that the random forest need not many trees to achieve good accuracy. In particular, we prove both theoretically and empirically that in practice the logarithmic in the size of the dataset number of random decision trees\footnote{Further in the paper by "logarithmic number of random decision trees" we always mean "logarithmic (in the size of the dataset) number of random decision trees".} suffices to achieve good performance. We also show that not only do there exist parameters of the setting (such as: the number of random trees in the forest, the height of the tree, etc.) under which one can effectively learn, but the setting is very robust. To be more precise, we empirically demonstrate that the parameters do not need to be chosen in the optimal way, in example one can choose far fewer trees to achieve good performance. We also show that majority voting and threshold averaging are good candidates for the differentially-private classifiers. Our experiments reveal that a simple majority voting rule is competitive with the threshold averaging rule and simultaneously they both outperform the probabilistic averaging rule. Furthermore, majority voting rule is much less sensitive to the choice of the parameters of the random forest (such as the number of the trees and the height of the tree) than the remaining two schemes.

This article is organized as follows. In Section 2 we describe previous work regarding random decision trees. We then introduce our model and the notion of differential privacy in Section 3. In Section 4 we present a differentially-private supervised algorithm that uses random decision trees.
Section 5 contains our theoretical analysis. We conclude the paper with experiments (Section 6) and a brief summary of our results (Section~\ref{sec:conclusions}).

\section{Prior work}

Random decision trees are considered as important methods in machine learning often used for supervised learning due to their simplicity, excellent practical performance and somewhat unreasonable effectiveness in practice. They became successful in a number of practical problems, e.g.~\cite{Shotton:2011:RHP:2191740.2192047,Amit:1997:SQR:263023.263042,Xiong:2012:RFM:2339530.2339680,Manik,kouzani2010multilabel,conf/kdd/YanTS07,journals/jcisd/SvetnikLTCSF03,citeulike:12009604,Criminisi:2013:DFC:2462584,conf/miccai/ZikicGC13} (there exist many more examples). The original random forests~\cite{breiman} were ensemble methods combining many CART-type~\cite{ig} decision trees using bagging~\cite{breiman1996bagging} (a convenient review of random forests can for instance be found in~\cite{Criminisi:2013:DFC:2462584}). They were inspired by some earlier published random approaches~\cite{Amit:1997:SQR:263023.263042,ho,Ho:1998:RSM:284980.284986,Dietterich:2000:ECT:350128.350131}. Despite their popularity, the statistical mechanism of random forests is difficult to analyze~\cite{Biau:2008:CRF:1390681.1442799,Biau:2012:ARF:2188385.2343682} and to these days remains largely ununderstood~\cite{Biau:2012:ARF:2188385.2343682,DBLP:conf/icml/DenilMF13,DBLP:conf/icml/DenilMF14}. Next we review the existing theoretical results in the literature.

A notable line of works provide an elegant analysis of the consistency of random forests~\cite{Biau:2012:ARF:2188385.2343682,DBLP:conf/icml/DenilMF13,DBLP:conf/icml/DenilMF14,Biau:2008:CRF:1390681.1442799,journals/jmlr/Meinshausen06,Lin02randomforests,breiman3,breiman2,journals/jmlr/Meinshausen06,citeulike:6833230}. Among these works, one of the most recent studies~\cite{Biau:2012:ARF:2188385.2343682} proves that the previously proposed random forest approach~\cite{breiman2} is consistent and achieves the rate of convergence which depends only on the number of strong features and not on the number of noise variables. Another recent paper~\cite{DBLP:conf/icml/DenilMF13} provides the first consistency result for online variant of random forests. The predecessor of this work~\cite{Domingos:2000:MHD:347090.347107} proposes the Hoeffding tree algorithm and prove that with high probability under certain assumptions the online Hoeffding tree converges to the offline tree. In our paper we focus on error bounds rather than the consistency analysis of random decision trees.

It has been noted~\cite{DBLP:conf/icml/DenilMF13} that the most famous theoretical result concerning random forests provides an upper-bound on the generalization error of the forest in terms of the correlation and strength of trees~\cite{breiman}. Simultaneously, the authors show that the generalization error converges almost surely to a limit as the number of trees in the forest becomes large. It should be noted however that the algorithm considered by the authors has data-dependent tree structure opposite to the algorithms in our paper. To be more specific, the original "random forests" method~\cite{breiman} selects randomly a subset of features and then it chooses the best splitting criteria from this feature subset. This affects efficiency since computing the heuristics (the best splitting criteria) is expensive~\cite{weifan2}. Furthermore, it also causes the tree structure to be data-dependent (another approach where the tree structure is data-dependent is presented in example in~\cite{Geurts:2006:ERT:1132034.1132040}) rather than fully random which poses a major problem when extending the method to the differentially-private setting since data-independent tree structure is important for preserving differential-privacy~\cite{jagannathan}. Opposite to this approach, in our algorithms we randomly draw the attribute in each tree node according to which we split and then we randomly choose a threshold used for splitting. This learning model is therefore much simpler. Our fully random approach is inspired by a methodology already described before in the literature~\cite{weifan2} (this work however has no theoretical analysis). Our theoretical results consider error bounds similarly to the original work on random forests~\cite{breiman}. The difference of approaches however does not allow to use the theoretical results from \cite{breiman} in our setting. Finally, note that in either \cite{breiman} or \cite{weifan2} only a single voting rule is considered, majority voting or probabilistic averaging respectively. In this paper we consider a wider spectrum of different voting approaches.


Next, we briefly review some additional theoretical results regarding random forests. A simplified analysis of random forests in one-dimensional settings was provided in the literature in the context of regression problems where minimax rate of convergence were proved~\cite{Genuer1,Genuer2}. Another set of results explore the connection of random forests with a specific framework of adaptive nearest-neighbor methods~\cite{Lin:Jeon:rand:2006}. Finally, for completeness we emphasize that there also exist some interesting empirical studies regarding random decision trees in the literature, e.g.~\cite{weifan3},~\cite{weifan} and~\cite{ho}, which however are not directly related to our work. 

Privacy preserving data mining has emerged as an effective method to solve the problem of data sharing in many fields
of computer science and statistics. One of the strongest currently used notions of privacy is the so-called differential privacy~\cite{nissim} (some useful tutorial material on differential privacy research can be found in~\cite{Yang:2012:DPD:2213836.2213910}). In this paper we are interested in the differentially-private setting in the context of random decision trees. It was first observed in \cite{jagannathan} that random decision trees may turn out to be an effective tool for constructing a differentially-private decision tree classifier. The authors showed a very efficient heuristic that averages over random decision trees and gives good practical results. Their work however lacks theoretical results regarding the quality of the differentially-private algorithm that is using random decision trees. In another published empirical study~\cite{6616536} the authors develop protocols to implement privacy-preserving random decision trees that enable efficient parallel and distributed privacy-preserving knowledge discovery. The very recent work~\cite{6968348} on differentially-private random forests shows experimental results demonstrating that quality functions such as information gain, max operator
and gini index gives almost equal accuracy regardless of their sensitivity towards the noise. Furthermore, they show that the accuracy of the classical random forest and its differentially-private counterpart is almost equal for various size of datasets. To the best of our knowledge none of the published works on differentially-private random decision trees provide any theoretical guarantees. Our paper provides strong theoretical guarantees of both non-differentially-private and differentially-private random decision trees. This is a major contribution of our work. We simultaneously develop a unified theoretical framework for analyzing both settings. 

\section{Preliminaries}

\subsection{Differential privacy}

Differential privacy is a model of privacy for database access mechanism.
It guarantees that small changes in a database (removal or addition of an element)
does not change substantially the output of the mechanism.


\begin{definition}(See~\cite{dwork}.)
A randomized algorithm $\mathcal{K}$ gives $\epsilon$-differential-privacy if for all
datasets $\mathcal{D}_1$ and $\mathcal{D}_2$  
differing on at most one element, and all $S \subseteq Range(\mathcal{K})$,
\begin{equation}
\mathbb{P}(\mathcal{K} (\mathcal{D}_{1}) \in S) \le \exp (\epsilon) \cdot \mathbb{P}(\mathcal{K} (\mathcal{D}_{2}) \in S).
\end{equation}
The probability is taken over the coin tosses of $\mathcal{K}$. 
\end{definition}

The smaller $\epsilon$, the stronger level of differential privacy is obtained.
Assume that the non-perturbed output of the mechanism can be encoded
by the function $f$. A mechanism $\mathcal{K}$ can compute a differentially-private 
noisy version of $f$ over a database $\mathcal{D}$ by adding noise with magnitude 
calibrated to the sensitivity of $f$.

\begin{definition}(See~\cite{nissim}.)
The global sensitivity $S(f)$  of a function $f$ is the smallest
number $s$ such that for all $\mathcal{D}_{1}$ and $\mathcal{D}_{2}$ which differ on at most one element, $| f(\mathcal{D}_{1}) - f(\mathcal{D}_{2})\ | \leq s$.
\end{definition}
Let $Lap(0, \lambda)$ denote the Laplace distribution with mean $0$ and 
standard deviation $\lambda$.  In other words, this is a random variable with probability density function given by the following formula:
$\frac{\lambda}{2}e^{-|x|\lambda}$. We will denote shortly by $g(\lambda)$ an independent copy of the $Lap(0, \lambda)$-random variable.

\begin{theorem}(See~\cite{nissim}.) \label{laplaciantheorem}
Let $f$ be a function on databases with range $R^{m}$, where $m$ is the number of rows of 
databases
\footnote{Number of rows of databases is the number of attributes of any data point from the databases.}. 
Then, the mechanism that
outputs $f(\mathcal{D} ) + (Y_1,\ldots, Y_m )$, where $Y_i$ are drawn i.i.d 
from $Lap(0,S(f)/\epsilon)$, satisfies $\epsilon$-differential-privacy.  
\end{theorem}
Stronger privacy guarantees and more sensitive functions need bigger variance of the Laplacian noise being added.
Differential privacy is preserved under composition, but with an extra loss of privacy for each conducted query.
 
\begin{theorem}(See~\cite{nissim}.) \label{compositiontheorem}
\textbf{(Composition Theorem)} \\The sequential application of mechanisms $\mathcal{K}_i$, each 
giving \\$\epsilon_{i}$-differential privacy, satisfies $\sum_i \epsilon_{i}$-differential-privacy.  
\end{theorem}
More information about differential privacy can be found in the work of \cite{scherrytalwar} and \cite{hall}.

\subsection{The model}
All data points are taken from $\mathcal{F}^{m}$, where $m$ is the number of the attributes and $\mathcal{F}$ is either a discrete set
or the set of real numbers. We assume that for every attribute $attr$ its smallest ($\min(attr)$) and largest possible value ($\max(attr)$)
are publicly available and that the labels are binary.
We consider only binary decision trees (all our results can be easily translated to the setting where inner nodes of the tree have more than two children). Therefore, if $\mathcal{F}$ is discrete then we will assume that $\mathcal{F} = \{0,1\}$, i.e. each attribute is binary.
In the continuous setting for each inner node of the tree we store the attribute according to which the selection is done and
the threshold value of this attribute. 
All decision trees considered in this paper are complete and of a fixed height $h$ that does not depend on the data. Let $T$ be a random decision tree and let $l$ be one of its leaves. We denote by $\theta_{l}$ the fraction of all training points
in $l$ with label $+$. If $l$ does not contain any of the training points we choose the value of $\theta_{l}$  uniformly at random from $[0,1]$.
The set $M$ of all possible decision trees is of size $|M|=m^{2^{h+1}-1}$ in the binary setting. It should be emphasized that it is true also in the continuous case.
In that setting the set of all possible threshold values for a node is infinite but needless to say, the set of all possible partitionings in the node is still finite.
Thus without loss of generality, we assume $M$ is finite. It can be very large
but it does not matter since we will never need the actual size of $M$ in our analysis.
For a given tree $T$ and given data point $d$ denote by $w_{d}^{T}$ the fraction of points (from the training set if $d$ is from this set and from the test set otherwise) with the same label as $d$ that end up in the same leaf
of $T$ as $d$. We call it the $\textit{weight of $d$ in $T$}$. Notice that a training point $d$ is classified correctly by $T$ in the single-tree setting iff its weight in $T$ is larger than $\frac{1}{2}$ (for a single decision tree we consider majority voting model for points classification).

The average value of $w_{d}^{T}$ over all trees of $M$ will be denoted as $w_{d}$ and called the \textit{weight of $d$ in $M$}.
We denote by $\sigma(d)$ the fraction of trees from $M$ with the property that most of the points of the leaf of the tree containing $d$ have the same label as $d$ (again, the points are taken from the training set if $d$ is from it and from the test set otherwise).
We call $\sigma(d)$ \textit{the goodness of $d$ in $M$}.
For a given dataset $\mathcal{D}$ the average tree accuracy $e(\mathcal{D})$ of a random decision tree model is an average accuracy of the random decision tree from $M$, where the accuracy is the fraction of data points that a given tree classifies correctly (accuracy is computed under assumption that the same distribution $\mathcal{D}$ was used in both: the training phase and test phase).


\begin{algorithm}[h]
\label{alg:non-diffRDT}       
\setlength\tabcolsep{1pt}
\begin{tabular}{ll}
\textbf{Input:} $Train$, $Test$: train and test sets,\\ 
\:\:\:\:\:\:\:\:\:\:\:\:\:\:\:\:$h$: height of the tree\\
\hline
\textbf{Random forest construction:}\\
$\:\:\:\:\:\:$construct   $k=\theta(\log(n))$ random decision trees  by\\
$\:\:\:\:\:\:$choosing for each inner node of the tree\\ 
$\:\:\:\:\:\:$independently at random its attribute (uniformly\\
$\:\:\:\:\:\:$from the set of all the attributes);\\
\\
$\:\:\:\:\:\:$in the continuous case for each chosen attribute\\
$\:\:\:\:\:\:$$attr$ choose independently at random a threshold\\
$\:\:\:\:\:\:$value uniformly from $[\min(attr),\max(attr)]$\\ 
\textbf{Training:}\\
$\:\:\:\:\:\:$\textbf{For} $d \in Train\:\:\:\{$\\
$\:\:\:\:\:\:\:\:\:\:$add $d$ to the forest by updating $\theta_{l}$ for every leaf\\
$\:\:\:\:\:\:\:\:\:\:$corresponding to $d$\:\:\:\}\\
\textbf{Testing:}\\
$\:\:\:\:\:\:$\textbf{For} $d \in Test\:\:\:\{$\\
$\:\:\:\:\:\:\:\:\:\:\:$\textbf{if} ($\textsf{majority voting})\:\:\:\{$\\
$\:\:\:\:\:\:\:\:\:\:\:\:\:\:\: $compute $num^{d}$ - the number of the trees\\
$\:\:\:\:\:\:\:\:\:\:\:\:\:\:\:\:\:\:\:\:\:\:\:\:\:\:\:\:\:\:\:\:\:\:\:\:\:\:\:\:\:\:\:\:\:\:\:\:$ classifying $d$ as $+$;$ \:\:\:$\\
$\:\:\:\:\:\:\:\:\:\:\:\:\:\:\: $classify $d$ as $+$ iff $num^{d} > \frac{k}{2}\:\:\:\}$\\
$\:\:\:\:\:\:\:\:\:\:\:$\textbf{if} ($\textsf{threshold averaging})\:\:\:\{$\\
$\:\:\:\:\:\:\:\:\:\:\:\:\:\:\:$compute $\theta^{d} = \frac{1}{k}\sum_{l \in \mathcal{L}} \theta_{l}$, where $\mathcal{L}$ is a set of all\\
$\:\:\:\:\:\:\:\:\:\:\:\:\:\:\:\:\:\:\:\:\:\:\:$leaves of the forest that correspond to $d$;\\
$\:\:\:\:\:\:\:\:\:\:\:\:\:\:\: $classify $d$ as $+$ iff $\theta^{d} > \frac{1}{2}\:\:\:\}$\\
$\:\:\:\:\:\:\:\:\:\:\:$\textbf{if} ($\textsf{probabilistic averaging})\:\:\:\{$\\
$\:\:\:\:\:\:\:\:\:\:\:\:\:\:\:$compute $\theta^{d} = \frac{1}{k}\sum_{l \in \mathcal{L}} \theta_{l}$, where $\mathcal{L}$ is a set of all\\
$\:\:\:\:\:\:\:\:\:\:\:\:\:\:\:\:\:\:\:\:\:\:\:$leaves of the forest that correspond to $d$;\\
$\:\:\:\:\:\:\:\:\:\:\:\:\:\:\: $classify $d$ as $+$ with probability $\theta^{d} \:$\\
$\:\:\:\:\:\:\:\:\:\:\:\:\:\:\:$\small{/*random tosses here are done independently}\\
$\:\:\:\:\:\:\:\:\:\:\:\:\:\:\:\:\:\:\:$\small{from all other previously conducted*/}$\:\:\:\}\:\}$\\
\hline
\textbf{Output:} Classification of all $d \in Test$
\end{tabular}
\caption{\textsf{Non-differentially-private RDT classifier}}
\label{alg:ndp}
\end{algorithm}

\begin{algorithm}[htp!]
\label{alg:diffRDT}       
\setlength\tabcolsep{1pt}
\begin{tabular}{ll}
\textbf{Input:} $Train$, $Test$: train and test sets,\\ 
\:\:\:\:\:\:\:\:\:\:\:\:\:\:\:\:$h$: height of the tree, $\eta$: privacy parameter\\
\hline
\textbf{Random forest construction:}$\:\:\:$as in Algorithm~\ref{alg:ndp}\\

\textbf{Training:}\\
$\:\:\:\:\:\:$\textbf{For} $d \in Train\:\:\:\{$\\
$\:\:\:\:\:\:\:\:\:\:\:\:$find the leaf $l$ for $d$ in every tree and\\
$\:\:\:\:\:\:\:\:\:\:\:\:$update $n_{l}^{+}$, $n_{l}^{-}$, where:\\
$\:\:\:\:\:\:\:\:\:\:\:\:\:\:\:\:\:\:$$n_{l}^{+}$ - the number of training points with\\
$\:\:\:\:\:\:\:\:\:\:\:\:\:\:\:\:\:\:\:\:\:\:\:\:\:\:\:\:$label $+$ belonging to that leaf;\\
$\:\:\:\:\:\:\:\:\:\:\:\:\:\:\:\:\:\:$$n_{l}^{-}$ - the number of training points with\\
$\:\:\:\:\:\:\:\:\:\:\:\:\:\:\:\:\:\:\:\:\:\:\:\:\:\:\:\:$label $-$ belonging to that leaf$\:\:\:\}$\\
$\:\:\:\:\:\:$\textbf{For every leaf} $\bm{l}\:\:\:\{$\\
$\:\:\:\:\:\:\:\:\:\:\:\:$calculate $n_{l}^{p,+} \!\!\!=\! n_{l}^{+} + g(\frac{\eta}{k})$ and $n_{l}^{p,-} \!\!\!=\! n_{l}^{-} + g(\frac{\eta}{k})$ \\
$\:\:\:\:\:\:\:\:\:\:\:\:$\textbf{if} ($n_{l}^{p,+} \!\!\!<\! 0$ or $n_{l}^{p,-} \!\!\!<\! 0$ or ($n_{l}^{p,+} \!\!\!=\! 0$ and $n_{l}^{p,-} \!\!\!=\! 0$))\\
$\:\:\:\:\:\:\:\:\:\:\:\:\:\:\:\:\:\:$choose $\theta^{p}_{l}$ uniformly at random from $[0,1];\:\:\:$\\
$\:\:\:\:\:\:\:\:\:\:\:\:$\textbf{else}\:\:\:\ let $\:\theta^{p}_{l} = \frac{n_{l}^{p,+}}{n_{l}^{p,+}+n_{l}^{p,-}};$\\
$\:\:\:\:\:\:\:\:\:\:\:\:$publish $\theta^{p}_{l}$ for every leaf$\:\:\:\}$\\
\textbf{Testing:} as in Algorithm~\ref{alg:ndp} but replace $\theta_{l}$ with $\theta^{p}_{l}$ $$\\
\hline
\textbf{Output:} Classification of all $d \in Test$
\end{tabular}
\caption{\textsf{$\eta$-Differentially-private RDT classifier}}
\label{alg:dp}
\end{algorithm}

\section{Algorithms}
Algorithm~\ref{alg:ndp} captures the non-differentially-private algorithm for supervised learning with random decision trees (RDT). Its differentially-private counterpart is captured in Algorithm~\ref{alg:dp}. We consider three versions of each algorithm: 
\begin{itemize}
\item majority voting
\item threshold averaging
\item probabilistic averaging. 
\end{itemize}
Only variables $n_{l}^{+},n_{l}^{-}$ stored in leaves depend on the data. This fact will play crucial role in the 
analysis of the differen-\\tially-private version of the algorithm where Laplacian error is added to the point counters at every leaf with variance calibrated to the number of all trees used by the algorithm.

\section{Theoretical results}

In this section we derive the upper-bounds on the empirical error (the fraction of the training data misclassified by the algorithm) and the generalization error (the fraction of the test data misclassified by the algorithm where the test data is taken from the same distribution as the training data) for all methods in Algorithm~\ref{alg:ndp} and~\ref{alg:dp}. 


We also show how to find the number of random decision trees to obtain good accuracy and, in the differentially-private setting, good privacy guarantees. 

We start with two technical results which, as we will see later, give an intuition why the random 
decision tree approach works very well in practice.

\begin{theorem}
\label{technicaltheorem_gamma}
Assume that the average tree accuracy of the set $M$ of all decision trees of height $h$ on the training/test set $\mathcal{D}$ is $e=1-\epsilon$ for some $0<\epsilon \leq \frac{1}{2}$. 
Then the average goodness $\sigma(d)$ of a training/test point $d$ in $M$ is at least $e \geq \frac{1}{2}$.
\end{theorem}

\begin{theorem}
\label{technicaltheorem}
Assume that the average tree accuracy of the set $M$ of all decision trees of height $h$ on the training/test set $\mathcal{D}$ is $e=1-\epsilon$ for some $0<\epsilon \leq \frac{1}{2}$. Then the average weight $w_{d}$ of a training/test point $d$ in $M$ is at least $e^{2} + (1-e)^{2} \geq \frac{1}{2}$.
\end{theorem}

The theorems above imply that if the average accuracy of the tree is better than random, then this is also reflected by
the average values of $w_{d}$ and $\sigma_{d}$. This fact is crucial for the theoretical analysis since
we will show that if the average values of $w_{d}$ and $\sigma_{d}$ are slightly better than random then this implies
very small empirical and generalization error.
Furthermore, for most of the training/test points $d$ their values of $\sigma_{d}$ and $w_{d}$ are well concentrated
around those average values and that, in a nutshell, explains why the random decision trees approach works well.
Notice that Theorem \ref{technicaltheorem_gamma} gives better quality guarantees than Theorem \ref{technicaltheorem}. 

We are about to propose several results regarding differen-\\tially-private learning with random decision trees.
They are based on careful structural analysis of the bipartite graph between the set of  decision trees and datapoints.
Edges of that bipartite graph connect datapoints with trees that correctly classified given datapoints.
In the differentially-private setting the key observation is that under relatively weak conditions one can assume that
the sizes of the sets of datapoints residing in leaves of the trees are substantial. Thus adding the Laplacian noise
will not perturb the statistics to an extent that would affect the quality of learning. All upper-bounds regarding the generalization error were obtained by combining this analysis with concentration results (such as Azuma's inequality).


\subsection{Non-differentially-private setting }

We start by providing theoretical guarantees in the non-differentially-private case. Below we consider majority voting and threshold averaging. The results for the probabilistic averaging are stated later in this subsection.

\begin{theorem} 
\label{alfa_setting_ndp_main_1}
Let $K>0$.
Assume that the average tree accuracy of the set $M$ of all decision trees of height 
$h$ on the training/test set $\mathcal{D}$ is $e=1-\epsilon$ for some $0<\epsilon \leq \frac{1}{2}$. 
Let $\mu$ be: the fraction of training/test points with goodness in $M$ at least $\sigma=\frac{1}{2} + \delta$ / $\sigma =\frac{1}{2} + \delta + \frac{1}{K}$ for 
$0 < \delta < \frac{1}{2}$ (in the majority version) or: the fraction of training/test points with weight in $M$ at least $w=\frac{1}{2} + \delta$ / $w = \frac{1}{2} + \delta + \frac{1}{K}$ for 
$0 < \delta < \frac{1}{2}$ (in the threshold averaging version).
Then Algorithm~\ref{alg:ndp} for every $C>0$ and $k=\frac{(1+C)\log(n)}{2\delta^{2}}$ selected random decision trees 
gives empirical error $err_{1} \leq 1 - \mu$ with probability $p_{1} \geq 1-\frac{1}{n^{C}}$. The generalization 
error $err_{2} \leq 1 - \mu$ will be achieved for $k=\frac{(1+C)\log(n)}{2(\frac{\delta}{2})^{2}}$ trees with probability $p_{2} \geq p_{1} - 2^{h+3}ke^{-2n\phi^{2}}$,
where $\phi = \frac{\delta}{2(4+\delta)2^{h}K}$.
Probabilities $p_{1}$ and $p_{2}$ are 
under random coin tosses used to construct the forest and the test set. 
\end{theorem}

Note that parameter $e$ is always in the range $[\frac{1}{2},1]$. The more decision trees that classify
data in the nontrivial way (i.e. with accuracy greater than $\frac{1}{2}$), the larger the value of $e$ is.
The result above in particular implies that if most of the points have goodness/weight in $M$ a little bit larger than $\frac{1}{2}$ then both errors are very close
to $0$. This is indeed the case - the average point's goodness/weight in $M$, as Theorem~\ref{technicaltheorem_gamma} and Theorem~\ref{technicaltheorem} say, is at least $e$ / $e^{2} + (1-e)^{2}$. 
The latter expression is greater than $\frac{1}{2}$
if the average tree accuracy is slightly bigger than the worst possible. 
Besides goodness/weight of most of the points, as was tested experimentally,
is well concentrated around that average goodness/weight.
We conclude that if the average accuracy of the decision tree is separated from $\frac{1}{2}$ (but not necessarily very close to $1$)
then it suffices to classify most of the data points correctly. The intuition behind this result is as follows: if the constructed forest of the decision trees
contains at least few "nontrivial trees" giving better accuracy than random 
then they guarantee correct classification of most of the points. 

If we know that the average tree accuracy is big enough then techniques used to prove Theorem~\ref{alfa_setting_ndp_main_1} give us more direct bounds on the empirical and generalization errors captured in Theorem~\ref{alfa_setting_ndp_main_2}. No assumptions regarding goodness/weight are necessary there.

\begin{theorem} 
\label{alfa_setting_ndp_main_2}
Let $K>0$.
Assume that the average tree accuracy of the set $M$ of all decision trees of height $h$ on the training/test 
set $\mathcal{D}$ is $e=1-\epsilon$ for some $0<\epsilon \leq \frac{1}{2}$. Then Algorithm~\ref{alg:ndp} for 
every $C>0$, $0 < \delta <\frac{1}{2}$ and $k=\frac{(1+C)\log(n)}{2\delta^{2}}$ selected 
random decision trees gives empirical error: 
$err_{1} \leq \frac{\epsilon}{\frac{1}{2}-\delta}$ (in the majority version) or:
$err_{1} \leq \frac{2\epsilon-2\epsilon^{2}}{0.5 - \delta}$ (in the threshold averaging version) 
with probability $p_{1} \geq 1-\frac{1}{n^{C}}$. The generalization error: 
$err_{2} \leq \frac{\epsilon+\frac{1}{K}}{\frac{1}{2}-\delta}$ (in the majority version) or:
$err_{2} \leq \frac{2(\epsilon+\frac{1}{K})-2(\epsilon+\frac{1}{K})^{2}}{0.5 - \delta}$ (in the threshold averaging version) will be achieved for
$k=\frac{(1+C)\log(n)}{2(\frac{\delta}{2})^{2}}$ trees
with probability $p_{2} \geq p_{1} - 2^{h+3}ke^{-2n\phi^{2}}$, where $\phi = \frac{\delta}{2(4+\delta)2^{h}K}$. 
Probabilities $p_{1}$ and $p_{2}$ are under random coin tosses used to construct the forest and the test set.
\end{theorem}


Theorems~\ref{alfa_setting_ndp_main_1} and~\ref{alfa_setting_ndp_main_2} show that logarithmic number of random decision trees in practice suffices to obtain high prediction accuracy with a very large probability. In particular, the upper-bound on the generalization error is about two times the average error of the tree. The existence of the tree with lower accuracy in the forest does not harm the entire scheme
since all trees of the forest play role in the final classification.

We now state our results (analogous to Theorem~\ref{alfa_setting_ndp_main_2}) for the probabilistic averaging setting. 
The following is true.

\begin{theorem}
\label{beta_setting_ndp_main}
Let $K>0$.
Assume that the average tree accuracy of the set $M$ of all decision trees of height $h$ on the training/test 
set $\mathcal{D}$ is $e=1-\epsilon$ for some $0<\epsilon \leq \frac{1}{2}$. Let $C>0$ be a constant. Let $0 < \delta, c < 1$. 
Then with probability at least $p_{1} = (1-\frac{1}{n^{C}})(1-e^{-2nc^{2}})$ the probabilistic averaging version of Algorithm~\ref{alg:ndp} gives 
empirical error $err_{1} \leq 2\epsilon -2\epsilon^{2}+\delta+c$ and with probability 
$p_{2} \geq p_{1} - 2^{h+3}ke^{-2n\phi^{2}}$, where $\phi = \frac{\delta}{2(4+\delta)2^{h}K}$, it gives generalization error 
$err_{2} \leq 2(\epsilon+\frac{1}{K}) -2(\epsilon+\frac{1}{K})^{2}+\delta+c$.
Probabilities $p_{1},p_{2}$ are under random tosses used to construct the forest and the test set.
\end{theorem}

Notice that this result is nontrivial for almost the entire range $[0,\frac{1}{2}]$ of $\epsilon$, and $\delta$ and $c$ close to $0$,
and large $K$. This is the case 
since note that $1-2\epsilon + 2\epsilon^{2} \geq \frac{1}{2}$ and the equality holds only for $\epsilon=\frac{1}{2}$.

\subsection{Differentially-private setting}

We begin this section with the proof that all three methods captured in Algorithm~\ref{alg:dp}, where the Laplacian noise is added to certain counts, are indeed $\eta$-differentially-private. 
\begin{proof}
Notice that in every method to obtain the forest of random decision trees with perturbed counters in leaves  we need $k$ queries to the private data
(this is true since the structure of the inner nodes of the trees does not depend at all on the data and data subsets corresponding to leaves are pairwise disjoint).
Furthermore, the values that are being perturbed by the Laplacian
noise are simple counts of global sensitivity $1$. Thus we can use use Theorem~\ref{laplaciantheorem} and Theorem~\ref{compositiontheorem}
to conclude that in order to obtain $\eta$-differential privacy of the entire system we need to add a $Lap(0,\frac{k}{\eta})$ to every count in the leaf.
This proves that our algorithms are indeed $\eta$-differentially-private.
\end{proof}

Next we show the theoretical guarantees we obtained in the differentially-private setting. As in the previous section, we first focus on the majority voting and threshold averaging, and then we consider the probabilistic averaging. 

\begin{theorem}
\label{alfa_setting_dp_main_1}
Assume that we are given a parameter $\eta>0$.  Let $K>0$.
Assume that the average tree accuracy of the set $M$ of all decision trees of height $h$ on the 
training/test set $\mathcal{D}$ is $e=1-\epsilon$ for some $0<\epsilon \leq \frac{1}{2}$. 
Let $\mu$ be the fraction of training/test points with:
goodness in $M$ at least $\sigma=\frac{1}{2} + \delta + \frac{1}{K}$ / $\sigma = \frac{1}{2} + \delta + \frac{2}{K}$ (in the majority version) or:
weight in $M$ at least $w=\frac{1}{2} + \delta + \frac{1}{K}$ / $w=\frac{1}{2} + \delta + \frac{2}{K}$ (in the threshold averaging version) 
for $0 < \delta < \frac{1}{2}$.
Then Algorithm~\ref{alg:dp} for $k$ selected random decision trees and differential privacy parameter $\eta$ 
gives empirical error $err_{1} \leq 1 - \mu$ with probability $p_{1} \geq 1-n
(e^{-\frac{k\delta^{2}}{2}} + e^{-\frac{k}{2}} + ke^{-\frac{\lambda n \eta}{k}})$ and generalization error $err_{2} \leq 1 - \mu$
with probability $p_{2} \geq p_{1} - 2^{h+3}ke^{-2n\phi^{2}}$, where: $\lambda = \frac{\delta}{24K \cdot 2^{h}}$
and $\phi = \frac{\delta}{2(4+\delta)2^{h}K}$.
Probabilities $p_{1}$ and $p_{2}$ are under random coin tosses used to construct the forest and the test set.
Furthermore, we always have: 
$\mu \geq 1 - \frac{\epsilon}{\frac{1}{2}-\delta-\frac{1}{K}}$ / $ \mu \geq 1 - \frac{\epsilon}{\frac{1}{2}-\delta-\frac{2}{K}}$ in the
majority version and: $ \mu \geq 1-\frac{2\epsilon - 2\epsilon^{2}}{\frac{1}{2}-\delta-\frac{1}{K}}$ / 
$\mu \geq 1-\frac{2\epsilon - 2\epsilon^{2}}{\frac{1}{2}-\delta-\frac{2}{K}}$ in the threshold averaging version.  
\end{theorem}

Notice that if the number of trees $k$ in the forest is logarithmic in $n$ then $p_{1}$ is close to one and so is $p_{2}$.


Again, as in the non-differentially-private case, we see that if there are many points of goodness/weight in $M$ close to 
the average goodness/weight then
empirical and generalization error are small. Notice also that increasing the number of the trees too much has an impact on the empirical error (term $ke^{-\frac{\lambda n \eta}{k}}$ in the lower bound on $p_{1}$). 
More trees means bigger variance of the
single Laplacian used in the leaf of the tree. This affects tree quality. The theorem above describes this phenomenon quantitatively.

\begin{table*}[t!]
\center
\caption{Comparison of the performance of random forests and rpart.} 
\setlength{\tabcolsep}{3.65pt}
\begin{tabular}{ c|c|c|c|c|c|c|c|c|c|c|c|c|c|c|c|c|c|c|}
\cline{4-16}
\multicolumn{3}{c|}{} & \multicolumn{13}{|c|}{\textbf{Method}}\\
\hline
\multicolumn{1}{|c|}{\multirow{2}{*}{Dataset}} & \multicolumn{1}{|c|}{\multirow{2}{*}{n}}  & \multicolumn{1}{|c||}{\multirow{2}{*}{m}}  & \textbf{rpart} & \multicolumn{3}{|c|}{\textbf{n-dpRFMV}} & \multicolumn{3}{|c|}{\textbf{n-dpRFTA}} & \multicolumn{3}{|c|}{\textbf{dpRFMV}} & \multicolumn{3}{|c|}{\textbf{dpRFTA}}\\
\cline{4-16}
\multicolumn{1}{|c|}{} & \multicolumn{1}{|c|}{} & \multicolumn{1}{|c||}{} & Error & Error & k & h & Error & k & h & Error & k & h & Error & k & h\\
\hline
\hline
\multicolumn{1}{|c|}{Ban\_Aut} & 1372 & \multicolumn{1}{|c||}{5} & 3.65\small{$\pm$0.99} & 3.09\small{$\pm$0.92} & 21 & 15 & 3.46\small{$\pm$0.97} & 17 & 9 & 5.44\small{$\pm$1.20} & 21 & 11 & 5.22\small{$\pm$1.18} & 7 & 12\\
\hline
\multicolumn{1}{|c|}{BTSC} & 748 & \multicolumn{1}{|c||}{5} & 18.92\small{$\pm$2.81} & 22.19\small{$\pm$2.98}  & 1 & 14 & 22.47\small{$\pm$2.99} & 1 & 14 & 23.42\small{$\pm$3.03} & 1 & 14 & 23.42\small{$\pm$3.03} & 1 & 13\\
\hline
\multicolumn{1}{|c|}{CVR} & 435 & \multicolumn{1}{|c||}{16} & 9.30\small{$\pm$2.73} & 9.05\small{$\pm$2.70} & 19 & 6 & 5.95\small{$\pm$2.22} & 13 & 9 & 8.10\small{$\pm$2.56} & 15 & 9 & 6.90\small{$\pm$2.38} & 15 & 9\\
\hline
\multicolumn{1}{|c|}{Mam\_Mass} & 961 & \multicolumn{1}{|c||}{6} & 21.88\small{$\pm$2.61} & 16.95\small{$\pm$2.37} & 9 & 12 & 16.21\small{$\pm$2.33} & 19 & 15 & 16.95\small{$\pm$2.37} & 5 & 12 & 17.37\small{$\pm$2.40} & 9 & 8\\
\hline
\multicolumn{1}{|c|}{Mushroom} & 8124 & \multicolumn{1}{|c||}{22} & 3.33\small{$\pm$0.39} & 0.83\small{$\pm$0.20} & 21 & 15 & 0.26\small{$\pm$0.11} & 13 & 14 & 4.69\small{$\pm$0.46} & 3 & 13 & 4.16\small{$\pm$0.43} & 3 & 15\\
\hline
\multicolumn{1}{|c|}{Adult} & 32561 & \multicolumn{1}{|c||}{123} & 17.75\small{$\pm$0.42} & 21.70\small{$\pm$0.45} & 3 & 14 & 21.58\small{$\pm$0.45} & 3 & 14 & 22.18\small{$\pm$0.45} & 3 & 11 & 21.72\small{$\pm$0.45} & 7 & 11\\
\hline
\multicolumn{1}{|c|}{Covertype} & 581012 & \multicolumn{1}{|c||}{54} & 26.90\small{$\pm$0.11} & 33.39\small{$\pm$0.12} & 21 & 15 & 30.80\small{$\pm$0.12} & 21 & 15 & 38.75\small{$\pm$0.13} & 3 & 13 & 37.82\small{$\pm$0.12} & 3 & 13\\
\hline 
\multicolumn{1}{|c|}{Quantum} & 50000 & \multicolumn{1}{|c||}{78} & 32.08\small{$\pm$0.41} & 34.81\small{$\pm$0.42} & 21 & 15 & 33.06\small{$\pm$0.41} & 19 & 14 & 39.91\small{$\pm$0.43} & 21 & 13 & 39.01\small{$\pm$0.43} & 13 & 9\\
  \hline 
\end{tabular} 
\label{tab:fivemethods}
\end{table*}

If the average tree accuracy is big enough then the following result becomes of its own interest. This result considers in particular the empirical error (similar result holds for the generalization error) of the threshold averaging version of Algorithm~\ref{alg:dp} (and also similar result holds for majority voting version of Algorithm~\ref{alg:dp}).
\begin{theorem}
\label{alfa_setting_dp_main_2}
Assume that we are given a parameter $\eta>0$.
Assume besides that the average tree accuracy of the set $M$ of all decision trees of height $h$ on the training set $\mathcal{D}$ is is $e=1-\epsilon$ for some $0<\epsilon \leq \frac{1}{2}$. Let $0<\delta<\frac{1}{2}$. 
Let $\gamma = \frac{1}{ 2^{h} \cdot 9600}$ and let $k_{opt}$ be the integer value for which the value of the function $f(k)=e^{-\frac{k}{200}}+2ke^{-\frac{\gamma \sqrt{n} \eta}{k}}$ is smallest possible.
Then with probability at least $p = 1-n(e^{-\frac{k_{opt}}{200}}+2k_{opt}e^{-\frac{\gamma \sqrt{n} \eta}{k_{opt}}}+e^{-\frac{n}{2}})$ 
the $\eta$-differentially-private threshold averaging version of Algorithm~\ref{alg:dp} gives empirical error at most $\frac{1}{8} + \frac{9}{2}\epsilon-5\epsilon^{2}$ for the forest with $k_{opt}$ randomly chosen decision trees.
Probability $p$ is under random coin tosses used to construct the forest.
\end{theorem}


Both theorems show that logarithmic number of random decision trees in practice suffices to obtain good accuracy and high level of
differential privacy. 

The next theorem considers the differentially-private probabilistic averaging setting. 

\begin{theorem}
\label{beta_setting_dp_main_1}
Assume that we are given a parameter $\eta>0$. Let $K,c>0$ and $0<\delta<1$. 
Assume that the average tree accuracy of the set $M$ of all decision trees of height $h$ on the training/test set $\mathcal{D}$ is $e=1-\epsilon$ for some $0<\epsilon \leq \frac{1}{2}$.  Let $\lambda = \frac{\delta}{24K \cdot 2^{h}}$.
Then for $k$ selected random decision trees the $\eta$-differentially-private probabilistic averaging version of Algorithm~\ref{alg:dp}
gives empirical error $err_{1} \leq 2(\epsilon+\frac{1}{K})-2(\epsilon+\frac{1}{K})^{2} + \delta + c$ with probability $p_{1} \geq 
(1-n(e^{-\frac{k\delta^{2}}{2}}+e^{-\frac{k}{2}}+ke^{-\frac{\lambda n \eta}{k}}))(1-e^{-2nc^{2}})$ and generalization error 
$err_{2} \leq 2(\epsilon+\frac{2}{K})-2(\epsilon+\frac{2}{K})^{2} + \delta + c$
with probability $p_{2} \geq p_{1} - 2^{h+3}ke^{-2n\phi^{2}}$, where: $\phi = \frac{\delta}{2(4+\delta)2^{h}K}$.
Probabilities $p_{1}$ and $p_{2}$ are under random coin tosses used to construct the forest and the test set.
\end{theorem}

As in the two previous settings, information about the average accuracy of just a single tree
gives strong guarantees regarding the classification quality achieved by the differentially-private version of the forest. The next result (analogous to Theorem \ref{alfa_setting_dp_main_2}) shows how to choose the optimal number of trees and that this number is again at most logarithmic in the data size. 


\begin{theorem}
\label{beta_setting_dp_main_2}
Assume that we are given a parameter $\eta>0$.
Assume besides that the average tree accuracy of the set $M$ of all decision trees of height $h$ on the training set $\mathcal{D}$ is $e=1-\epsilon$ for some $0<\epsilon \leq \frac{1}{2}$.
Let $\gamma = \frac{1}{2^{h} \cdot 9600}$ and let $k_{opt}$ be the integer value for which the value of the function $f(k)=e^{-\frac{k}{200}}+2ke^{-\frac{\gamma \sqrt{n} \eta}{k}}$ is smallest possible.
Then with probability at least $p = 1-n(e^{-\frac{k_{opt}}{200}}+2k_{opt}e^{-\frac{\gamma \sqrt{n} \eta}{k_{opt}}}+e^{-\frac{n}{2}})(1-e^{-\frac{n}{200}})$ the $\eta$-differentially-private probabilistic averaging version of Algorithm~\ref{alg:dp} gives empirical error at most  $\frac{1}{5} + \frac{19}{10}\epsilon - 2\epsilon^{2}$  for the forest with $k_{opt}$ randomly chosen decision trees. Probability $p$ is under random coin tosses used to construct the forest.
\end{theorem}


\section{Experiments}

The experiments were performed on the benchmark data-\\sets\footnote{downloaded from \textsf{http://osmot.cs.cornell.edu/kddcup/,\\ \textsf{http://archive.ics.uci.edu/ml/datasets.html}, 
} and \textsf{http://www.csie.ntu.edu.tw/$\sim$cjlin/libsvmtools/datasets}}: \textit{Banknote Authentication} (\textit{Ban\_Aut}), \textit{Blood Transfusion Service Center} (\textit{BTSC}), \textit{Congressional Voting Records} (\textit{CVR}), \textit{Mammographic Mass} (\textit{Mam\_Mass}), \textit{Mushroom}, \textit{Adult}, \textit{Covertype} and \textit{Quantum}. $90\%$ of each dataset was used for training and the remaining part for testing. Furthermore, $10\%$ of the training dataset was used as a
validation set. All code for our experiments is publicly released. 

We first compare the test error ($\%$) obtained using five different methods: open-source implementation of CART called \textit{rpart}~\cite{rpart}, non-differentially-private (\textit{n-dp}) and diffe-\\rentially-private (\textit{dp}) random forest with majority voting (\textit{RFMV}) and threshold averaging (\textit{RFTA}). For all methods except \textit{rpart} we also report the number of trees in the forest ($k$) and the height of the tree ($h$) for which the smallest validation error was obtained, where we explored: $h \in \{1, 2, 3, \dots, 15\}$ and $k \in \{1,3,5, \dots, 21\}$. In all the experiments the differential privacy parameter $\eta$ was set to $\eta = 1000/n_{tr}$, where $n_{tr}$ is the number of training examples. Table~\ref{tab:fivemethods} captures the results. For each experiment we report average test error over $10$ runs. We also show the binomial symmetrical $95\%$ confidence intervals for our results. The performance of random forest with probabilistic averaging (\textit{RFPA}) was significantly worse than the competitive methods (\textit{RFMV}, \textit{RFTA}, \textit{rpart}) and is not reported in the table. The performance of \textit{RFPA} will however be shown in the next set of results.
\begin{figure*}[htp!]
\center
\begin{tabular}{cc}
\textbf{\textit{Banknote Authentication}}\\
\hspace{-0.1in}\includegraphics[width = 1.45in]{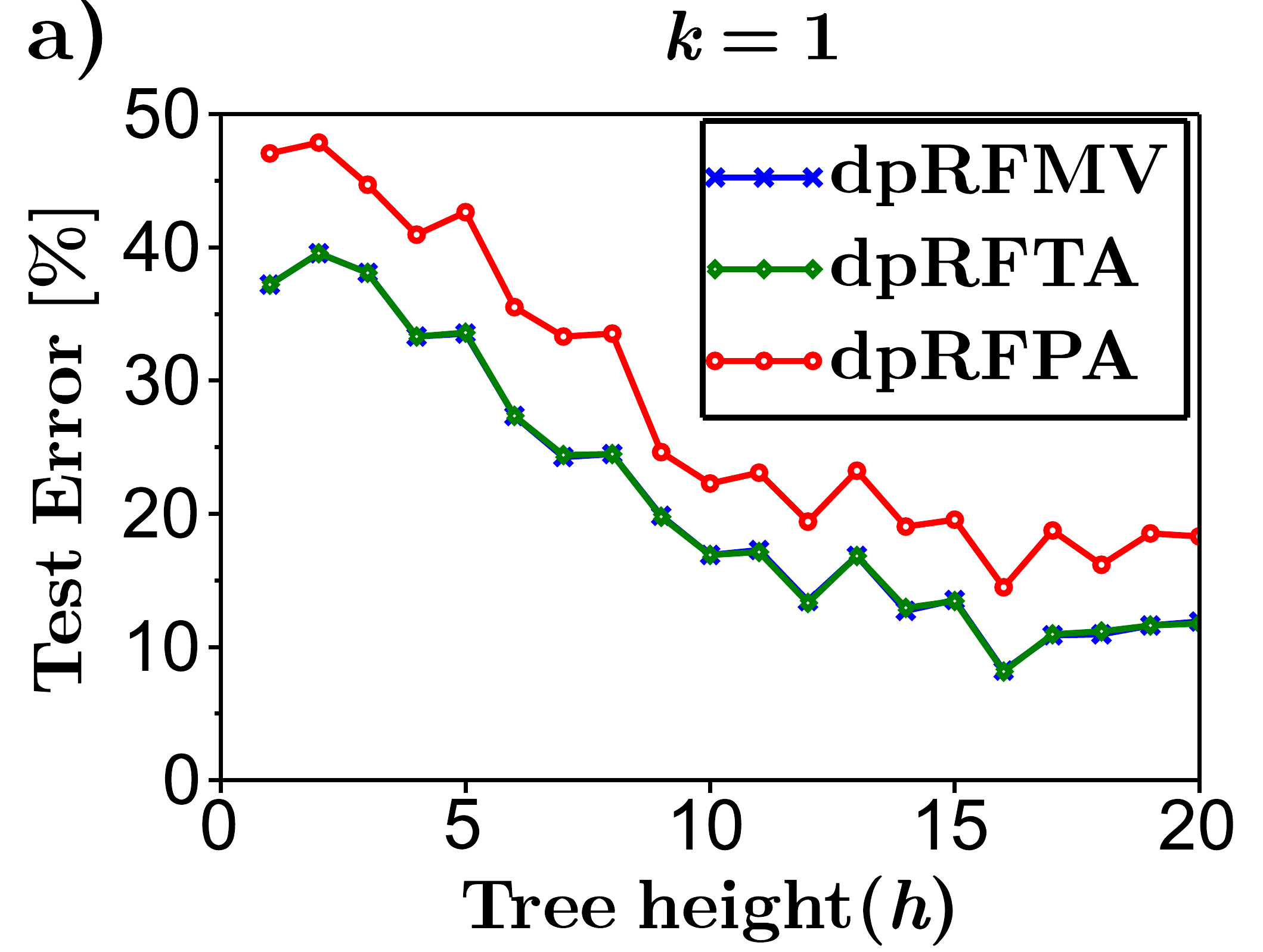} 
\hspace{-0.04in}\includegraphics[width = 1.45in]{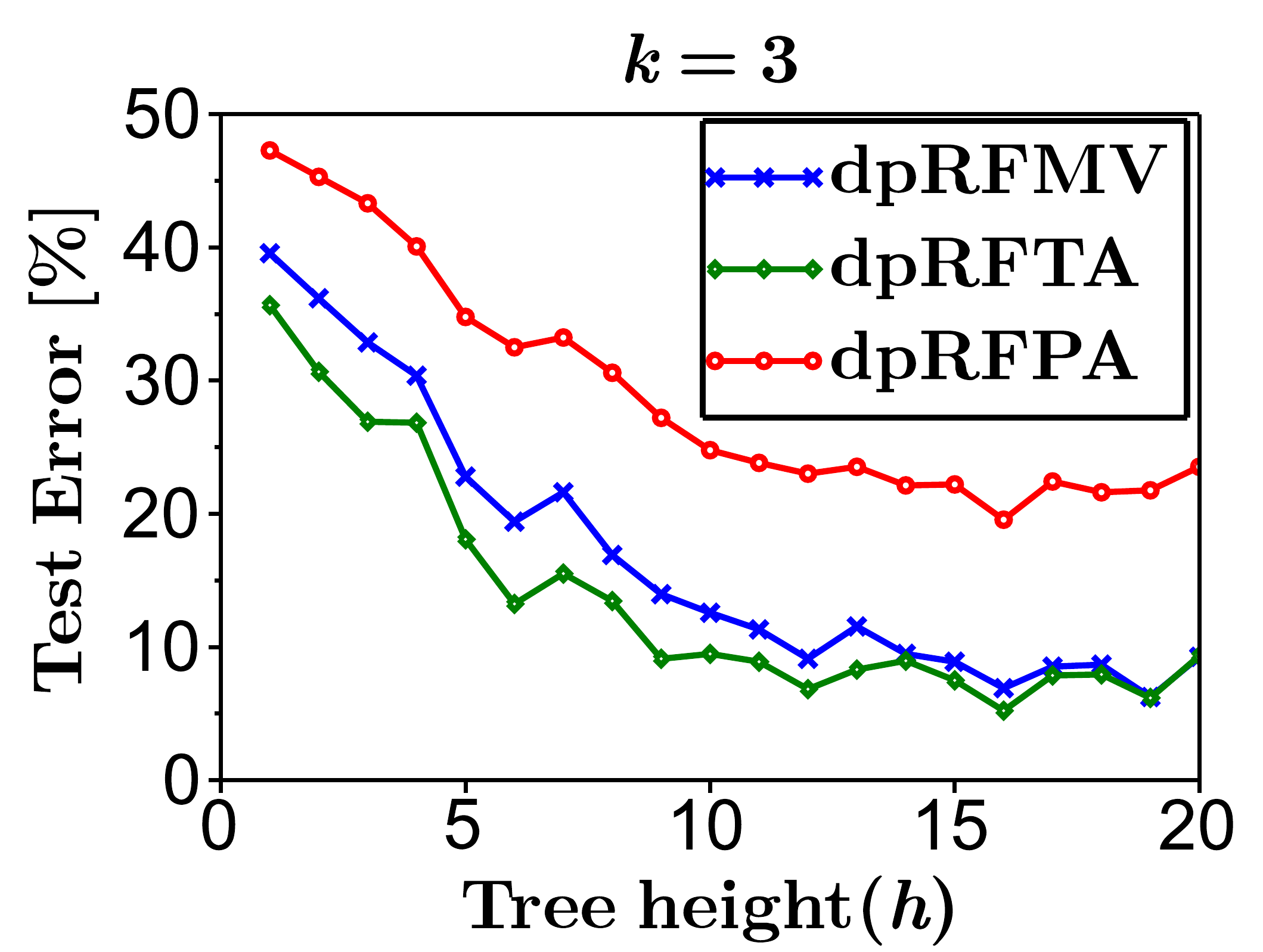} 
\hspace{-0.035in}\includegraphics[width = 1.45in]{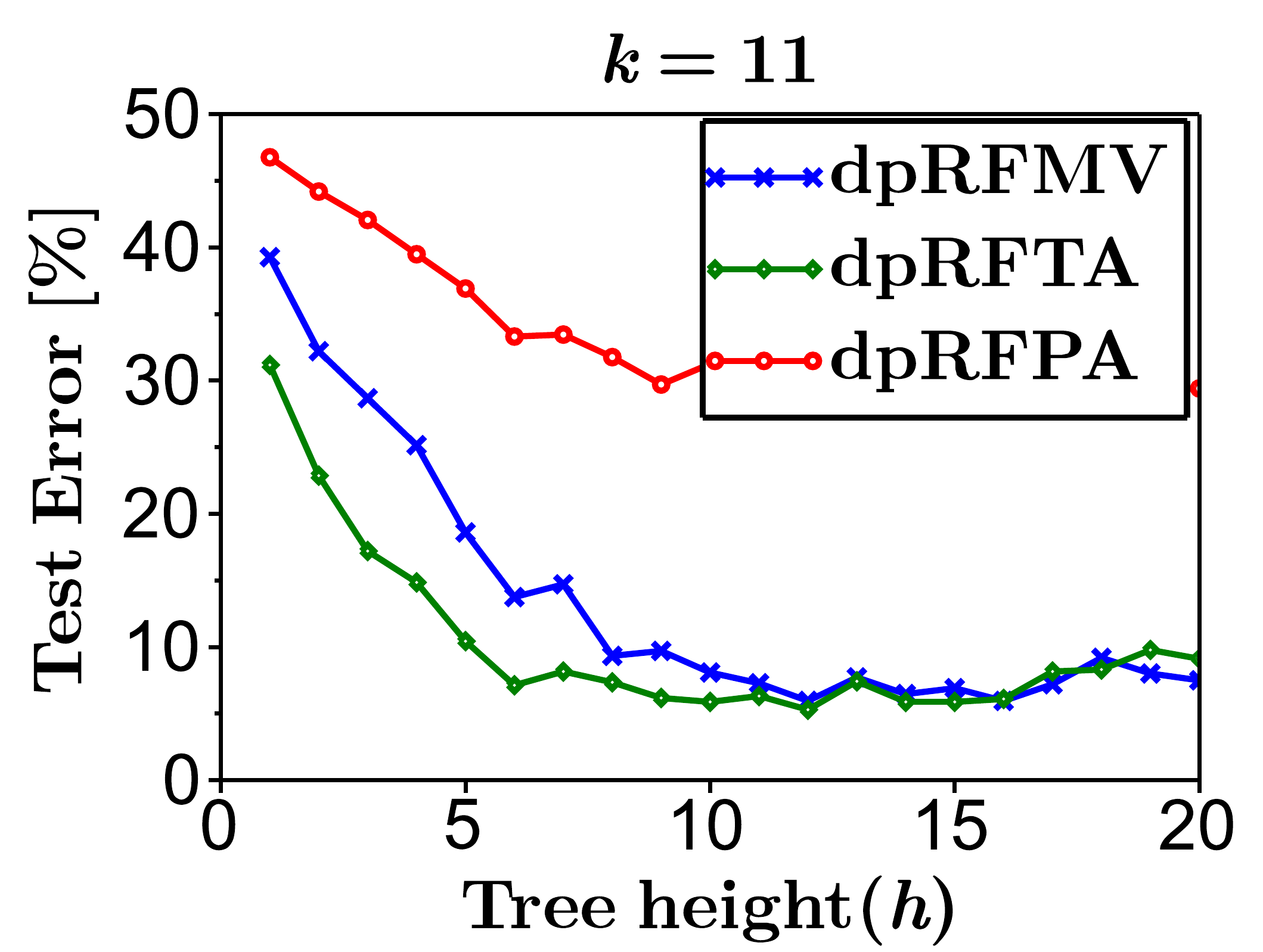} 
\hspace{-0.04in}\includegraphics[width = 1.45in]{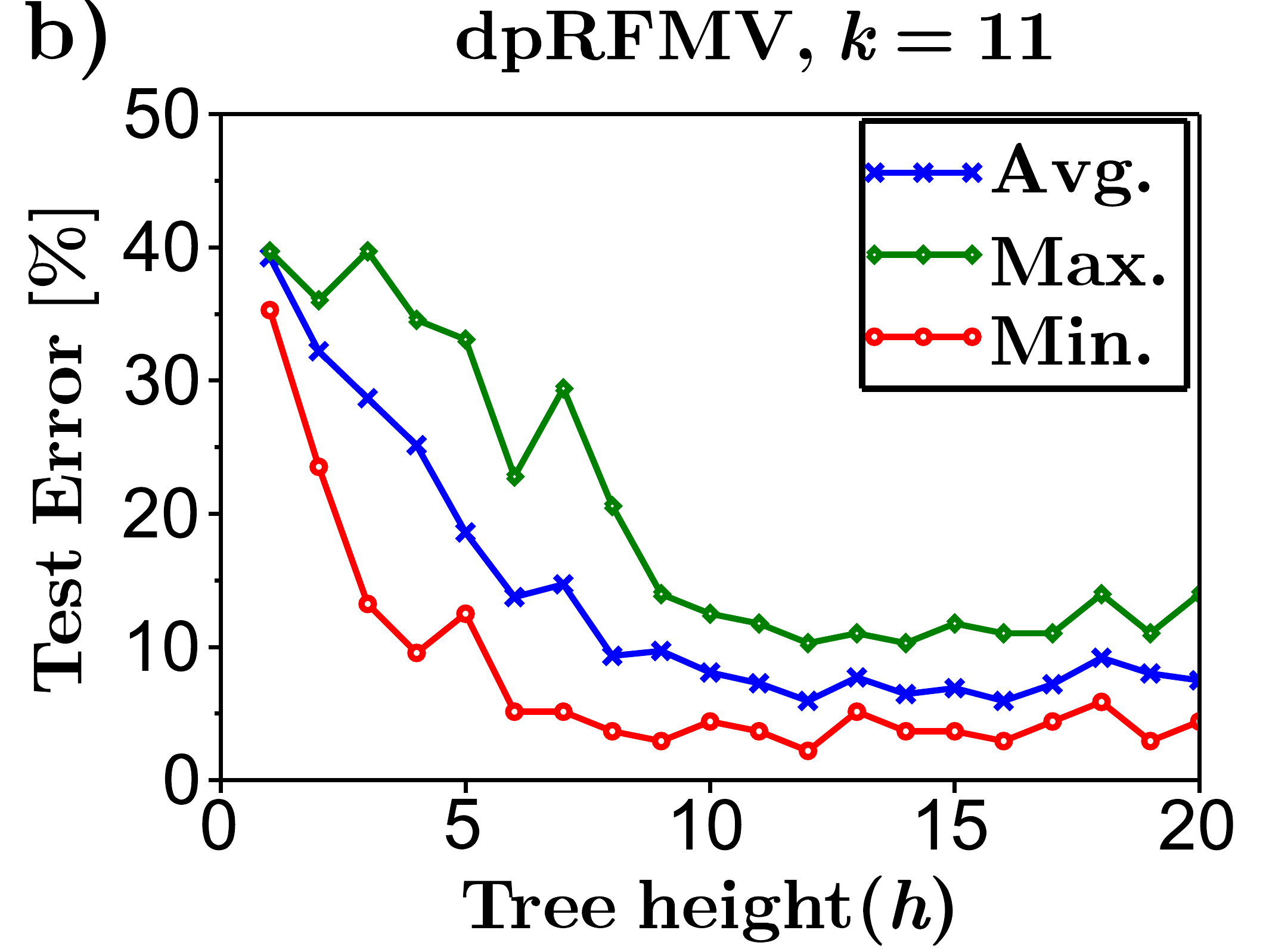}\\
\hspace{-0.1in}\includegraphics[width = 1.45in]{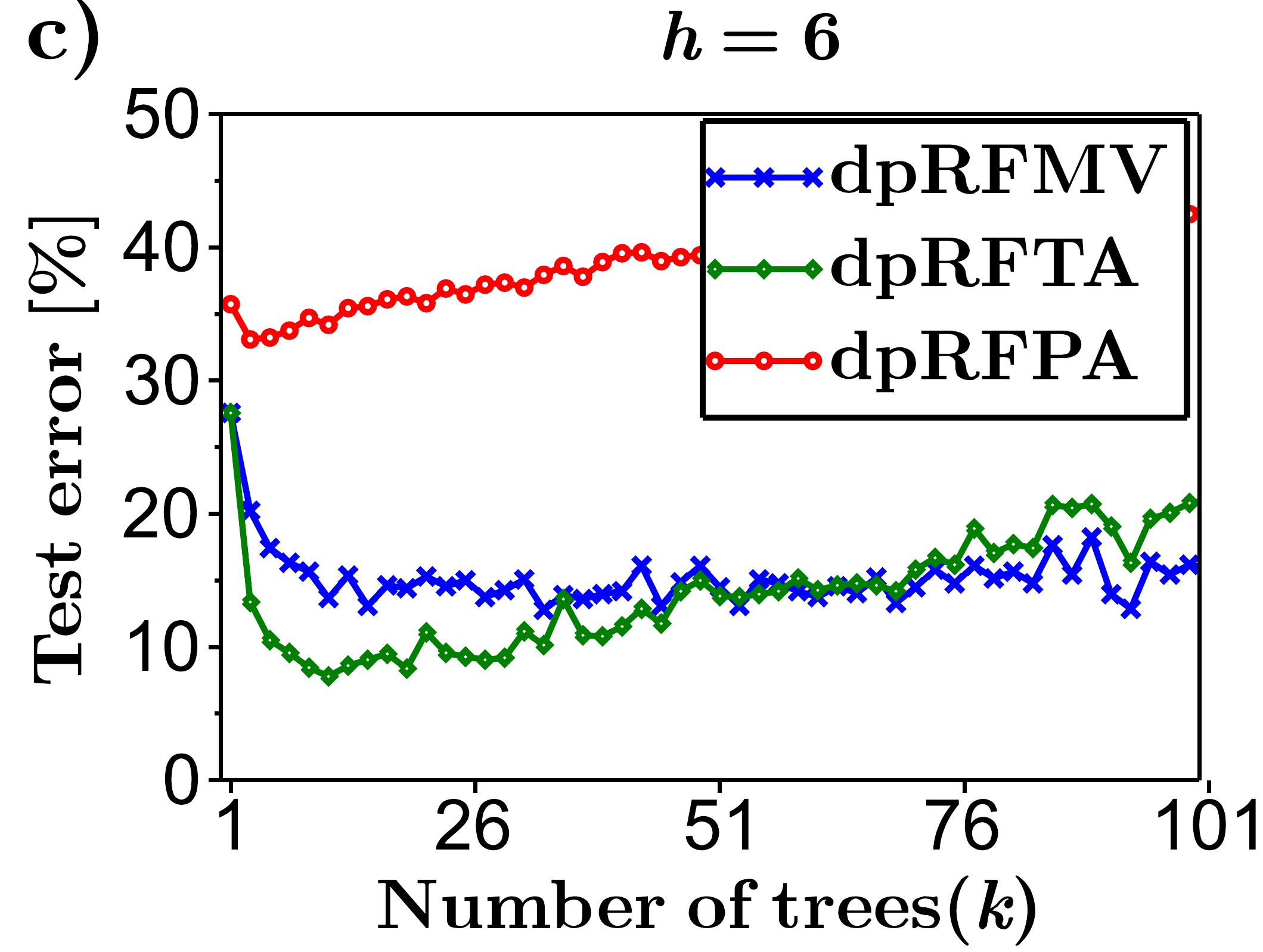} 
\hspace{-0.04in}\includegraphics[width = 1.45in]{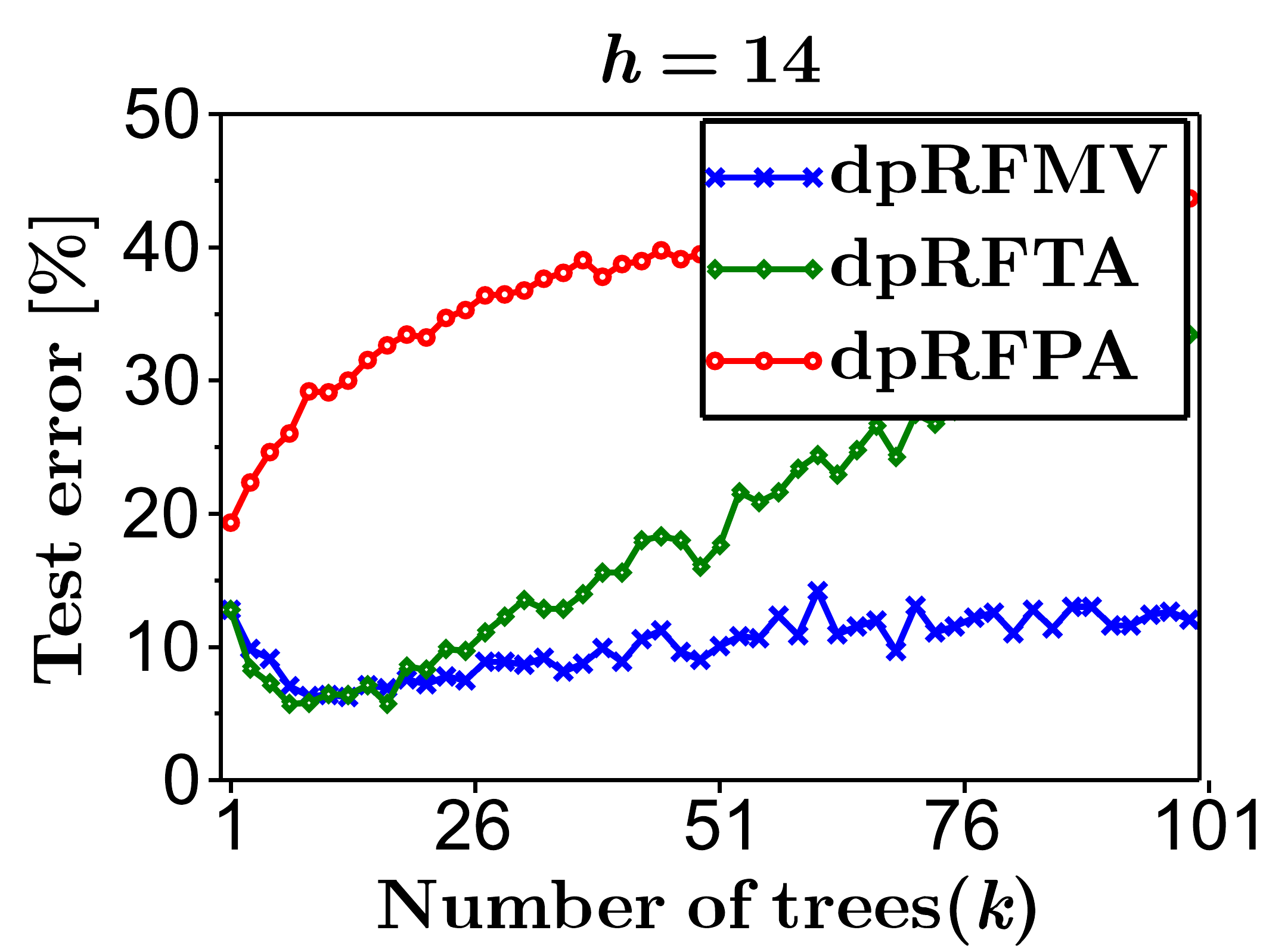} 
\hspace{-0.035in}\includegraphics[width = 1.45in]{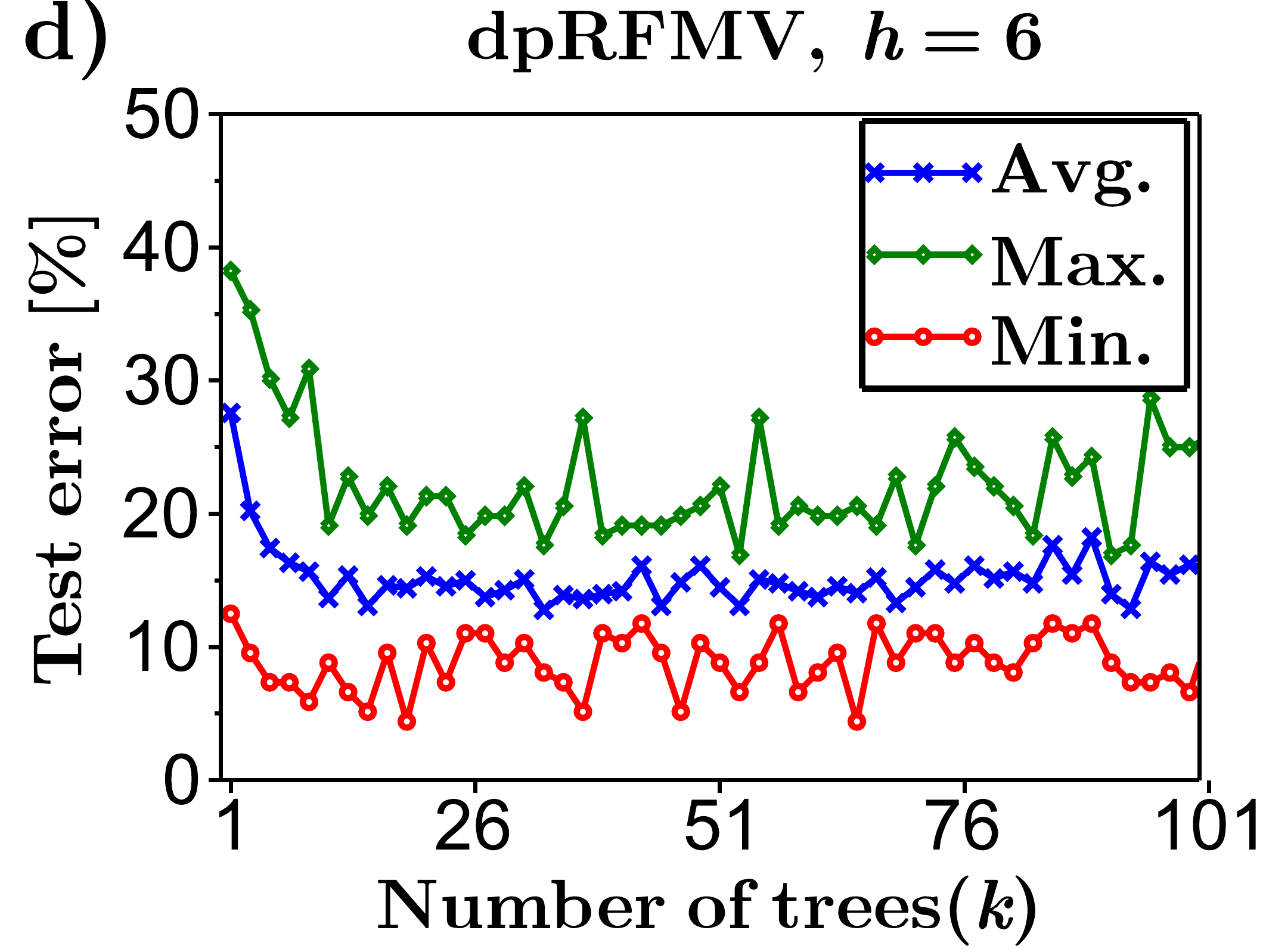} 
\hspace{-0.04in}\includegraphics[width = 1.45in]{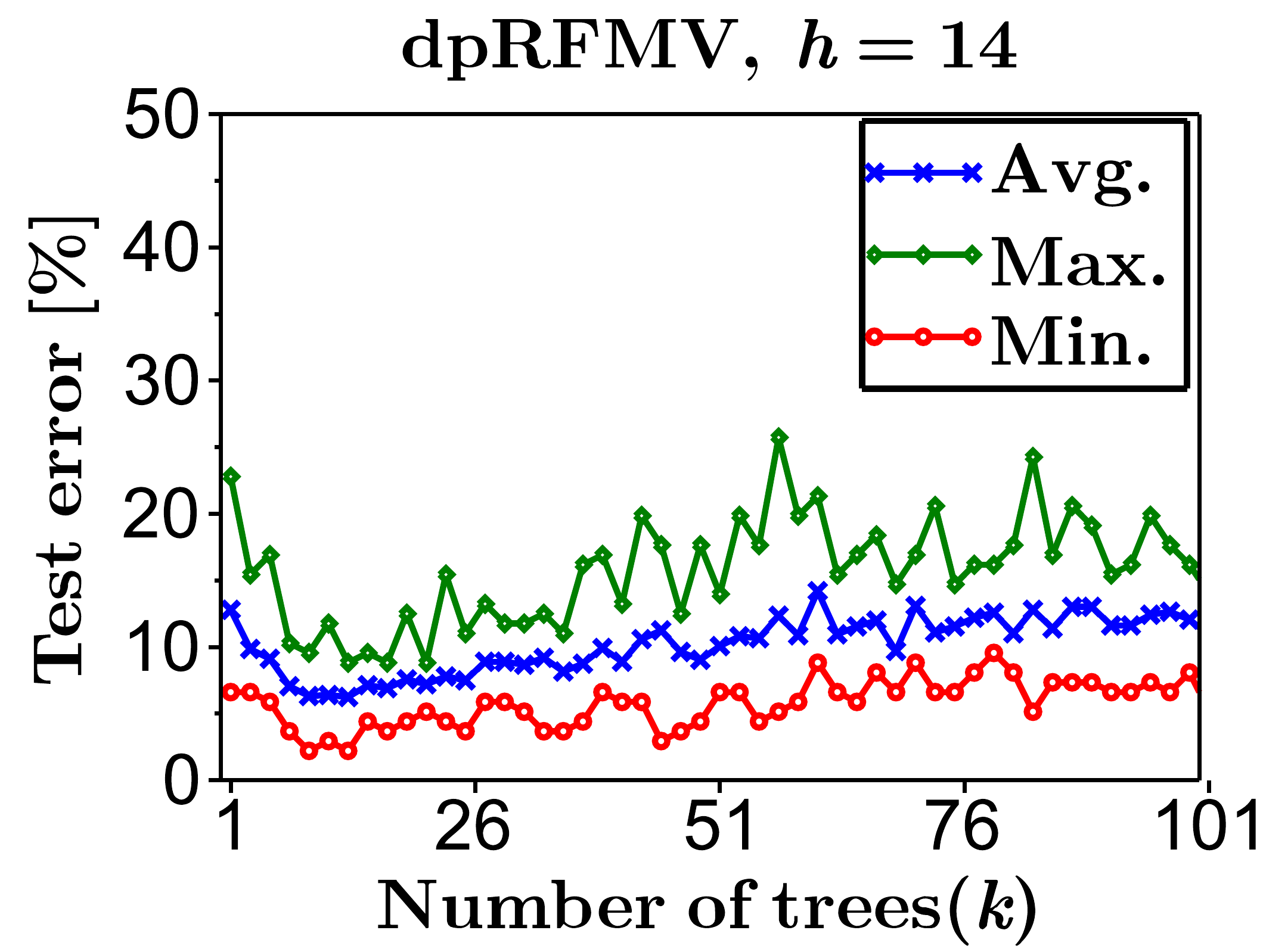}\\

\textbf{\textit{Congressional Voting Records}}\\
\hspace{-0.1in}\includegraphics[width = 1.45in]{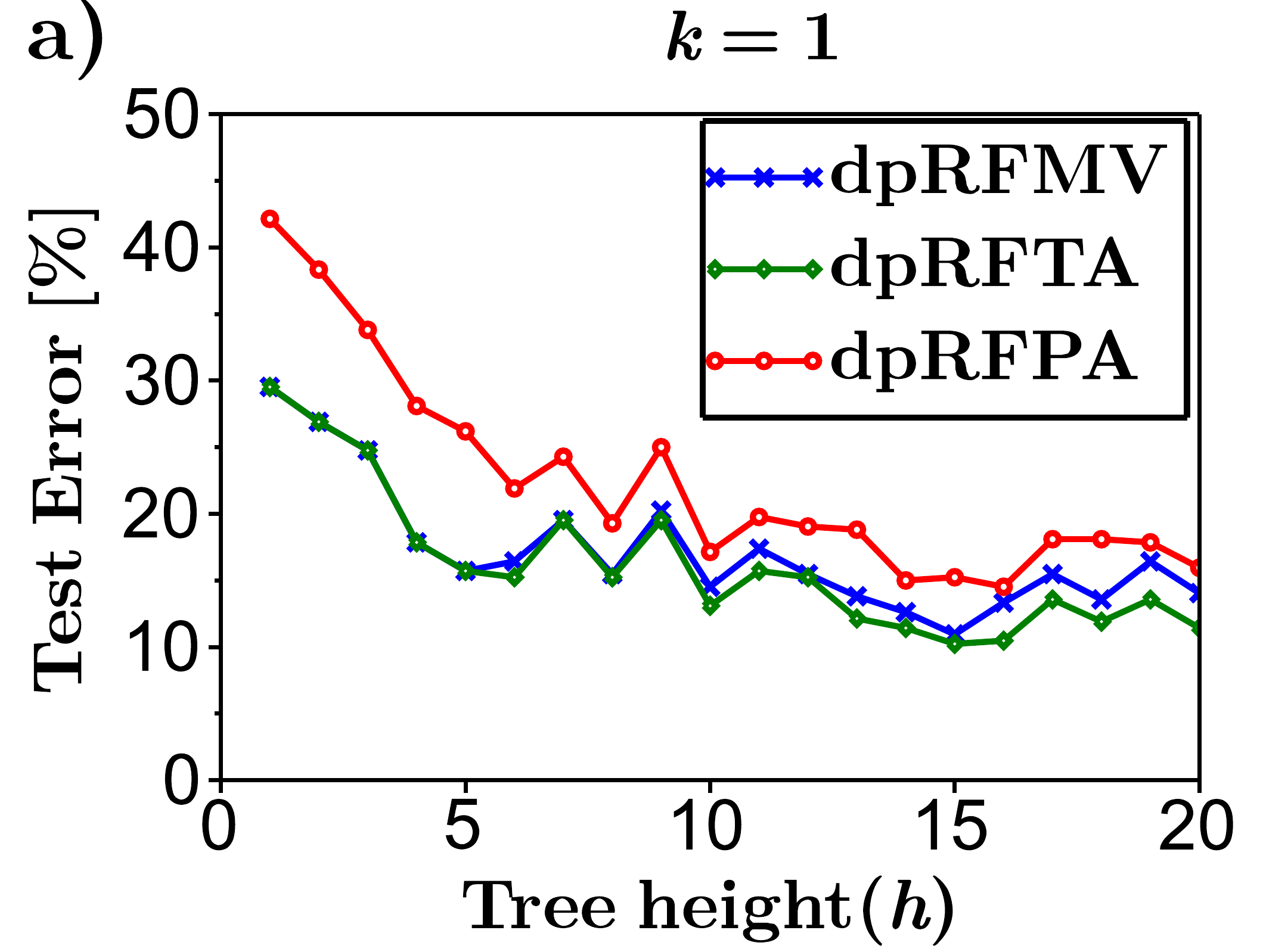} 
\hspace{-0.04in}\includegraphics[width = 1.45in]{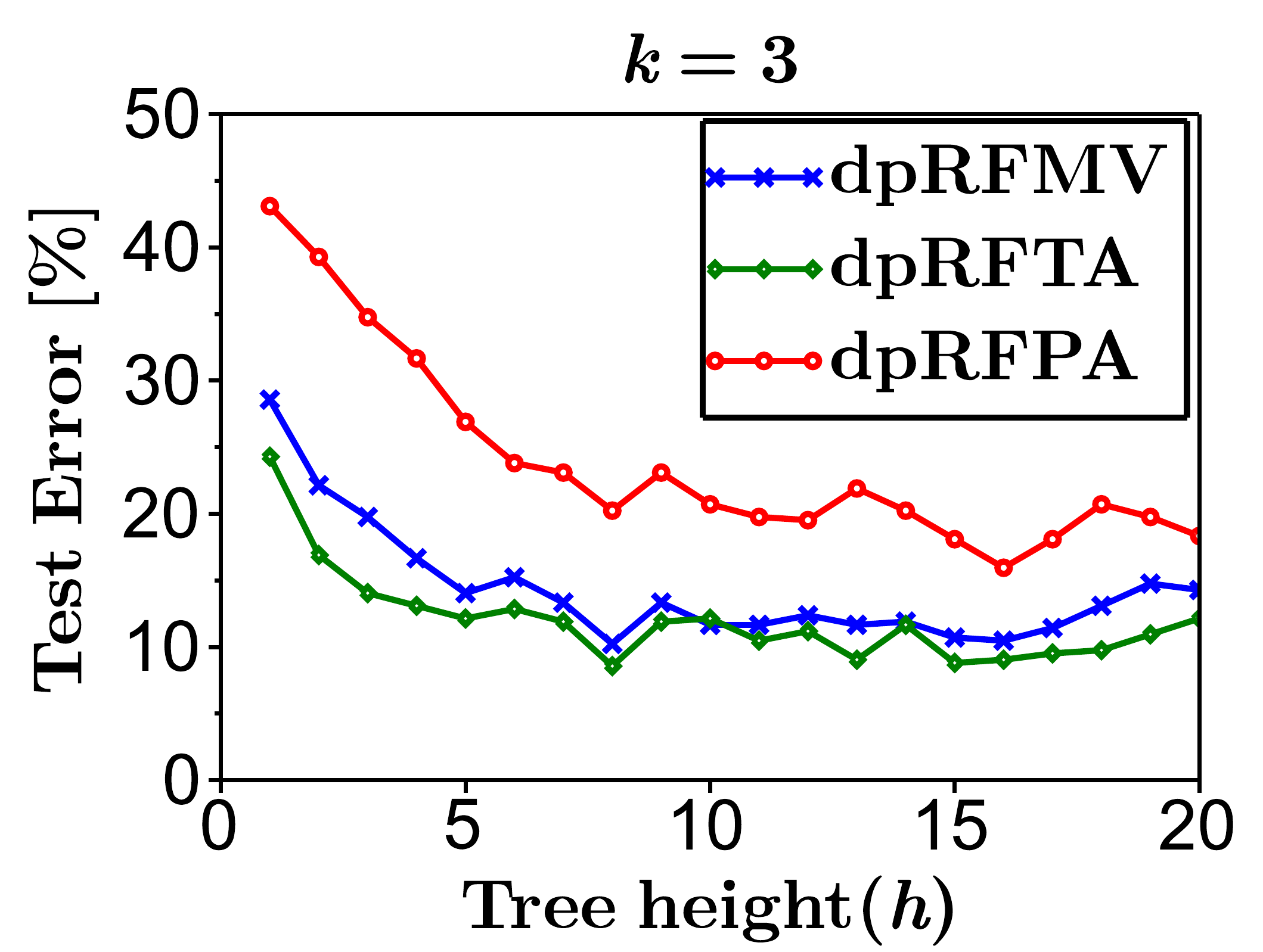} 
\hspace{-0.035in}\includegraphics[width = 1.45in]{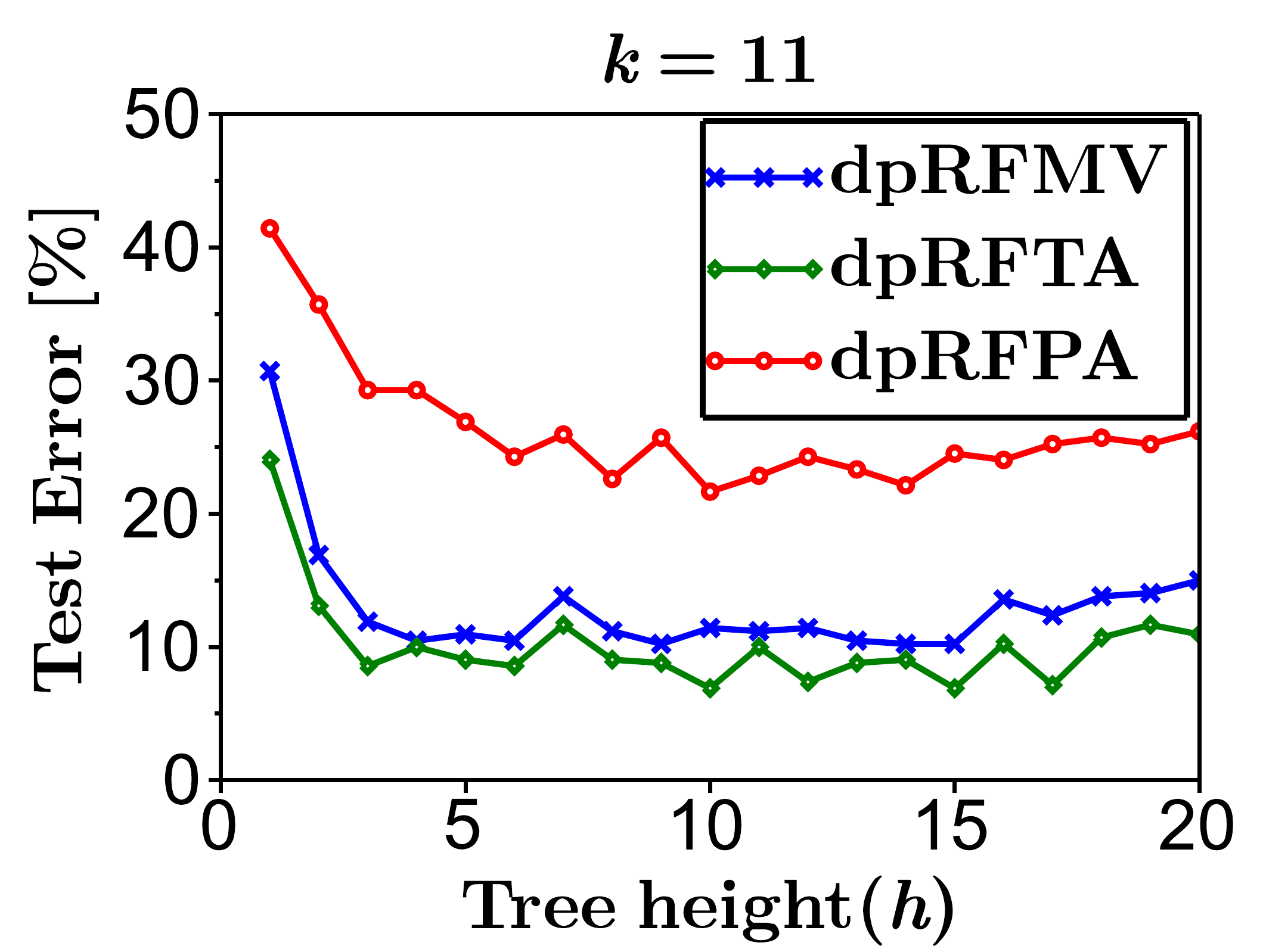} 
\hspace{-0.04in}\includegraphics[width = 1.45in]{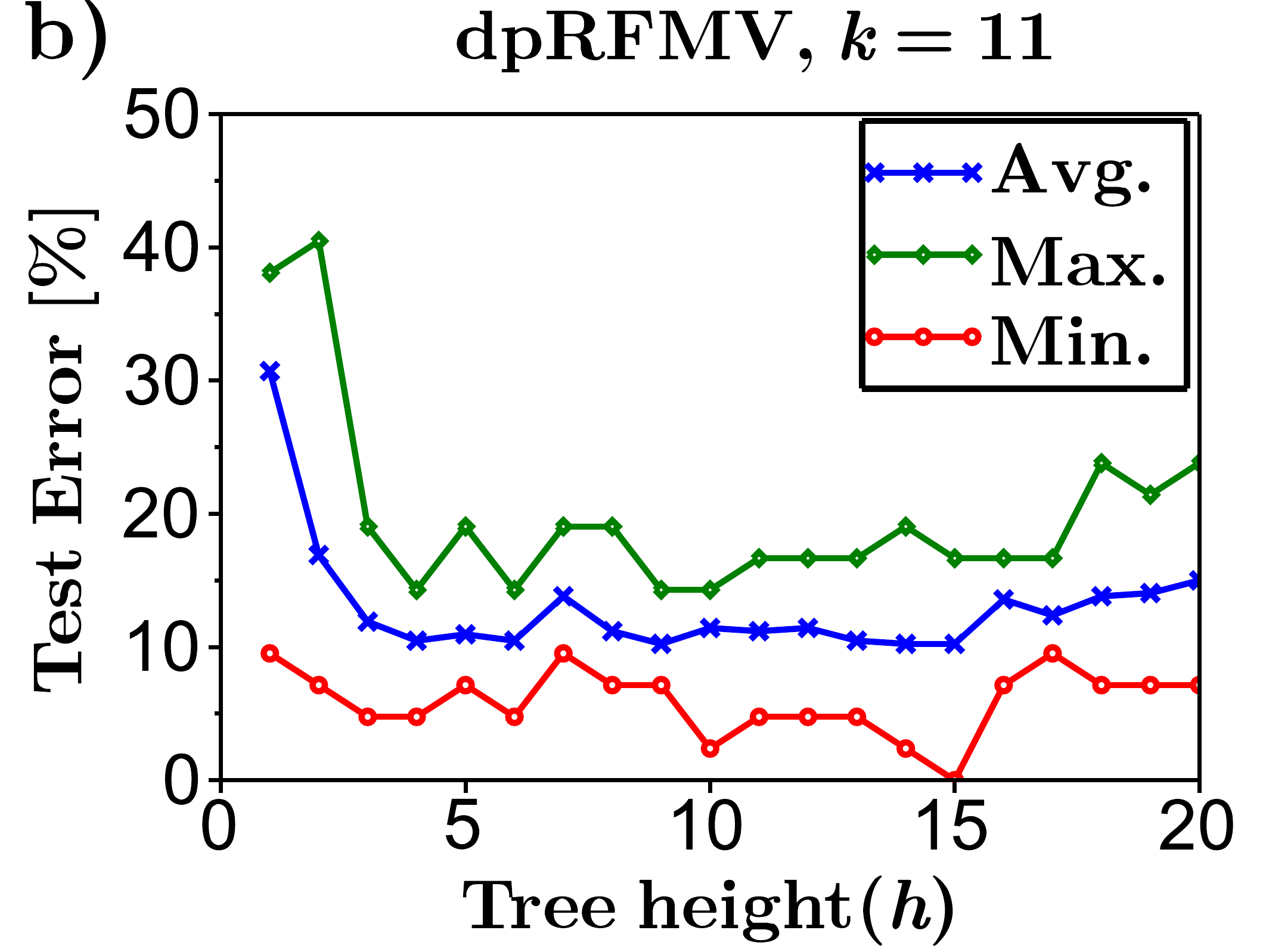}\\  
\hspace{-0.1in}\includegraphics[width = 1.45in]{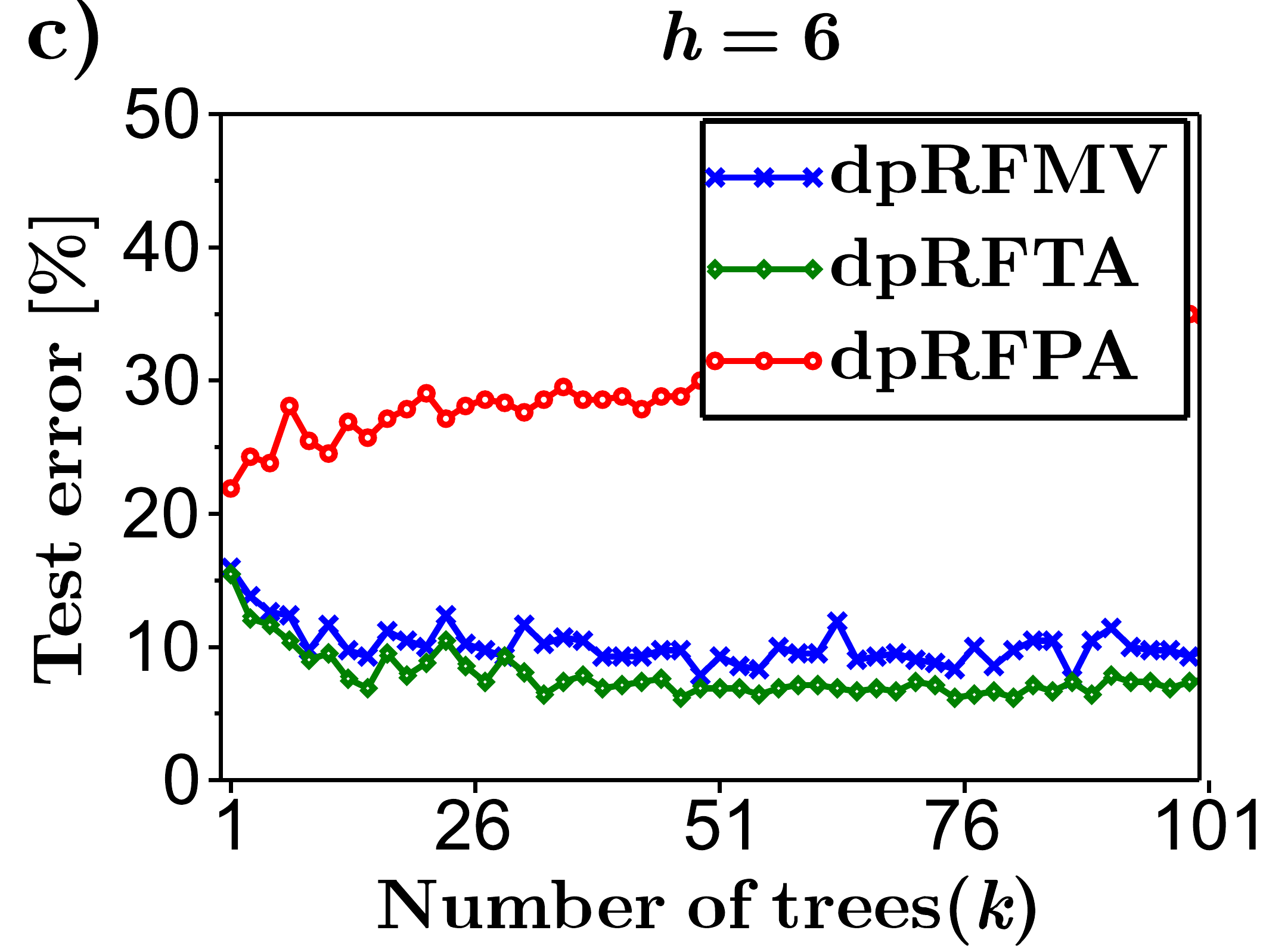} 
\hspace{-0.04in}\includegraphics[width = 1.45in]{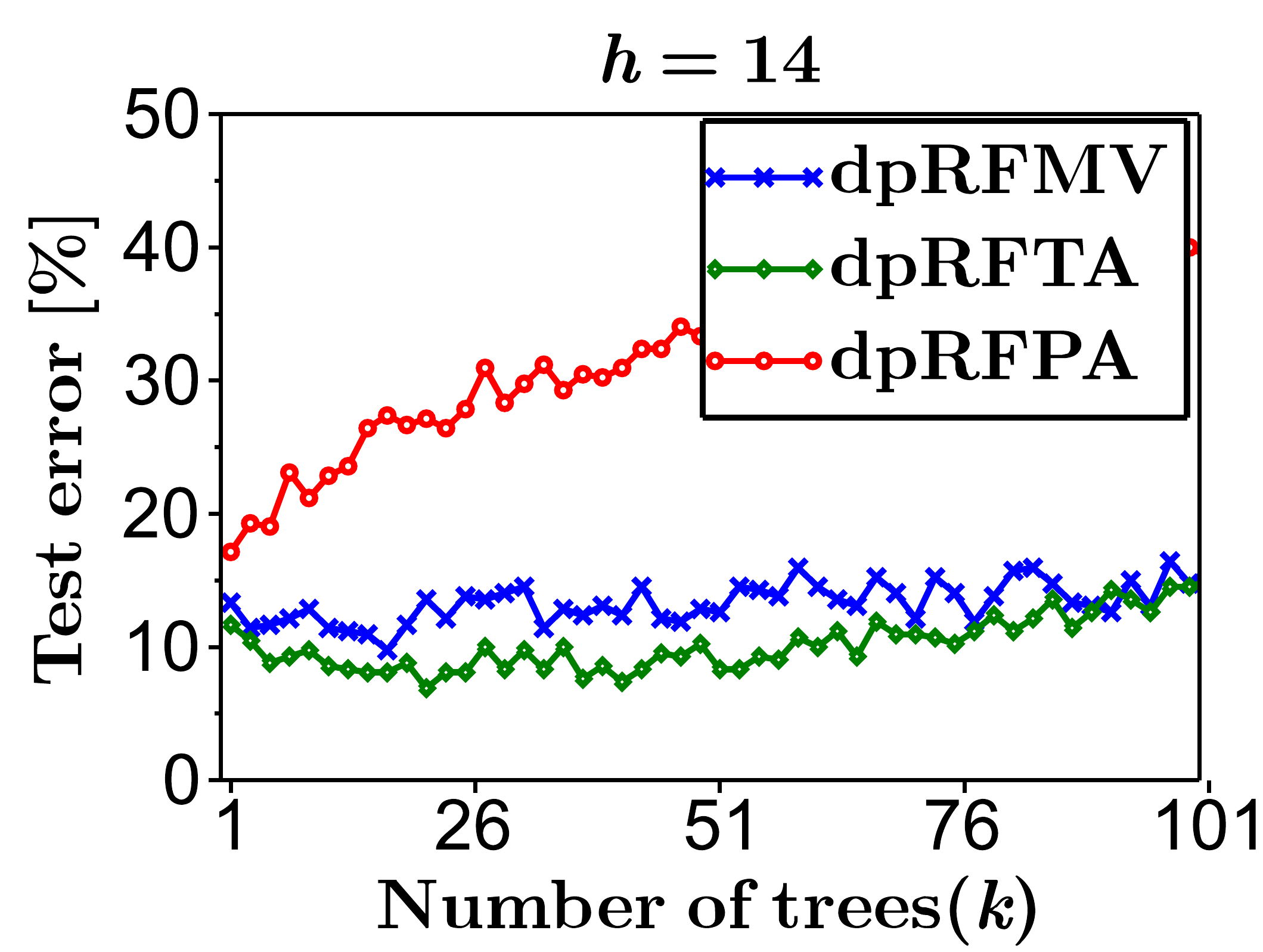} 
\hspace{-0.035in}\includegraphics[width = 1.45in]{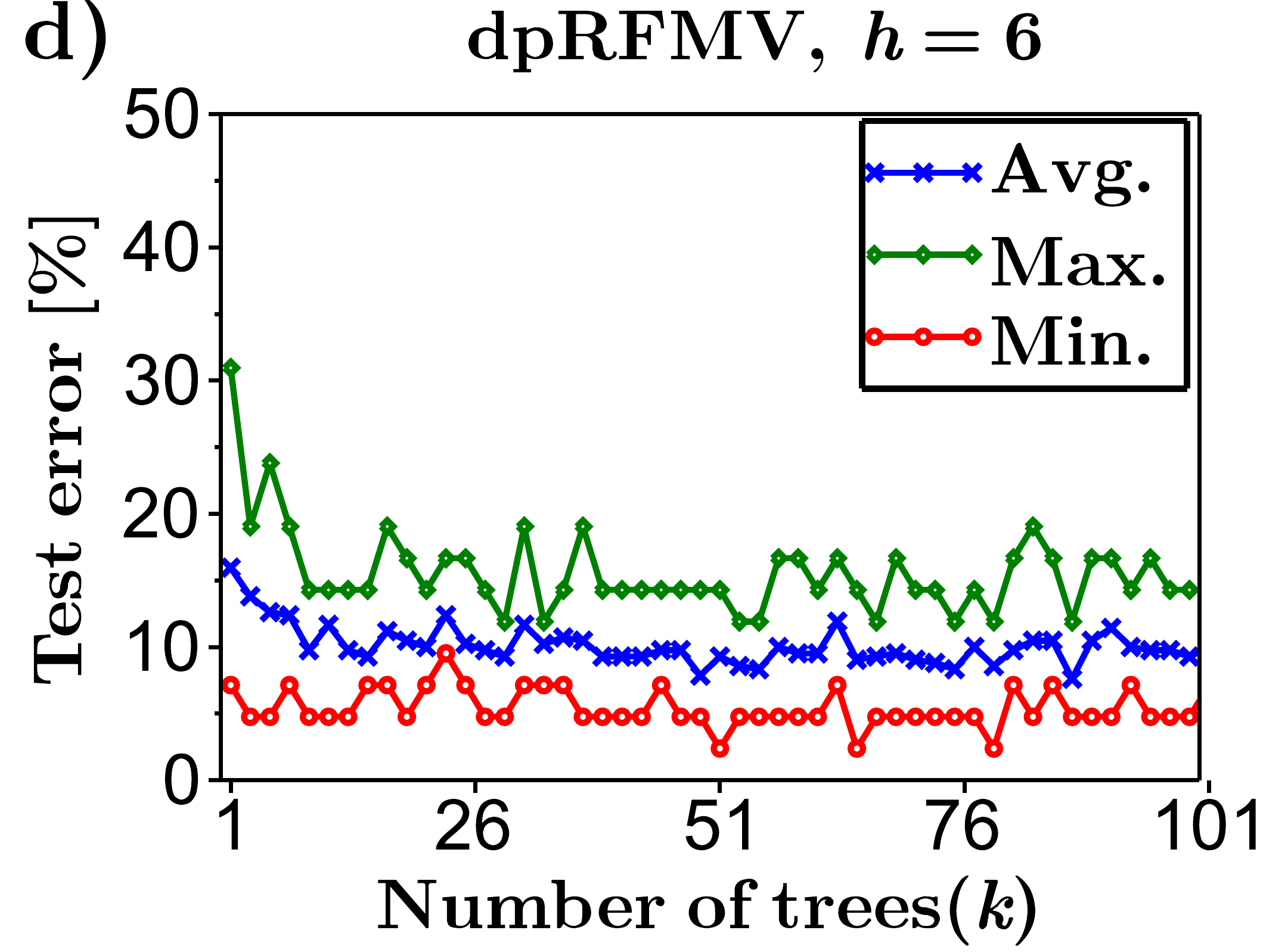} 
\hspace{-0.04in}\includegraphics[width = 1.45in]{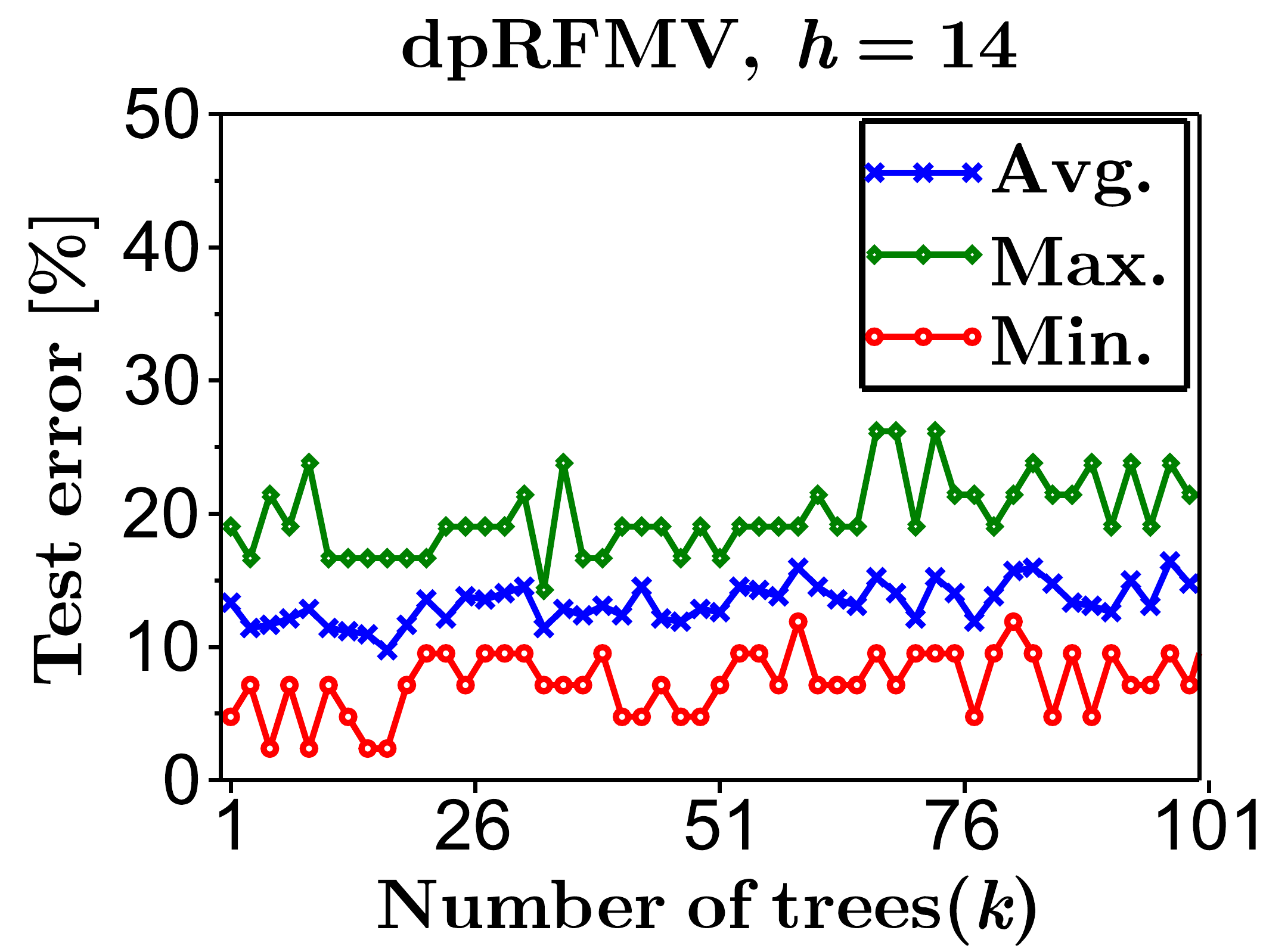}\\

\textbf{\textit{Mammographic Mass}}\\
\hspace{-0.1in}\includegraphics[width = 1.45in]{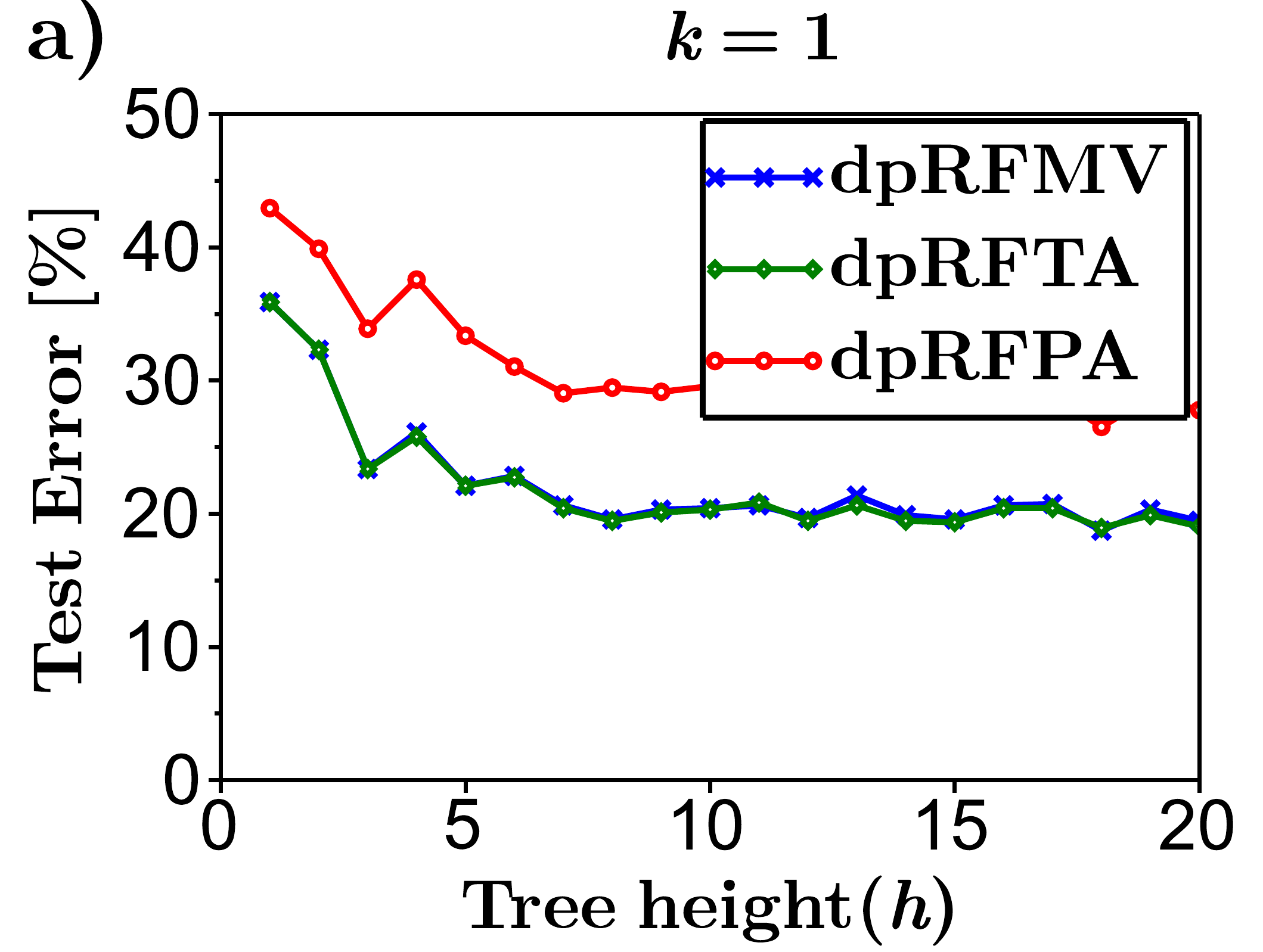} 
\hspace{-0.04in}\includegraphics[width = 1.45in]{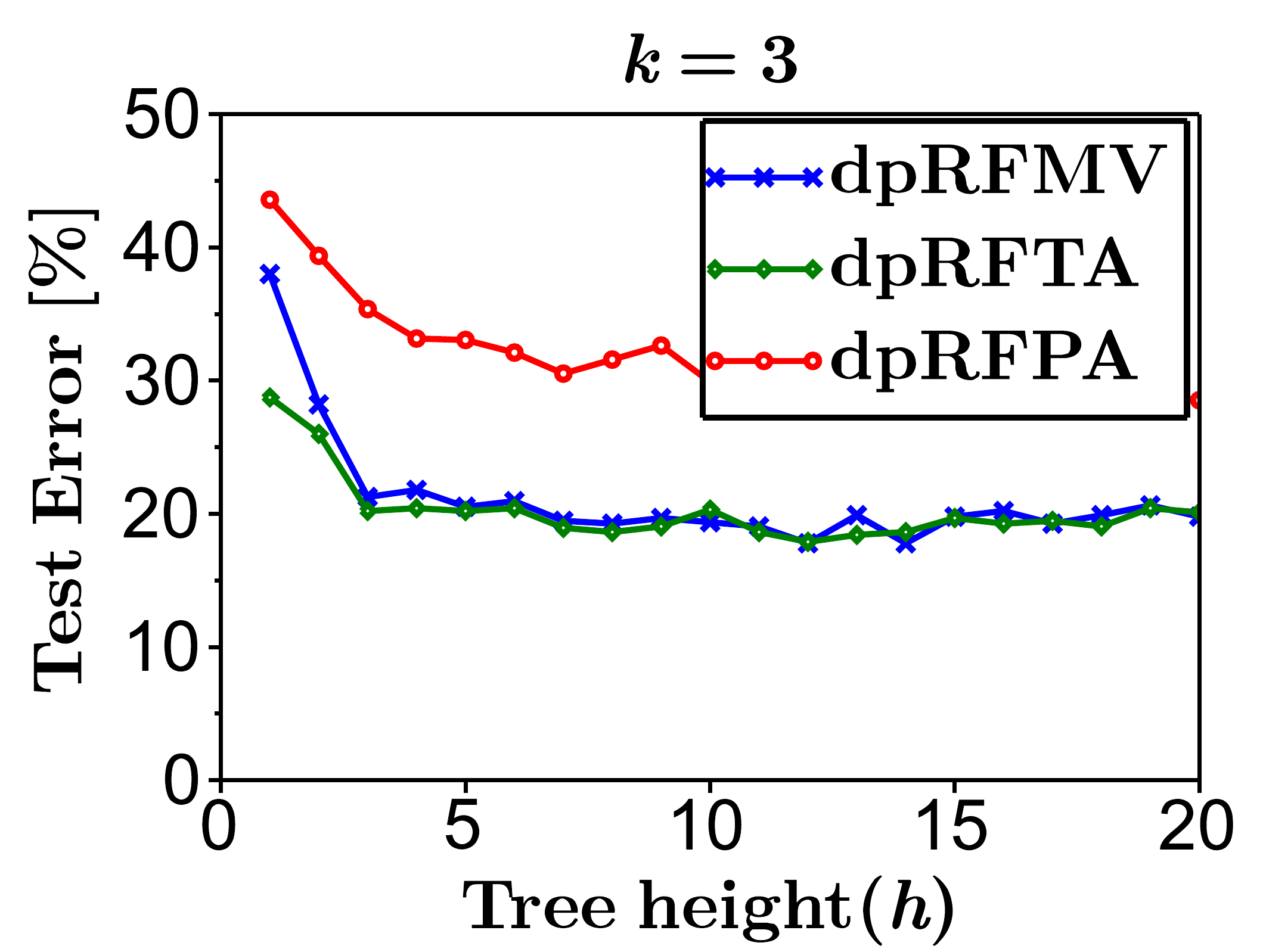} 
\hspace{-0.035in}\includegraphics[width = 1.45in]{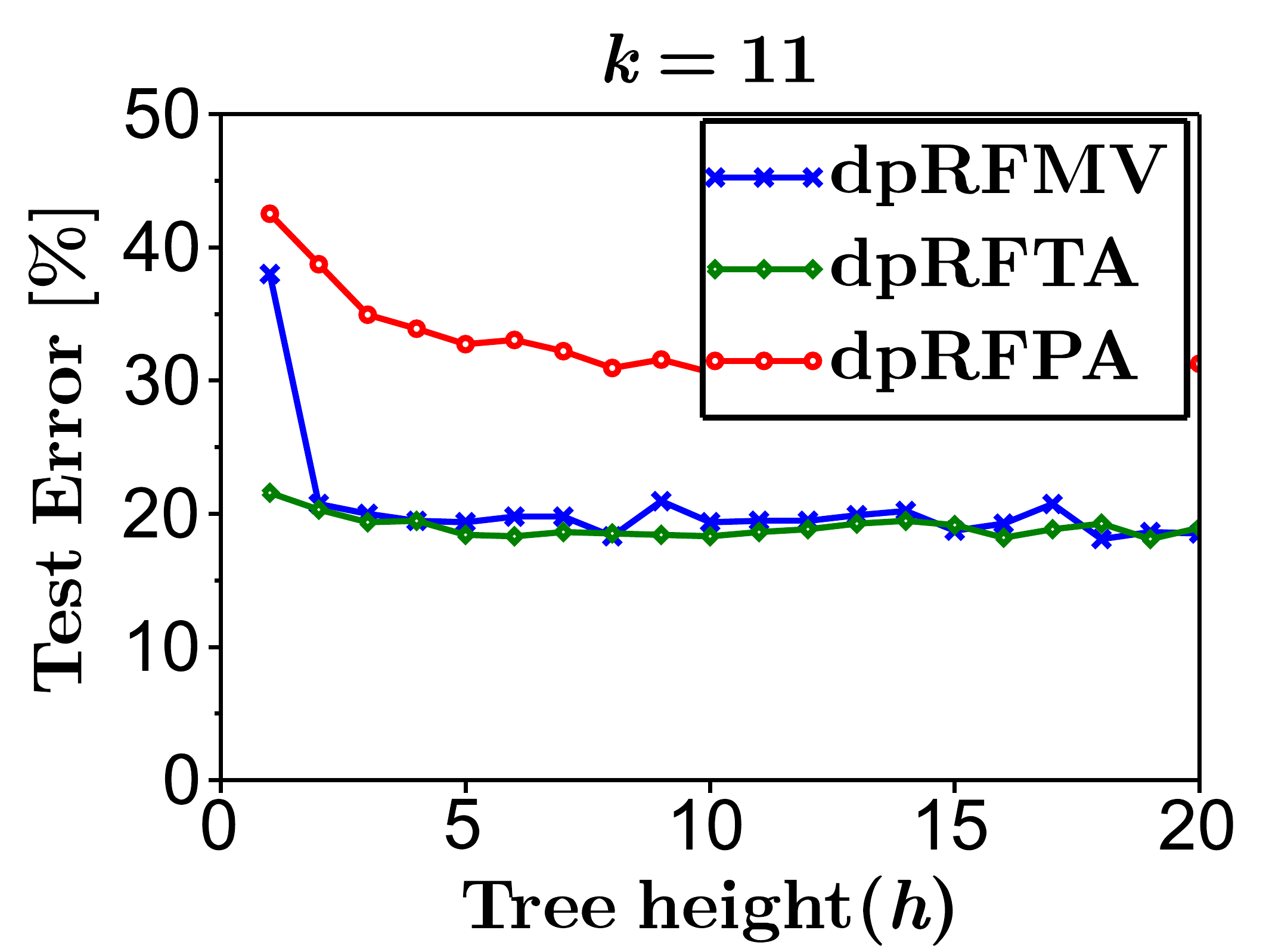} 
\hspace{-0.04in}\includegraphics[width = 1.45in]{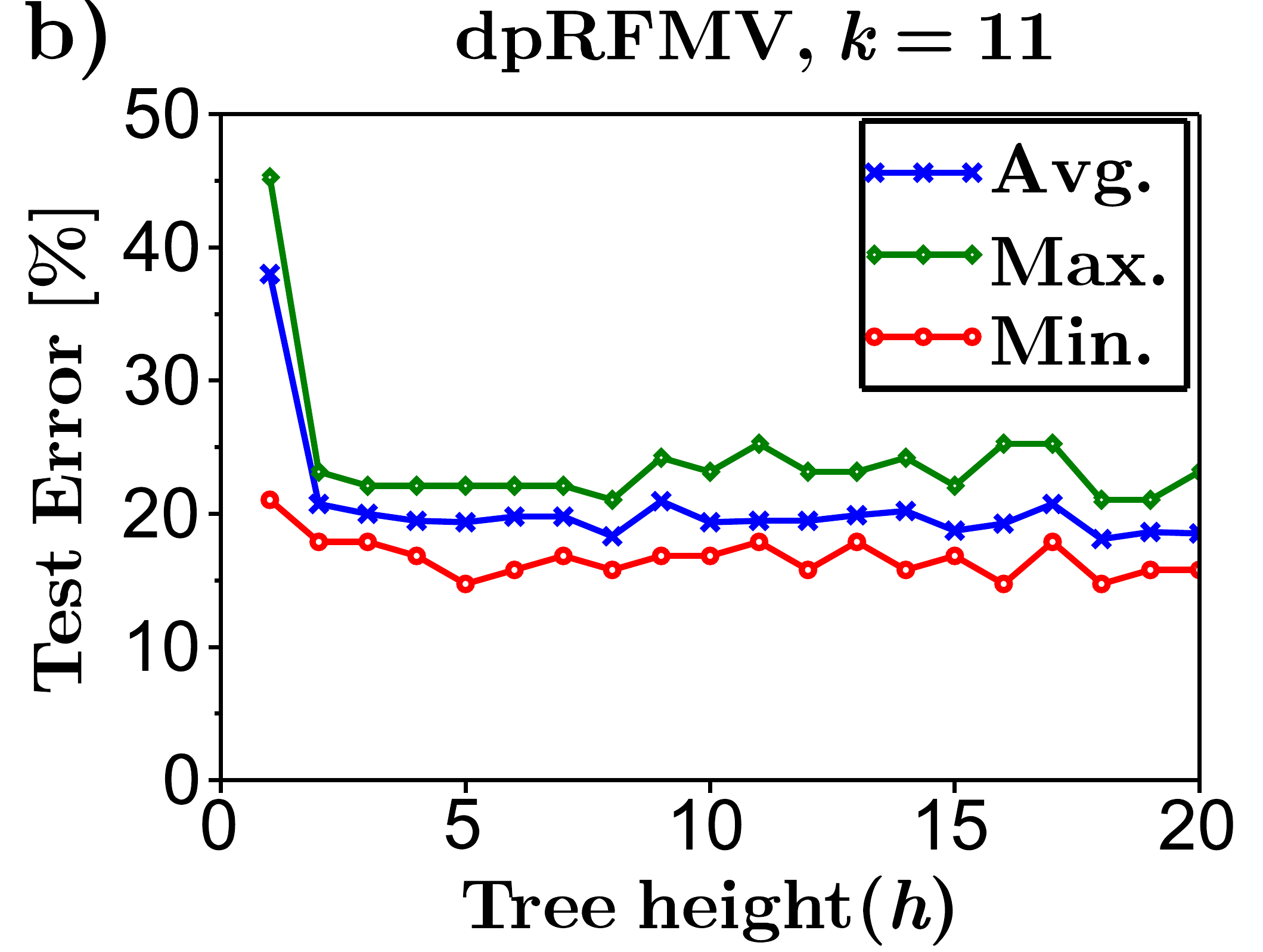}\\  
\hspace{-0.1in}\includegraphics[width = 1.45in]{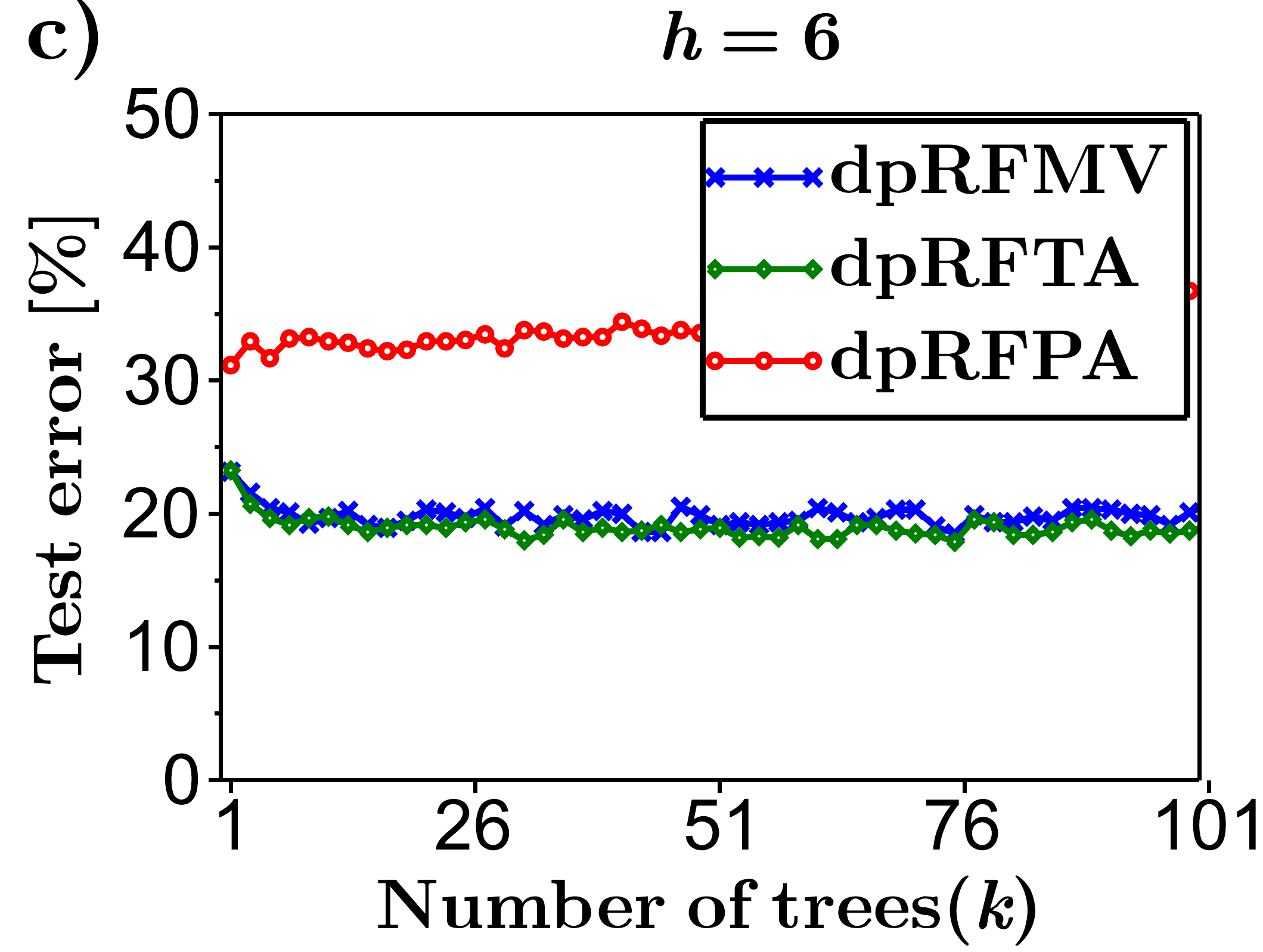} 
\hspace{-0.04in}\includegraphics[width = 1.45in]{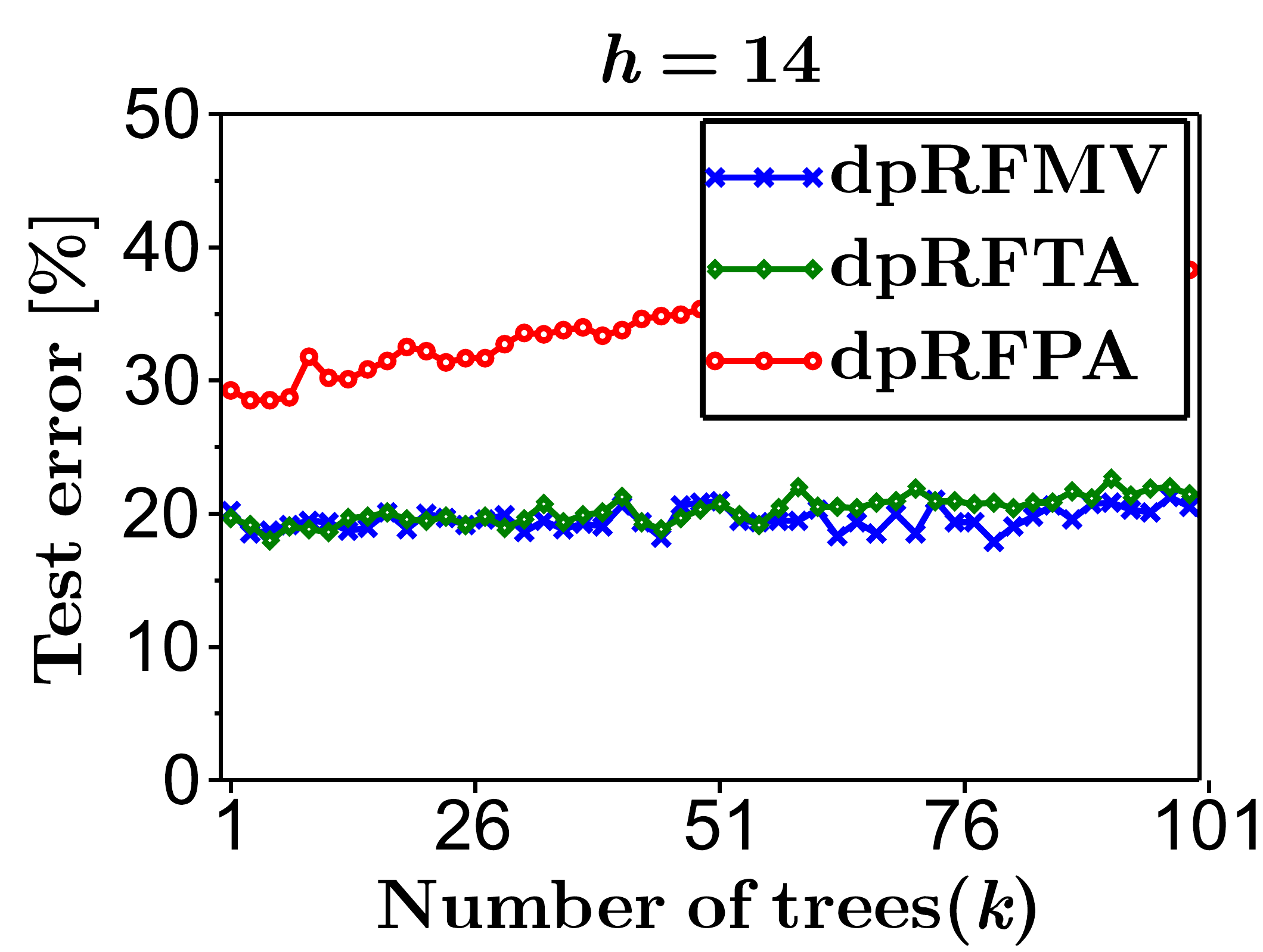} 
\hspace{-0.035in}\includegraphics[width = 1.45in]{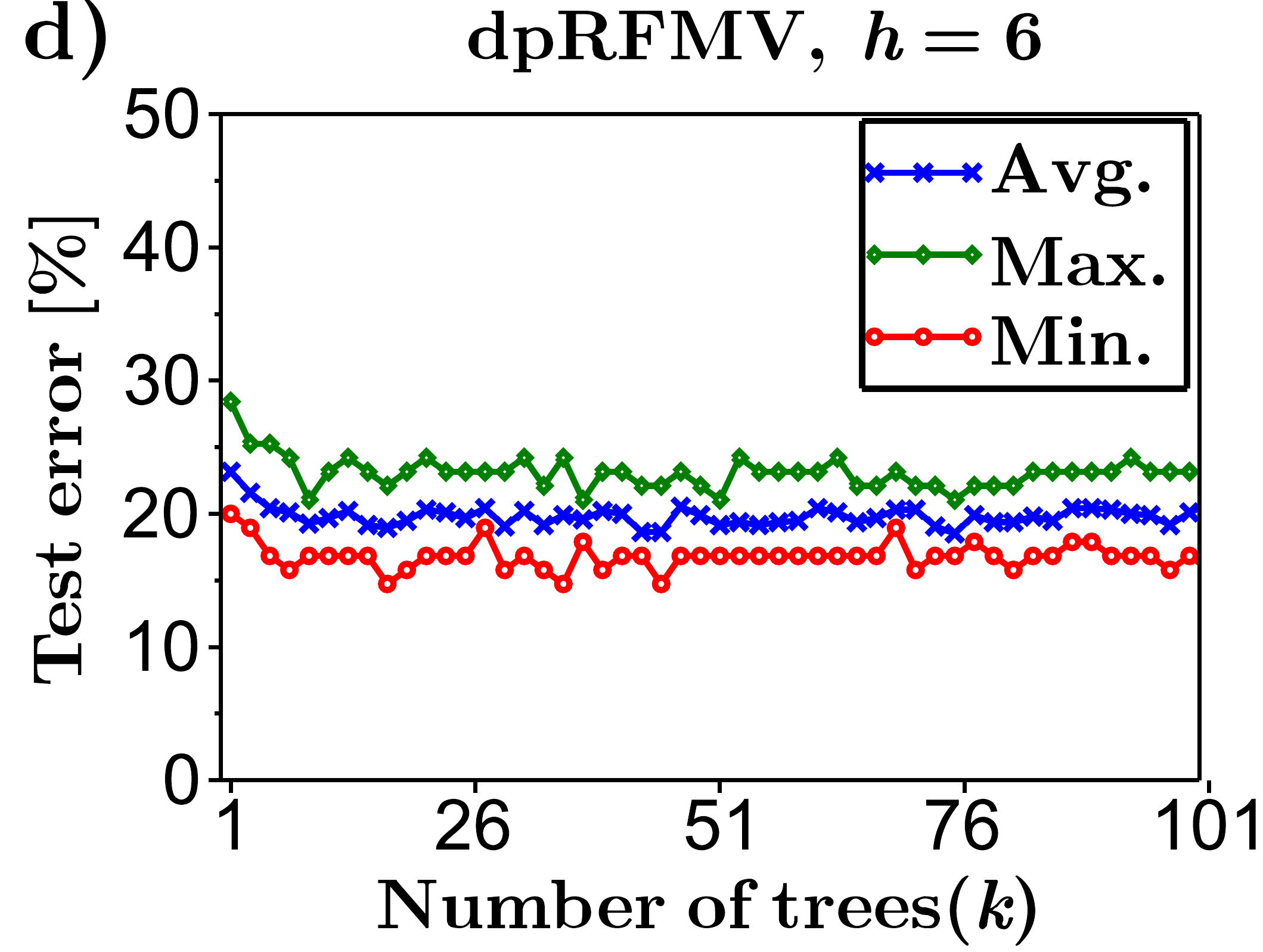} 
\hspace{-0.04in}\includegraphics[width = 1.45in]{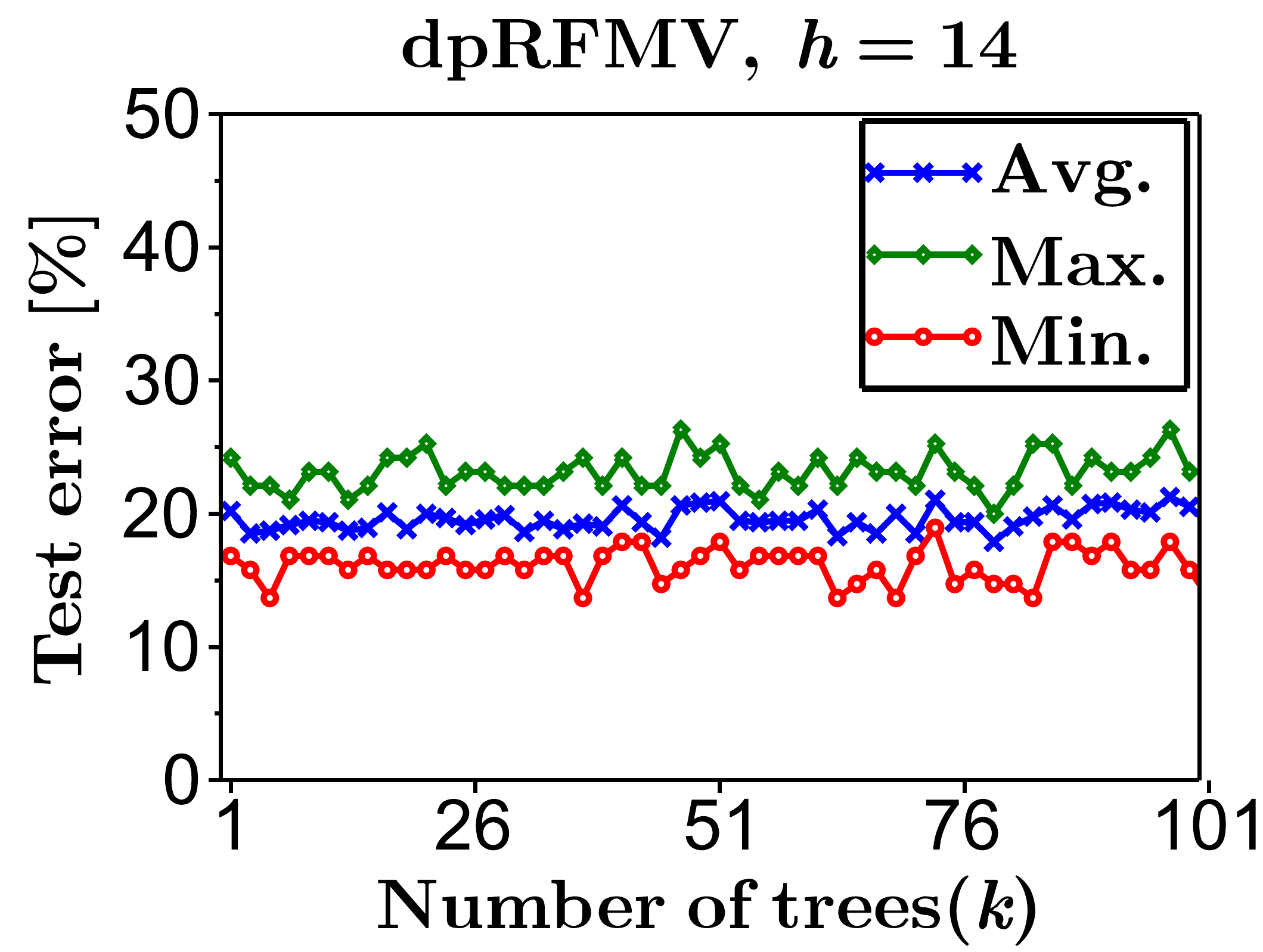}\\

\textbf{\textit{Mushroom}}\\
\hspace{-0.1in}\includegraphics[width = 1.45in]{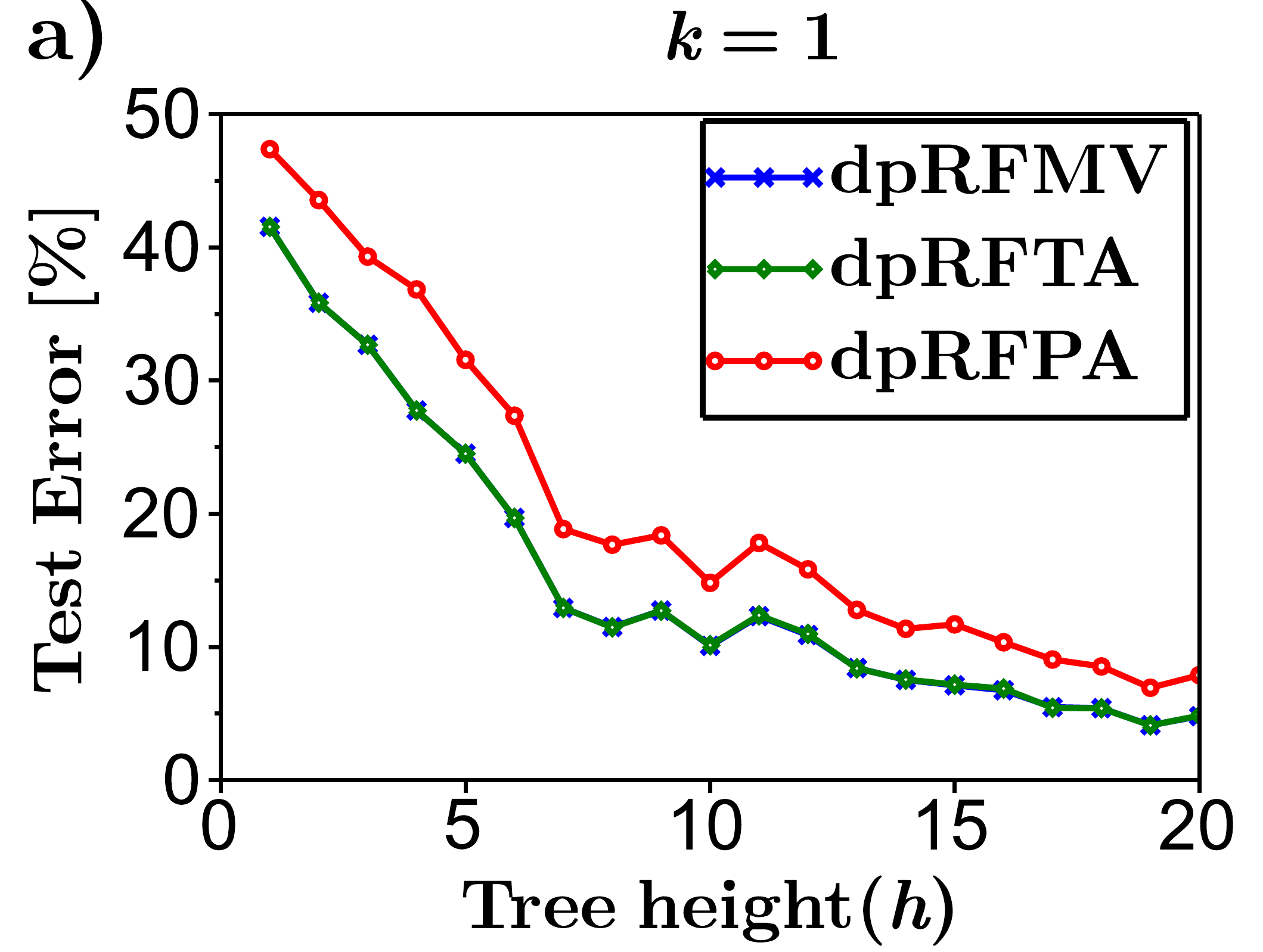} 
\hspace{-0.04in}\includegraphics[width = 1.45in]{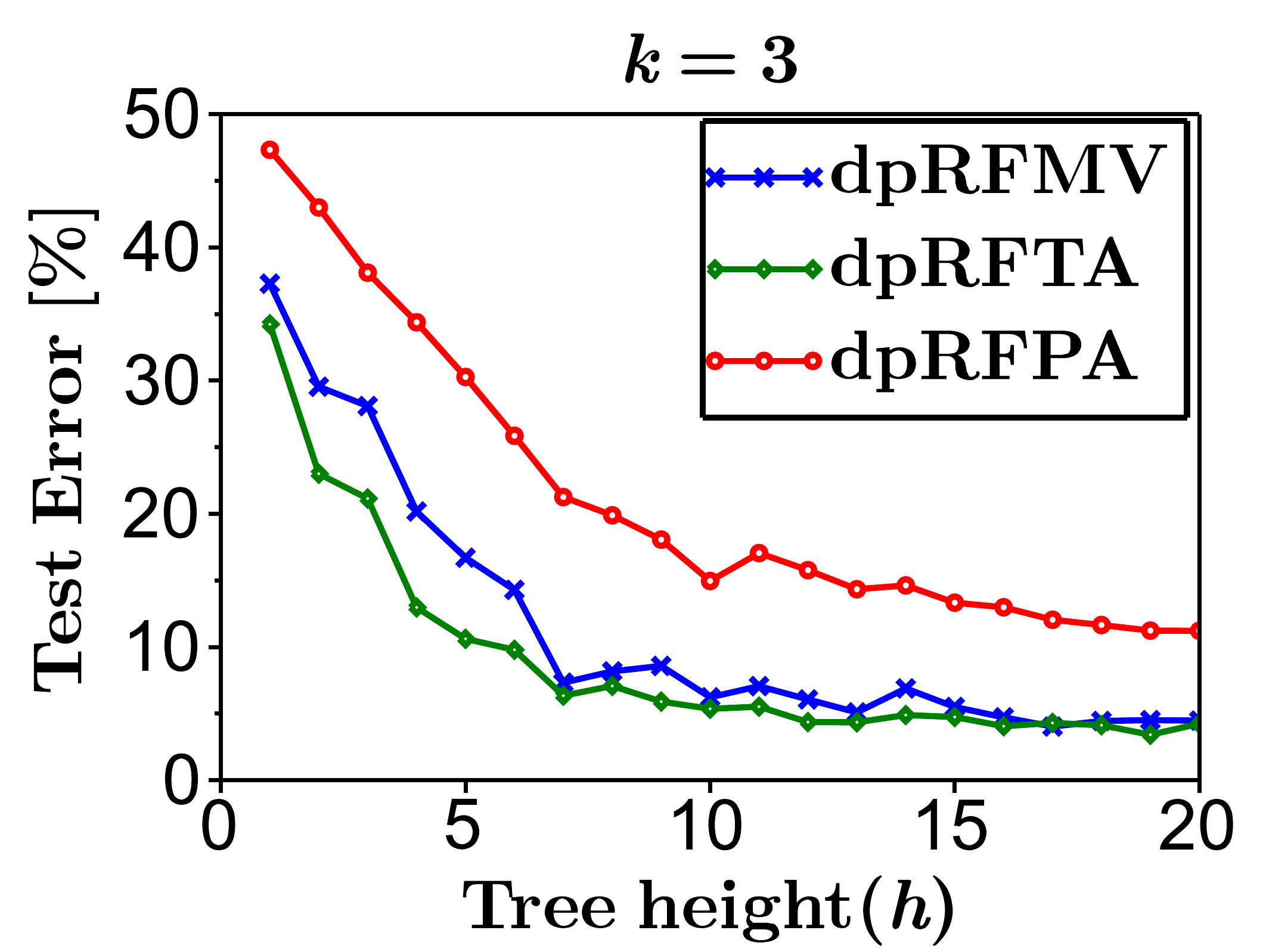} 
\hspace{-0.035in}\includegraphics[width = 1.45in]{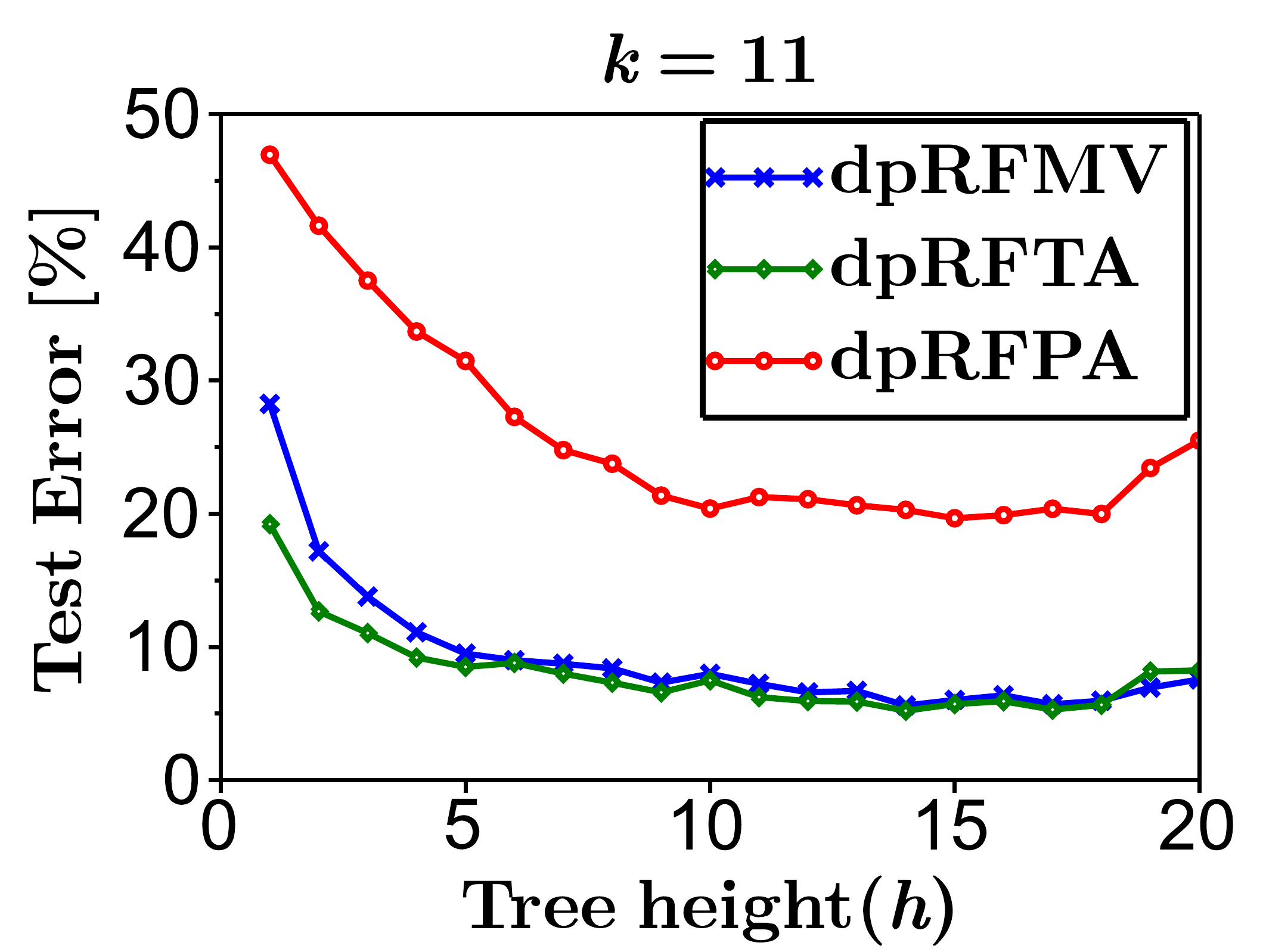} 
\hspace{-0.04in}\includegraphics[width = 1.45in]{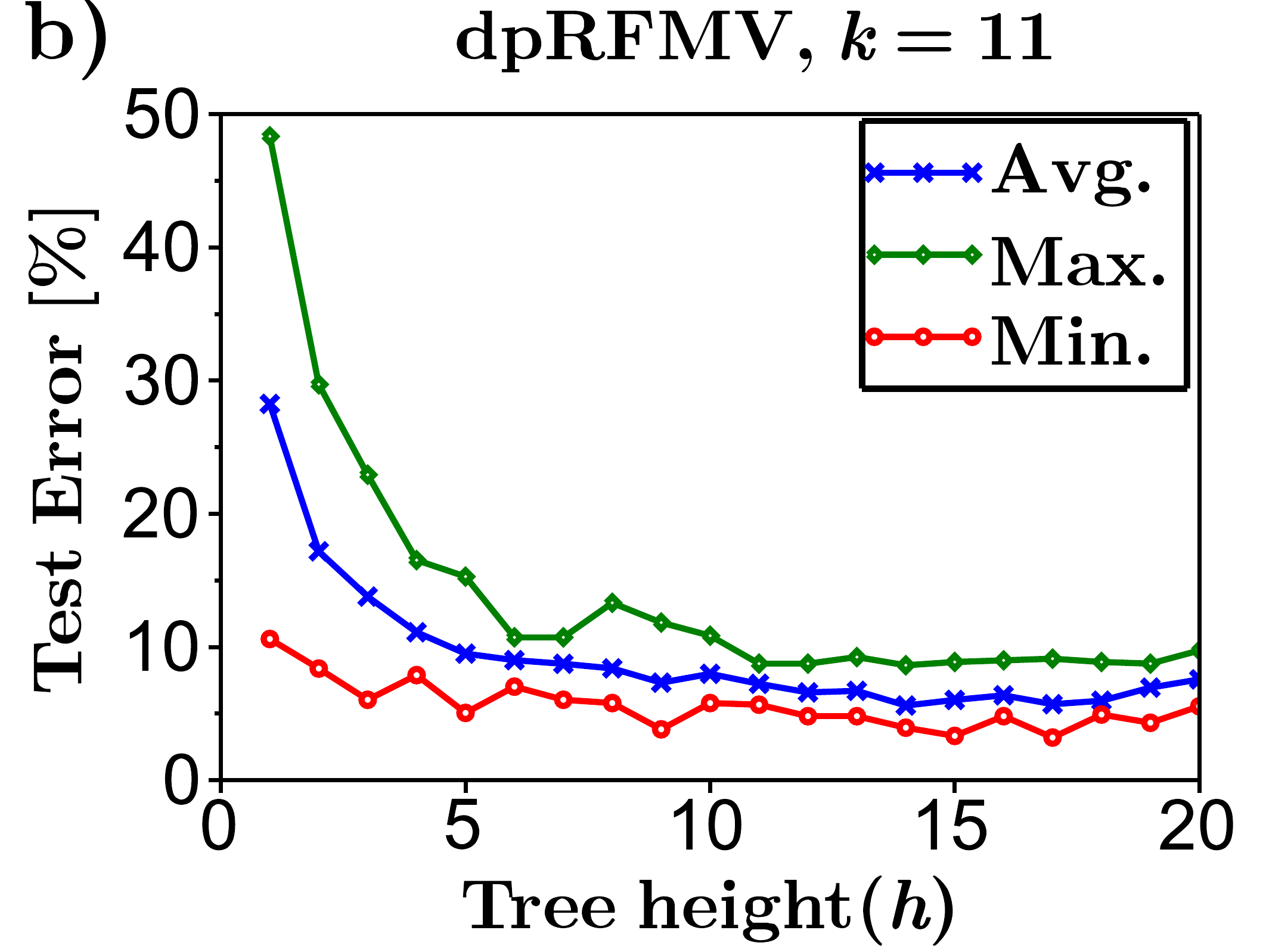}\\  
\hspace{-0.1in}\includegraphics[width = 1.45in]{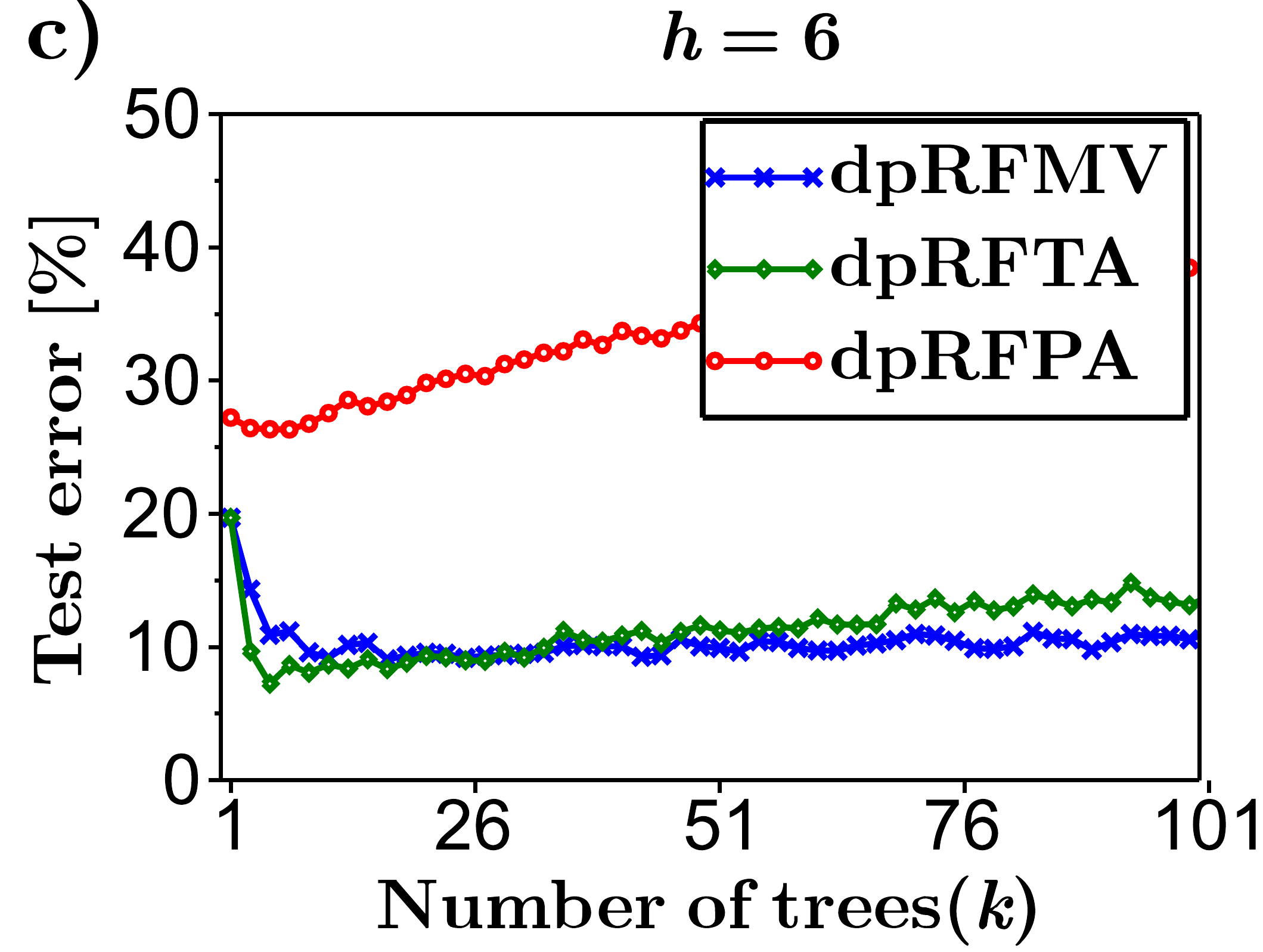} 
\hspace{-0.04in}\includegraphics[width = 1.45in]{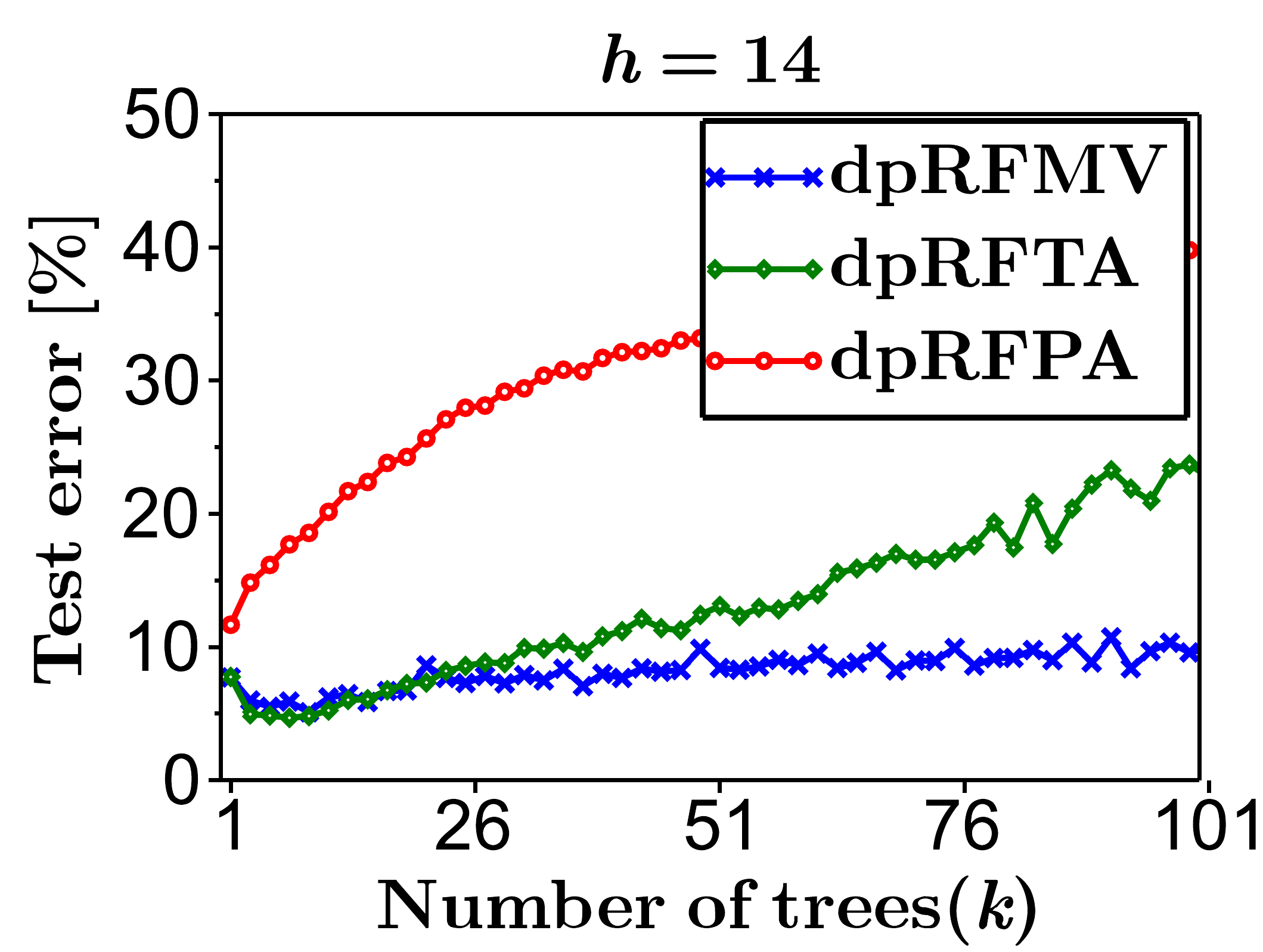} 
\hspace{-0.035in}\includegraphics[width = 1.45in]{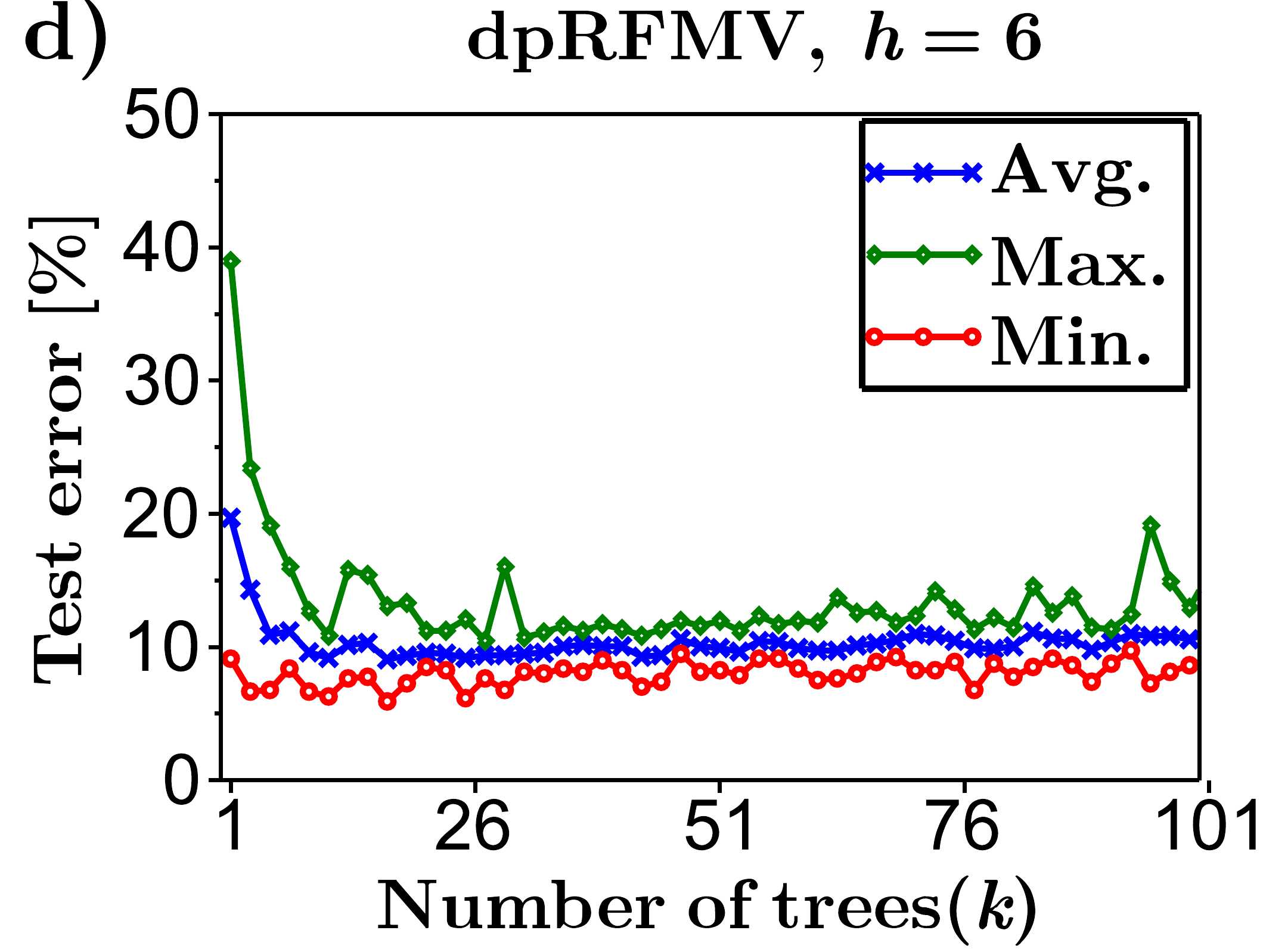} 
\hspace{-0.04in}\includegraphics[width = 1.45in]{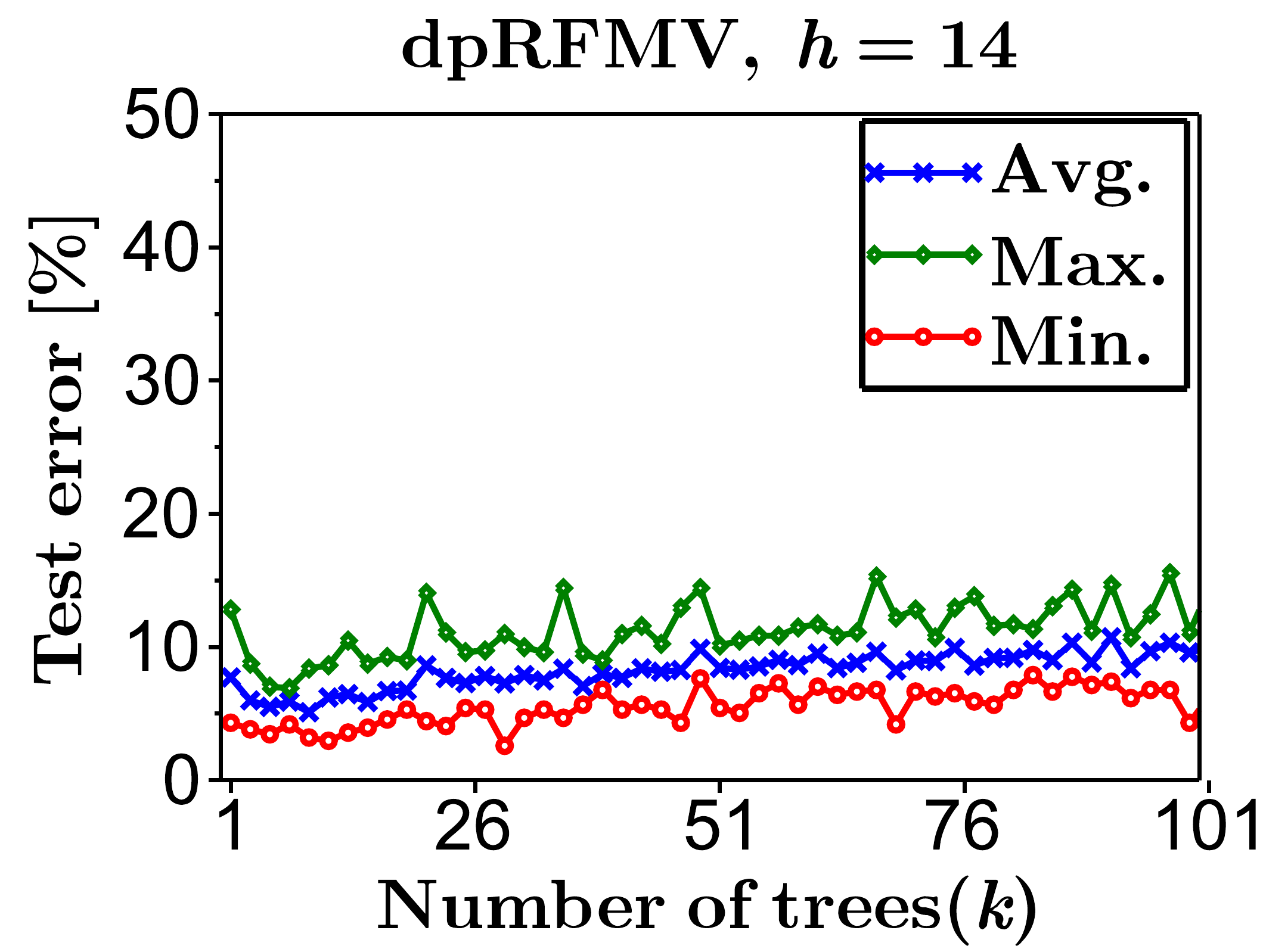}\\
\end{tabular}
\vspace{-0.1in}
\caption{Comparison of dpRFMV, dpRFTA and dpRFPA. $\eta = 1000/n_{tr} = 0.137$ for selected datasets. Test error resp. vs. \textbf{a)} $h$ across various settings of $k$ and vs. \textbf{c)} $k$ across various settings of $h$; Minimal, average and maximal test error resp. vs. $h$ (\textbf{b)}) and vs. $k$ (\textbf{d)}) for dpRFMV.}
\label{fig:one}
\end{figure*}
\begin{figure*}[htp!]
  \center
\textbf{\textit{Banknote Authentication}}\\
\includegraphics[width = 2.8in,height = 1.15in]{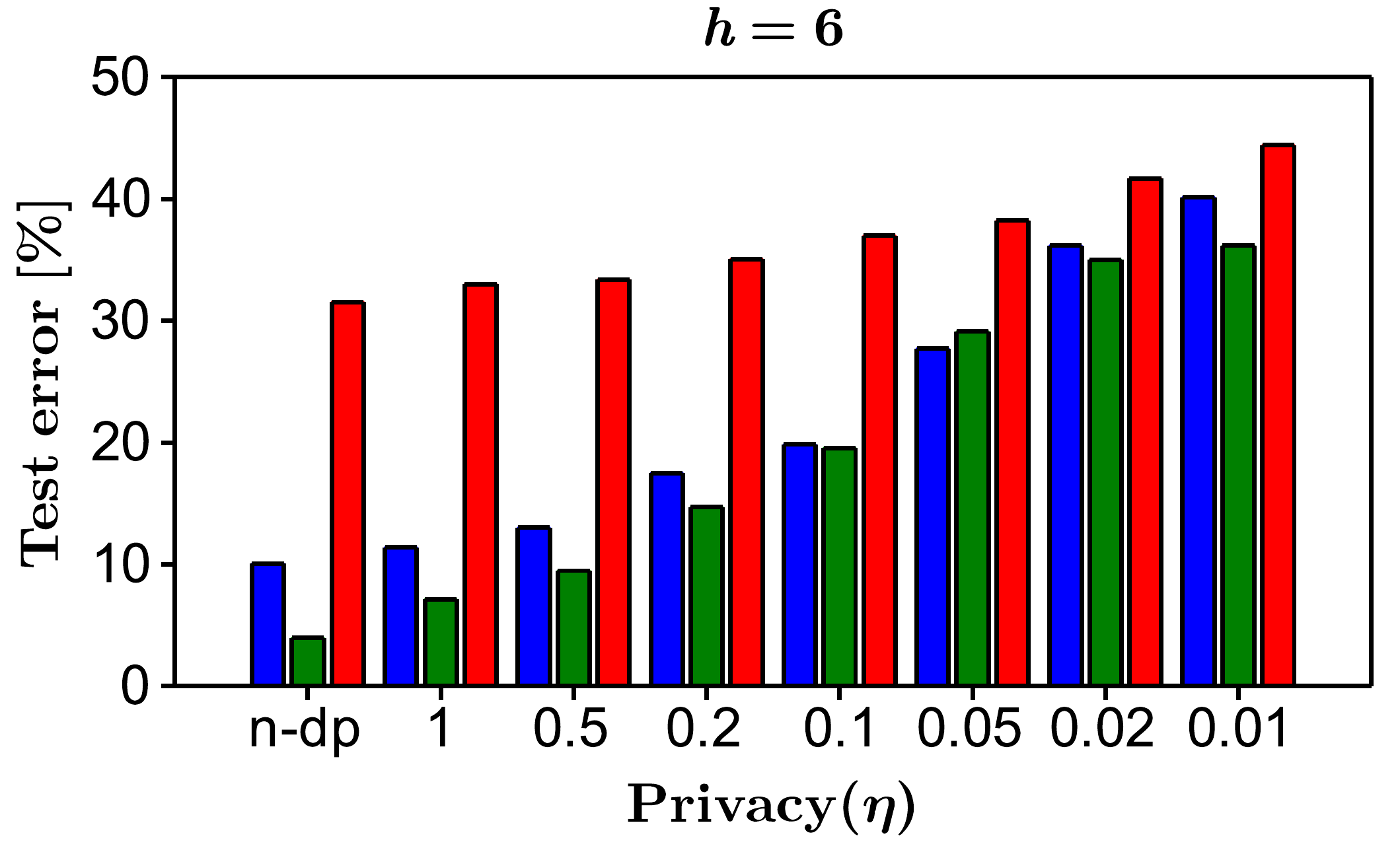} 
\includegraphics[width = 2.8in,height = 1.15in]{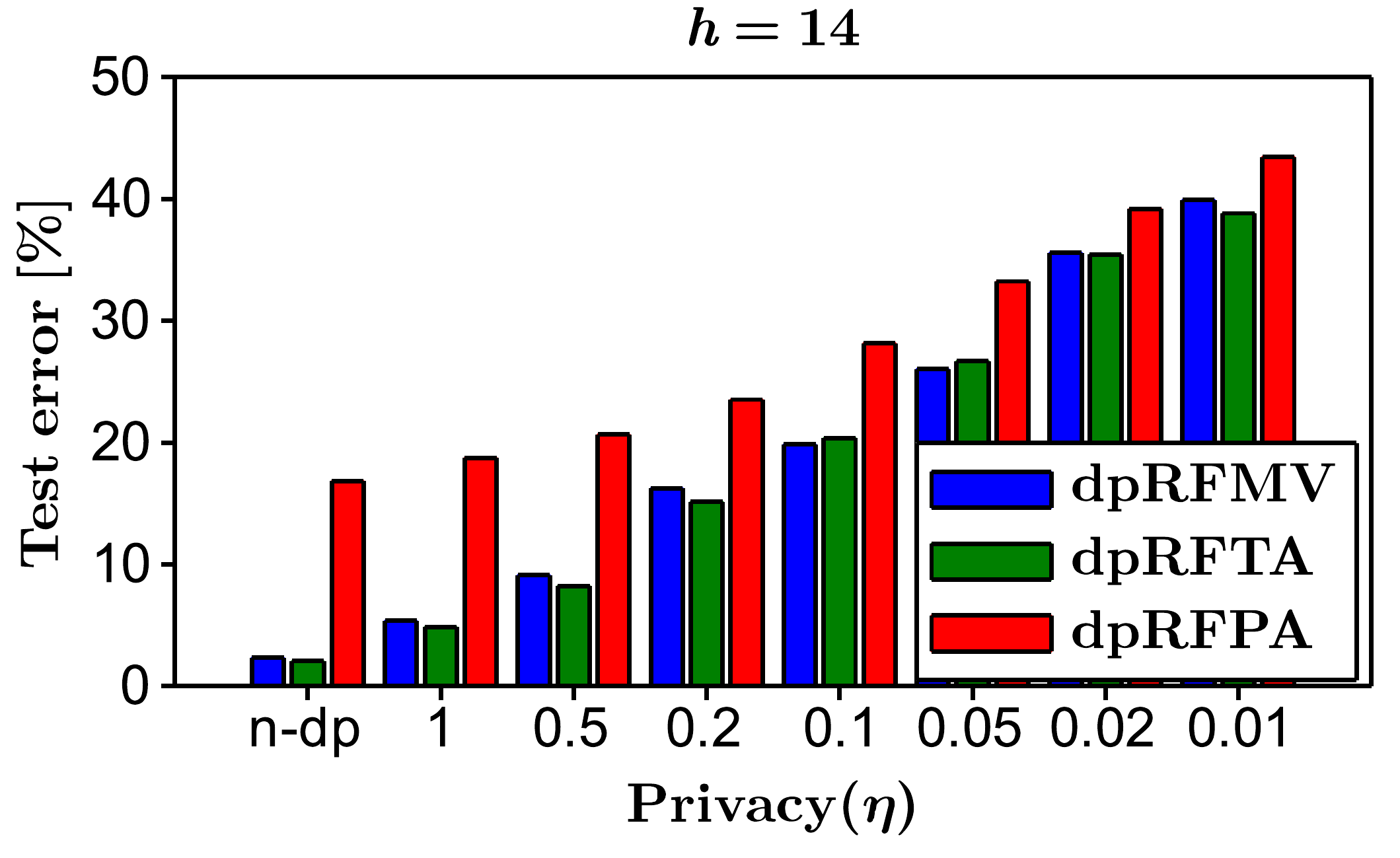}\\
\includegraphics[width = 1.49in]{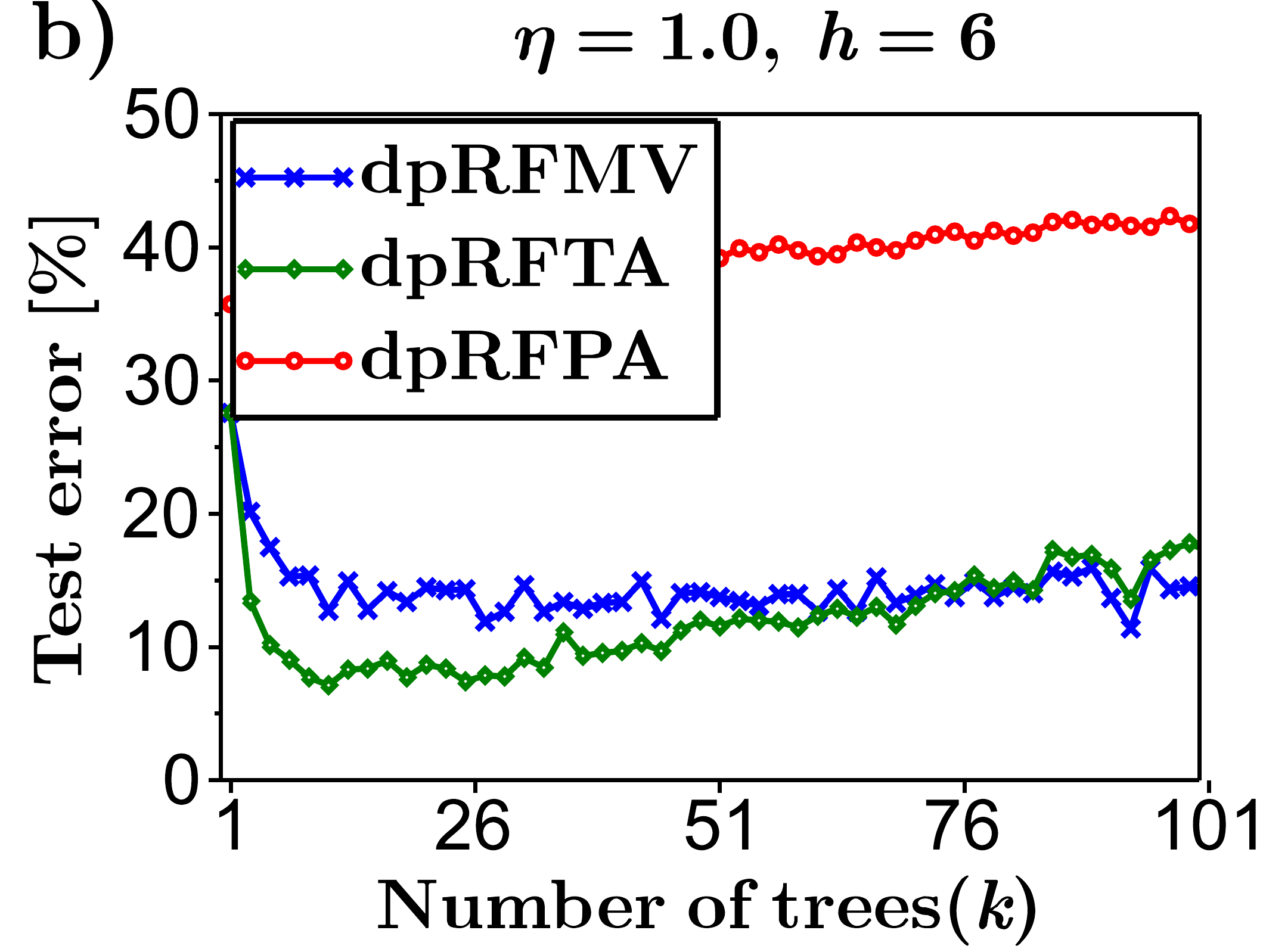} 
\hspace{-0.03in}\includegraphics[width = 1.49in]{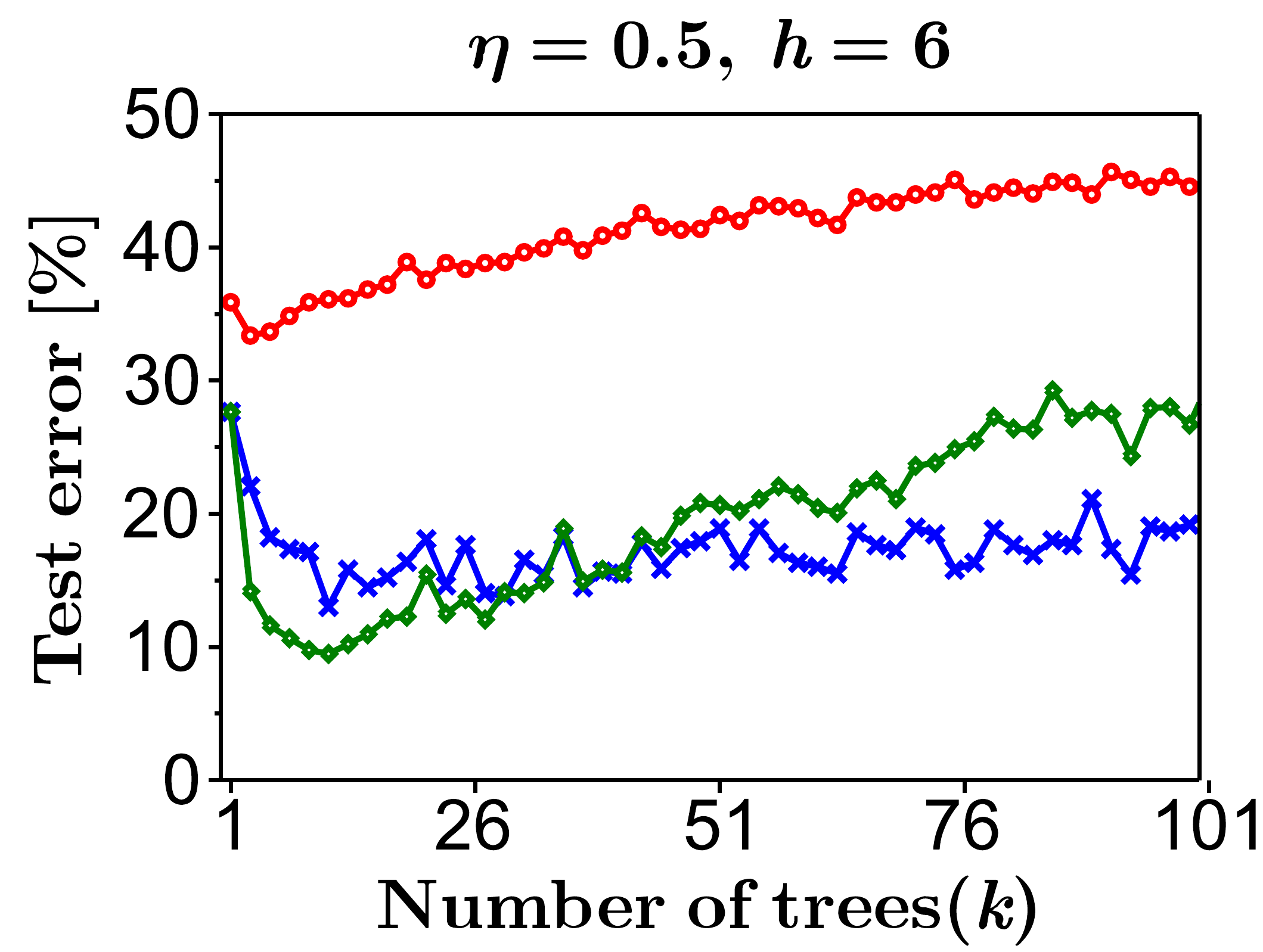} 
\hspace{-0.03in}\includegraphics[width = 1.49in]{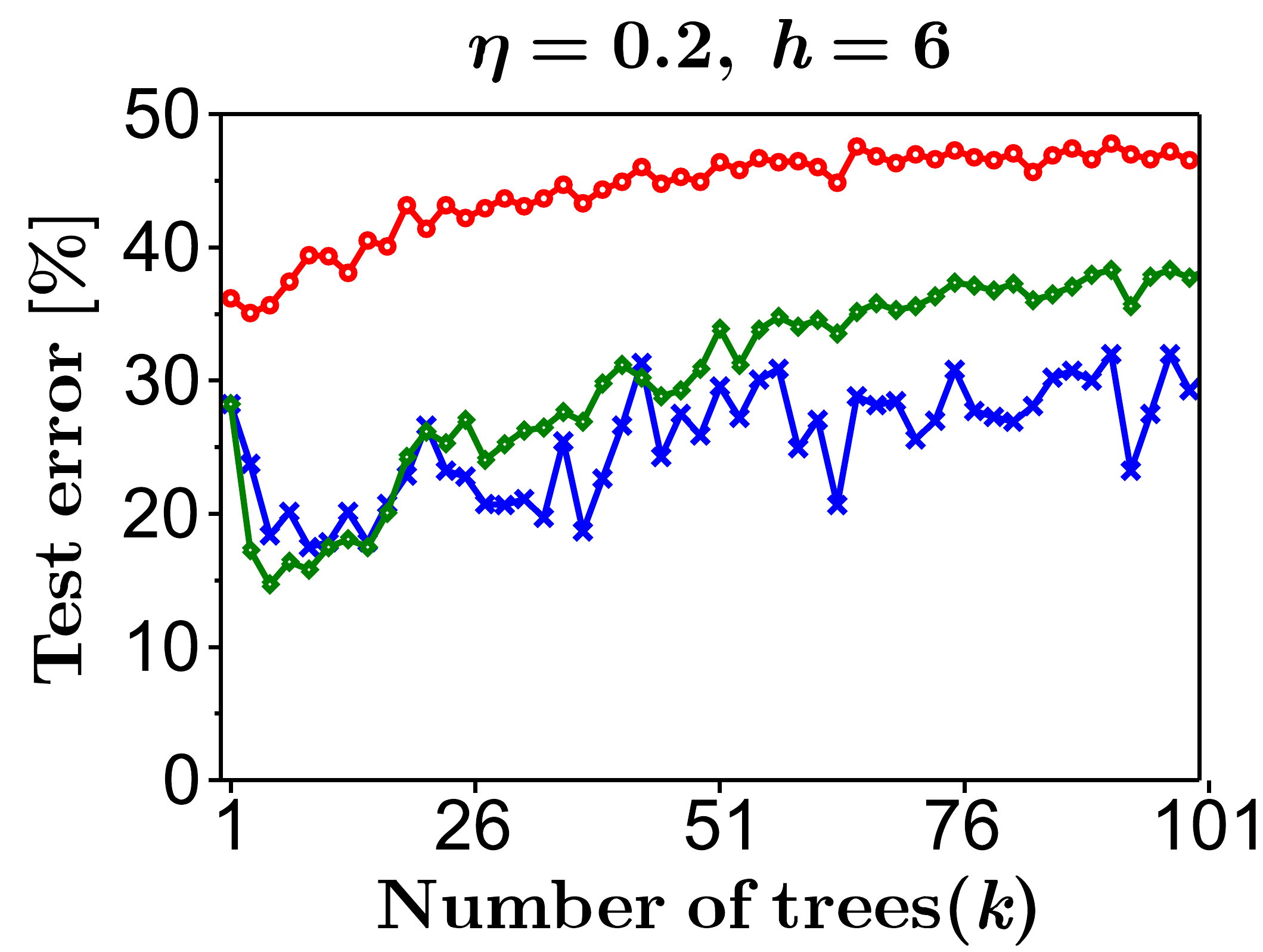} 
\hspace{-0.03in}\includegraphics[width = 1.49in]{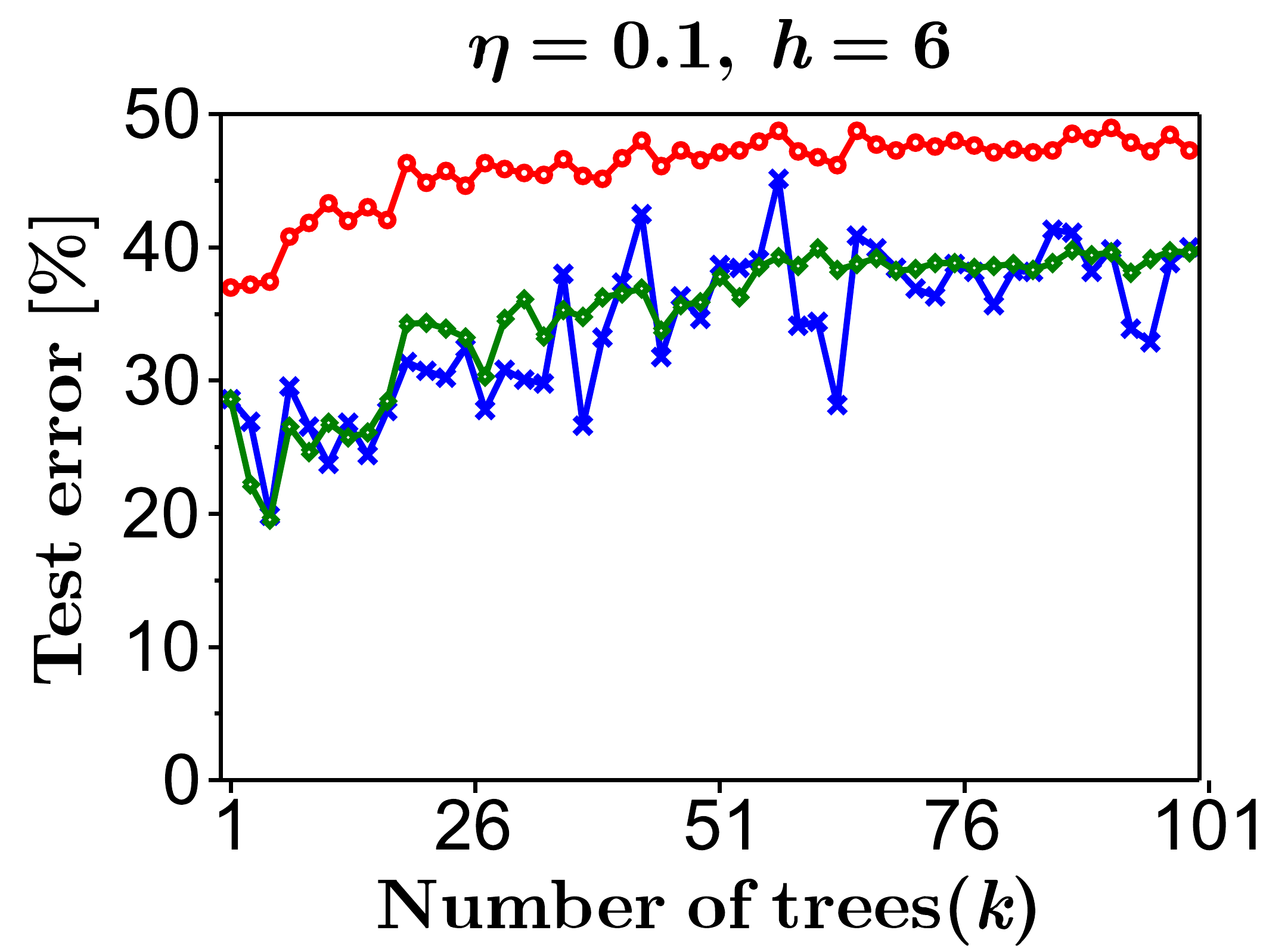}\\

\textbf{\textit{Congressional Voting Records}}\\
\includegraphics[width = 2.8in,height = 1.15in]{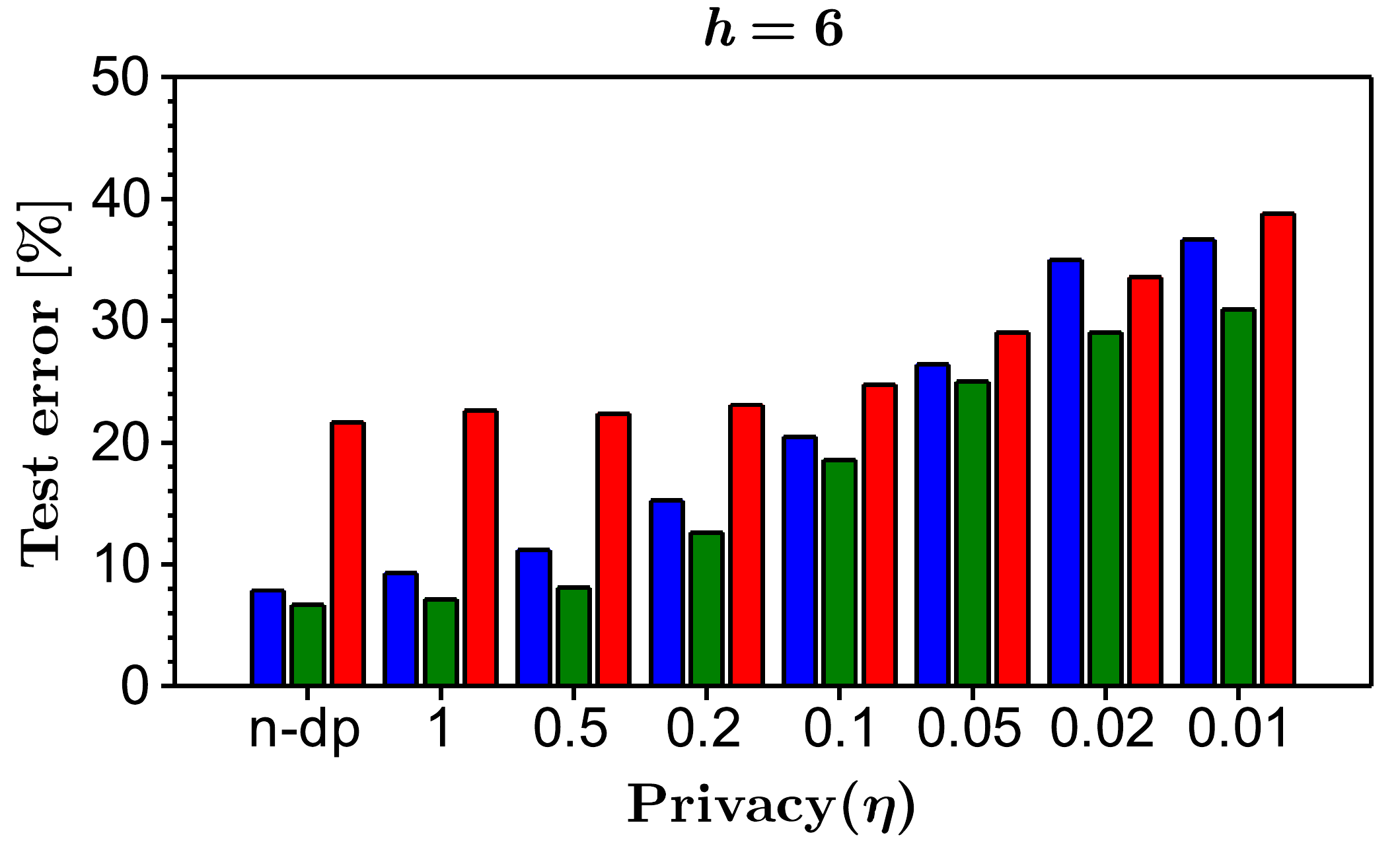} 
\includegraphics[width = 2.8in,height = 1.15in]{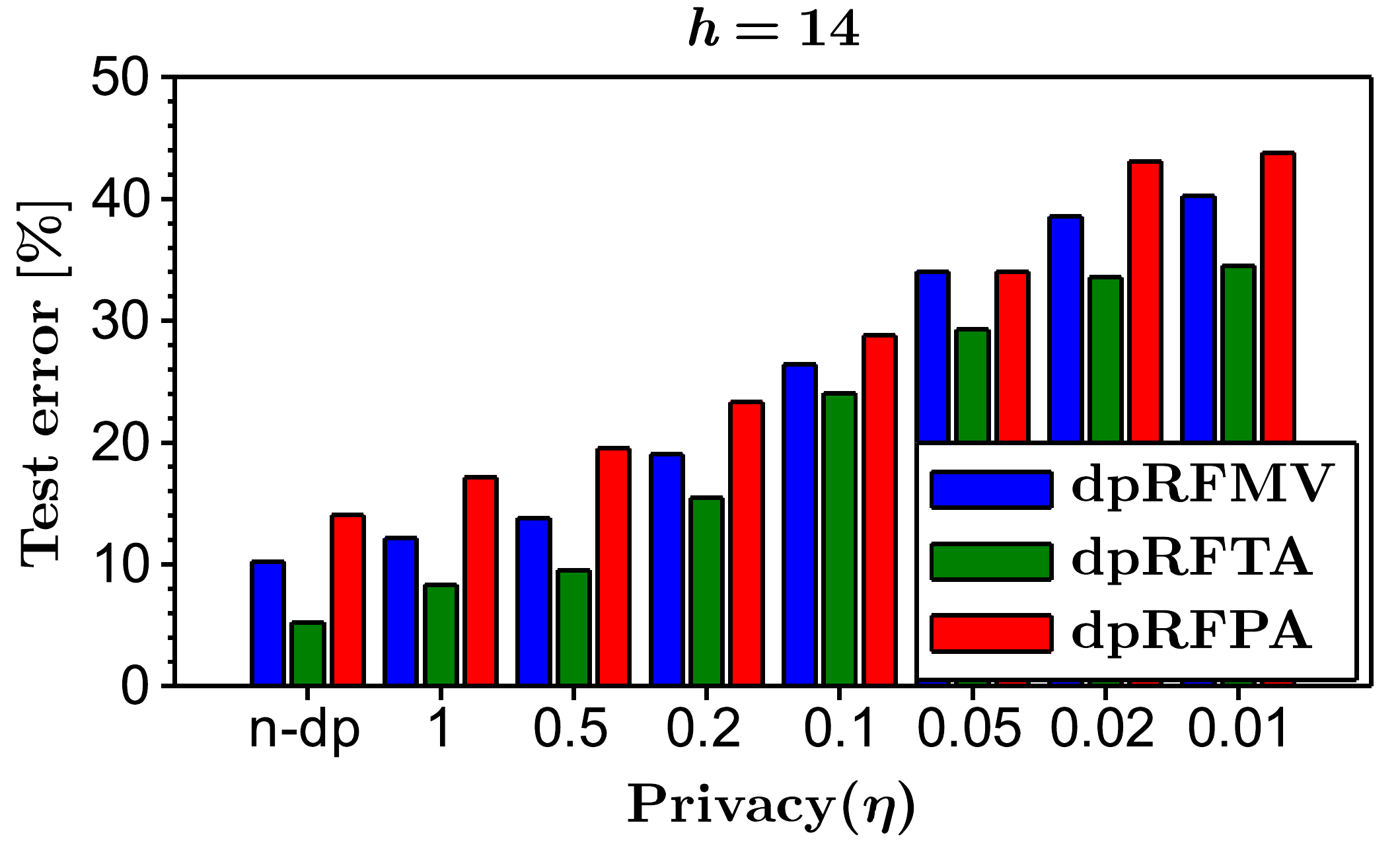}\\
\includegraphics[width = 1.49in]{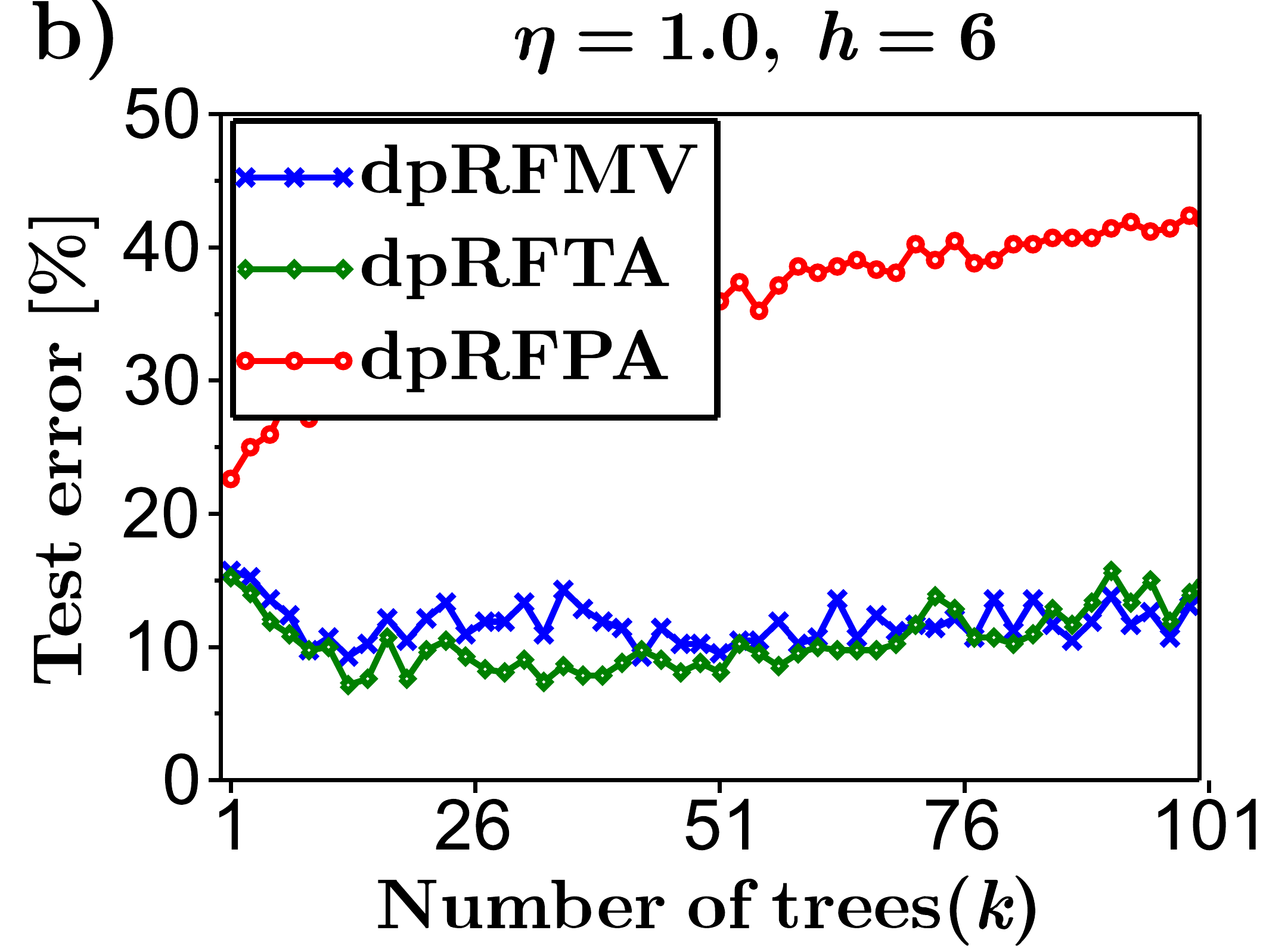} 
\hspace{-0.03in}\includegraphics[width = 1.49in]{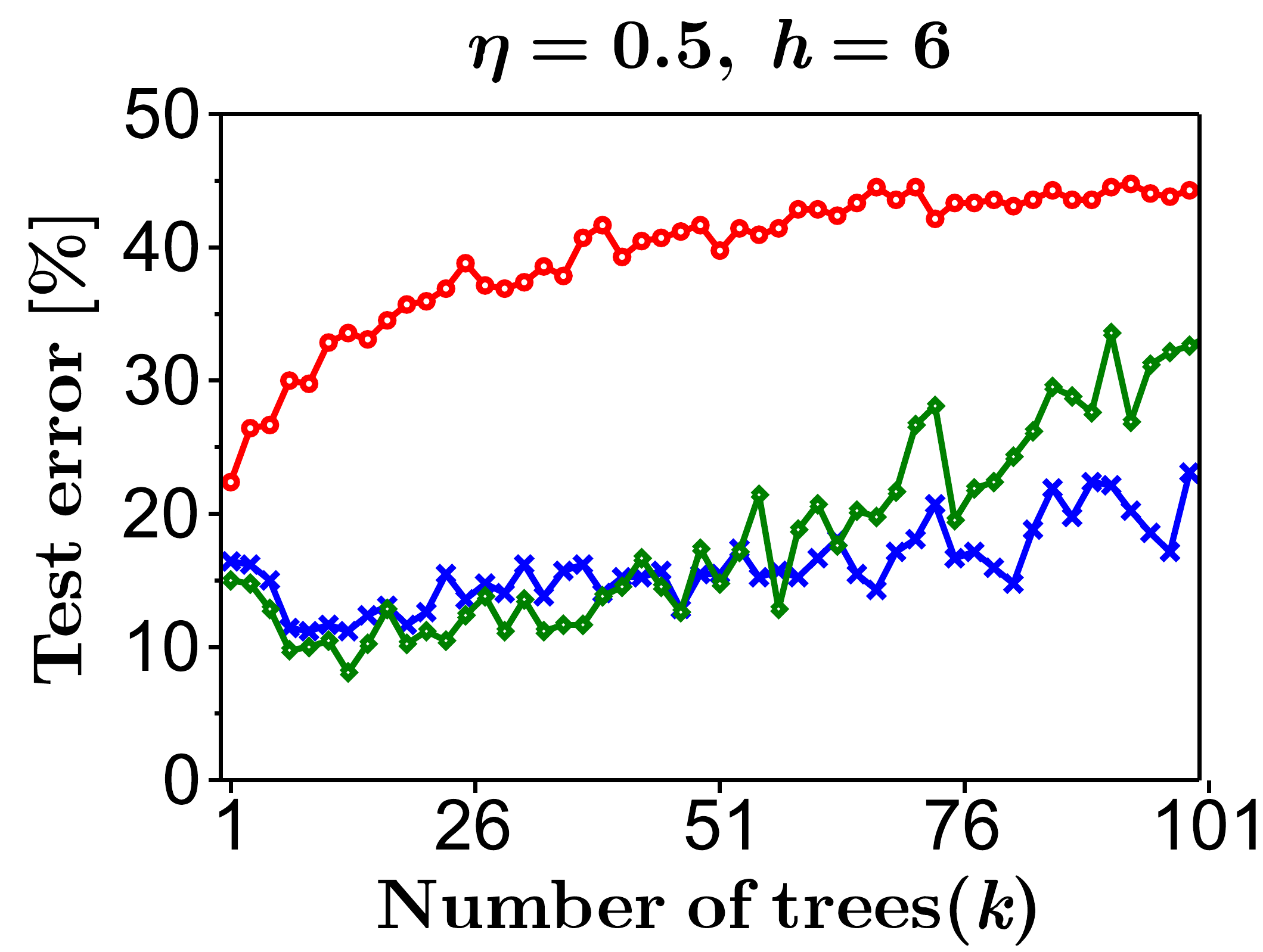} 
\hspace{-0.03in}\includegraphics[width = 1.49in]{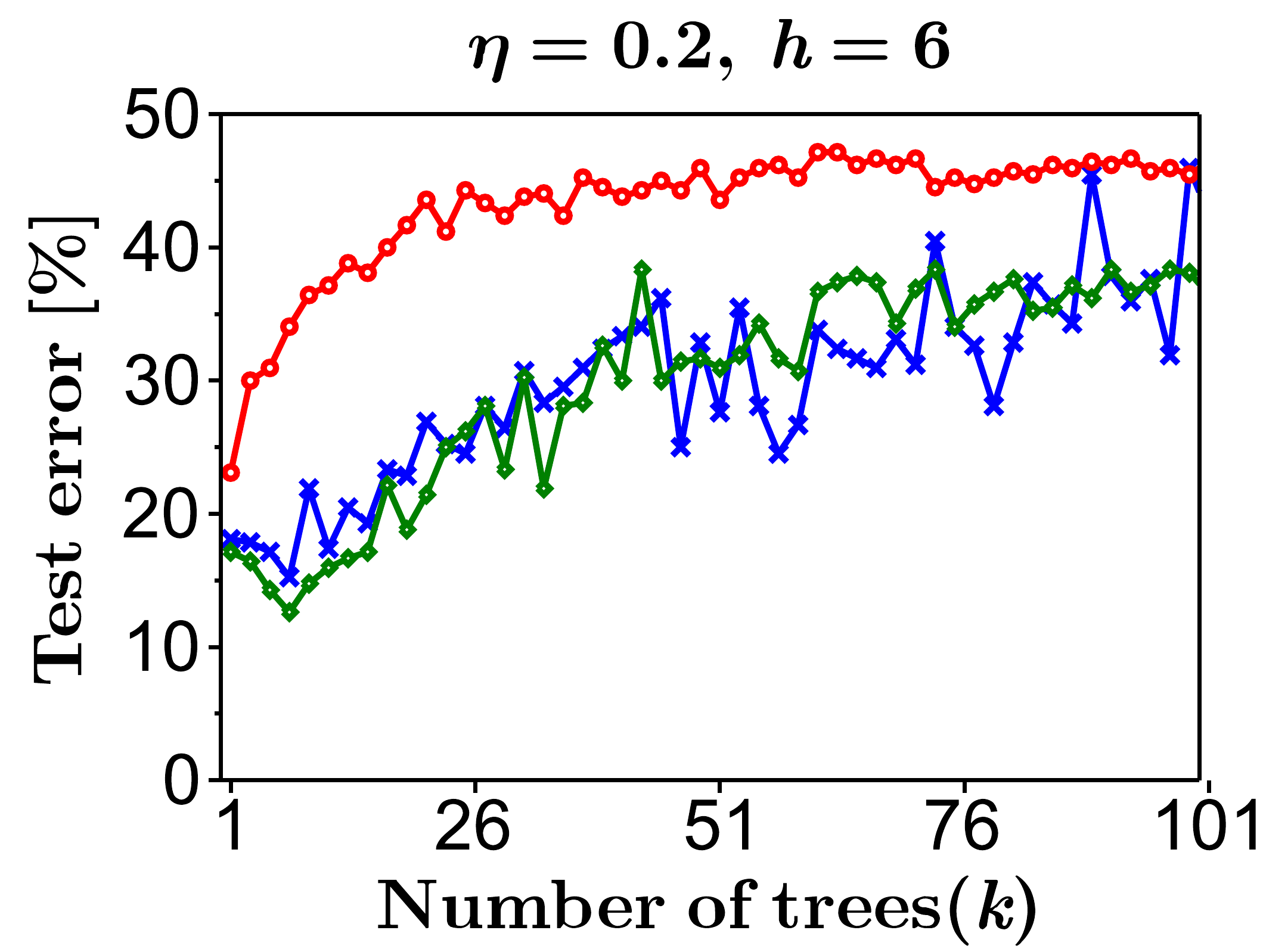} 
\hspace{-0.03in}\includegraphics[width = 1.49in]{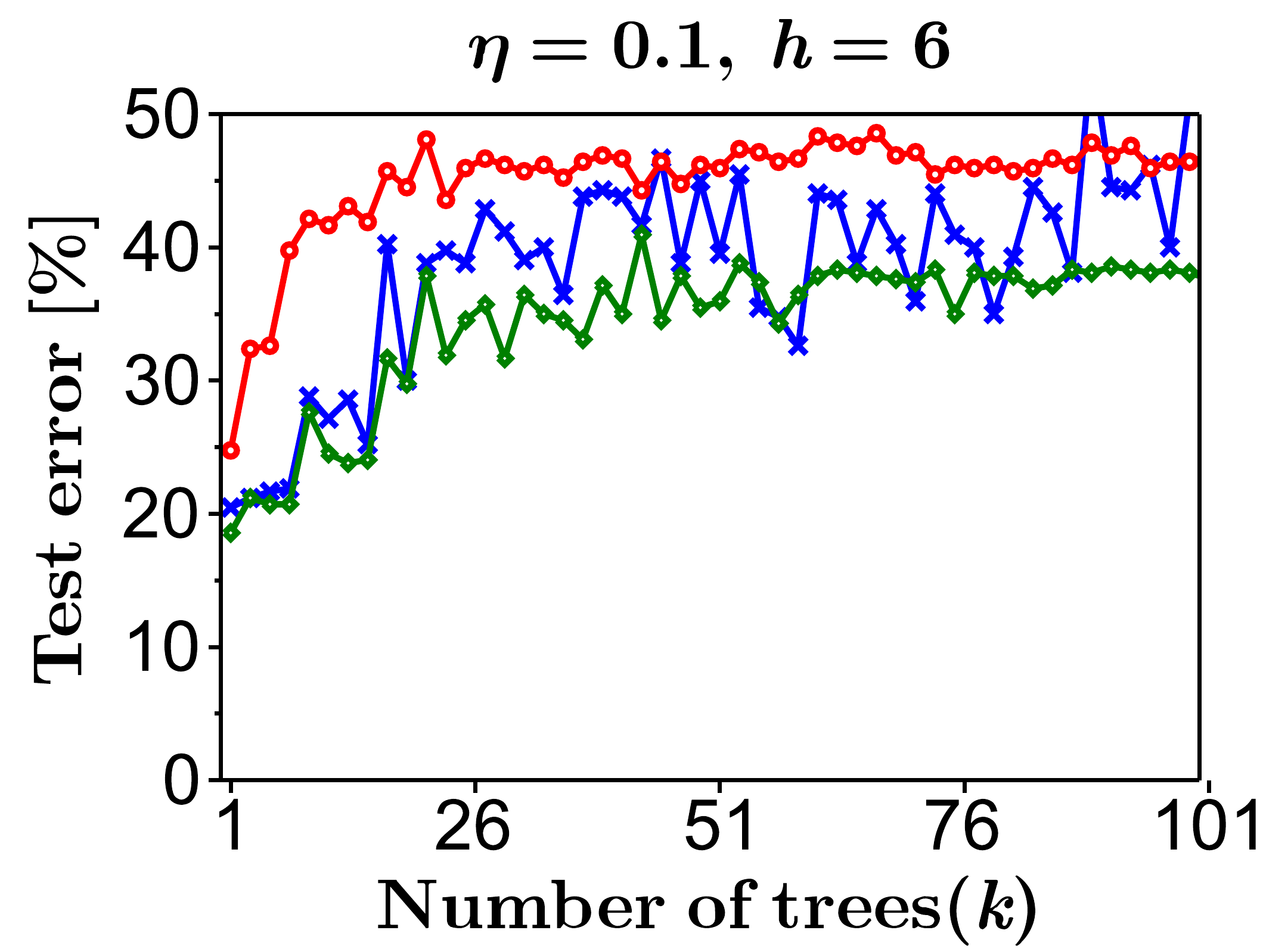}\\

\textbf{\textit{Mammographic Mass}}\\
\includegraphics[width = 2.8in,height = 1.15in]{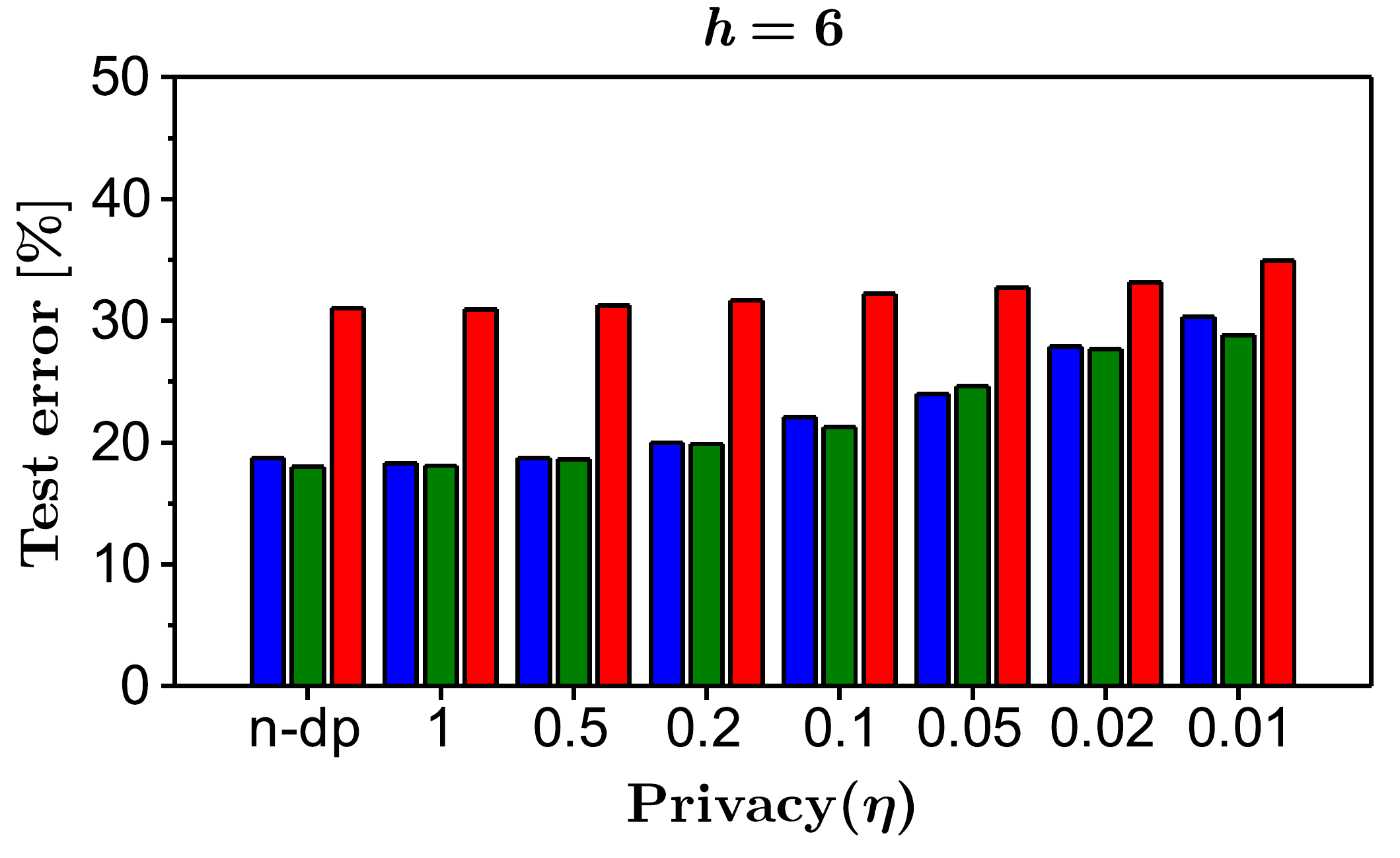} 
\includegraphics[width = 2.8in,height = 1.15in]{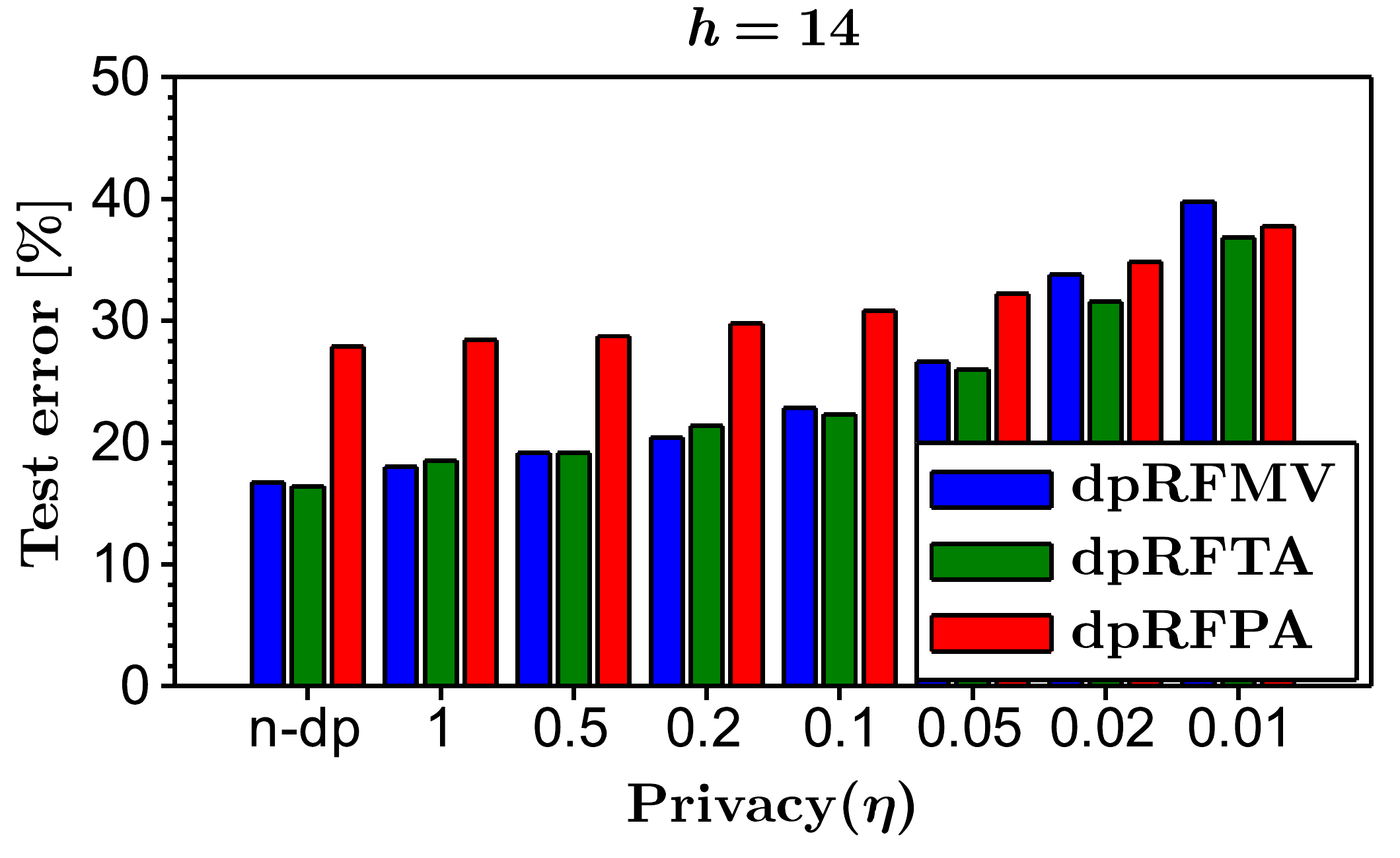}\\
\includegraphics[width = 1.49in]{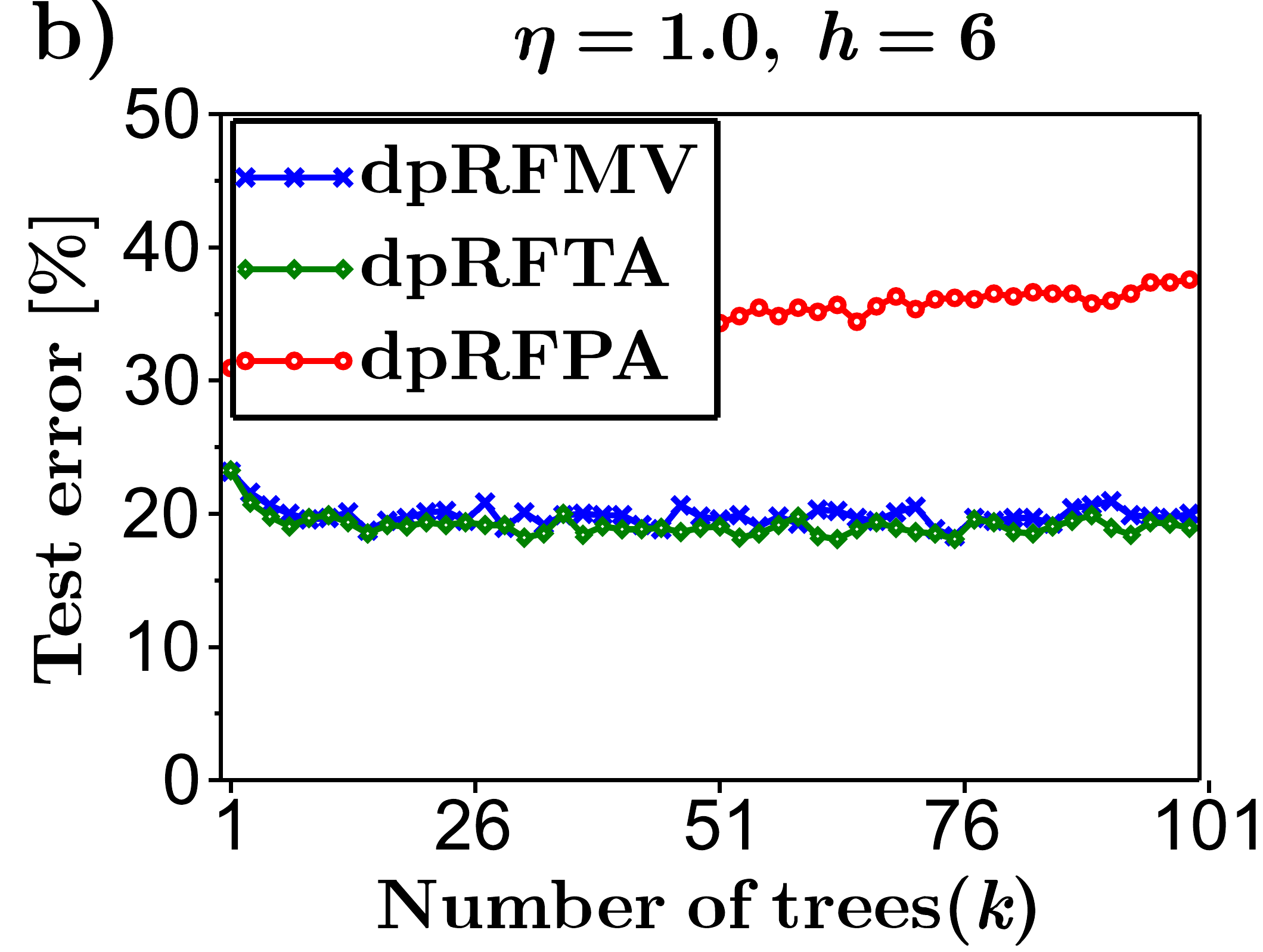} 
\hspace{-0.03in}\includegraphics[width = 1.49in]{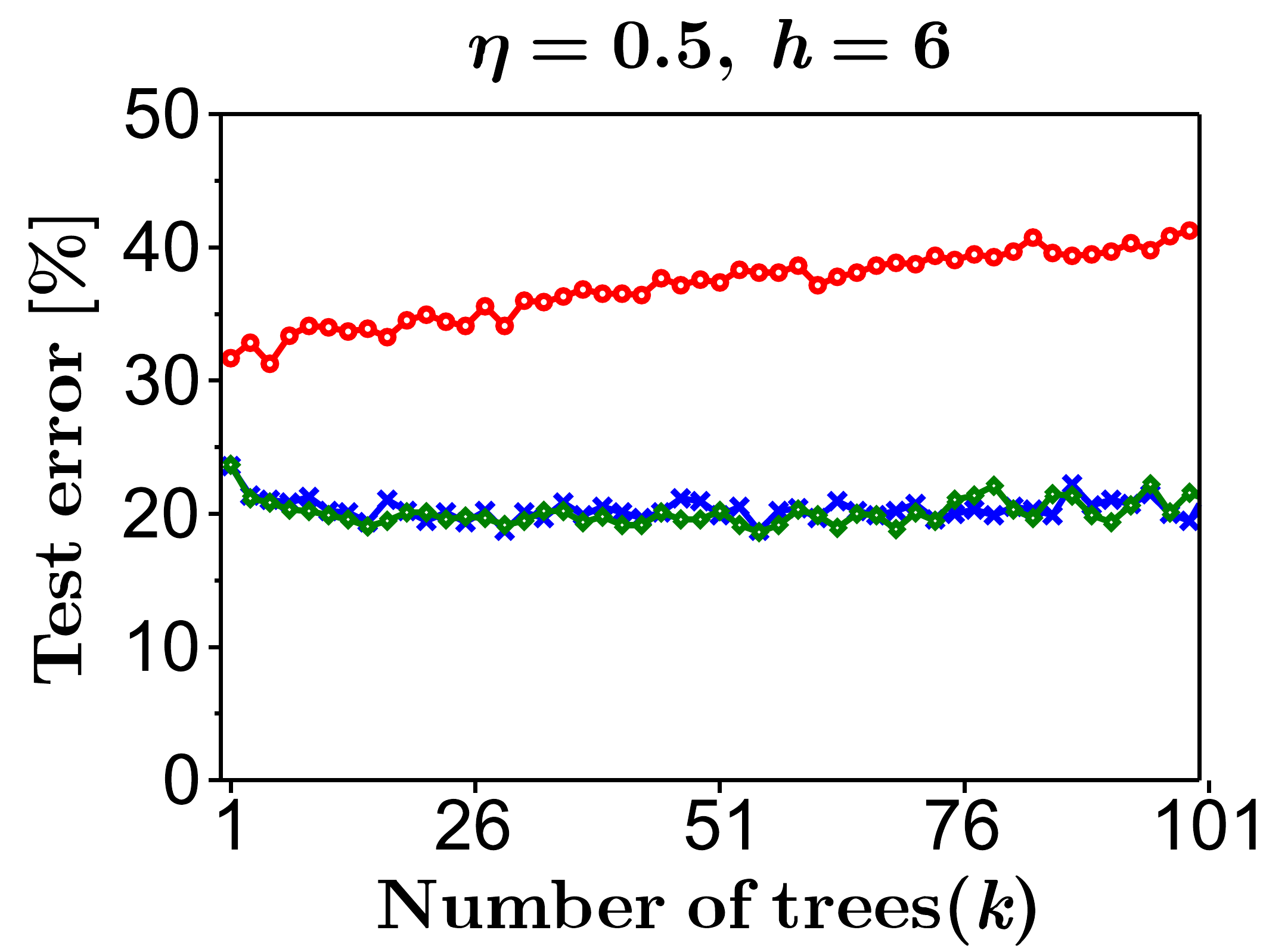} 
\hspace{-0.03in}\includegraphics[width = 1.49in]{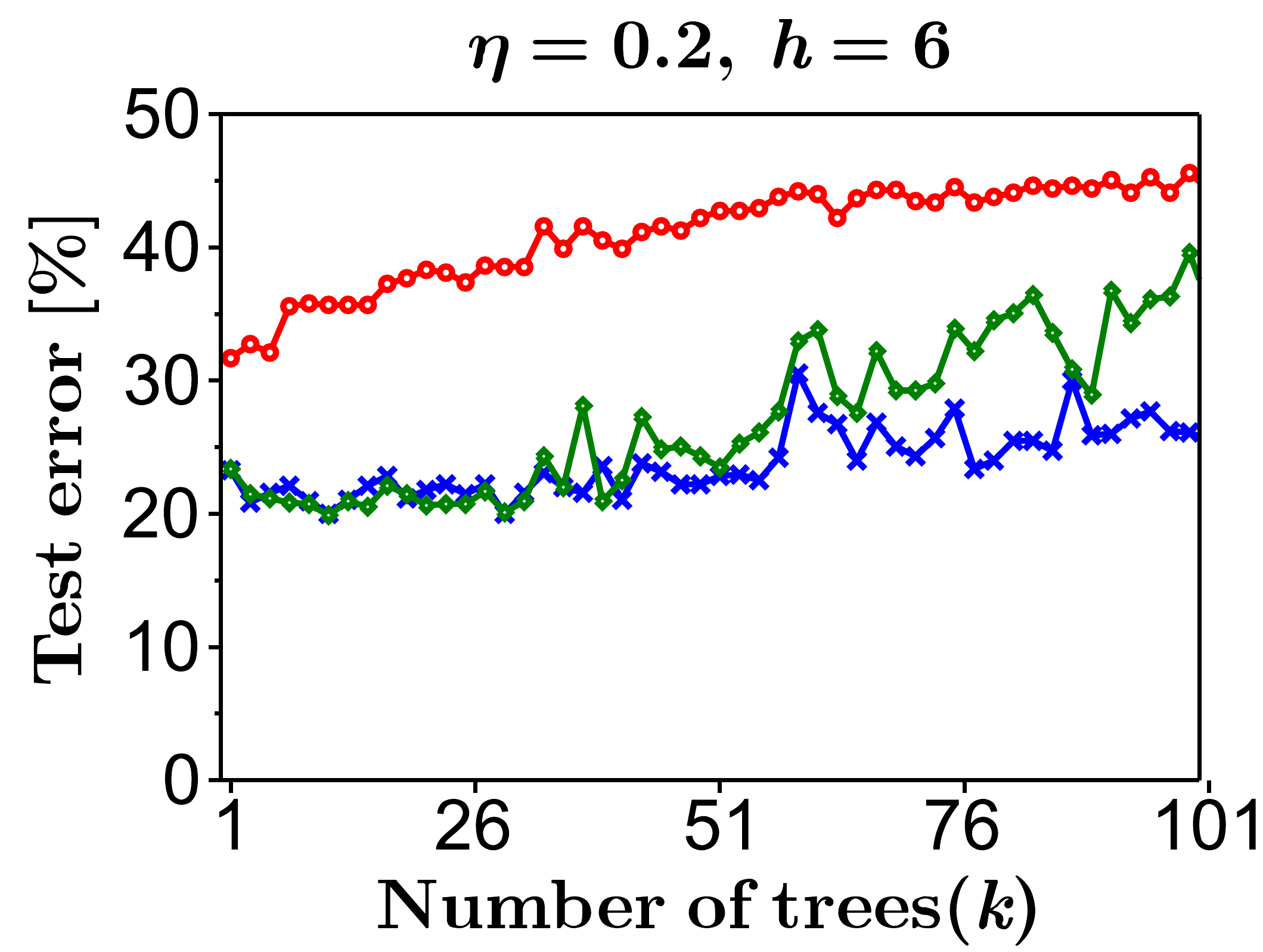} 
\hspace{-0.03in}\includegraphics[width = 1.49in]{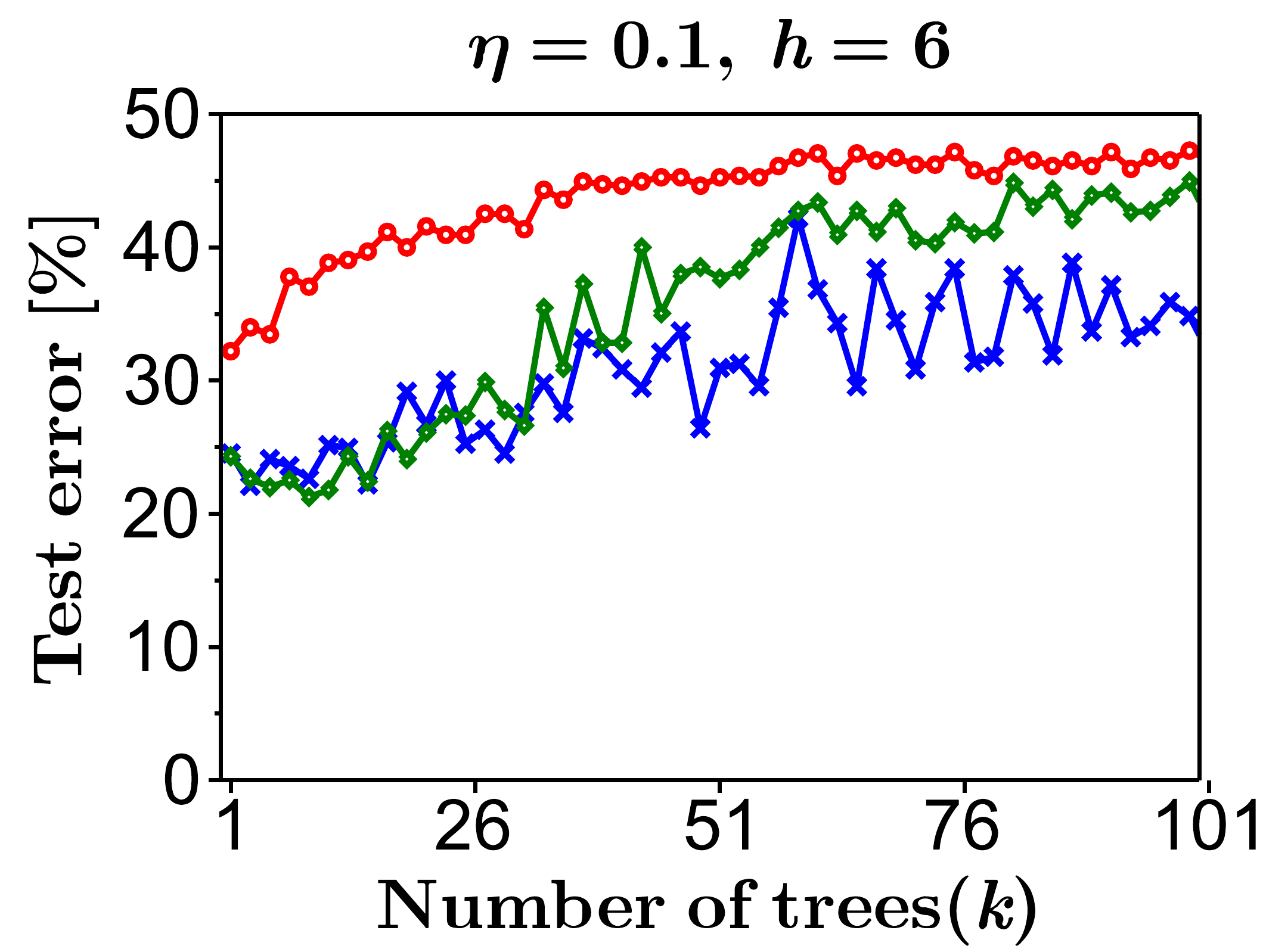}\\

\textbf{\textit{Mushroom}}\\
\includegraphics[width = 2.8in,height = 1.15in]{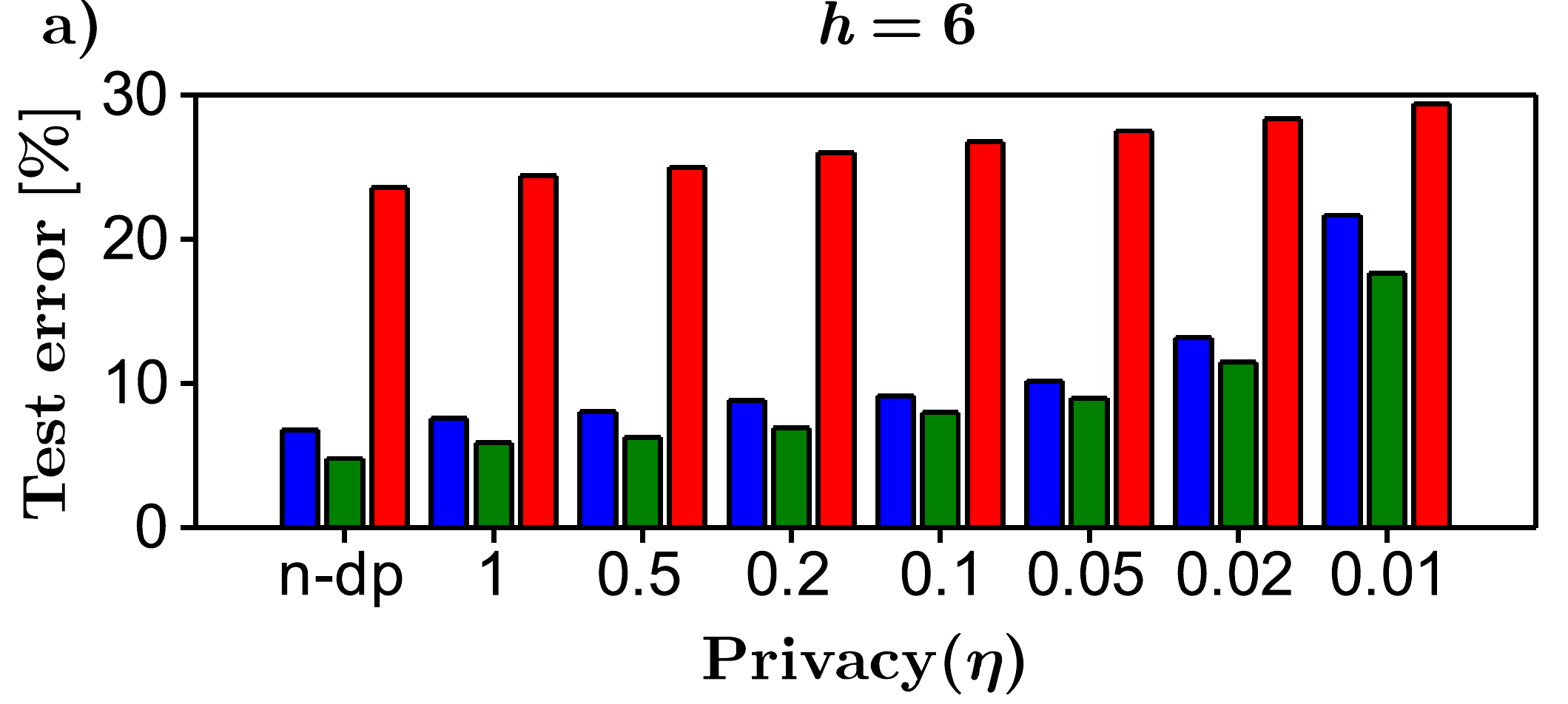} 
\includegraphics[width = 2.8in,height = 1.15in]{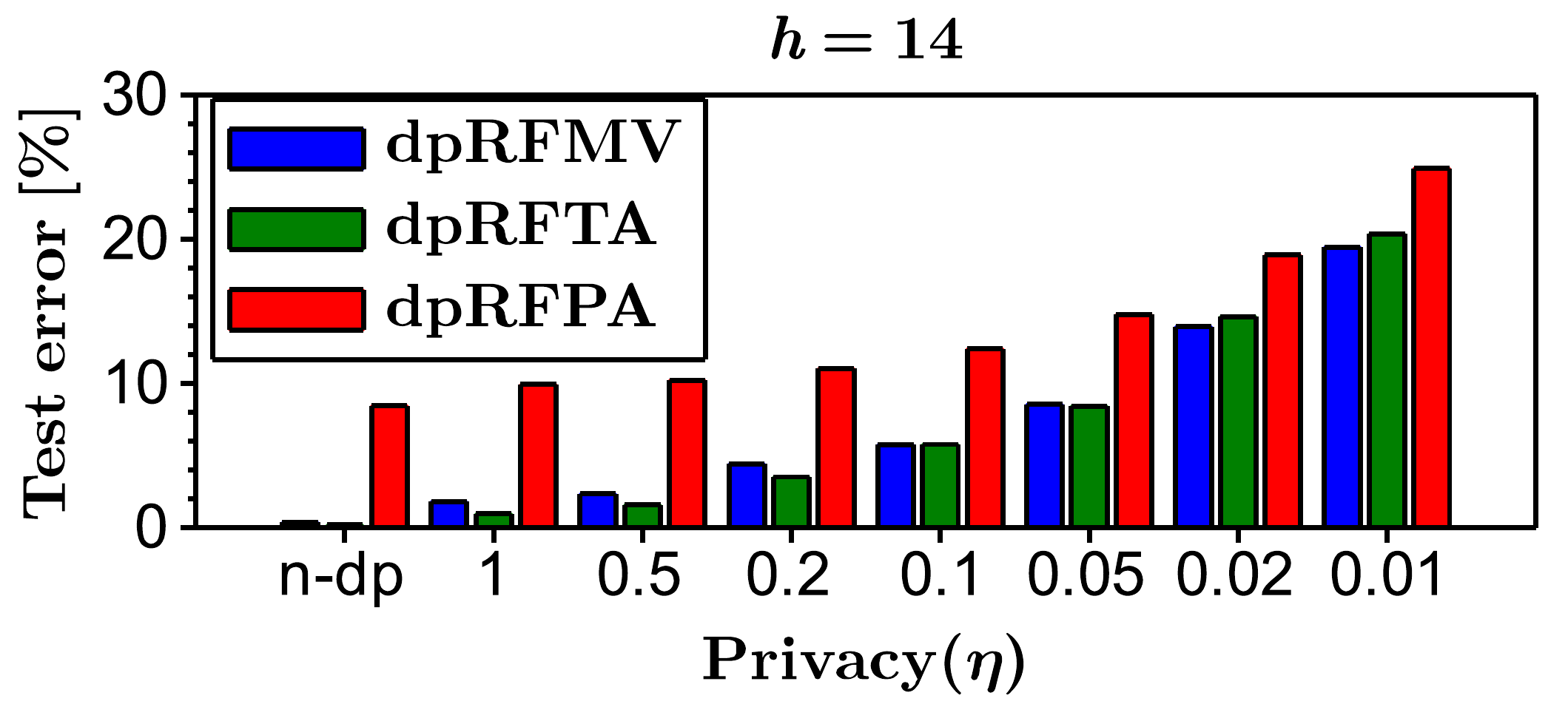}\\
\includegraphics[width = 1.49in]{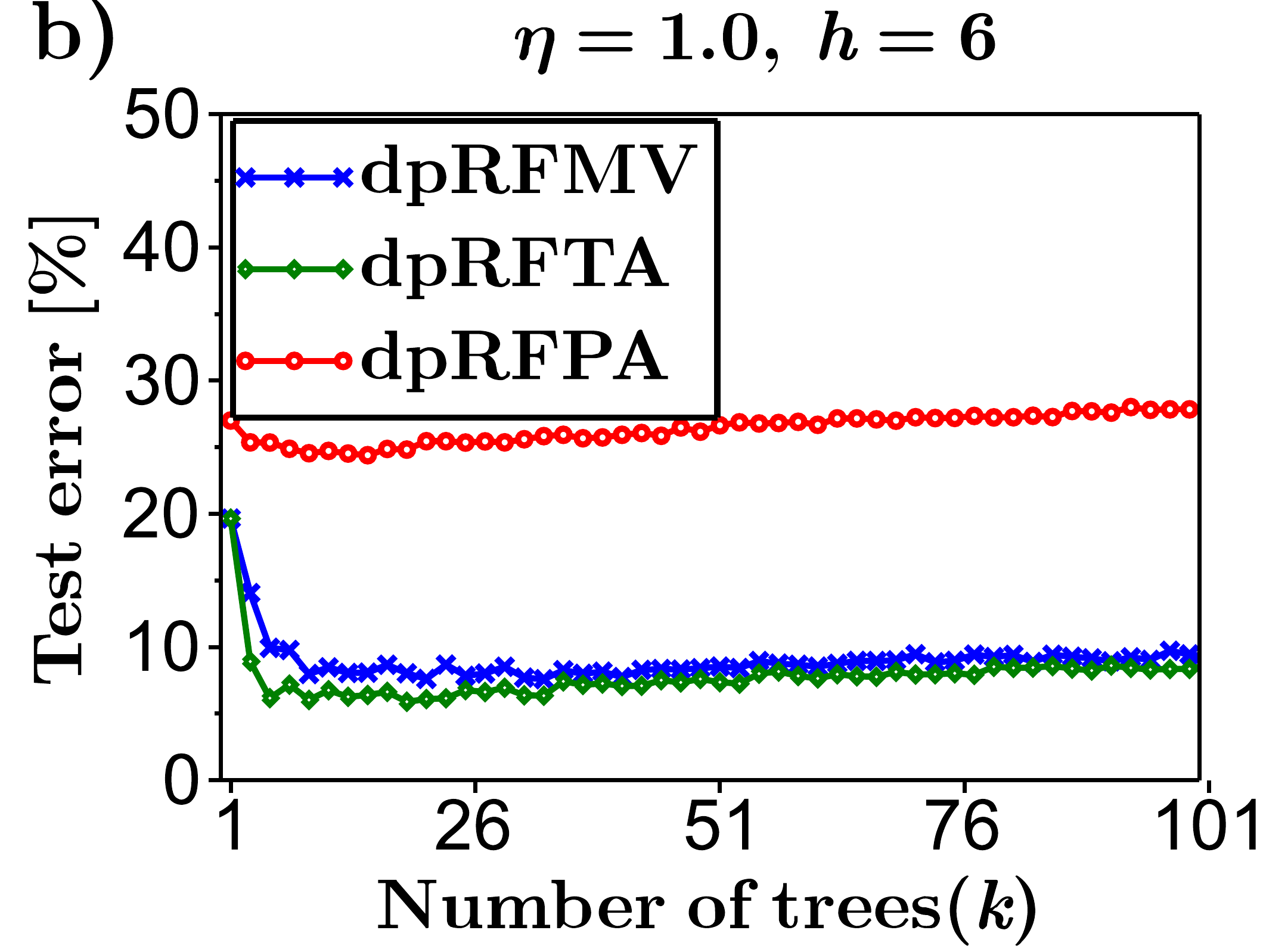} 
\hspace{-0.03in}\includegraphics[width = 1.49in]{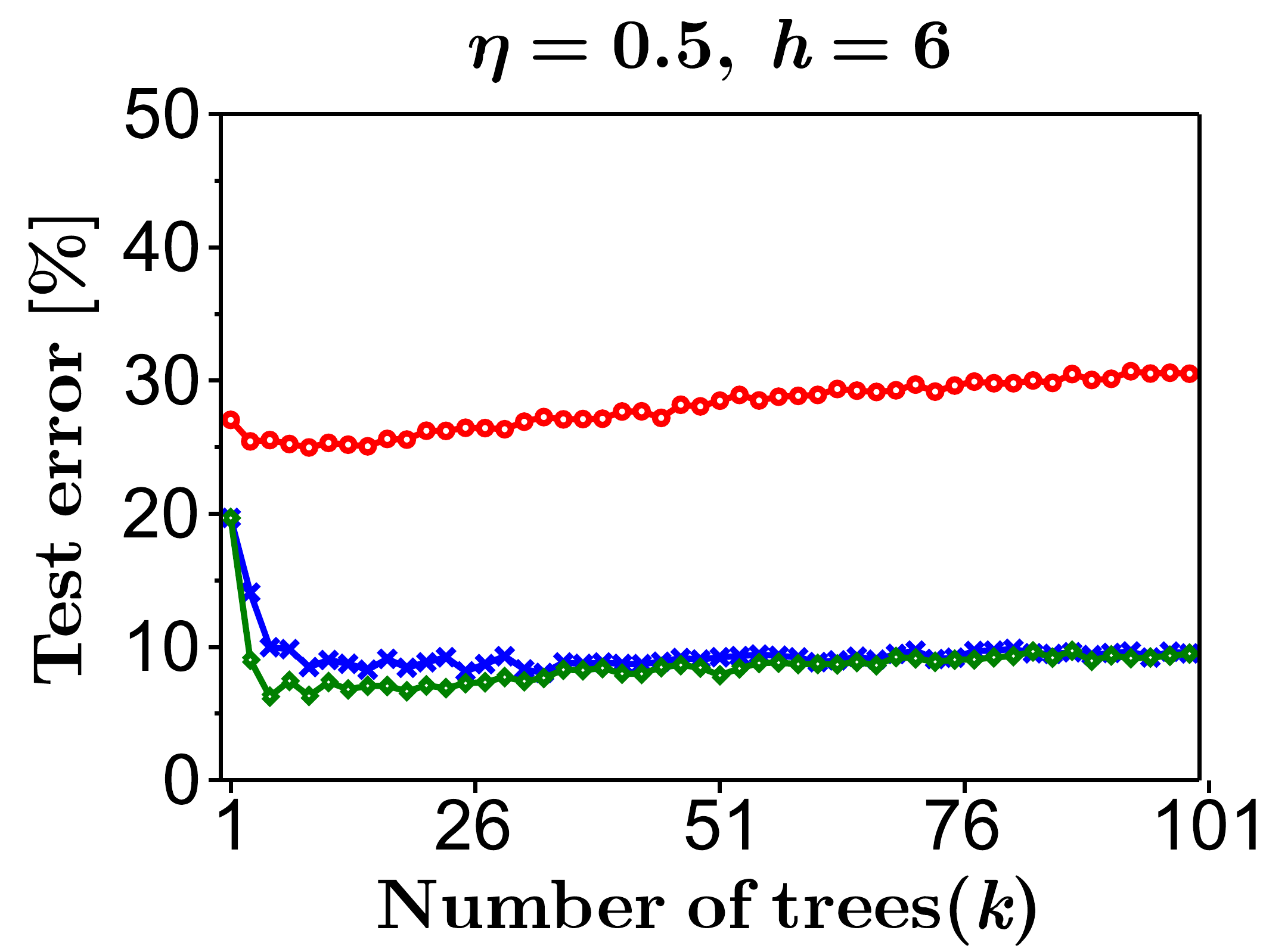} 
\hspace{-0.03in}\includegraphics[width = 1.49in]{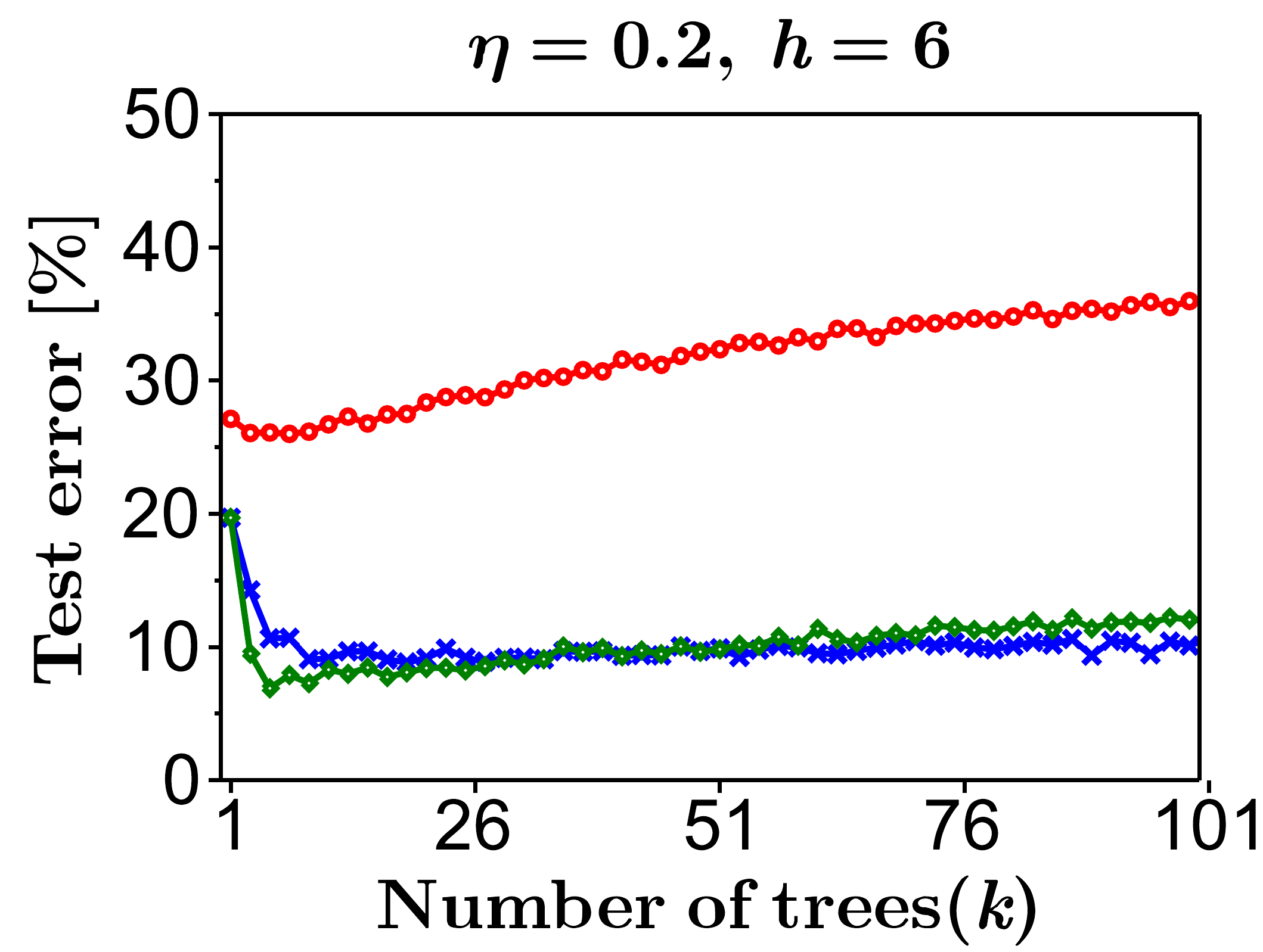} 
\hspace{-0.03in}\includegraphics[width = 1.49in]{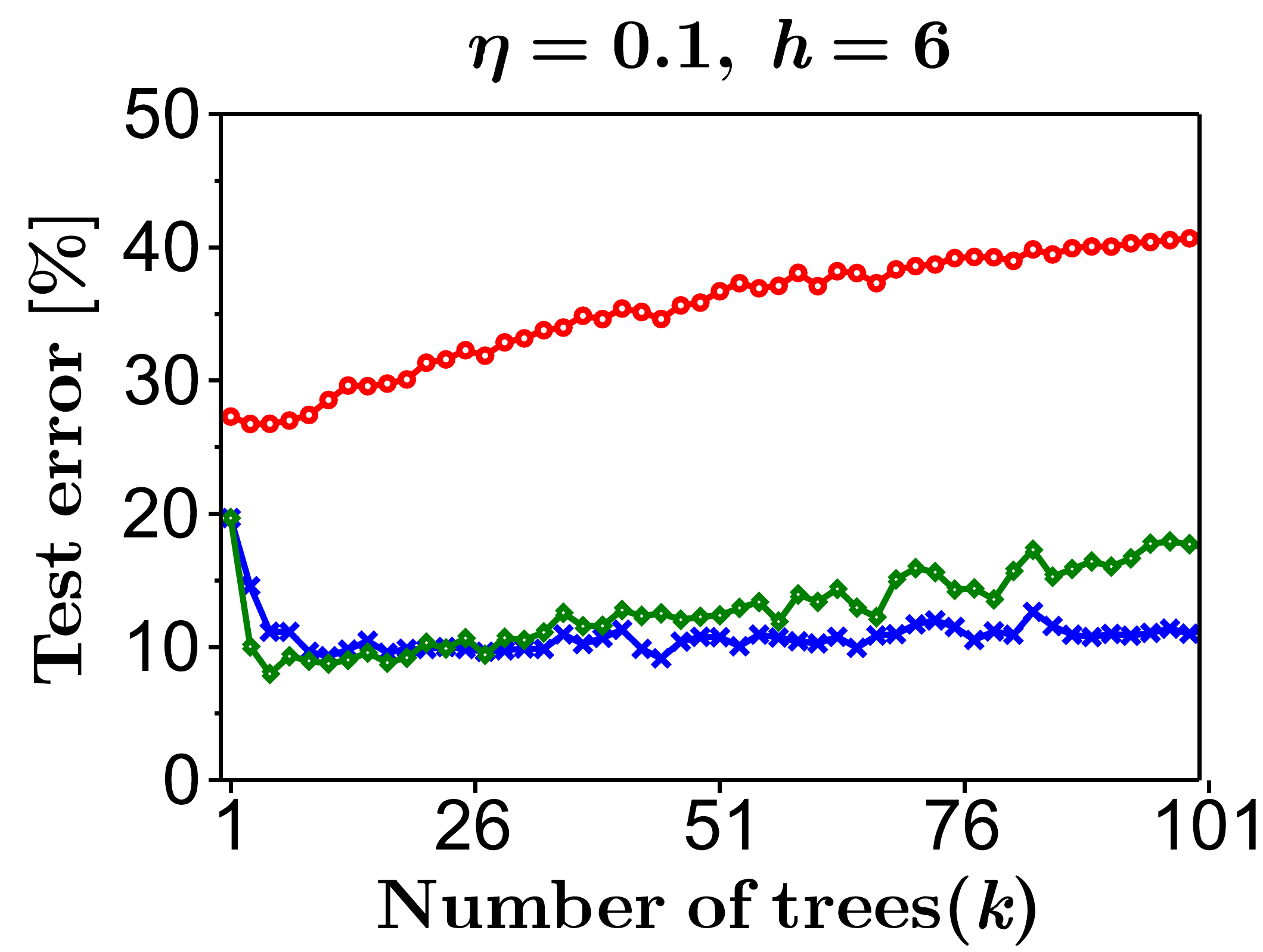}\\
\vspace{-0.1in}
\caption{Comparison of dpRFMV, dpRFTA and dpRFPA for selected datasets. \textbf{a)} Test error vs. $\eta$ for two settings of $h$. \textbf{b)} Test error vs. $k$ for fixed $h$ and across different settings of $\eta$.}
\label{fig:three}
\end{figure*}

Next set of results\footnote{All figures in this section should be read in color.} (Figure~\ref{fig:one} and~\ref{fig:three}) is reported for an exemplary datasets (\textit{Banknote Authentication}, \textit{Congressional Voting Records}, \textit{Mammographic Mass} and \textit{Mushroom}) and for the following methods: \textit{dpRFMV}, \textit{dpRFTA} and \textit{dpRFPA}. Note that similar results were obtained for the remaining datasets. In Figure~\ref{fig:one}a we report the test error vs. $h$ for selected settings of $k$\footnote{Recall that in case when the forest contains only one tree ($k=1$) majority voting and threshold averaging rules are equivalent thus the blue curve overlaps with the green curve on the plot then.}. In Figure~\ref{fig:one}b we also show minimal, average and maximal test error vs. $h$ for dpRFMV, whose performance was overall the best. Similarly, in Figure~\ref{fig:one}c we report the test error vs. $k$ for two selected settings of $h$ and in Figure~\ref{fig:one}d we also show minimal, average and maximal test error vs. $k$ for dpRFMV.

Finally, in Figure~\ref{fig:three}a we report test error for various settings of $\eta$ and two selected settings of $h$. For each experiment $k$ was chosen from the set $\{1,2,\dots,101\}$ to give the smallest validation error. Additionally, in Figure~\ref{fig:three}b we show how the test error changes with $k$ for a fixed $h$ and various levels of $\eta$.

Figure~\ref{fig:three}a shows that in most cases \textit{dpRFTA} outperforms remaining differentially-private classifiers, however it requires careful selection of the forest parameters ($h$ and $k$) in order to obtain the optimal performance as is illustrated on Figure~\ref{fig:one}c and~\ref{fig:three}b. This problem can be overcome by using \textit{dpRFMV} which has comparable performance to \textit{dpRFTA} but is much less sensitive to the setting of the forest parameters. Therefore \textit{dpRFMV} is much easier to use in the differentially-private setting.    

\section{Conclusions}
\label{sec:conclusions}

In this paper we first provide novel theoretical analysis of supervised learning with non-differentially-private random decision trees in three cases: majority voting, threshold averaging and probabilistic averaging. Secondly we show that the algorithms we consider here can be easily adapted to the setting where high privacy guarantees must be achieved. We furthermore provide both theoretical and experimental evaluation of the differentially-private random decision trees approach. To the best of our knowledge, the theoretical analysis of the differentially-private random decision trees was never done before. Our experiments reveal that majority voting and threshold averaging are good differentially-private classifiers and that in particular majority voting exhibits less sensitivity to forest parameters. 

\bibliographystyle{unsrt}
\bibliography{KDD_MAKY}  

\newpage
\onecolumn

\vbox{\hsize\textwidth
\linewidth\hsize \vskip 0.1in 
\centering
{\LARGE\bf  Differentially- and non-differentially-private random decision trees\\(Supplementary Material) \par}
\vskip 0.3in minus 0.1in}
\vspace{-0.2in}

\section{Empirical and generalization errors}

\subsection{Preliminaries}

We will  prove here results regarding empirical and generalization errors of all the variants of the algorithm mentioned in the paper as well
as Theorem~\ref{technicaltheorem_gamma} and Theorem~\ref{technicaltheorem}. Without loss of generality we will assume that all attributes are binary (taken from the set $\{0,1\}$). It can be easily noticed that the proofs can
be directly translated to the continuous case. We leave this simple exercise to the reader.

Let us introduce first some useful notation that will be very helpful in the proofs we present next.

We denote by $n$ the size of the dataset (training or test) $\mathcal{T}$.
Let us remind that $m$ is the number of attributes of any given data point, $h$ is the height of the random decision tree and $M$ is the set of all random decision trees under consideration. 

We focus first on classifying with just one decision tree. Fix some decision tree $T_{j}$ and one of its leaves. Assume that it contains $a$ points with label: $-$ and $b$ points with label: $+$. We associate label $-$ with that leaf if $a>b$ and label $1$ otherwise.
To classify a given point using that tree we feed our tree with that point and assign to a point a label of the corresponding leaf. 
Denote by $m^{i}$ the number of data points that were correctly classified by a tree $T_{i}$. Denote $e^{i}=\frac{m^{i}}{n}$. We call $e^{i}$ the \textit{quality} (or \textit{accuracy}) of the tree $T_{i}$. Note that obviously we always have: $e^{i} \geq \frac{1}{2}$, since for every leaf of any given tree the majority of the data points from that leaf are classified correctly. Denote: $e=\frac{1}{|M|}\sum_{i=1}^{|M|}e^{i}$. 
We call $e$ the average tree accuracy. This parameter measures how well data points are classified on average by a complete decision tree of a given height $h$. Note that $e \geq \frac{1}{2}$. 
Denote $t=2^{h}$. Parameter $t$ is the number of leaves of the decision tree. 

For $i=1,2,...,|M|$ and $j=1,2,...,t$ denote by $n^{i}_{j}$ the number of points from the dataset in the $j^{th}$ leaf of a decision tree $T_{i}$. 
Denote by $m^{i}_{j}$ the number of points from the dataset in the $j^{th}$ leaf of the decision tree $T_{i}$ that were classified correctly. Denote $e^{i}_{j}=\frac{m^{i}_{j}}{n^{i}_{j}}$ for $n^{i}_{j}>0$ and $e^{i}_{j}=1$ for $n^{i}_{j}=0$. Note that $e^{i}_{j} \geq \frac{1}{2}$ for every $i,j$. Note also that we have: $n=n^{i}_{1}+...+n^{i}_{t}$ and $m^{i}=m^{i}_{1}+...+m^{i}_{t}$.
Denote by $a^{i}_{j}$ the number  of data points in the $j^{th}$ leaf of the decision tree $T_{i}$ that are of label $0$. Denote by $b^{i}_{j}$ the number  of data points in the $j^{th}$ leaf of the decision tree $T_{i}$ that are of label $1$.  

We will use frequently the following structure in the proofs.
Let $\mathcal{G}$ be a bipartite graph with color classes: $\mathcal{A}$, $\mathcal{B}$ and weighted edges. Color class $\mathcal{A}$ consists of $n$ points from the dataset. Color class $\mathcal{B}$ consists of $2t|M|$ elements of the form $y^{i,b}_{j}$, where $i \in \{1,2,...,|M|\}$, $b \in \{0,1\}$ and $j \in \{1,2,...,t\}$.\\
Data point $x \in \mathcal{A}$ is adjacent to $y^{i,1}_{j}$ iff it belongs to larger of the two groups (these with labels: $0$ and $1$) of the data points that are in the $j^{th}$ leaf of the decision tree $T_{i}$. An edge joining $x$ with $y^{i,1}_{j}$ has weight $e^{i}_{j}$.
Data point $x \in \mathcal{A}$ is adjacent to $y^{i,0}_{j}$ iff it belongs to smaller of the two groups of the data points that are in the $j^{th}$ leaf of the decision tree $T_{i}$. An edge joining $x$ with $y^{i,0}_{j}$ has weight $1-e^{i}_{j}$.
Note that the degree of a vertex $y^{i,1}_{j}$ is $m^{i}_{j}$ and the degree of a vertex $y^{i,0}_{j}$ is $n^{i}_{j}-m^{i}_{j}$.

In the proofs we will refer to the size of the set of decision trees under consideration as: $|M|$ or $k$ (note that
$k$ is used in the main body of the paper).

We are ready to prove Theorem~\ref{technicaltheorem_gamma} and Theorem~\ref{technicaltheorem}.

\begin{proof}
We start with the proof of Theorem~\ref{technicaltheorem}.
Note that from the definition of $w_{d}$ we get:
$$\sum_{d \in \mathcal{T}} w_{d} = \frac{1}{|M|}\sum_{i=1}^{|M|}\sum_{j=1}^{t}(m^{i}_{j}e^{i}_{j}+(n^{i}_{j}-m^{i}_{j})(1-e^{i}_{j})).$$
Therefore, using formula on $m^{i}_{j}$, we get:
$$\sum_{d \in \mathcal{T}} w_{d} = \frac{1}{|M|}\sum_{i=1}^{|M|}\sum_{j=1}^{t}(n^{i}_{j}(e^{i}_{j})^{2}+n^{i}_{j}(1-e^{i}_{j})^{2}).$$
Note that we have: $\sum_{i=1}^{|M|}\sum_{j=1}^{t} n^{i}_{j} = n|M|$.
From Jensen's inequality, applied to the function $f(x)=x^{2}$, we get:
$\sum_{i=1}^{|M|}\sum_{j=1}^{t} \frac{n^{i}_{j}}{|M|n}(e^{i}_{j})^{2} \geq (\sum_{i=1}^{|M|}\sum_{j=1}^{t} \frac{n^{i}_{j}e^{i}_{j}}{|M|n})^{2}=(\frac{\sum_{i=1}^{|M|}\sum_{j=1}^{t}m^{i}_{j}}{|M|n})^{2}=(\frac{en|M|}{n|M|})^{2}=e^{2}$, where $e$ is the average quality of the system of all 
complete decision trees of height $h$ (the average tree accuracy).
Similarly, $\sum_{i=1}^{|M|}\sum_{j=1}^{t} \frac{n^{i}_{j}}{|M|n}(1-e^{i}_{j})^{2} \geq (1-e)^{2}$. Thus we get:
$$\sum_{d \in \mathcal{T}} w_{d} \geq n(e^{2}+(1-e)^{2}).$$

That completes the proof of Theorem~\ref{technicaltheorem}. The proof of Theorem~\ref{technicaltheorem_gamma} is even simpler.
Notice that for any data point $d$ the expression $\sigma(d) \cdot |M|$ counts the number of decision trees from $M$ that classified 
$d$ correctly (follows directly from the definition of $\theta$). Thus we have: 
$\sum_{d \in \mathcal{T}} \sigma(d) \cdot |M| = \sum_{i=1}^{|M|}m^{i}$. Therefore 
$\frac{1}{n}\sum_{d \in \mathcal{T}} \sigma(d)= \frac{1}{|M|}\sum_{i=1}^{|M|} e^{i}$ and we are done.
\end{proof}

We need one more technical result, the Azuma's inequality:

\begin{lemma}
\label{Azuma}
Let $\{W_{n},n \geq 1\}$ be a martingale with mean $0$ and suppose that for some non-negative constants: $\alpha_{i},\beta_{i}$ we have: $-\alpha_{i} \leq W_{i}-W_{i-1} \leq \beta_{i}$ for $i=2,3,...$. Then for any $n \geq 0$, $a>0$:
\[\mathbb{P}(W_{n} \geq a) \leq e^{-\frac{2a^{2}}{\sum_{i=1}^{n} (\alpha_{i}+\beta_{i})^{2}}} \:\:\:\text{and}\:\:\: \mathbb{P}(W_{n} \leq -a) \leq e^{-\frac{2a^{2}}{\sum_{i=1}^{n}(\alpha_{i}+\beta_{i})^{2}}}.
\]
\end{lemma}

\subsection{Majority voting and threshold averaging setting - empirical error}

We will now prove  parts of theorems:~\ref{alfa_setting_ndp_main_1} and~\ref{alfa_setting_ndp_main_2} regarding empirical errors. 

\begin{proof}
Again, we start with the analysis of the threshold averaging.
Take $i^{th}$ random decision tree $T^{R}_{i}$, where $i \in \{1,2,...,k\}$. For a given data point $d$ from the training set let $X^{d}_{i}$ be a random variable defined as follows.
If $d$ does not belong to any leaf of $T^{R}_{i}$ then let $X^{d}_{i}=0$. Otherwise let $a^{R}_{i}$ be the number of points from the training set with label $0$ in that leaf  and let $b^{R}_{i}$  be the number of points from the training set with label $1$ in that leaf. If $d$ has label $0$ then we take $X^{d}_{i}=\frac{a^{R}_{i}}{a^{R}_{i}+b^{R}_{i}}$. Otherwise we take $X^{d}_{i}=\frac{b^{R}_{i}}{a^{R}_{i}+b^{R}_{i}}$. 
Denote $X^{d}=\frac{X^{d}_{1}+...+X^{d}_{k}}{k}$.
When from the context it is clear to which data point we refer to we will skip upper index and simply write $X$ or $X_{i}$ respectively.\\ 
Fix some point $d$ from the training set.
Note that if $X > \frac{1}{2}$ then point $d$ is correctly classified.   
Notice that the weight of the point $d$ denoted as $w_{d}$ is nothing else but the sum of weights of all the edges of $\mathcal{G}$ incident to $d$
divided by the number of all trees (or the average weight of an edge indecent to $d$ if we consider real-valued attributes). Note that we have $EX = w_{d}$ and that from Theorem~\ref{technicaltheorem} we get:
$$\sum_{d \in \mathcal{T}} w_{d} \geq n(e^{2}+(1-e)^{2}).$$

Take $0 < \delta < \frac{1}{2}$. Denote by $\mu$ the fraction of points $d$ from the training data such that $w_{d} \geq \frac{1}{2}+\delta$. From the lower bound on $\sum_{d \in \mathcal{T}} w_{d}$, we have just derived, we get: $(\frac{1}{2}+\delta)(1-\mu)n+\mu n \geq n(e^{2}+(1-e)^{2})$, which gives us:
$$\mu \geq 1 - \frac{2\epsilon-2\epsilon^{2}}{0.5-\delta},$$ where $\epsilon = 1-e$.\\
Take point $d$ from the training set such that $w_{d} \geq \frac{1}{2} +\delta.$
Denote by $p_{d}$ the probability that $d$ is misclassified. We have:
$$p_{d} \leq \mathbb{P}(\frac{X_{1}+...+X_{k}}{k} \leq w_{d} - \delta).$$
Denote: $Z_{i}=X_{i}-w_{d}$ for $i=1,2,...,k$.
We have: 
$$p_{d} \leq \mathbb{P}(Z_{1}+...+Z_{k} \leq - k\delta).$$
Note that, since $w_{d}=EX$ and random variables $X_{i}$ are independent, we can conclude that $\{Z_{1},Z_{1}+Z_{2},...,Z_{1}+Z_{2}+...+Z_{k}\}$ is a martingale. 
Note also that $-\alpha_{i} \leq Z_{i} \leq \beta_{i}$ for some $\alpha_{i},\beta_{i}>0$ such that $\alpha_{i}+\beta_{i}=1$.

Using Lemma~\ref{Azuma}, we get: 
$$\mathbb{P}(Z_{1}+...+Z_{k} \leq -k\delta) \leq e^{-\frac{2(k\delta)^{2}}{k}}.$$

Therefore the probability that at least one of $ \mu n $ points $d$ for which $w_{d} \geq \frac{1}{2}+\delta$ will be misclassified by the set of $k$ random decision trees is, by union bound, at most: $\mu n e^{-2k\delta^{2}} \leq n e^{-2k\delta^{2}} $.
That, for $k=\frac{(1+C)\log(n)}{2\delta^{2}}$, completes the proof of the upper bound on the empirical error from theorems:~\ref{alfa_setting_ndp_main_1} and ~\ref{alfa_setting_ndp_main_2} since we have  already proved that $\mu \geq 1 - \frac{2\epsilon-2\epsilon^{2}}{0.5-\delta}$.
The proof of the majority voting version goes along exactly the same lines. This time, instead of Theorem \ref{technicaltheorem}, we use
Theorem \ref{technicaltheorem_gamma}. We know that $\sum_{d \in \mathcal{T}} \sigma(d) \geq ne$, where $e = 1 - \epsilon$. 
Denote the fraction of points $d$ with $\sigma(d) \geq \frac{1}{2} + x$ for $0<x<\frac{1}{2}$ by $\mu^{x}$.
Then, by the argument similar to the one presented above,we have:
\begin{equation}
\label{majority_ineq}
\mu^{x} \geq 1 - \frac{\epsilon}{0.5 - x}.
\end{equation}
All other details of the proof for the
majority voting are exactly the same as for the threshold averaging scheme.
\end{proof}

Next we prove  parts of Theorem~\ref{alfa_setting_dp_main_1} regarding empirical error and Theorem~\ref{alfa_setting_dp_main_2}.

\begin{proof}
Let $K>0$ be a constant.
We first consider the threshold averaging scheme.
Take a decision tree $T_{i}$. Denote by $S_{T_{i}}$ the set of points $d$ from the training set with the following property: point $d$ belongs in $T_{i}$ do the leaf that contains at least $\frac{n}{K2^{h}}$ points. Note that since each $T_{i}$ has exactly $2^{h}$ leaves, we can conclude that $|S_{T_{i}}| \geq n(1-\frac{1}{K})$. 
In this proof and proof of theorems:~\ref{beta_setting_dp_main_1} and ~\ref{beta_setting_dp_main_1} (presented in the next section) we will consider graph $\mathcal{G}^{D}$ that is obtained from $\mathcal{G}$ by deleting edges adjacent to those vertices of the color class $\mathcal{B}$ that correspond to leaves containing less than $\frac{n}{K2^{h}}$ points from the training set.
Take point $d$ from the training set with $w^{t}_{d} \geq \frac{1}{2}+\delta$, where $w^{t}_{d}$ is the average weight of an edge incident
to $d$ in $\mathcal{G}^{D}$. Notice that $w_{d} \geq \frac{1}{2}+\delta + \frac{1}{K}$ implies: $w^{t}_{d} \geq \frac{1}{2}+\delta$.
We say that a decision tree $T_{i}$ is \textit{$d$-good} if the leaf of $T_{i}$ to which $d$ belongs contains at least $\frac{n}{K2^{h}}$ points from the training set. Let us now define $X^{d}_{i}$. If $i^{th}$ chosen random decision tree is $d$-good then $X^{d}_{i}$ is defined as in the proof of Theorem~\ref{alfa_setting_ndp_main_1}. Otherwise we put $X^{d}_{i}=0$. Denote $Z_{i}=X^{d}_{i}-w^{t}_{d}$.
Note that the probability $p_{d}$ that point $d$ is misclassified by selected random decision trees is $p_{d} \leq \mathbb{P}(\frac{Z_{1}+...+Z_{k}}{k}+\frac{\sum_{j \in \mathcal{I}} R_{j}}{|\mathcal{I}|} \leq -\delta)$, where $\mathcal{I}$ is the set of indices corresponding to those chosen random decision trees that are $d$-good and random variables $R_{j}$ are correction terms for $d$-good random decision trees that must be introduced in order to take into account added Laplacians (if $\mathcal{I} = \emptyset$ then we assume that the value of the expression $\frac{\sum_{j \in \mathcal{I}} R_{j}}{|\mathcal{I}|}$ is $0$).
Note also that set $\{R_{j}$, $Z_{j}:j=1,2,...,k\}$ is a set of independent random variables. We get:
$$
p_{d} \leq \mathbb{P}(\frac{Z_{1}+...+Z_{k}}{k} \leq -\frac{\delta}{2})+\mathbb{P}(\frac{\sum_{j \in \mathcal{I}} R_{j}}{|\mathcal{I}|} \leq -\frac{\delta}{2}).
$$

Since from the Azuma's inequality we get: $\mathbb{P}(\frac{Z_{1}+...+Z_{k}}{k} \leq -\frac{\delta}{2}) \leq e^{-\frac{k\delta^{2}}{2}}$, we have:

\begin{equation}
\label{inequality1}
p_{d} \leq e^{-\frac{k\delta^{2}}{2}} + \mathbb{P}(\frac{\sum_{j \in \mathcal{I}} R_{j}}{|\mathcal{I}|} \leq -\frac{\delta}{2})
\end{equation}

We will now estimate the expression $p_{r}=\mathbb{P}(\frac{\sum_{j \in \mathcal{I}} R_{j}}{|\mathcal{I}|} \leq -\frac{\delta}{2})$. 

For $i \in \mathcal{I}$ denote by $\mathcal{A}_{i}$ an event that each of the two perturbation errors added to the leaf containing point $d$ was of magnitude at most $\frac{\sqrt{n}}{K2^{h}}\delta_{1}$, where $\delta_{1}=\frac{\delta}{24}$. Denote $\mathcal{A}=\bigcap_{i \in \mathcal{I}} \mathcal{A}_{i}$. Denote by $\mathcal{A}^{c}$ the complement of $\mathcal{A}$.
We have:
$\mathbb{P}(\frac{\sum_{j \in \mathcal{I}} R_{j}}{|\mathcal{I}|} \leq -\frac{\delta}{2}) = \mathbb{P}(\frac{\sum_{j \in \mathcal{I}} R_{j}}{|\mathcal{I}|} \leq -\frac{\delta}{2}|\mathcal{A})\mathbb{P}(\mathcal{A})+\mathbb{P}(\frac{\sum_{j \in \mathcal{I}} R_{j}}{|\mathcal{I}|} \leq -\frac{\delta}{2}|\mathcal{A}^{c})(1-\mathbb{P}(\mathcal{A}))$.
Thus we get:
\begin{equation}
\label{inequality2}
p_{r} \leq \mathbb{P}(\frac{\sum_{j \in \mathcal{I}} R_{j}}{|\mathcal{I}|} \leq -\frac{\delta}{2}|\mathcal{A}) + (1-\mathbb{P}(\mathcal{A})).
\end{equation}

Now take one of the chosen random decision trees $T_{i}$ with $i \in \mathcal{I}$. Take its leaf that contains given point $d$ from the training set.  Assume that this leaf contains $r$ points from the training set with some fixed label $l \in \{-,+\}$ and that it altogether contains $n_{a}$ points. Note that 
from the definition of $\mathcal{I}$ we have: $n_{a} \geq \frac{n}{K2^{h}}$. Let $g_{1},g_{2}$ be two independent Laplacian random variables, each of density function $\frac{\eta}{2k}e^{-\frac{|x|\eta}{k}}$. We would like to estimate the following random variable $\Theta = \frac{r+g_{1}}{n_{a}+g_{1}+g_{2}}-\frac{r}{n_{a}}$ for an event $\mathcal{A}$.
Note that in particular we know that $|g_{1}|,|g_{2}| \leq \frac{\delta_{1}n_{a}}{\sqrt{n}}$. Simple calculation gives us:
\begin{equation}
\label{inequality3}
|\Theta| \leq \frac{\delta}{4\sqrt{n}}.
\end{equation}

Now consider truncated probability space $\Omega | \mathcal{A}$ and truncated random variables $R^{t}_{i} = R_{i} |\mathcal{A}$ for $i \in \mathcal{I}$.
We have: $\mathbb{P}(\sum_{i \in \mathcal{I}} R_{i} \leq -\frac{|\mathcal{I}|\delta}{2}|\mathcal{A}) = \mathbb{P}(\sum_{i \in \mathcal{I}} R^{t}_{i} \leq -\frac{|\mathcal{I}|\delta}{2})$.
Using inequality~\ref{inequality3}, we get: 
\begin{equation}
\label{equality0}
|R^{t}_{i}| \leq \frac{\delta}{4\sqrt{n}},
E|R^{t}_{i}| \leq \frac{\delta}{4\sqrt{n}}.
\end{equation}

Thus we can use Azuma's inequality once more, this time to find the upper bound on the expression: $\mathbb{P}(\sum_{i \in \mathcal{I}} R^{t}_{i} \leq -\frac{|\mathcal{I}|\delta}{2})$ (we assume here that the random decision trees have been selected thus $\mathcal{I}$ is given). Without loss of generality we can assume that $\mathcal{I} \neq \emptyset$.
We have: $\mathbb{P}(\sum_{i \in \mathcal{I}} R^{t}_{i} \leq -\frac{|\mathcal{I}|\delta}{2}) = \mathbb{P}(\sum_{i \in \mathcal{A}} (R^{t}_{i}-ER^{t}_{i}) \leq -\frac{\mathcal{I}\delta}{2}-\sum_{i \in \mathcal{I}} ER^{t}_{i}) \leq \mathbb{P}(\sum_{i \in \mathcal{I}} (R^{t}_{i}-ER^{t}_{i}) \leq -\frac{|\mathcal{I}|\delta}{4}) \leq e^{-\frac{2|\mathcal{I}|(\frac{\delta}{4})^{2}}{(\frac{\delta}{4\sqrt{n}}+\frac{\delta}{4\sqrt{n}})^{2}}}$.
Therefore we get:
\begin{equation}
\label{inequality4}
p_{r} \leq e^{-\frac{n}{2}} + (1-\mathbb{P}(\mathcal{A})).
\end{equation}

It remains to bound the expression: $(1-\mathbb{P}(\mathcal{A}))$.
Let $g$ be a Laplacian random variable with density function $\frac{\eta}{2k}e^{-\frac{|x|\eta}{k}}$.
Note that from the union bound we get: $1-\mathbb{P}(\mathcal{A}) \leq 2k \mathbb{P}(|g| > \frac{\sqrt{n}\delta}{24K2^{h}})$, where factor $2$ in the expression $2k \mathbb{P}(g > \frac{\sqrt{n}\delta}{24K2^{h}})$ comes from the fact that for a given data point $d$ we need to add perturbation error in two places in the leaf of the chosen random decision tree corresponding to $d$.\\
Denote $\gamma = \frac{\delta}{24K2^{h}}$.
We have:
\begin{equation}
\label{inequality5}
p_{r} \leq e^{-\frac{n}{2}} + 4k \int\limits_{\gamma \sqrt{n}}^{\infty} \frac{\eta}{2k}e^{-\frac{x\eta}{k}}dx.
\end{equation}

Evaluation of the RHS-expression gives us:

\begin{equation}
\label{inequality6}
p_{r} \leq e^{-\frac{n}{2}} +2ke^{-\frac{\lambda \sqrt{n} \eta}{k}}, \:\:\:\:\:\text{where}\:\:\: \lambda = \frac{\delta}{24K2^{h}}.
\end{equation}

Thus we can conclude that the probability $p_{d}$ that the fixed point $d$ from the training set will be misclassified by the set of $k$ randomly chosen random decision trees satisfies:
\begin{equation}
\label{important_inequality}
p_{d} \leq  e^{-\frac{k\delta^{2}}{2}} + e^{-\frac{n}{2}} + 2ke^{-\frac{\gamma \sqrt{n} \eta}{k}}.
\end{equation}

Note that by the similar argument to the one presented in the proof of Theorem~\ref{alfa_setting_ndp_main_1} and Theorem~\ref{alfa_setting_ndp_main_2}, we can conclude that at least $n(1-\frac{2(\epsilon+\frac{1}{K})-2(\epsilon+\frac{1}{K})^{2}}{0.5-\delta})$
points $d$ from the training data satisfy: $w^{t}_{d} \geq \frac{1}{2} +\delta$. Let $\mu^{t}$ be a fraction of points with this property.
As we observed earlier, if the points $d$ satisfies: $w_{d} \geq \frac{1}{2} +\delta + \frac{1}{K}$ then it also satisfies:
$w^{t}_{d} \geq \frac{1}{2} +\delta$. Thus $\mu \geq \mu^{t}$. We also have: $\mu^{t} \geq 1-\frac{2(\epsilon+\frac{1}{K})-2(\epsilon+\frac{1}{K})^{2}}{0.5-\delta}$. Thus $\mu \geq 1-\frac{2(\epsilon+\frac{1}{K})-2(\epsilon+\frac{1}{K})^{2}}{0.5-\delta}$.
We replace $\epsilon$ by $\epsilon+\frac{1}{K}$ in the formula derived in the proof of Theorem~\ref{alfa_setting_ndp_main_1} since now for any fixed decision tree we do not take into account points that belong to leaves with less that $\frac{n}{K2^{h}}$ points from the training set. For every given decision tree $T_{i}$ there are at most $\frac{n}{K}$ points $d$ from the training set such that $T_{i}$ is not $d$-good. 
Note that, by union bound, the probability that at least one from the $n \mu$ points $d$ with $w^{t}_{d} \geq \frac{1}{2} +\delta$ is misclassified is at most $n \mu p_{d} \leq n p_{d}$. To see how Theorem~\ref{alfa_setting_dp_main_2} and the part of Theorem~\ref{alfa_setting_dp_main_1} regarding empirical error follow now, take $K=40$ and $\delta=\frac{1}{10}$.
The proof of the majority voting version is very similar. We use inequality \ref{majority_ineq} (that was derived from
Theorem \ref{technicaltheorem_gamma}) but all other details are exactly the same. Therefore we will not give it in details here
since it would basically mean copying almost exactly the proof that we have just showed.
\end{proof}

\subsection{Probabilistic averaging setting - empirical error}

Let us switch now to the probabilistic averaging setting. In practice, as was shown in the experimental section, it is the
least effective method. However for the completeness of our theoretical analysis and since for very large datasets
theoretical guarantees regarding also this setting can be obtained, we focus on it now.

We will first focus on the part of Theorem~\ref{beta_setting_ndp_main} regarding empirical error.

\begin{proof}
We already know that: $\sum_{d \in \mathcal{T}} w_{d} \geq n(e^{2}+(1-e)^{2})$, where $e$ is the average quality.
Assume that $k$ random decision trees have been selected.
Denote by $Y_{d}$ the indicator of the event that a fixed data point $d$ from the training set will be correctly classified. We have:
$$ Y_{d} = 
\left\{ \begin{array}{rcl}
1 & \mbox{with probability}
& X^{d} \\ 0 & \mbox{with probability} & 1-X^{d}, \\
\end{array}\right.$$
where $X^{d}$ is random variable defined in the proof of theorems:~\ref{alfa_setting_ndp_main_1} and ~\ref{alfa_setting_ndp_main_2}. Note that after random decision trees have been selected, $X^{d}$ has a deterministic value.
Note also that random variables $Y_{d}$ are independent and $EY_{d}=X^{d}$. Thus, we can use Lemma~\ref{Azuma} in the very similar way as in the proof of theorems:~\ref{alfa_setting_ndp_main_1} and ~\ref{alfa_setting_ndp_main_2} to get that for any given $c>0$:

\begin{equation}
\label{inequality0}
\mathbb{P}(\sum_{d \in \mathcal{T}} (Y_{d}-X^{d}) \leq -nc) \leq e^{-2nc^{2}}.
\end{equation}

Let us focus now on the process of choosing random decision trees.
Fix parameter $\delta>0$. Fix some point $d$ from the training set. Using Lemma~\ref{Azuma} in exactly the same way as in the proof of theorems:~\ref{alfa_setting_ndp_main_1} and ~\ref{alfa_setting_ndp_main_2}, we conclude that $\mathbb{P}( X^{d} < w_{d} - \delta) \leq e^{-2k\delta^{2}}$. Therefore, by the union bound, with probability at least $(1-ne^{-2k\delta^{2}})$ we have: $\sum_{d \in \mathcal{T}} X^{d} \geq \sum_{d \in \mathcal{T}} (w_{d} - \delta)$. Thus, according to the lower bound for $\sum_{d \in \mathcal{T}} w_{d}$ we presented at the beginning of the proof, we get that with probability at least $(1-ne^{-2k\delta^{2}})$ the following holds: $\sum_{d \in \mathcal{T}} X^{d} \geq n(1-2\epsilon+2\epsilon^{2}-\delta)$, where $\epsilon = 1 - e$.
Note that random variables $Y_{d}$ are independent from random variables $X_{d}$. We can conclude, using inequality~\ref{inequality0}, that with probability at least $(1-ne^{-2k\delta^{2}})(1-e^{-2nc^{2}})$ at least $n(1-2\epsilon+2\epsilon^{2}-\delta-c)$ points will be correctly classified. Now we can take  $k=\frac{(1+C)\log(n)}{2\delta^{2}}$ and that completes the proof.
Again, as in the previous proof, the majority voting scheme requires only minor changes in the presented proof so we will leave to the reader.
\end{proof}

Lets focus now on  parts of theorems: ~\ref{beta_setting_dp_main_1} and ~\ref{beta_setting_dp_main_2} regarding empirical errors.

\begin{proof}

Proofs of statements regarding empirical errors go along exactly the same lines as presented proof of the part of 
Theorem~\ref{beta_setting_ndp_main} (regarding empirical error). The changes in the statement, due to the added perturbation error, follow from the proof of bounds on the empirical error from theorems:~\ref{alfa_setting_dp_main_1} and ~\ref{alfa_setting_dp_main_2} . Therefore we will not give the entire proof but only mention few things. 

In comparison with the statement of Theorem~\ref{beta_setting_ndp_main}, in the expression on the upper bound on empirical error the term $\epsilon$ is replaced by $\epsilon+\frac{1}{K}$. This is, as explained in the proof of Theorem~\ref{alfa_setting_dp_main_1} (regarding empirical error), due to the fact that while dealing with weights of edges in graph $\mathcal{G}^{D}$ we do not take into account points from the training set 
corresponding to leaves with too few data points. 
To see how Theorem~\ref{beta_setting_dp_main_2} can be derived, take $K=40$, $\delta=\frac{1}{10}$, $c=\frac{1}{20}$. Again, as for Theorem~\ref{alfa_setting_dp_main_2}, Theorem~\ref{beta_setting_dp_main_2} follows now by simple calculations.

\end{proof}

\vspace{-0.3in}
\subsection{Generalization error}

We will now prove upper bounds regarding generalization error for all the theorems presented in the previous paragraphs. We do it for all of them in the same section 
since all the proofs are very similar. Besides, right now, when we have already developed tools for obtaining upper bounds on the empirical error,
we can use them to simplify our analysis regarding generalization error.
Random decision trees give strong bounds on the generalization error since they do not lead to data overfitting. The internal structure of each constructed tree (i.e. the set of its inner nodes) does not depend at all on the data. 
This fact is crucial in obtaining strong guarantees on the generalization error. 
All the experiments presented in the main body of the paper measured generalization error of the random tree approach and stand for the
empirical verification that this method is a good learning technique in the setting requiring high privacy guarantees.
Below is the proof of the presented upper bounds on the generalization error.

\begin{proof}
Consider test set of $n$ points. 
Whenever we refer to the weight or goodness of the test point $d$, this is in respect to the test set (see: definition of goodness
and other terms in the description of the model).
Let $\phi>0$ be a small constant and denote by $\mathcal{E}_{\phi}$ an event that for the selected forest $\mathcal{F}$
of random decision trees the non-perturbed counts in all leafs (for each leaf we count points with label $+$ and $-$ in that leaf)
for the test set and training set differ by at most $2\phi n$.
We start by finding a lower bound on $\mathbb{P}(\mathcal{E}_{\phi})$.
Let us fix a forest, a particular tree of that forest and a particular leaf of that tree.
Denote by $X_{i}$ a random variable that takes value $1$ if $i^{th}$ point of the training set corresponds to that leaf and $0$ otherwise.
Similarly, denote by $Y_{i}$ a random variable that takes value $1$ if $i^{th}$ point of the test set corresponds to that leaf and $0$
otherwise.
Denote by $X_{i}^{+}$ a random variable that takes value $1$ if $i^{th}$ point of the training set corresponds to that leaf and has
label $+$ and is $0$ otherwise. Similarly, denote by $Y_{i}^{+}$ a random variable that takes value $1$ if $i^{th}$ point of the
test set corresponds to that leaf and has label $+$ and is $0$ otherwise.
Denote by $p_{1}$ the probability that $i^{th}$ point of the training/test set corresponds to that leaf and
by $p_{2}$ the probability that $i^{th}$ point of the training/test set corresponds to that leaf and has label $+$.
Notice that $p_{1},p_{2}$ are the same for the training and test set since we assume that training and test set are 
taken from the same distribution.
Since all the random variables introduced above are independent, we can conclude using Azuma's inequality that
\[\mathbb{P}(X_{1} + ... + X_{n} \in [n(p_{1}-\phi),n(p_{1}+\phi)]) \geq 1 - 2e^{-2n\phi^{2}}.
\]
Similarly, 
\[\mathbb{P}(Y_{1}+...+Y_{n} \in [n(p_{1}-\phi),n(p_{1}+\phi)]) \geq 1 - 2e^{-2n\phi^{2}}.
\]
Therefore, by the union bound
\[\mathbb{P}(|(X_{1}+...+X_{n})-(Y_{1}+...+Y_{n})| \leq 2\phi n) \geq 1 - 4e^{-2n\phi^{2}}.
\]
By the same analysis we can show that 
\[\mathbb{P}(X_{1}^{+} + ... + X_{n}^{+} \in [n(p_{2}-\phi),n(p_{2}+\phi)]) \geq 1 - 2e^{-2n\phi^{2}}
\]
and 
\[\mathbb{P}(Y_{1}^{+} + ... + Y_{n}^{+} \in [n(p_{2}-\phi),n(p_{2}+\phi)]) \geq 1 - 2e^{-2n\phi^{2}}.
\]
Thus we also have
\[\mathbb{P}(|(X_{1}^{+}+...+X_{n}^{+})-(Y_{1}^{+}+...+Y_{n}^{+})| \leq 2\phi n) \geq 1 - 4e^{-2n\phi^{2}}.
\]
We can conclude that the probability of the following event:
\begin{equation*}
|(X_{1}+...+X_{n})-(Y_{1}+...+Y_{n})| \leq 2\phi n \:\:\:\text{and}\:\:\: |(X_{1}^{+}+...+X_{n}^{+})-(Y_{1}^{+}+...+Y_{n}^{+})| \leq 2\phi n
\end{equation*}
is at least $1 - 8e^{-2n\phi^{2}}$.
If we now take the union bound over all $2^{h}k$ leafs of the forest then we obtain: 
$\mathbb{P}(\mathcal{E}_{\phi}) \geq 1 - 2^{h+3}ke^{-2n\phi^{2}}$.
We will now consider average weights $w_{d}$ of the test points.
The analysis for the majority voting uses $\sigma(d)$ and is completely analogous.
Assume now that all the counts for all the leaves for the test and training set differ by at most $2\phi$.
As in the analysis of the empirical error in the differentially-private setting, 
lets focus on those leaves of the forest that contain at least $\frac{n}{2^{h}K}$ of the test points each,
for a constant $K>0$.
Take a leaf $l$ with this property. Denote by $x^{1}$ the number of test points corresponding to that leaf
and with label $+$. Denote by $x^{2}$ the number of training points corresponding to that leaf and with 
label $+$. Denote by $y^{1}$ the number of all test points corresponding to that leaf and by $y^{2}$ the number
of all training points corresponding to that leaf.
We want to find an upper bound on the expression $q=|\frac{x^{1}}{y^{1}} - \frac{x^{2}}{y^{2}}|$.
Simple algebra gives us: $q \leq \frac{2\phi n (x^{1}+y^{1})}{y^{1}(y^{1}-2\phi n)}$.
If we now take $\zeta = \frac{2\phi}{\theta}$, where $\theta = \frac{1}{2^{h}K}$ then we get:
$q \leq \frac{2\zeta}{1-\zeta}$. Let us take $\zeta$ such that: $\frac{2\zeta}{1-\zeta} \leq \frac{\delta}{2}$,
where $\delta>0$ is a positive constant. Thus we want: $\zeta \leq \frac{\delta}{4+\delta}$, i.e.
$\phi \leq \frac{\delta \theta}{2(4+\delta)}$.
Take $\phi = \frac{\delta \theta}{2(4+\delta)}$. We can conclude that with probability at least $\mathbb{P}(\mathcal{E}_{\phi})$
the difference between ratios of counts in leaves containing at least $\frac{n}{\theta}$ test points for the test and training set 
is at most $\frac{\delta}{2}$.
This in particular implies that if we consider test point $d$ and a truncated bipartite graph $G^{d}$ (but this time with respect
to the test set, not training set) then weights of $d$ in $G^{d}$ and its corresponding version for the training set differ by
at most $\frac{\delta}{2}$.

We are almost done. Consider first majority voting/threshold averaging scheme.
The only changes we need to introduce in the statement of 
Theorem \ref{alfa_setting_ndp_main_1} for the empirical error
is to subtract from $p_{1}$ the probability that $\mathcal{E}_{\phi}$ does not hold to obtain a lower bound on $p_{2}$, 
add factor $\frac{1}{K}$ to the expression 
on $w$ (since we are using the truncated model) and change $\delta$ by $\frac{\delta}{2}$ in the expression on number of 
random decision trees used.
Similarly, in the statement of Theorem \ref{alfa_setting_ndp_main_2} we need to replace $\epsilon$ in the expression on 
$err_{1}$ by $\epsilon + \frac{1}{K}$ to obtain an upper bound on $err_{2}$ (again, because we are using truncation argument) and make the same
change in the number of decision trees as the one above. To obtain a lower bound on $p_{2}$
it suffices to subtract the probability that $\mathcal{E}_{\phi}$ does not hold. 
Let us focus now on Theorem \ref{alfa_setting_dp_main_1}.
Again we need to add extra factor $\frac{1}{K}$ to the expression on $w$ and subtract probability that $\mathcal{E}_{\phi}$ does 
not hold to obtain a lower bound on $p_{2}$.

Now lets consider probabilistic averaging scheme. Take the statement of Theorem \ref{beta_setting_ndp_main} first.
We make similar correction to those mentioned earlier to get a lower bound on $p_{2}$. Besides in the upper bound
on $err_{1}$ we need to replace $\epsilon$ by $\epsilon+\frac{1}{K}$ to obtain an upper bound on $err_{2}$.
In Theorem \ref{beta_setting_dp_main_1} we need to add one extra term $\frac{1}{K}$ in the upper bound on $err_{1}$
to obtain an upper bound on $err_{2}$ and again modify $p_{1}$ in the same way as before to obtain a lower bound on $p_{2}$.
\end{proof}

\section{Experiments on the remaining datasets}

In this section we enclose the experimental results we obtained for all the remaining benchmark datasets. The plots have similar form to the ones shown in the main body of the paper. 

\begin{figure}[h]
\center
\begin{tabular}{cc}
\hspace{-0.1in}\includegraphics[width = 1.54in]{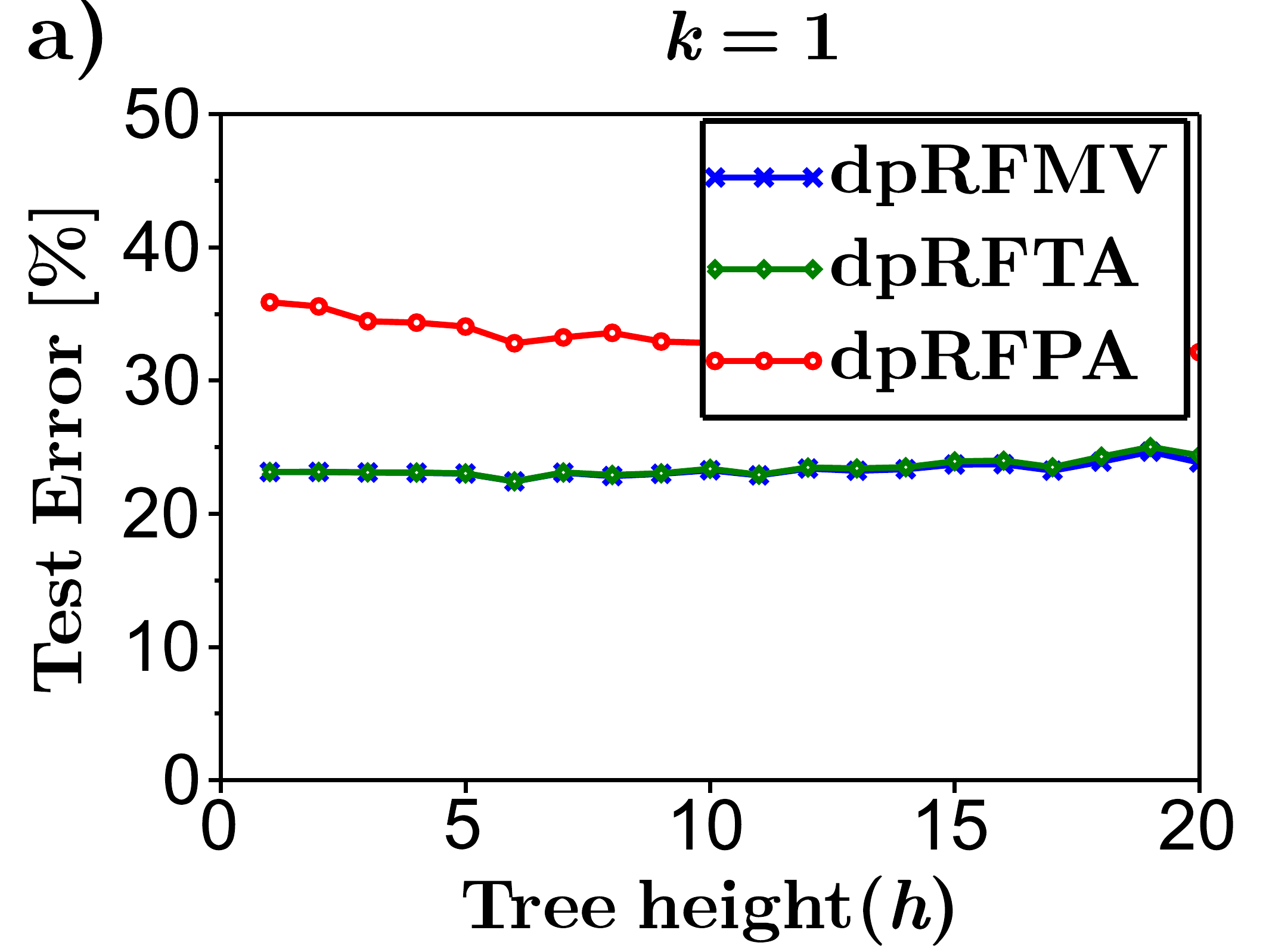} 
\hspace{-0.04in}\includegraphics[width = 1.54in]{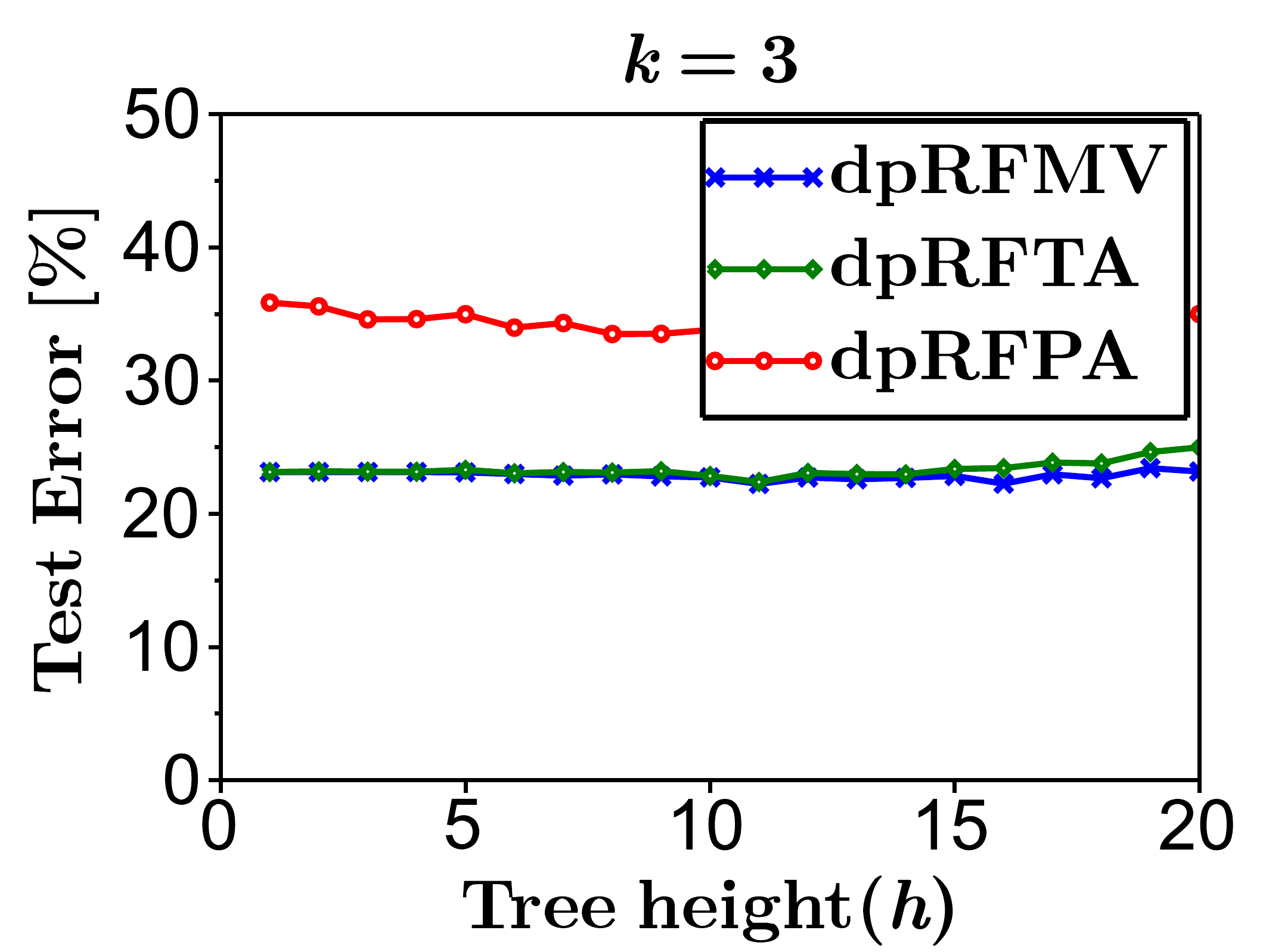} 
\hspace{-0.035in}\includegraphics[width = 1.54in]{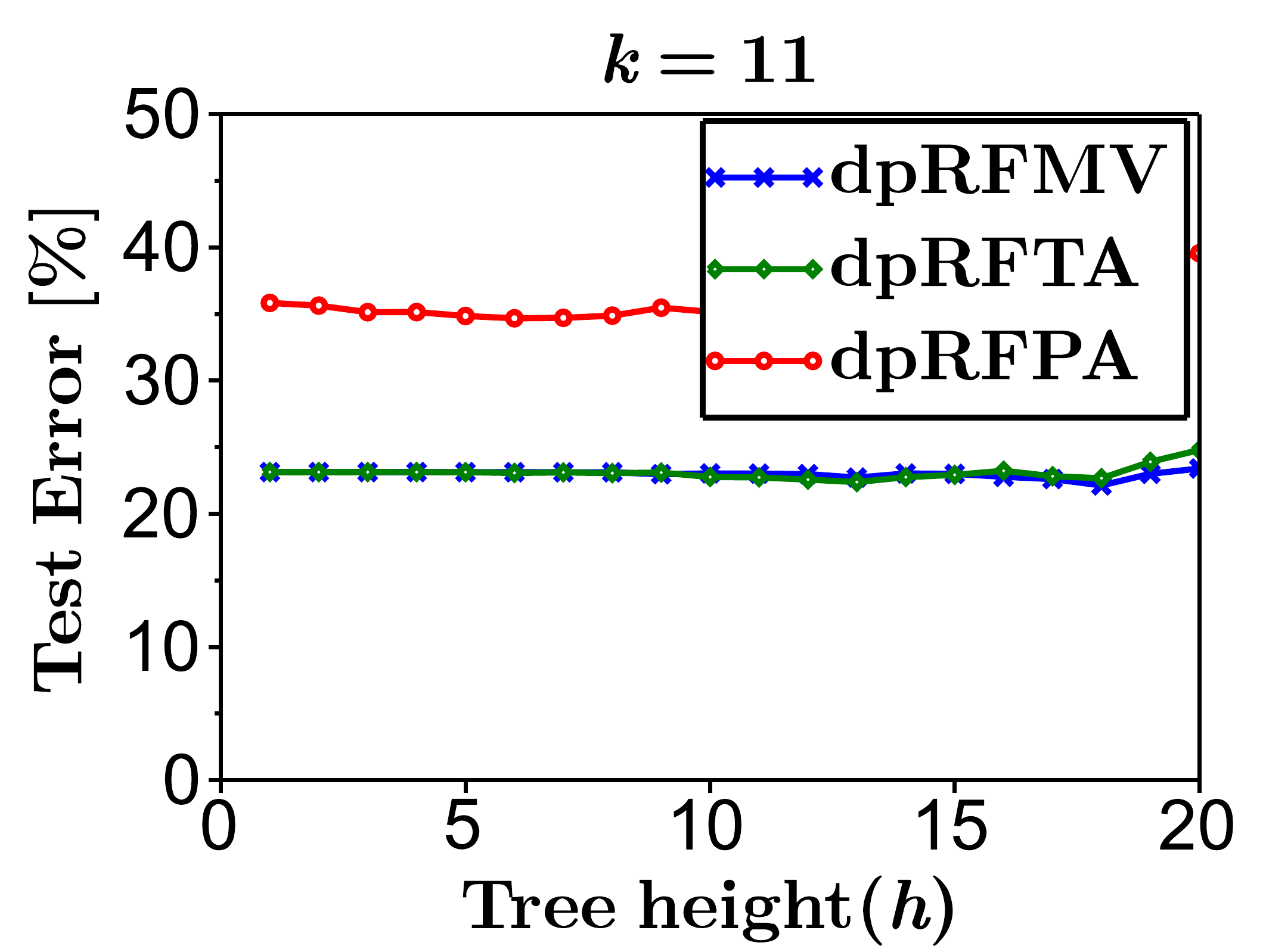} 
\hspace{-0.04in}\includegraphics[width = 1.54in]{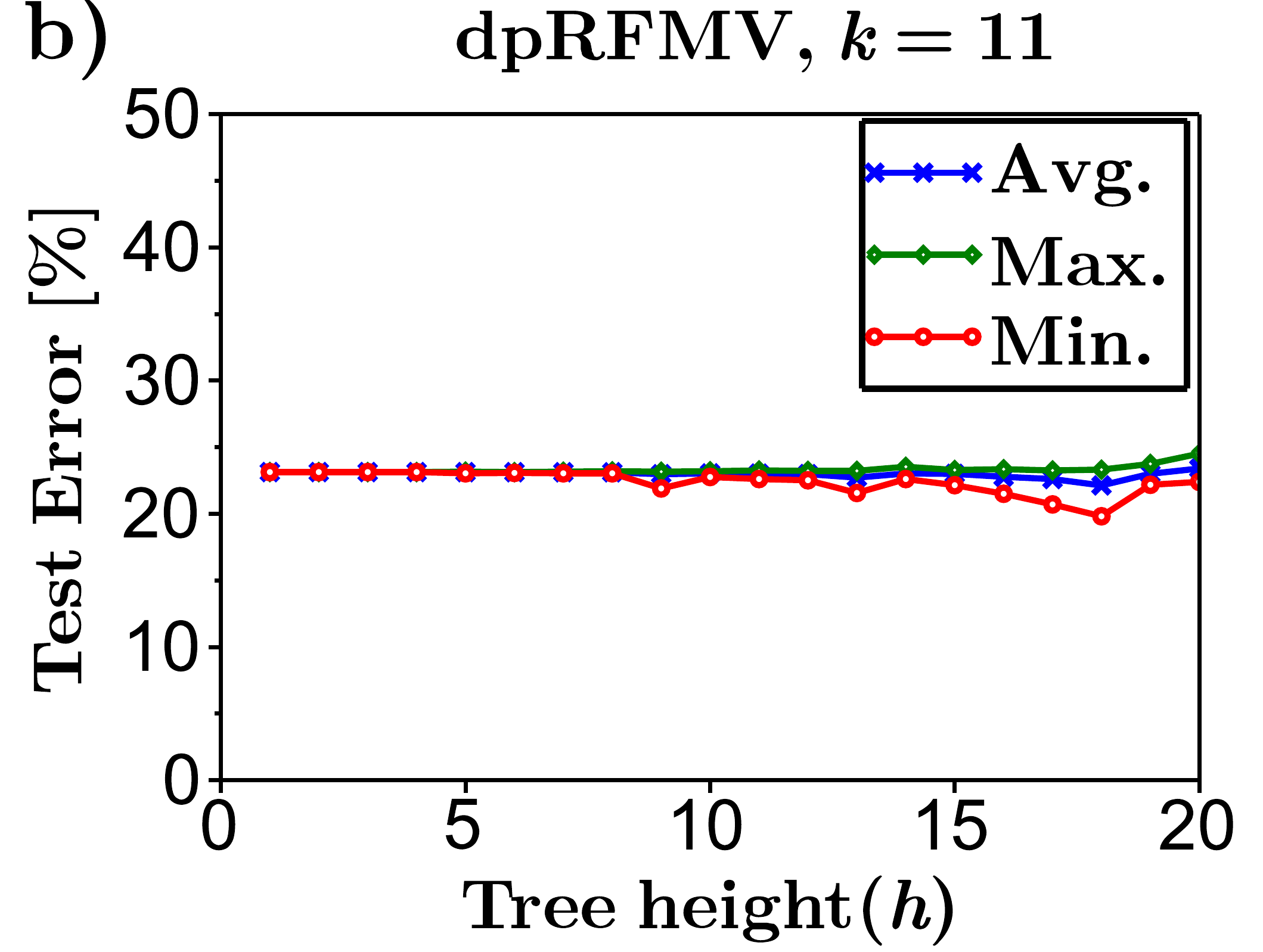}\\  
\hspace{-0.1in}\includegraphics[width = 1.54in]{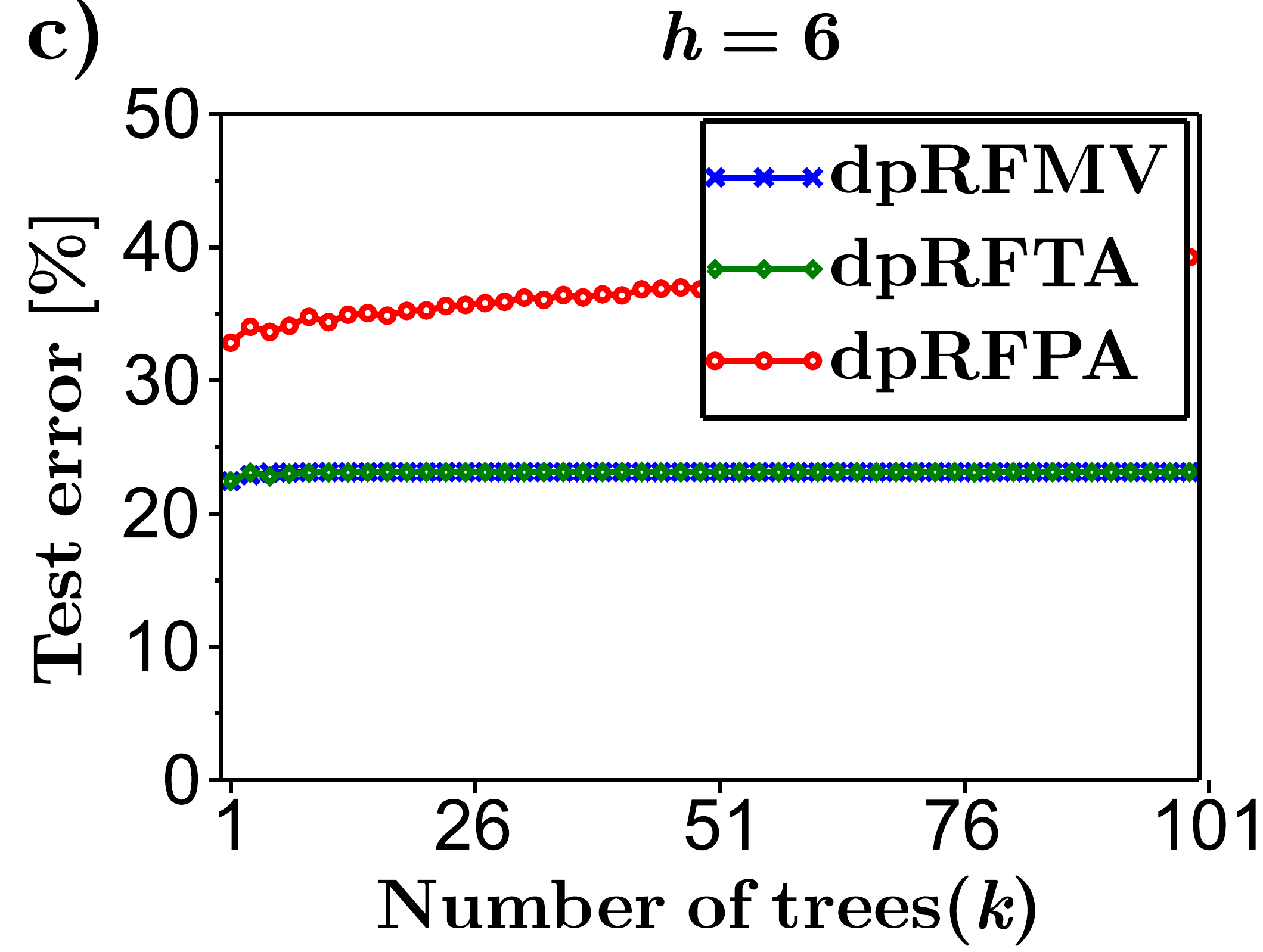} 
\hspace{-0.04in}\includegraphics[width = 1.54in]{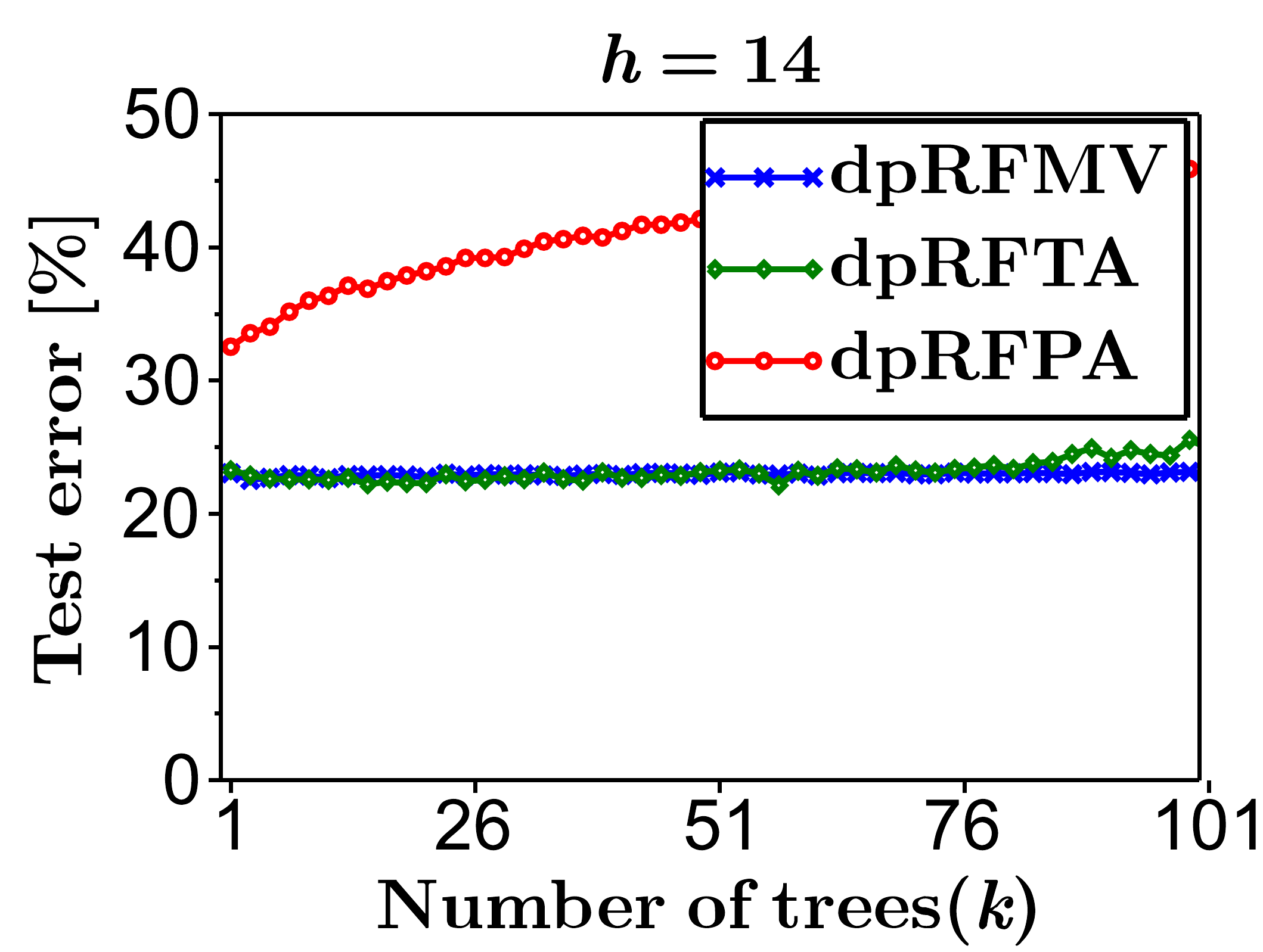} 
\hspace{-0.035in}\includegraphics[width = 1.54in]{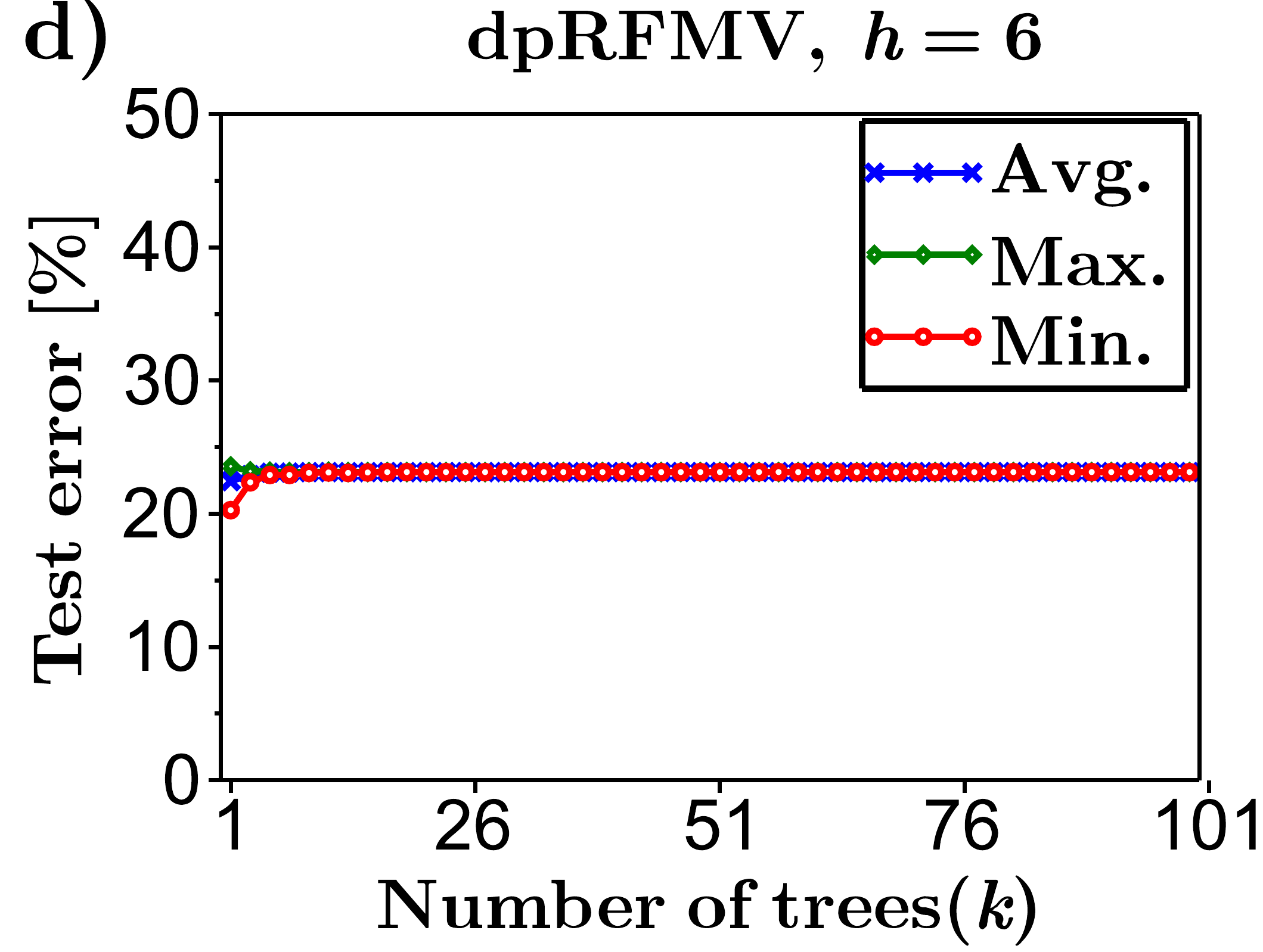} 
\hspace{-0.04in}\includegraphics[width = 1.54in]{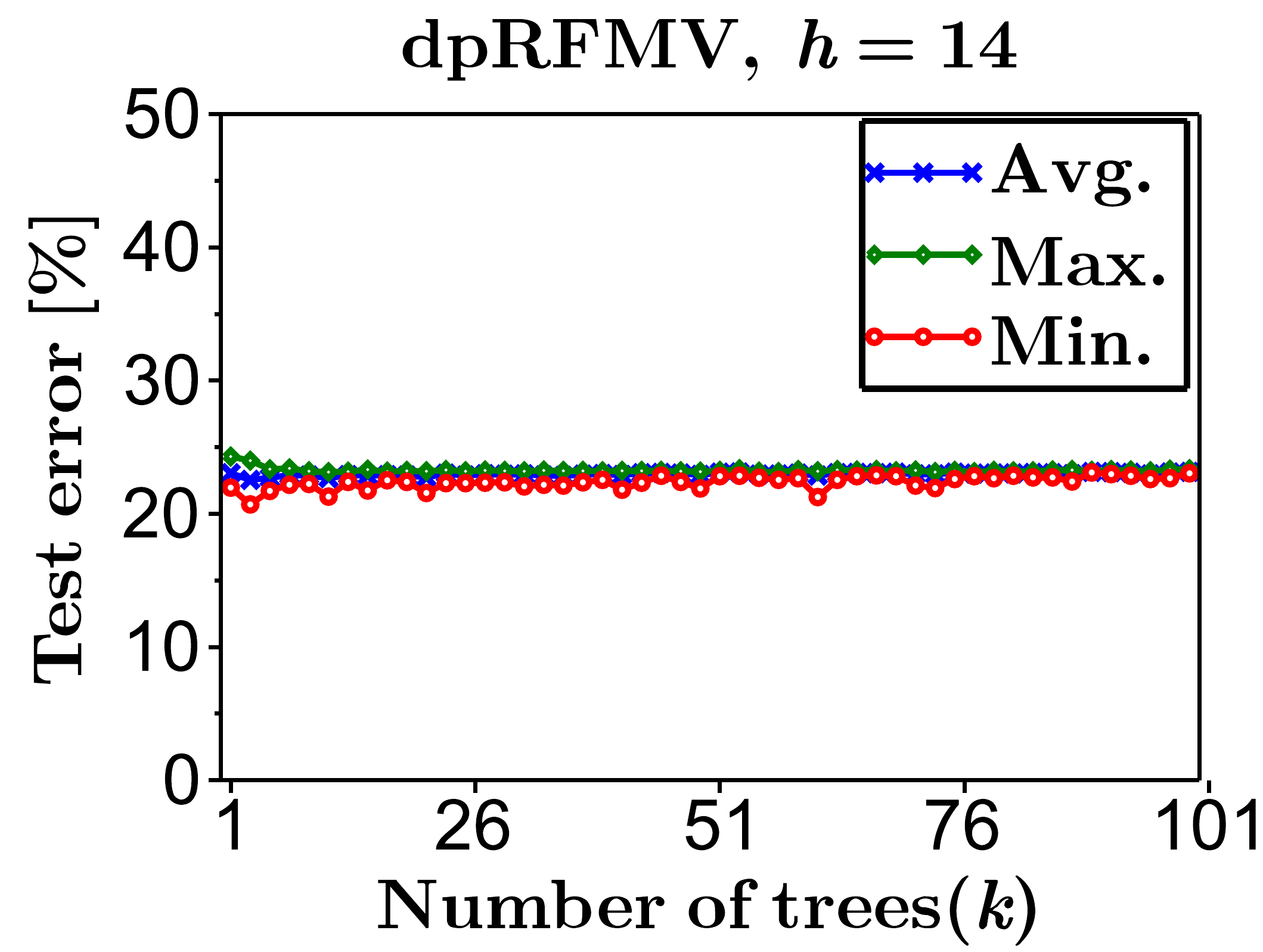}
\end{tabular}
\caption{\textit{adult} dataset. Comparison of dpRFMV, dpRFTA and dpRFPA. $\eta = 1000/n_{tr}$. Test error resp. vs. \textbf{a)} $h$ across various settings of $k$ and vs. \textbf{c)} $k$ across various settings of $h$; Minimal, average and maximal test error resp. vs. $h$ (\textbf{b)}) and vs. $k$ (\textbf{d)}) for dpRFMV.}
\label{fig:adult_one}
\end{figure}

\begin{figure}[h]
  \center
\includegraphics[width = 2.8in,height = 1.25in]{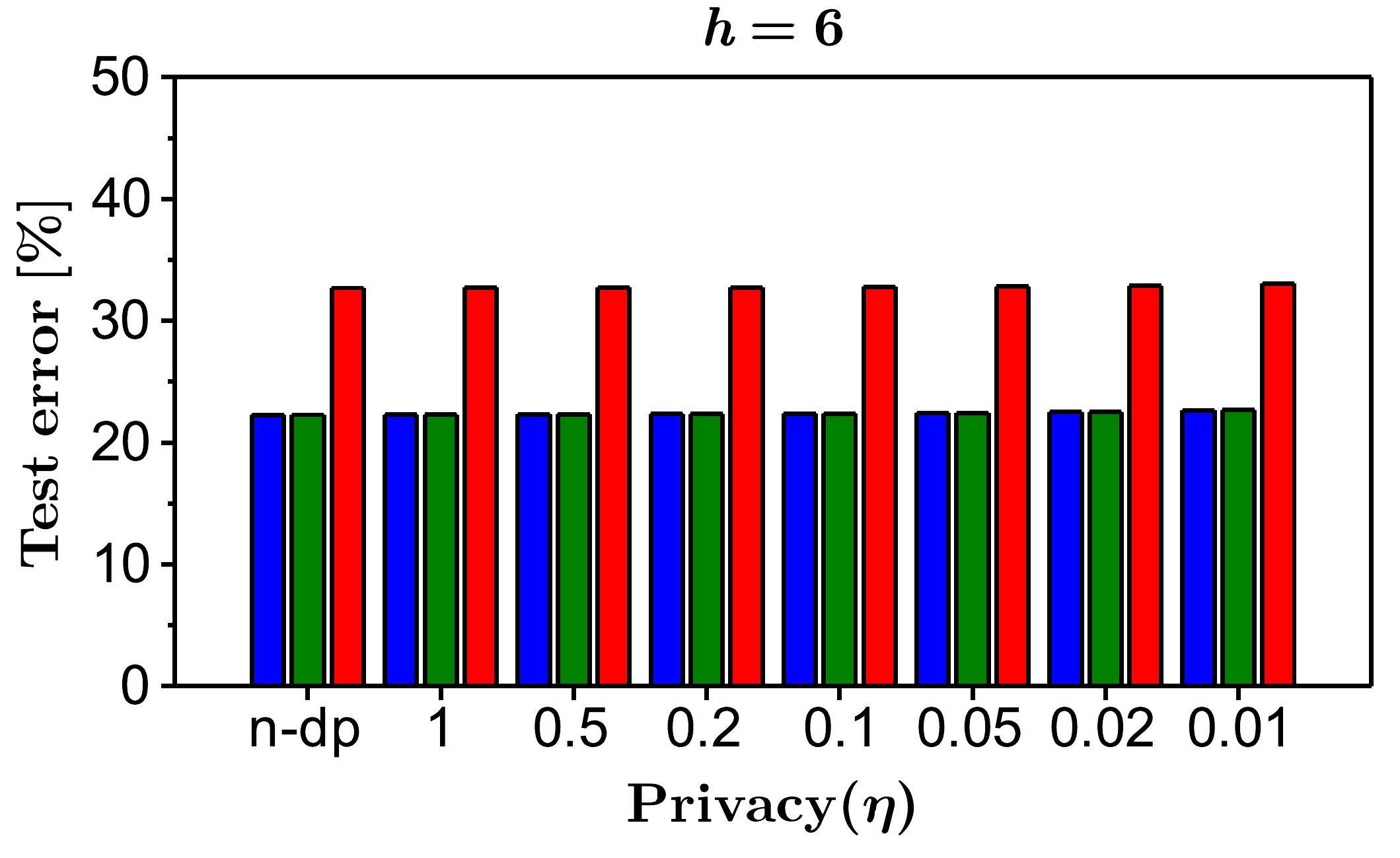} 
\includegraphics[width = 2.8in,height = 1.25in]{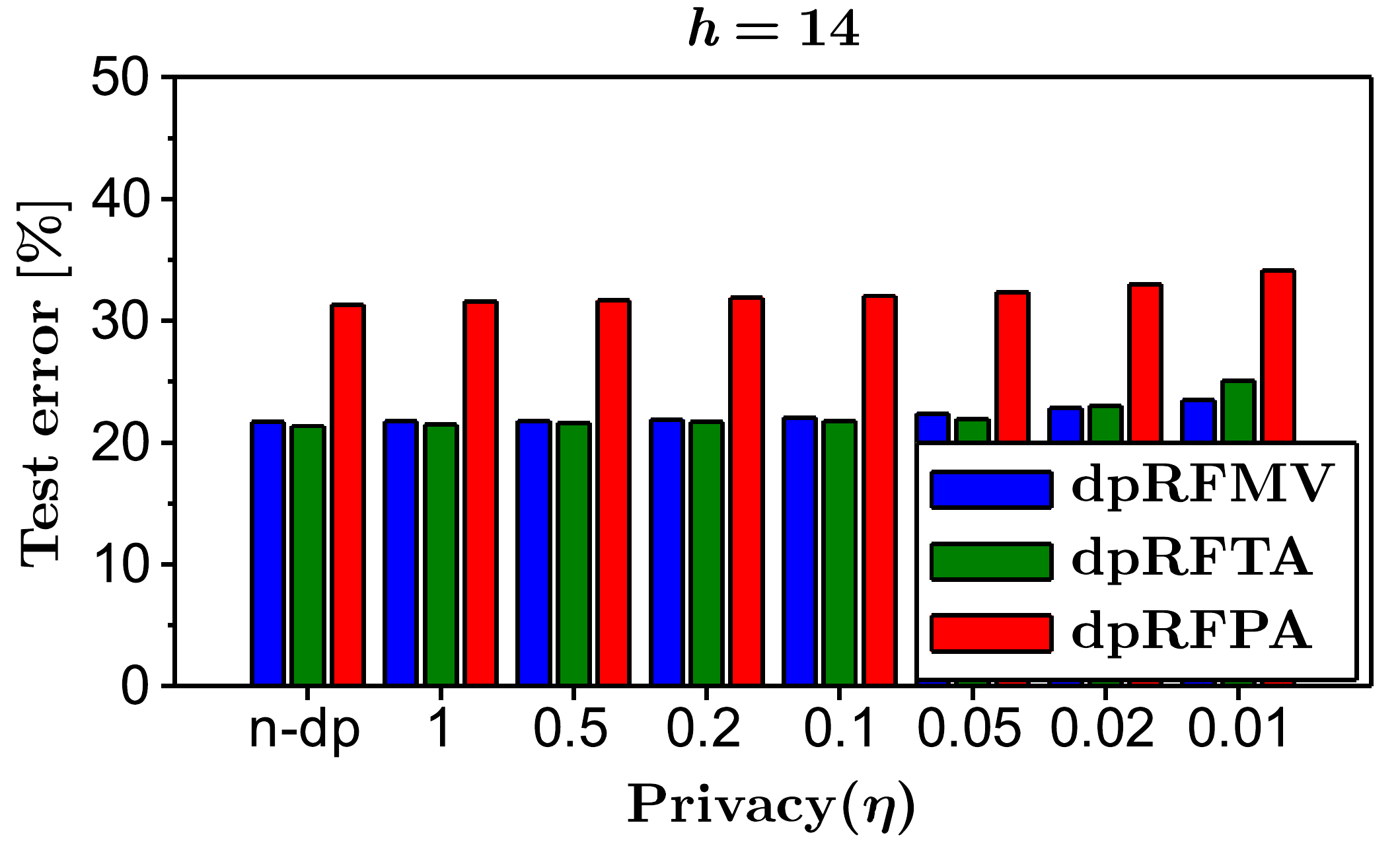}\\
\includegraphics[width = 1.6in]{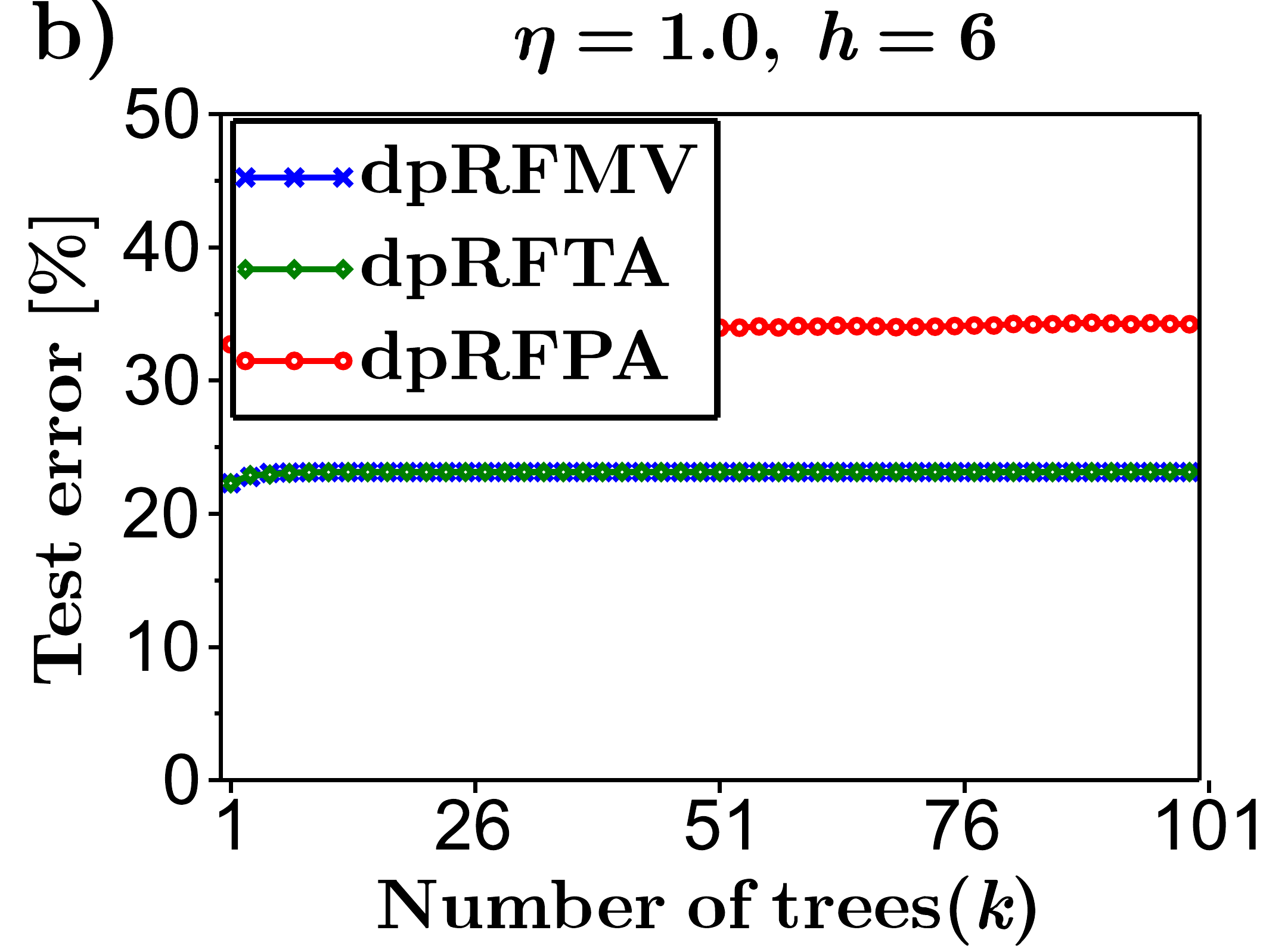} 
\hspace{-0.03in}\includegraphics[width = 1.6in]{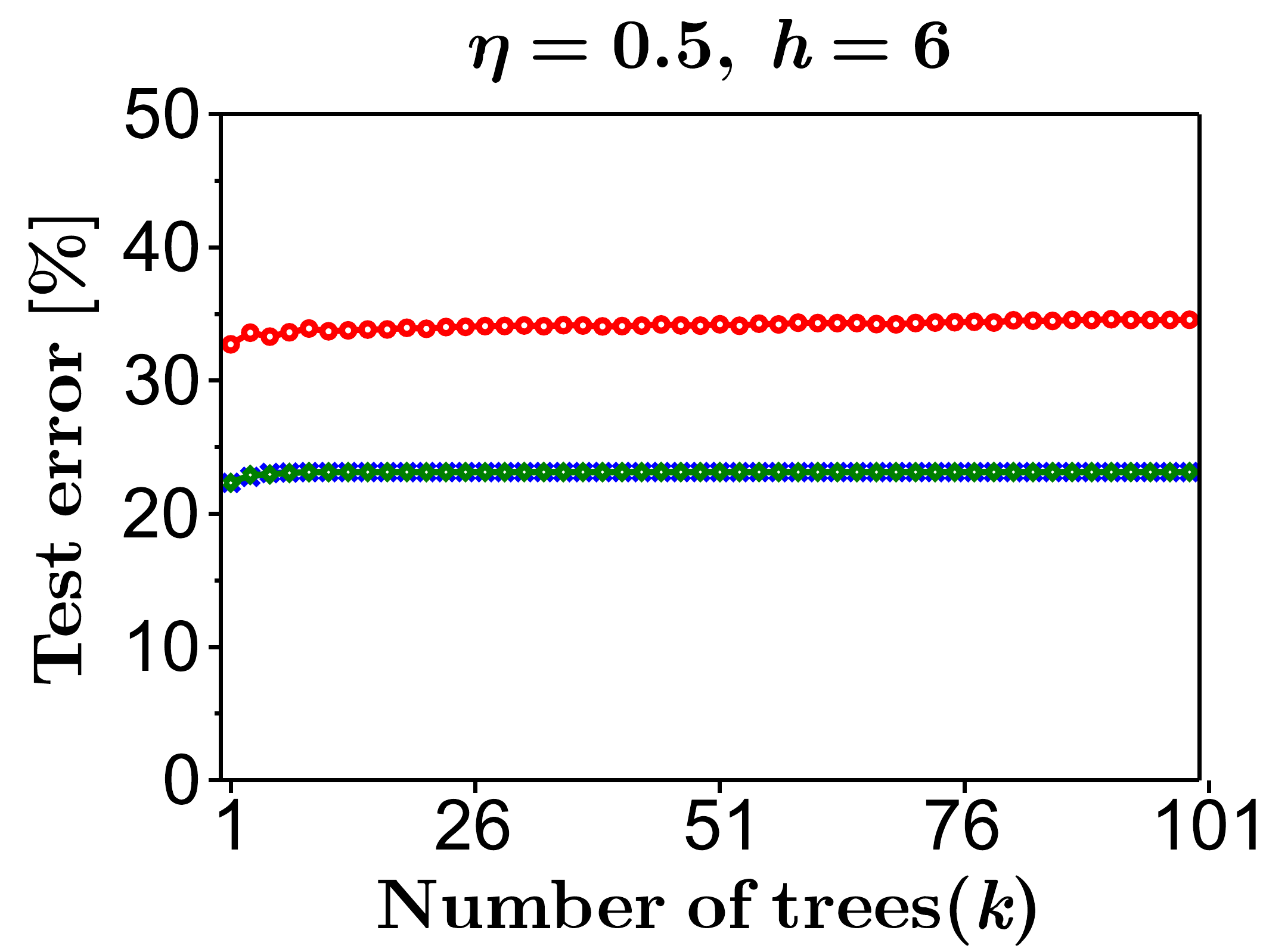} 
\hspace{-0.03in}\includegraphics[width = 1.6in]{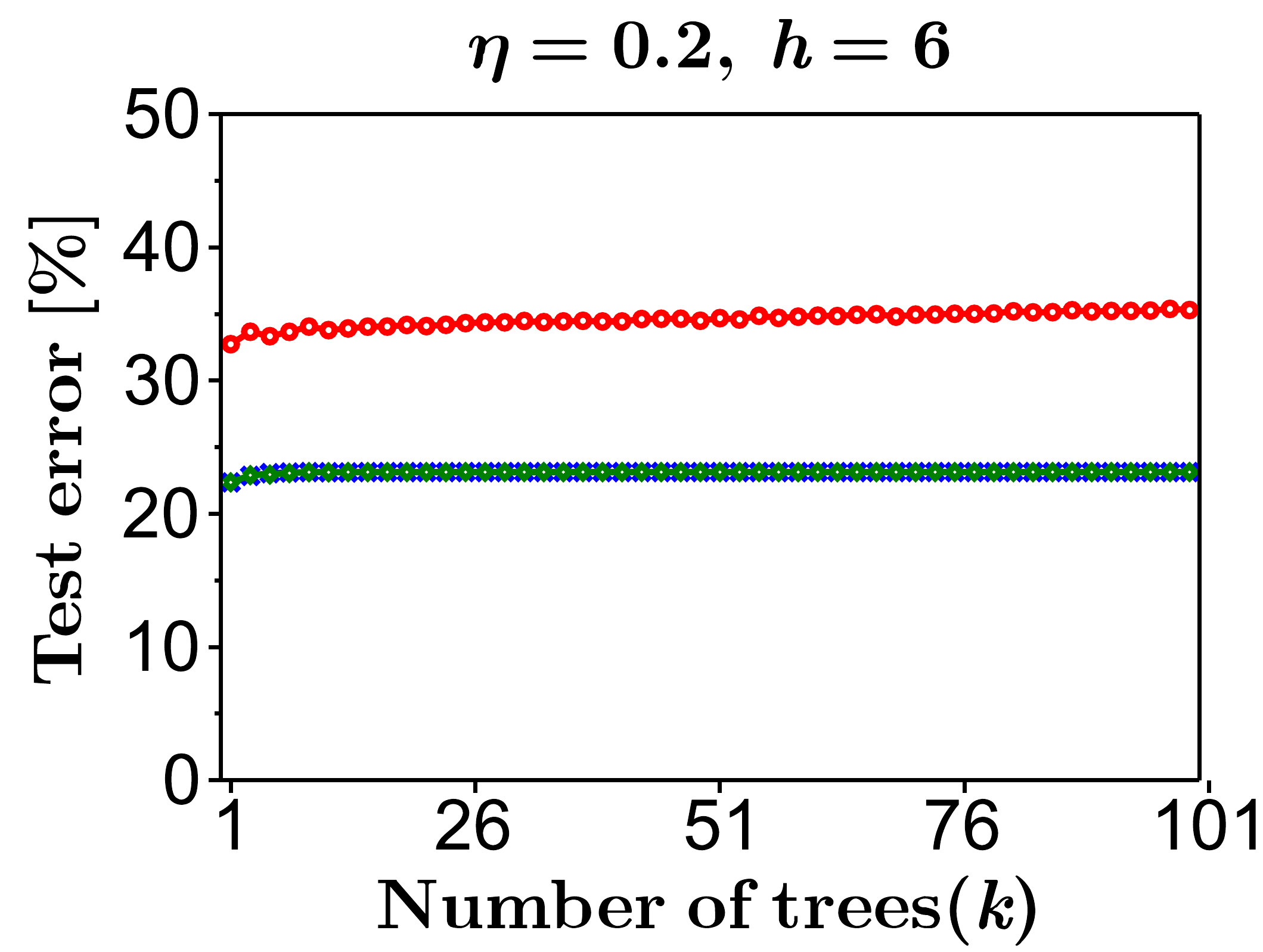} 
\hspace{-0.03in}\includegraphics[width = 1.6in]{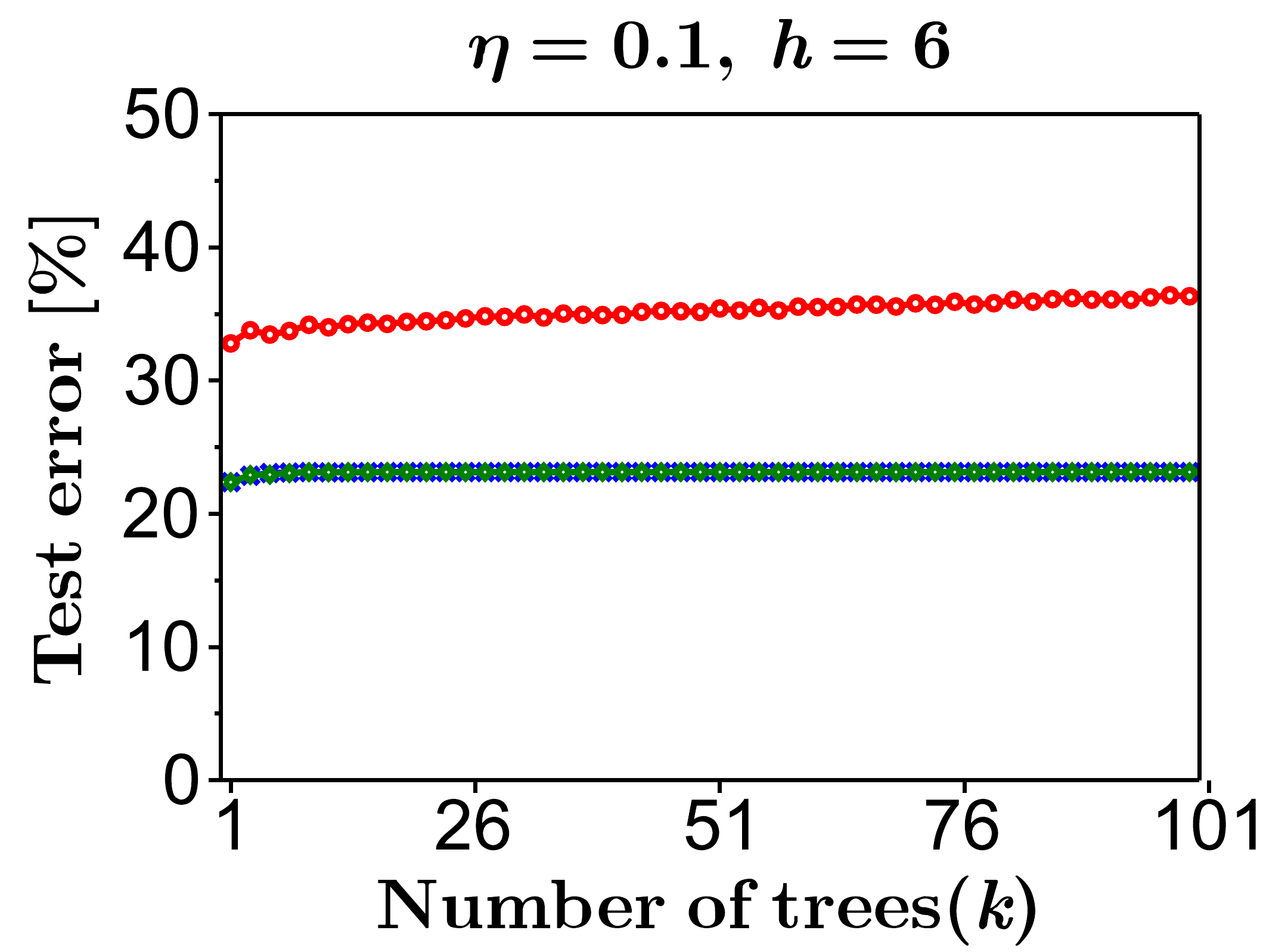}
\caption{\textit{adult} dataset. Comparison of dpRFMV, dpRFTA and dpRFPA. \textbf{a)} Test error vs. $\eta$ for two settings of $h$. \textbf{b)} Test error vs. $k$ for fixed $h$ and across different settings of $\eta$.}
\label{fig:adult_two}
\end{figure}

\begin{figure}[h]
\center
\begin{tabular}{cc}
\hspace{-0.1in}\includegraphics[width = 1.54in]{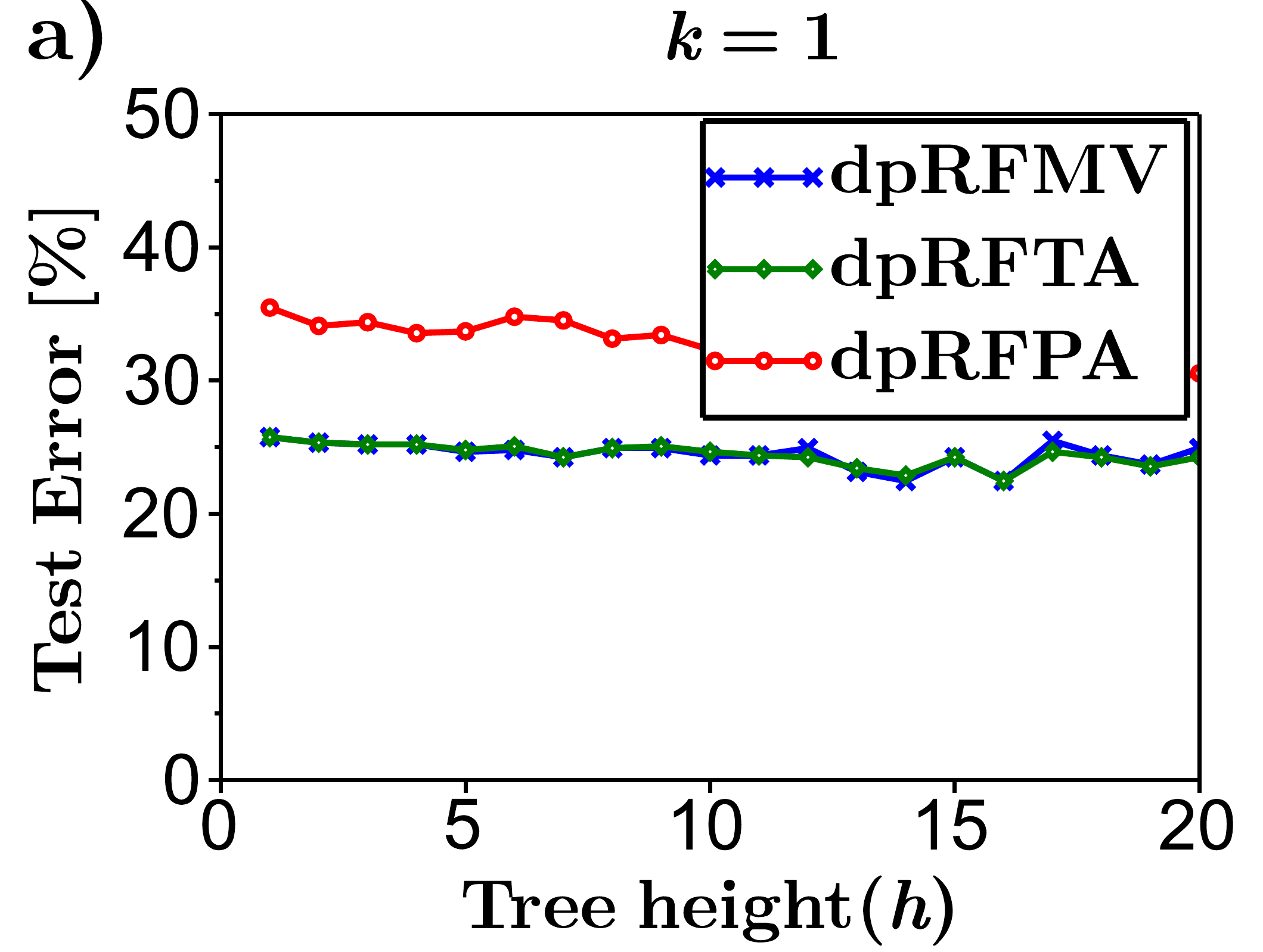} 
\hspace{-0.04in}\includegraphics[width = 1.54in]{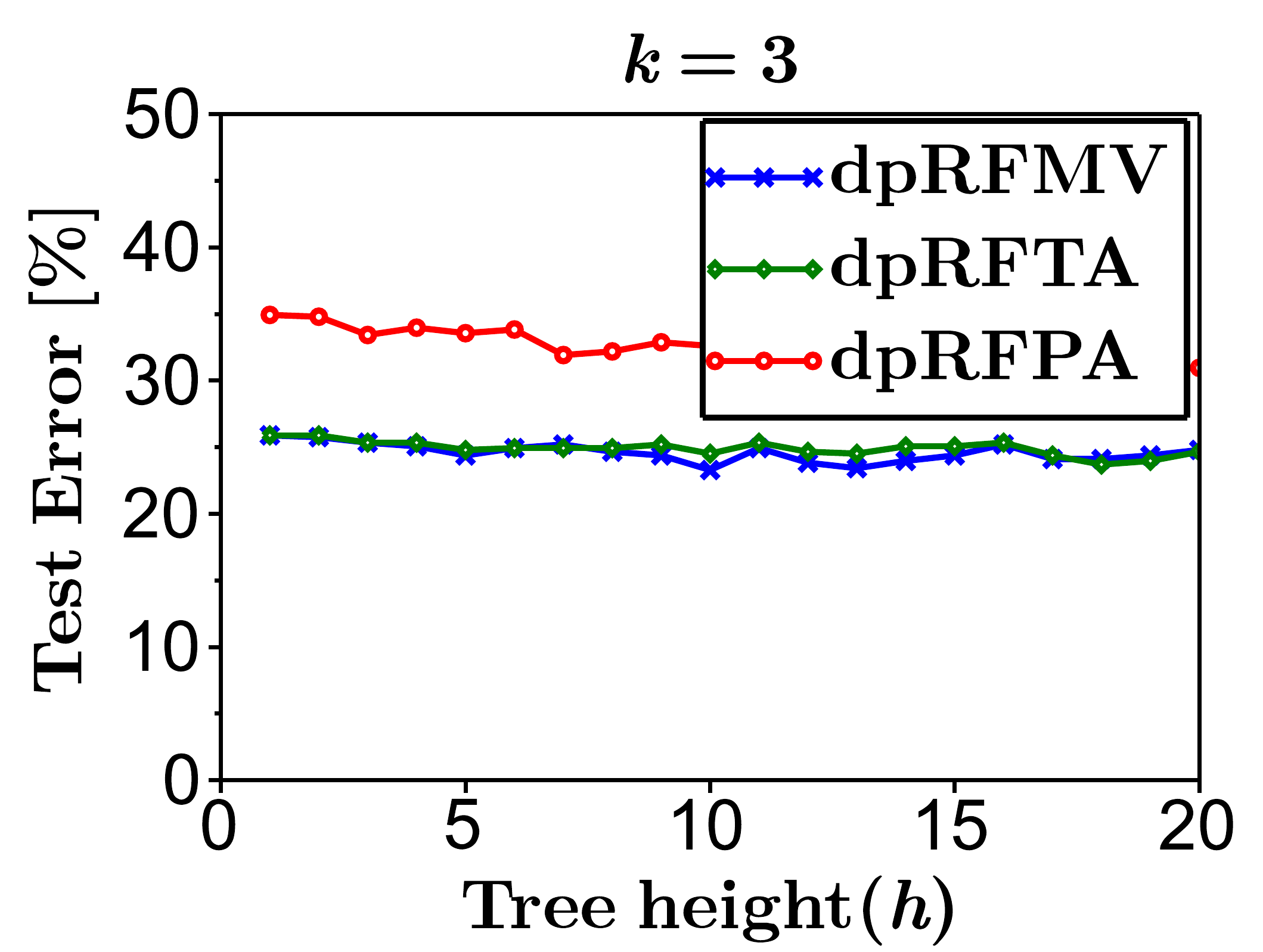} 
\hspace{-0.035in}\includegraphics[width = 1.54in]{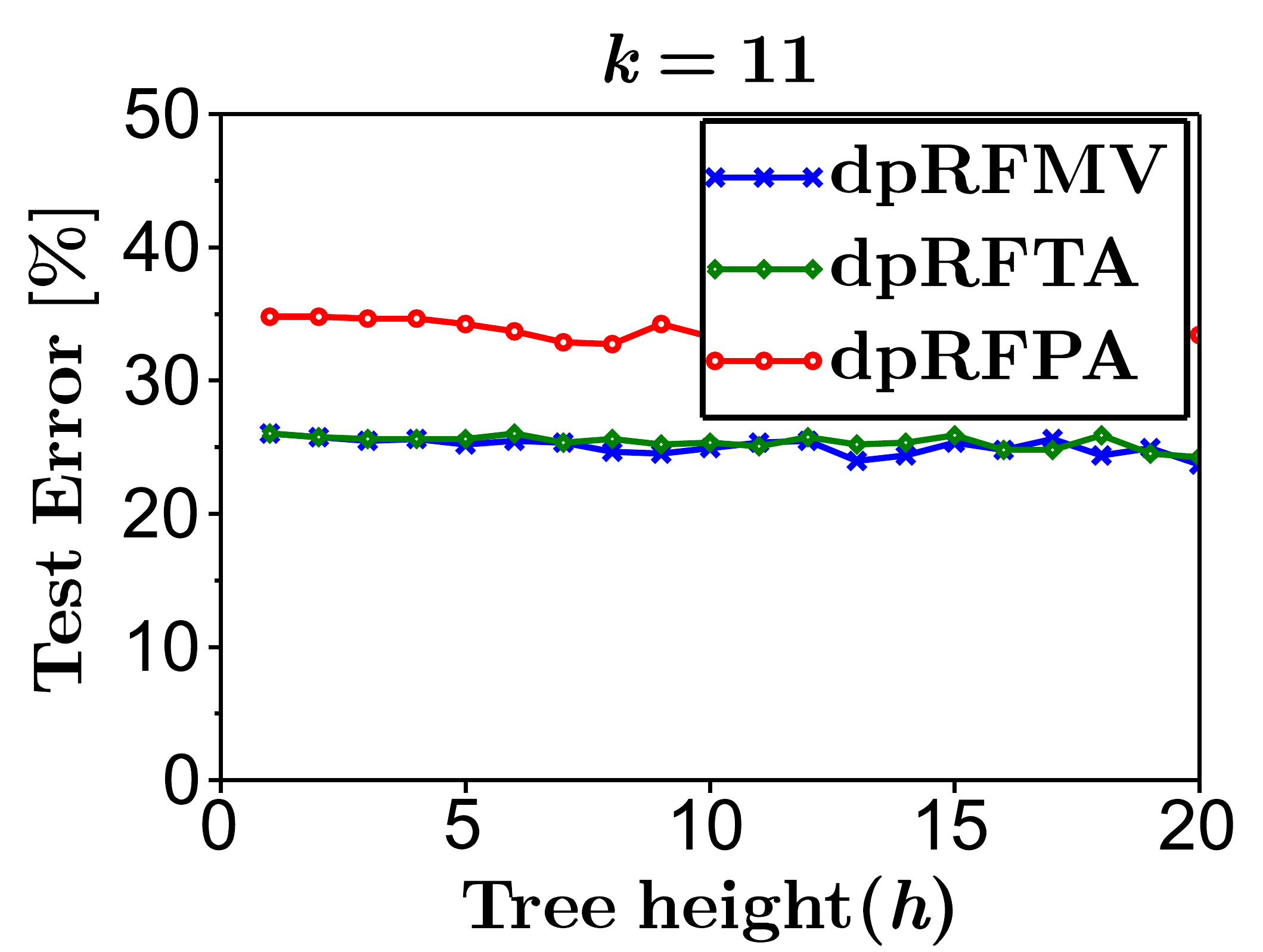} 
\hspace{-0.04in}\includegraphics[width = 1.54in]{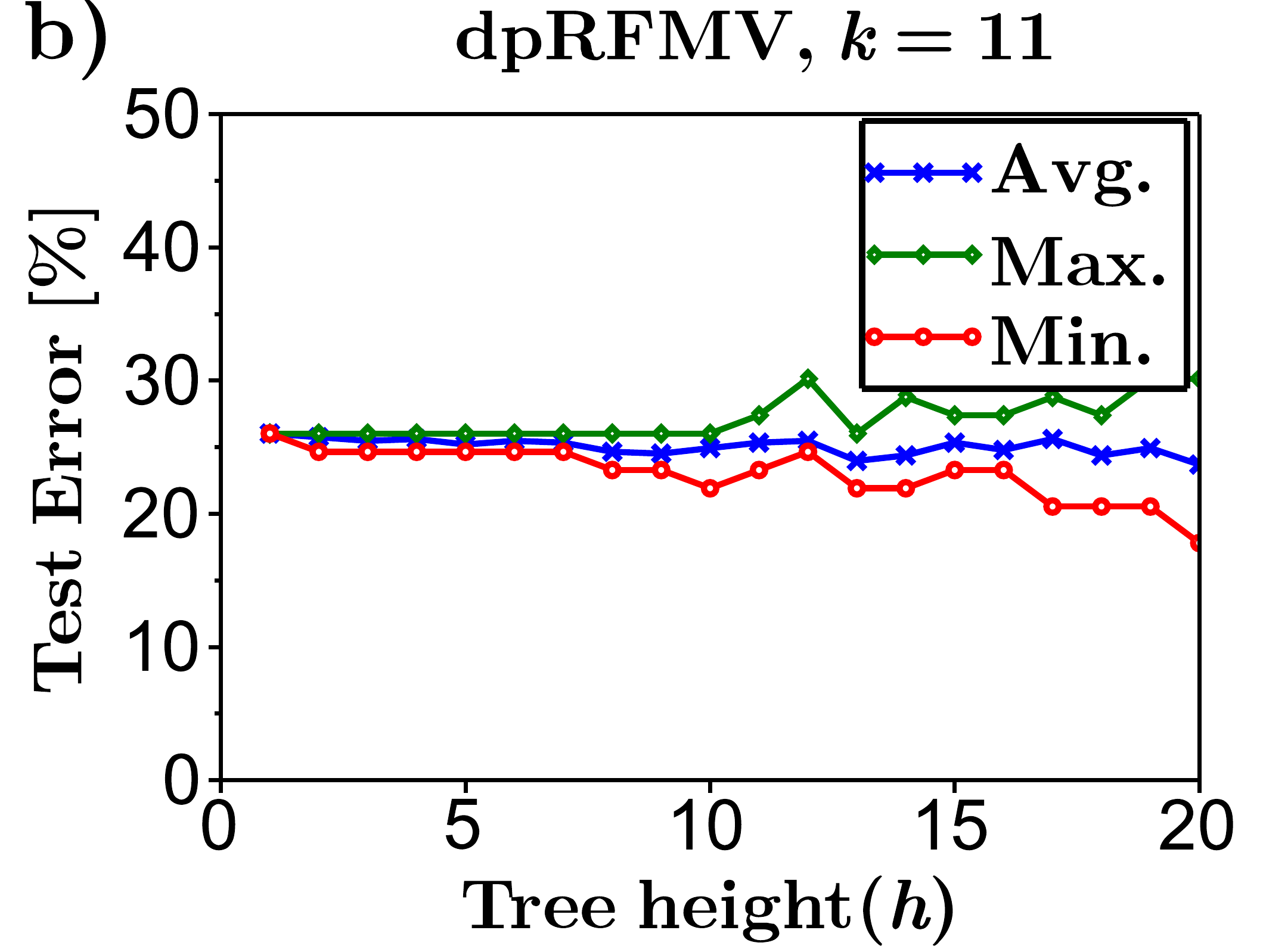}\\  
\hspace{-0.1in}\includegraphics[width = 1.54in]{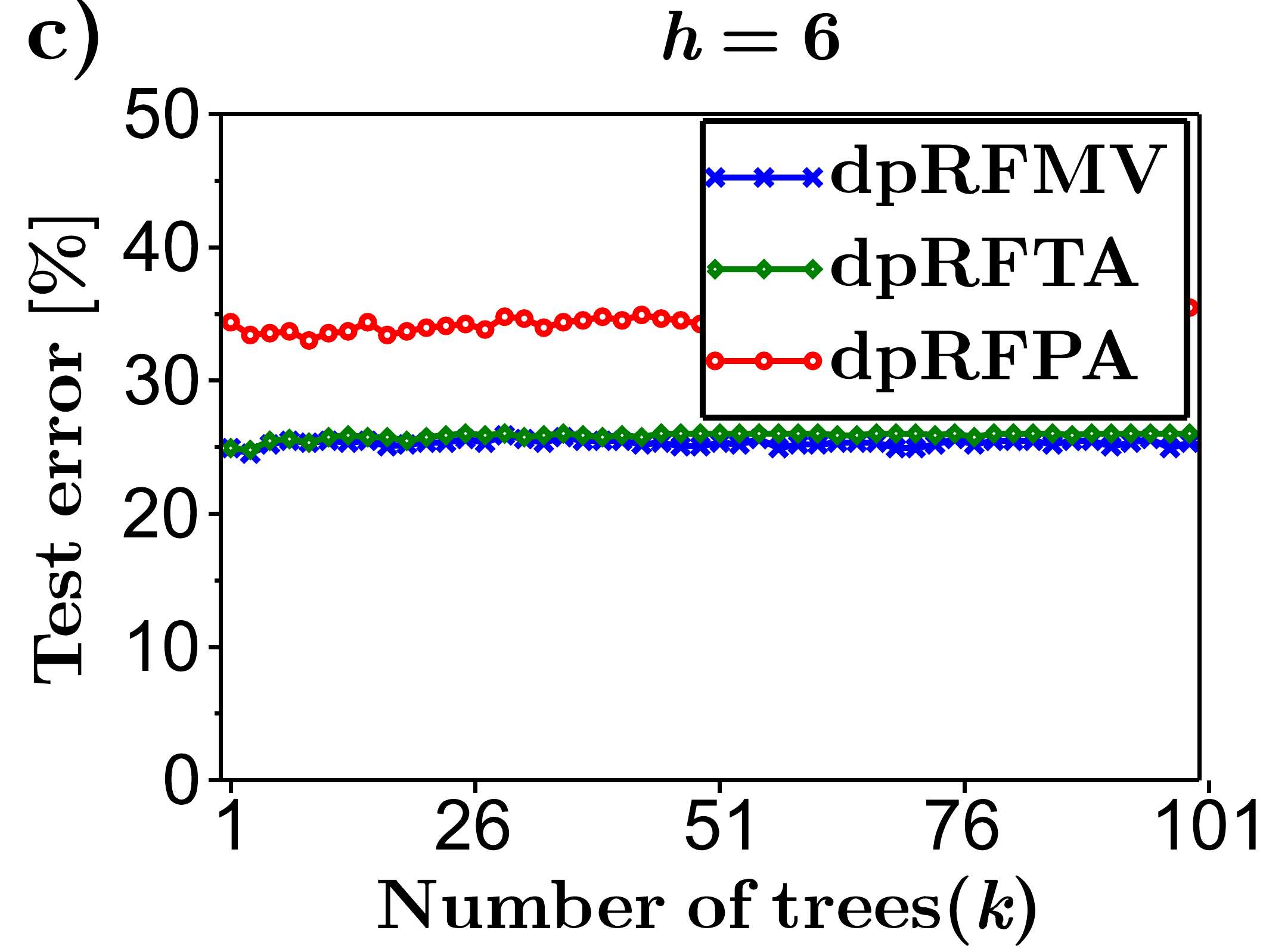} 
\hspace{-0.04in}\includegraphics[width = 1.54in]{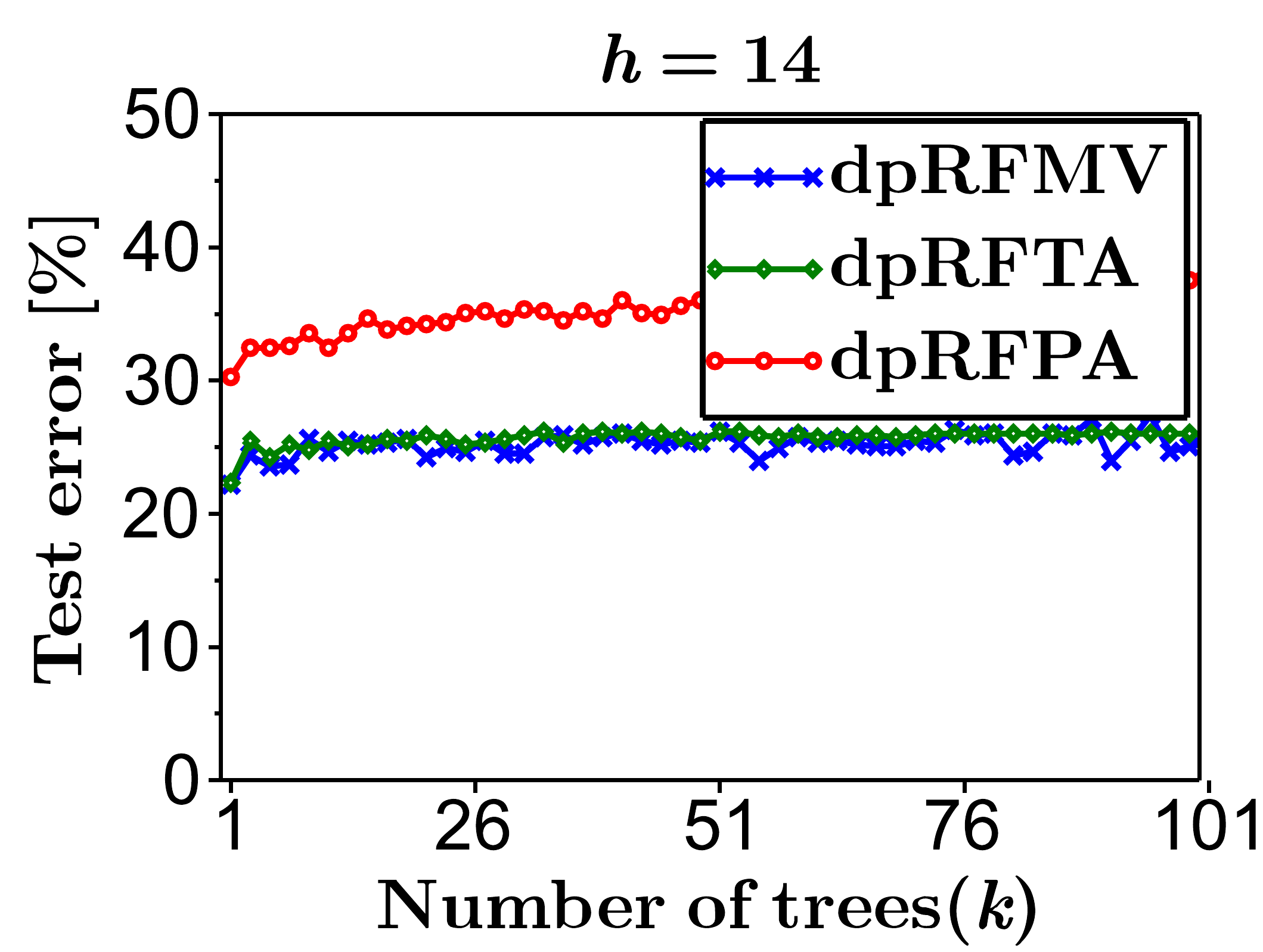} 
\hspace{-0.035in}\includegraphics[width = 1.54in]{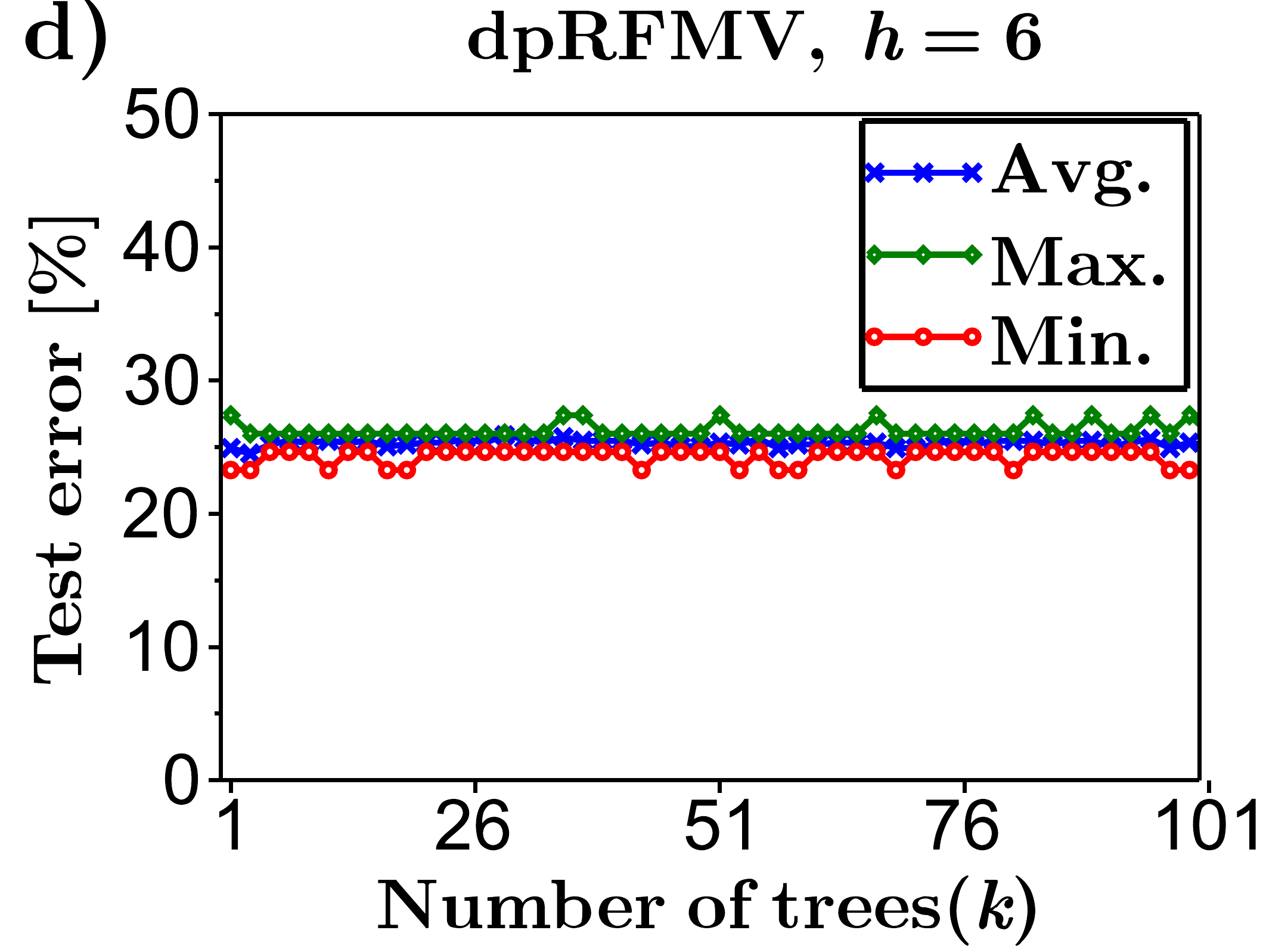} 
\hspace{-0.04in}\includegraphics[width = 1.54in]{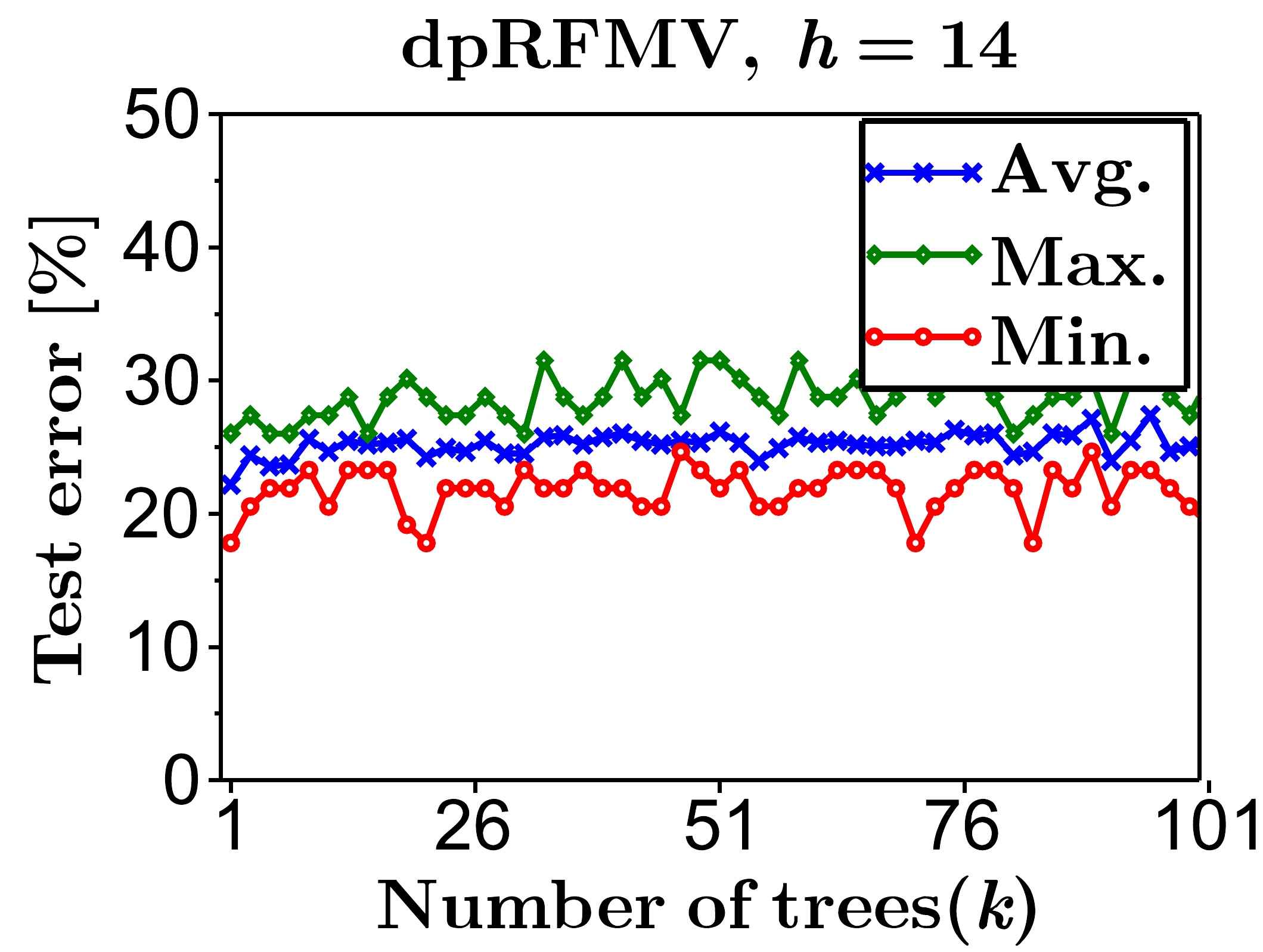}
\end{tabular}
\caption{\textit{BTSC} dataset. Comparison of dpRFMV, dpRFTA and dpRFPA. $\eta = 1000/n_{tr}$. Test error resp. vs. \textbf{a)} $h$ across various settings of $k$ and vs. \textbf{c)} $k$ across various settings of $h$; Minimal, average and maximal test error resp. vs. $h$ (\textbf{b)}) and vs. $k$ (\textbf{d)}) for dpRFMV.}
\label{fig:btsc_one}
\vspace{-0.04in}
\end{figure}

\begin{figure}[h]
  \center
\includegraphics[width = 2.8in,height = 1.25in]{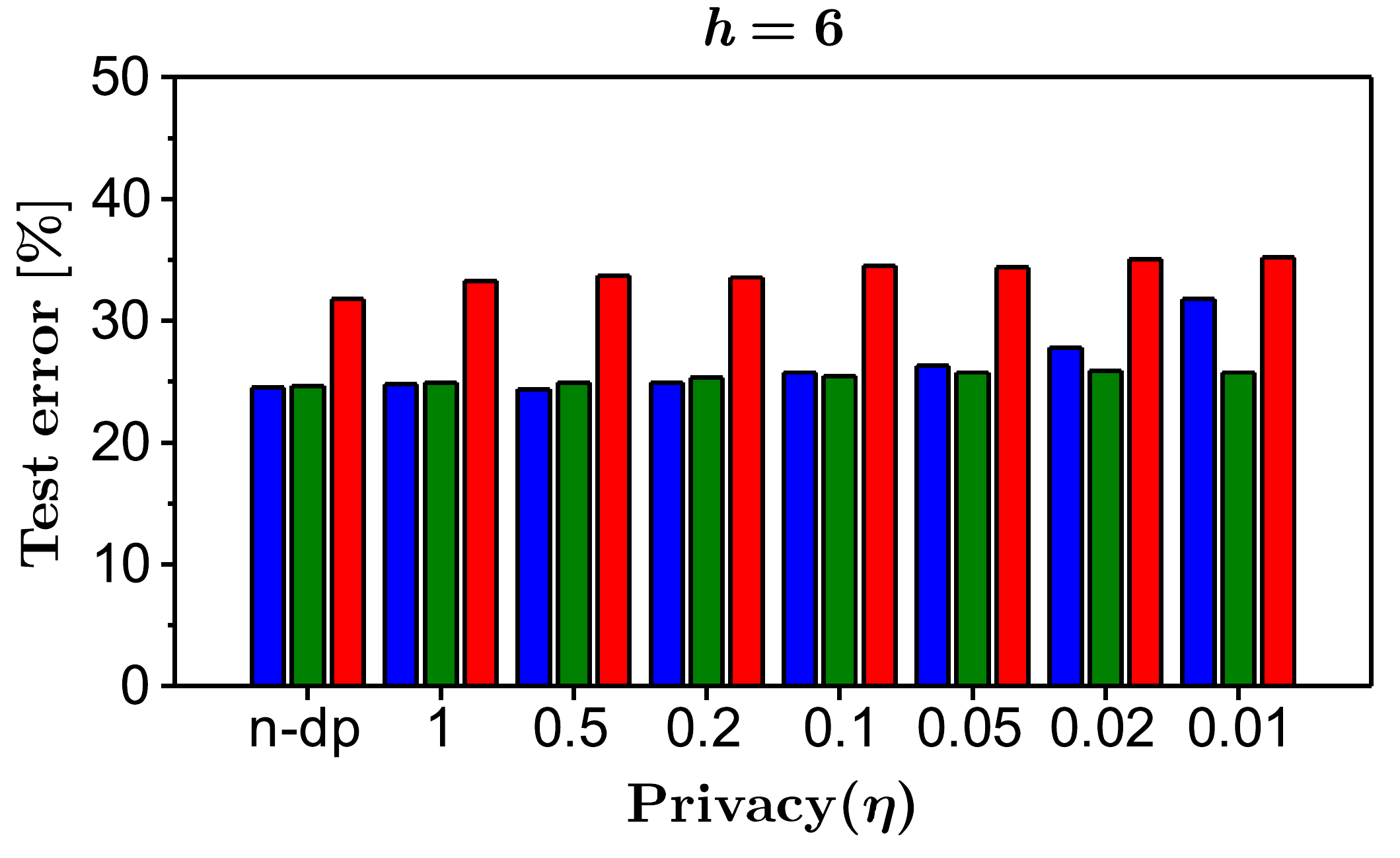} 
\includegraphics[width = 2.8in,height = 1.25in]{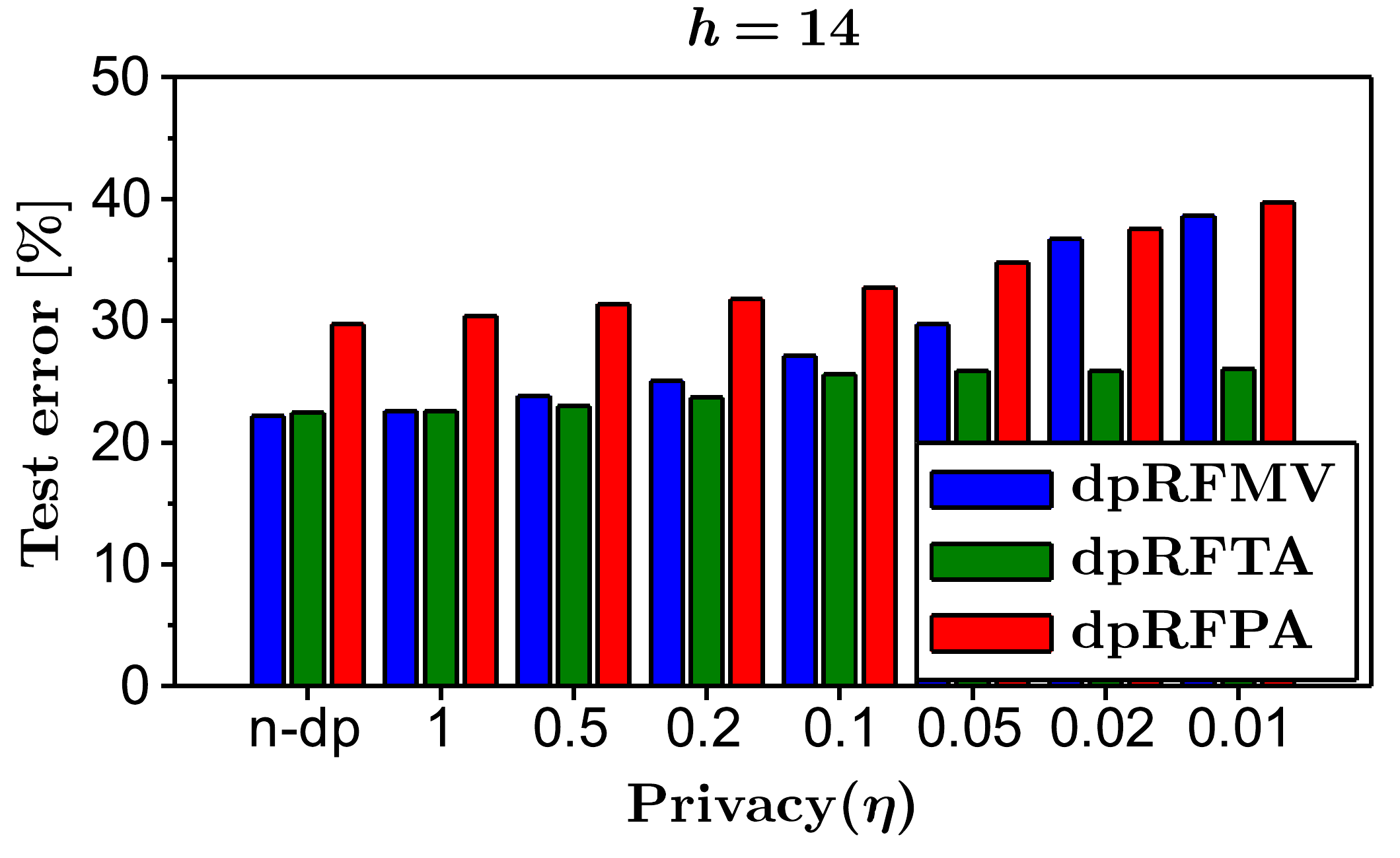}\\
\includegraphics[width = 1.6in]{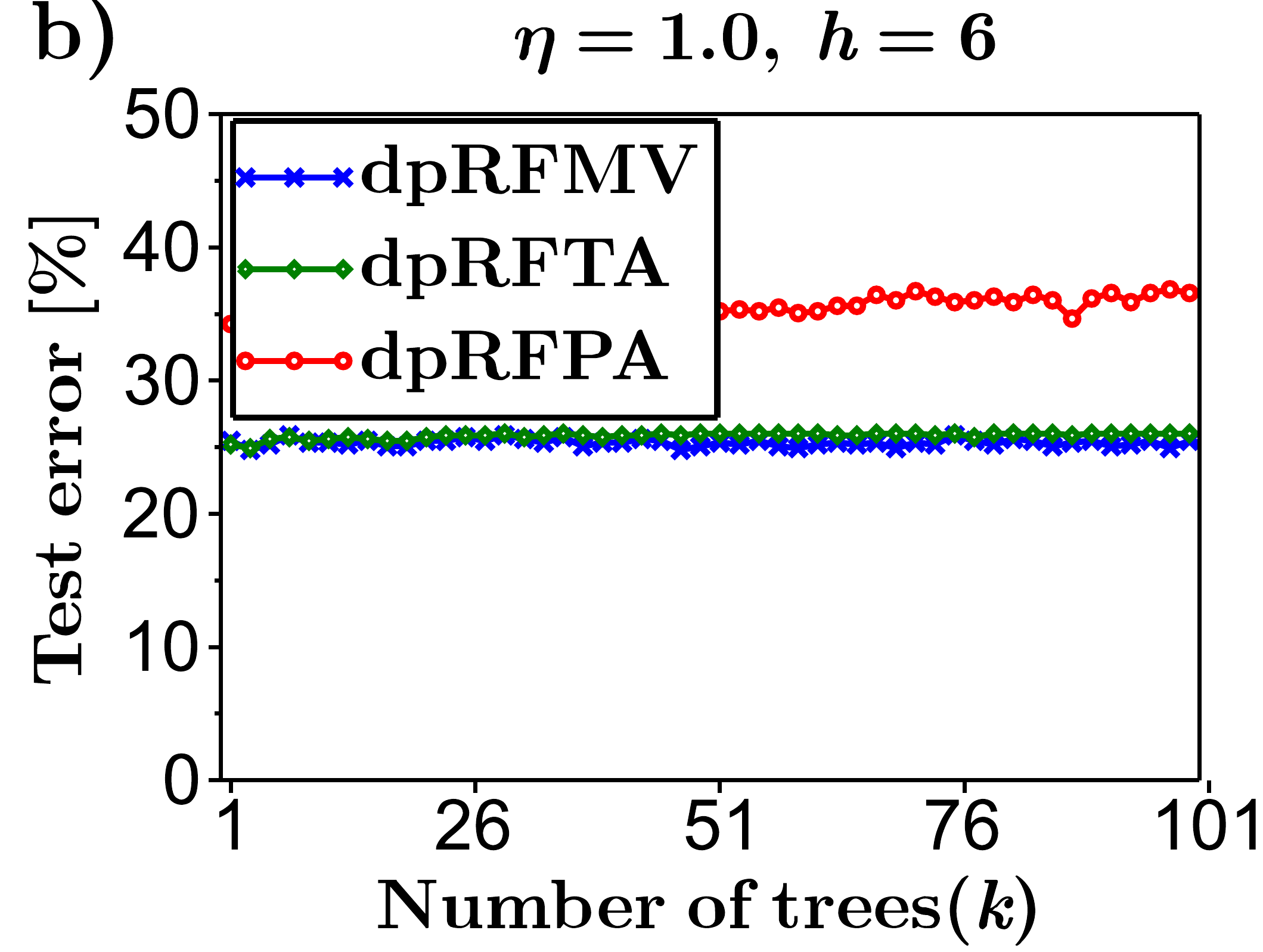} 
\hspace{-0.03in}\includegraphics[width = 1.6in]{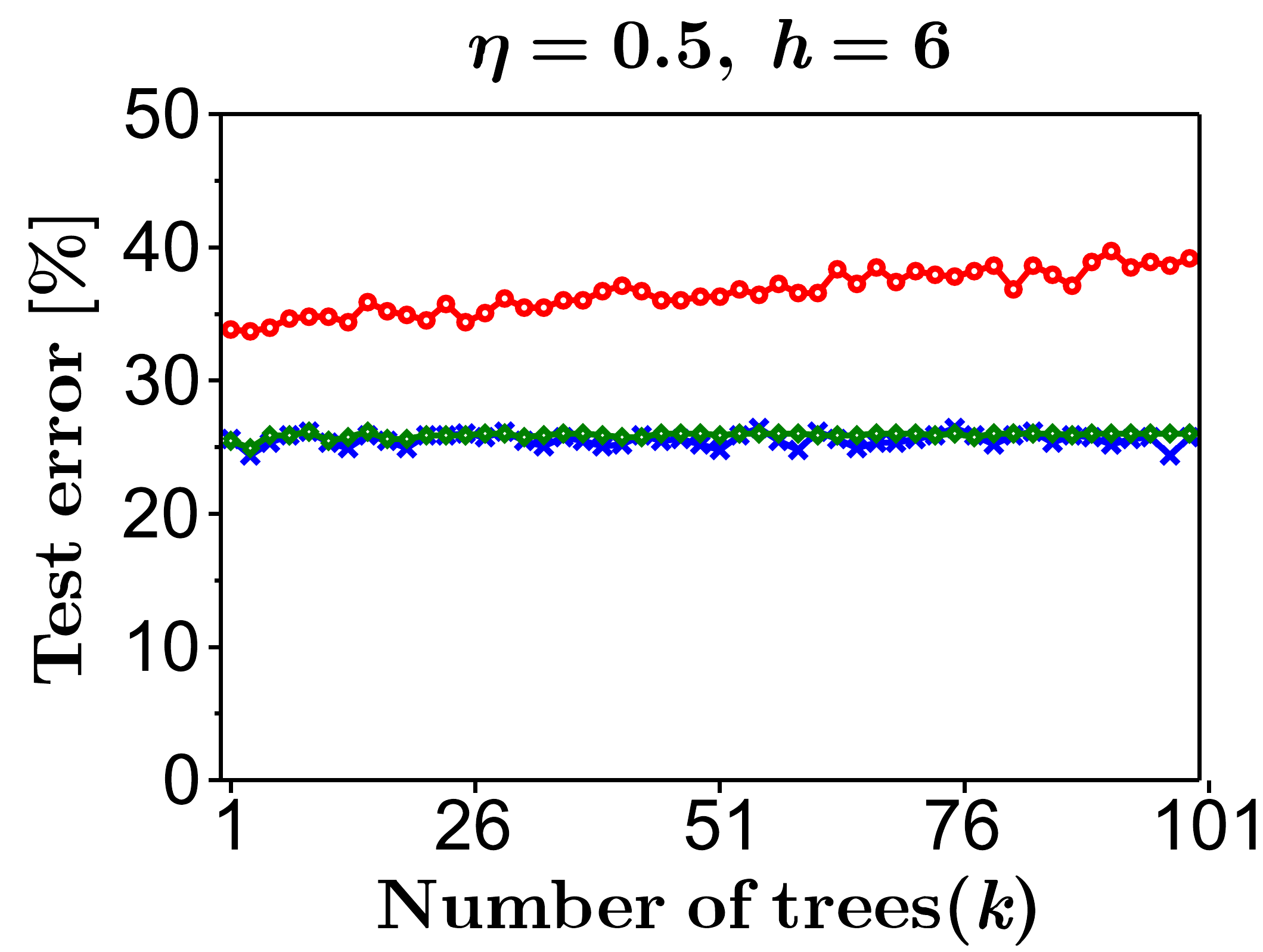} 
\hspace{-0.03in}\includegraphics[width = 1.6in]{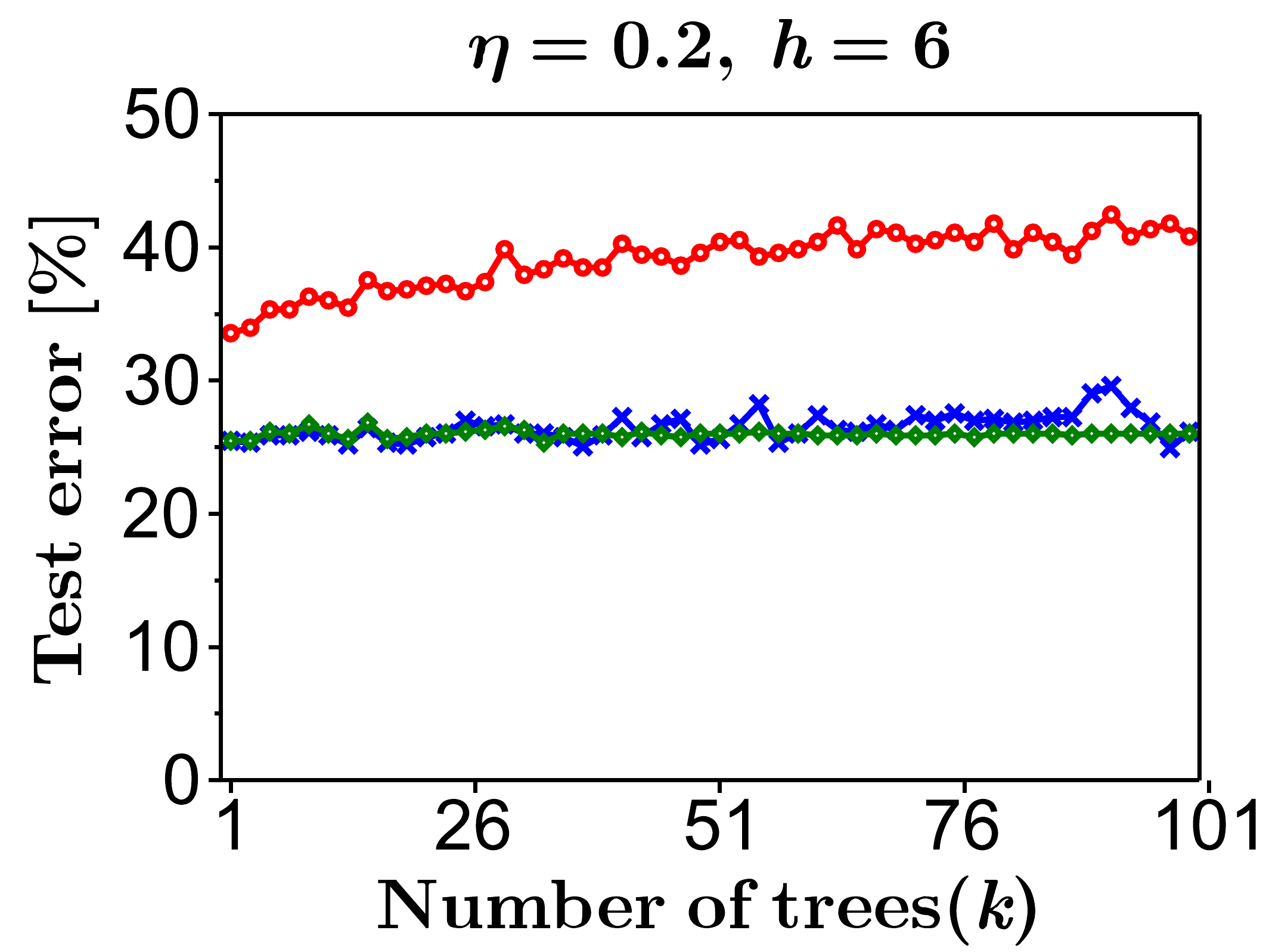} 
\hspace{-0.03in}\includegraphics[width = 1.6in]{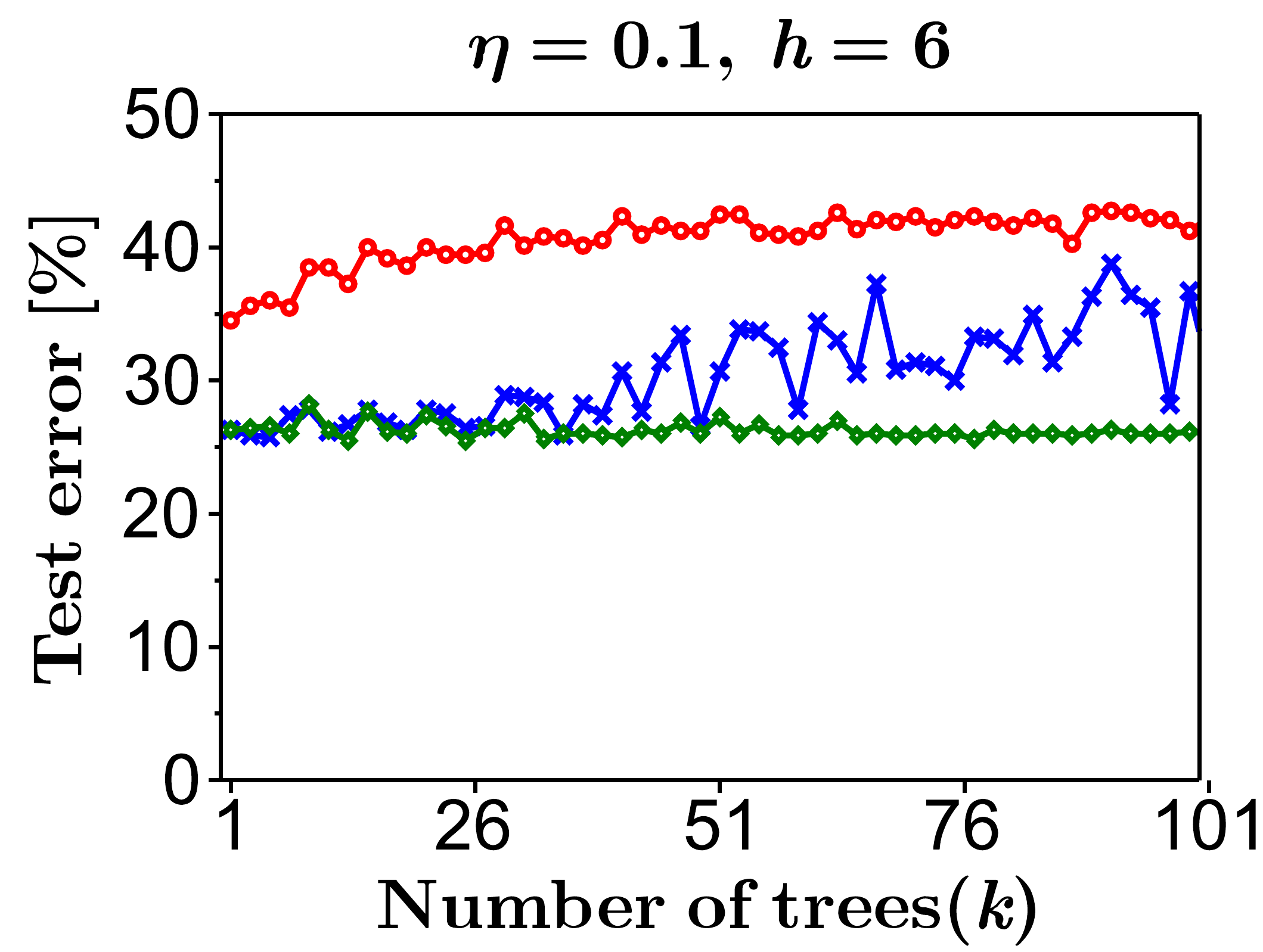}
\caption{\textit{BTSC} dataset. Comparison of dpRFMV, dpRFTA and dpRFPA. \textbf{a)} Test error vs. $\eta$ for two settings of $h$. \textbf{b)} Test error vs. $k$ for fixed $h$ and across different settings of $\eta$.}
\label{fig:btsc_two}
\end{figure}

\begin{figure}[h]
center
\begin{tabular}{cc}
\hspace{-0.1in}\includegraphics[width = 1.54in]{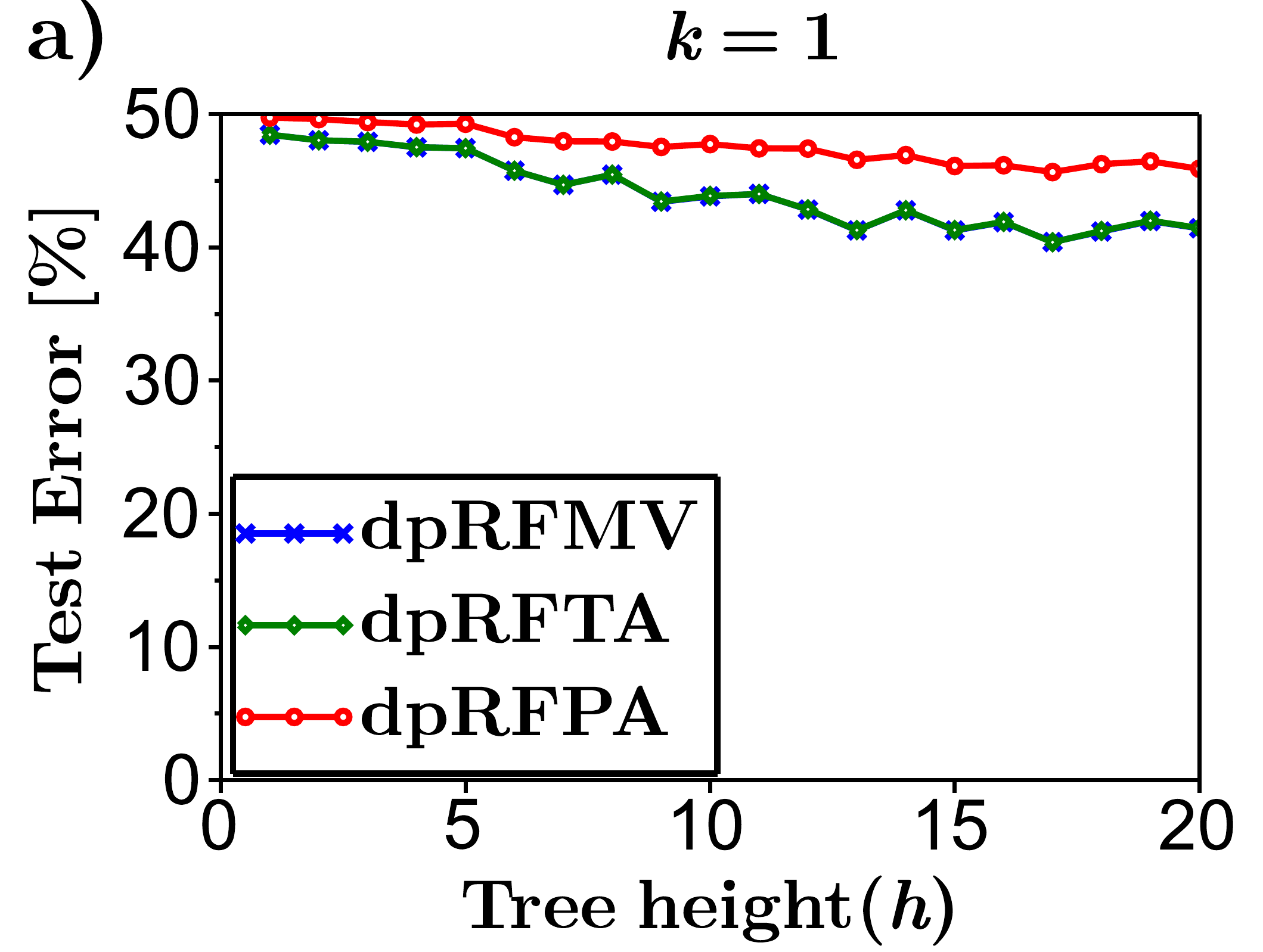} 
\hspace{-0.04in}\includegraphics[width = 1.54in]{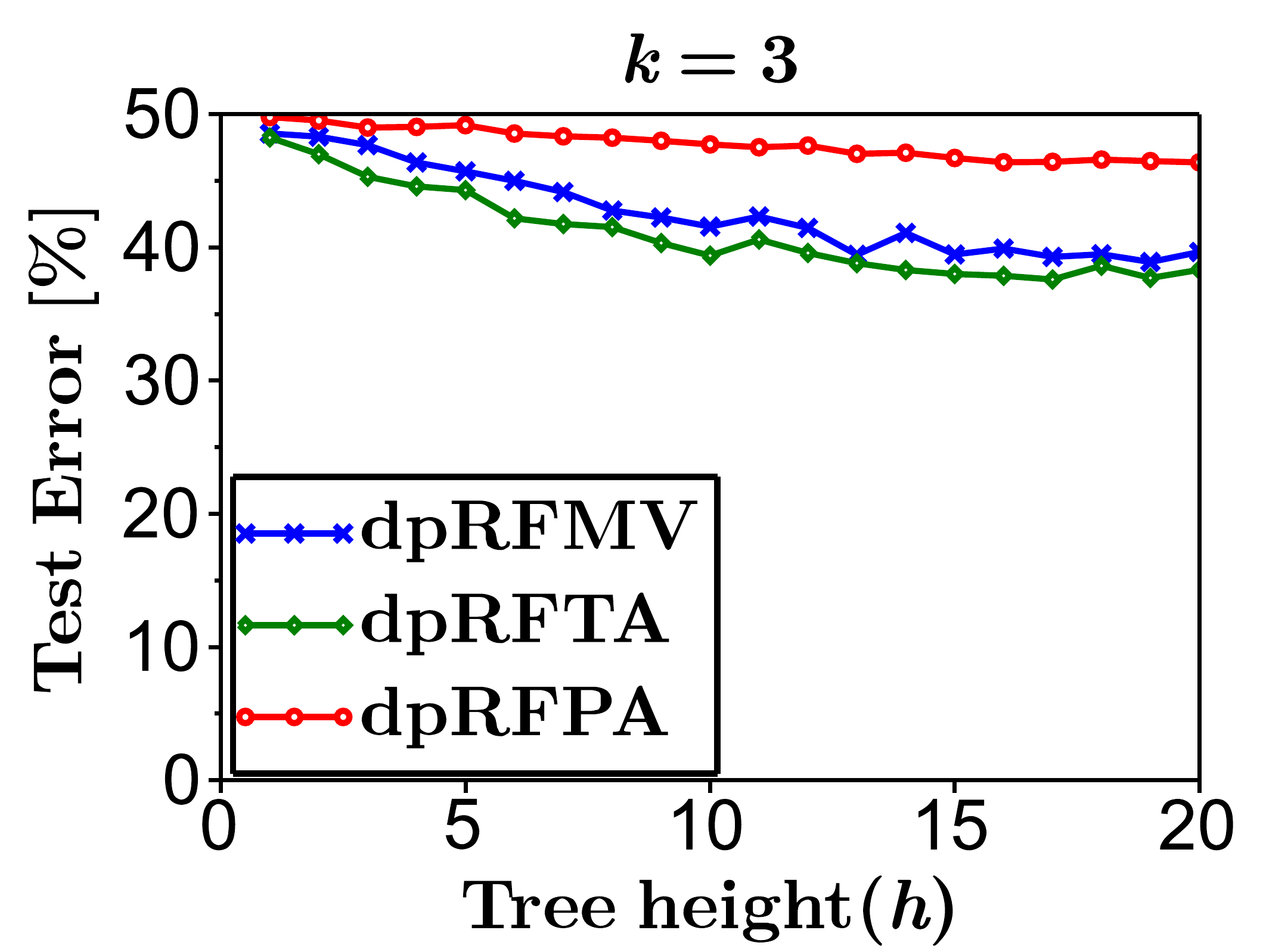} 
\hspace{-0.035in}\includegraphics[width = 1.54in]{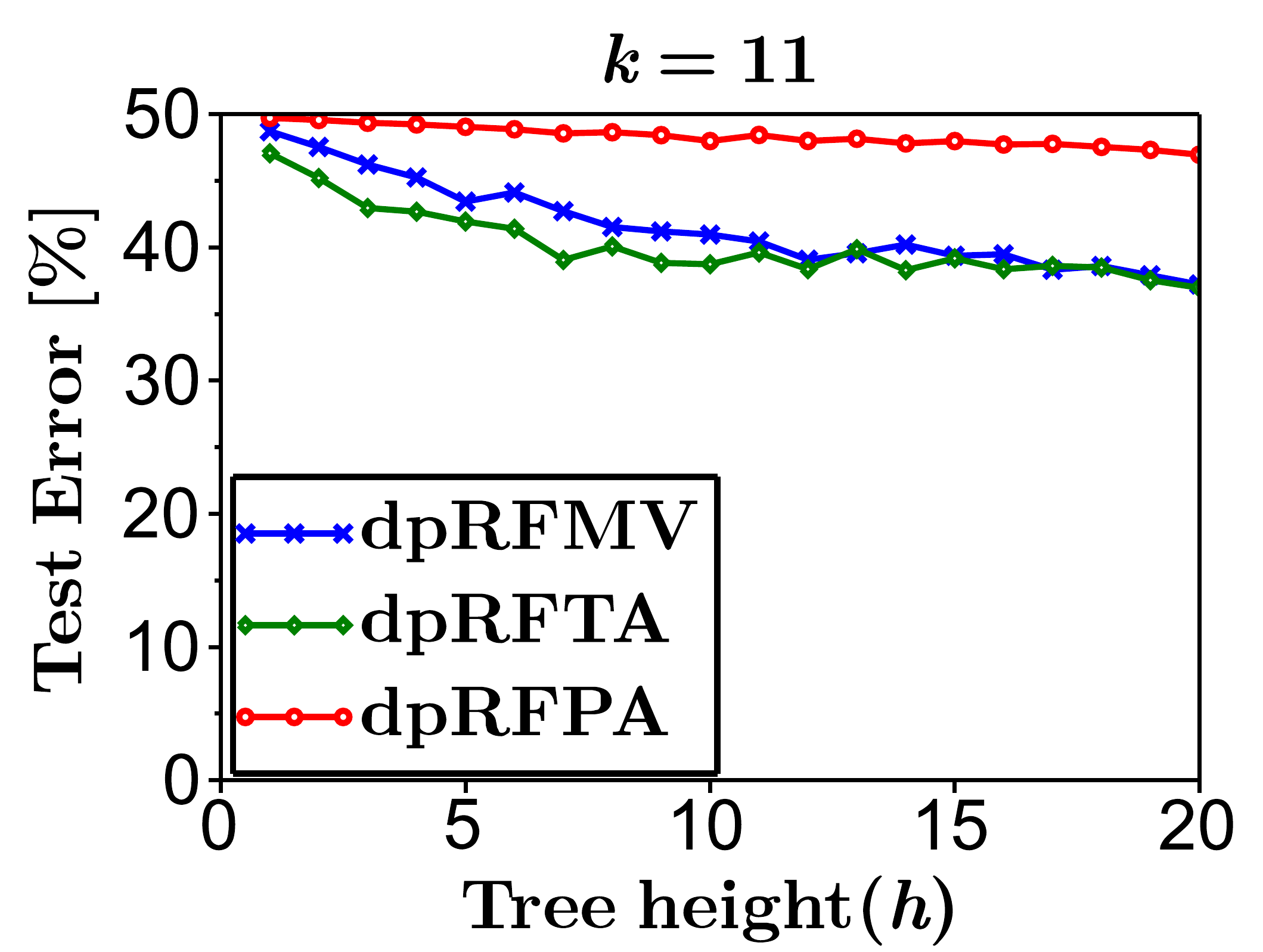} 
\hspace{-0.04in}\includegraphics[width = 1.54in]{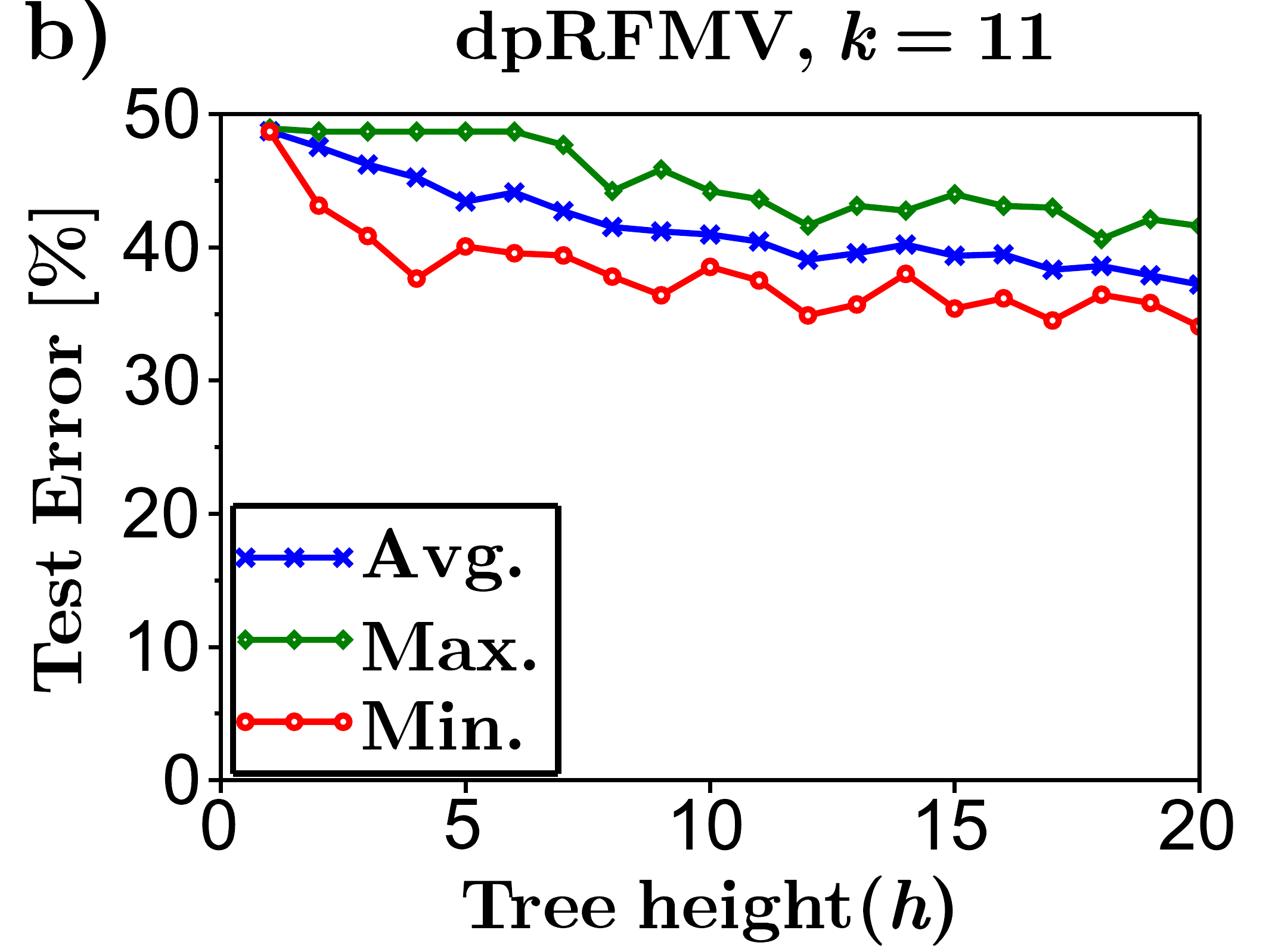}\\  
\hspace{-0.1in}\includegraphics[width = 1.54in]{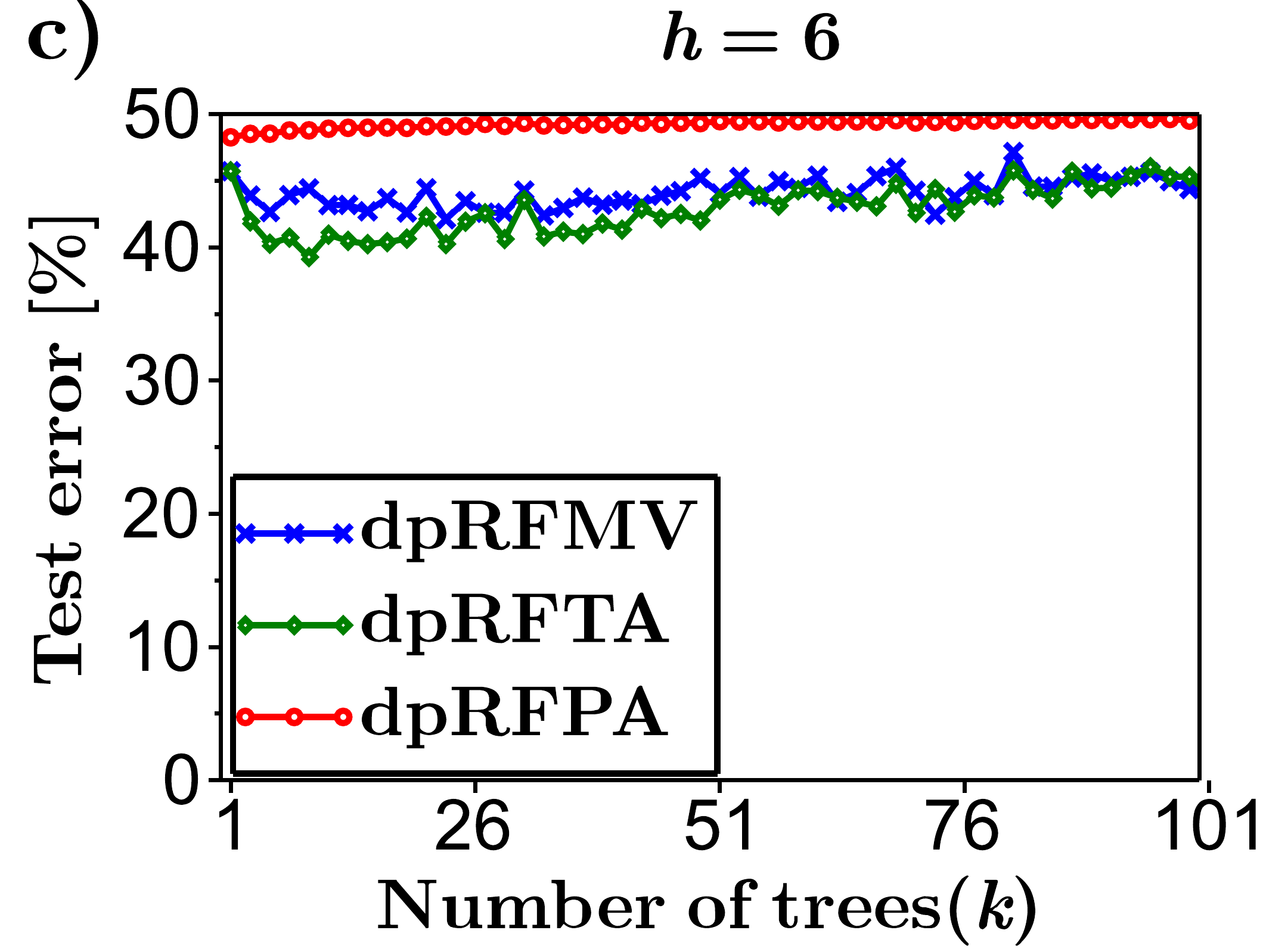} 
\hspace{-0.04in}\includegraphics[width = 1.54in]{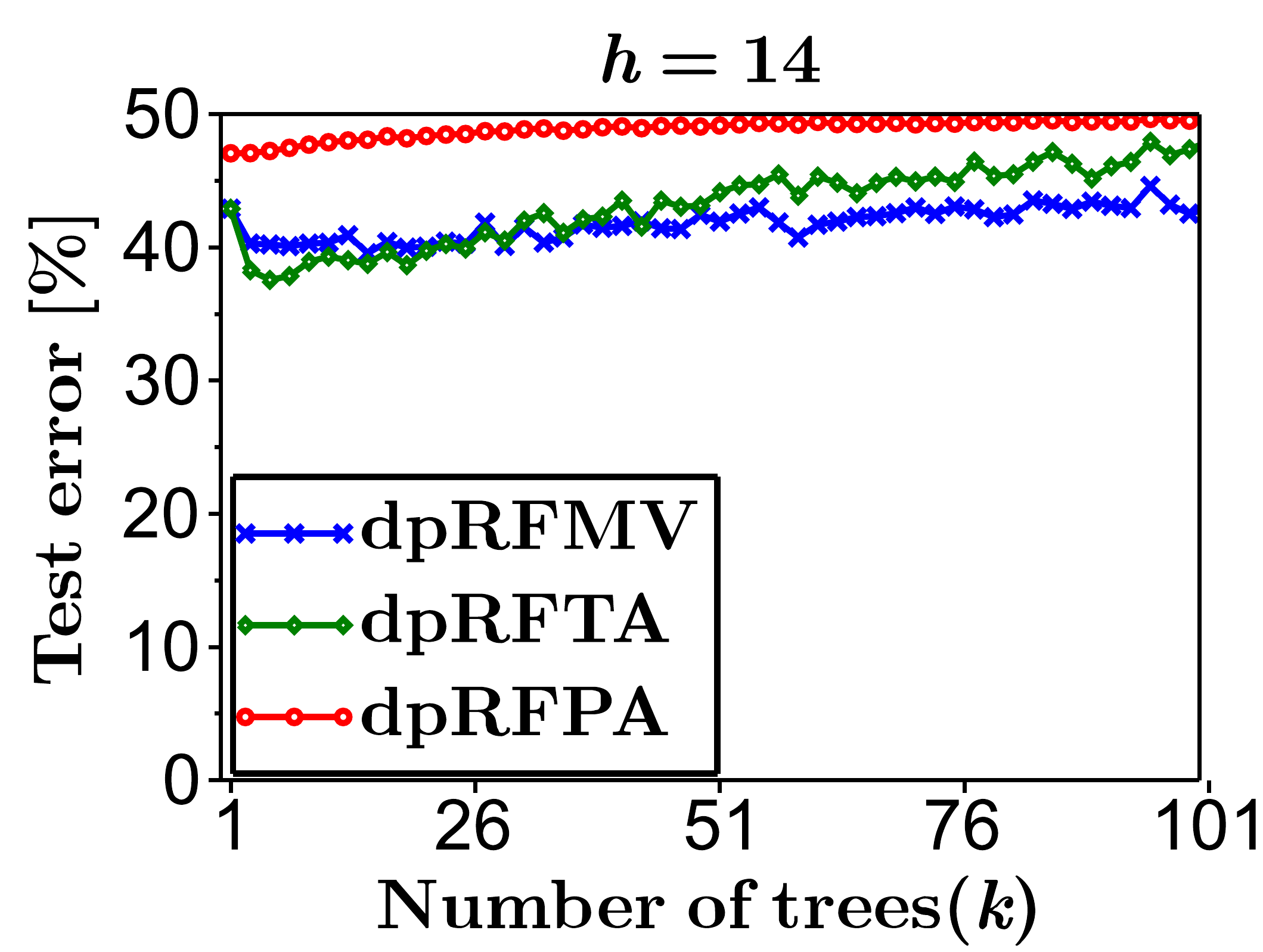} 
\hspace{-0.035in}\includegraphics[width = 1.54in]{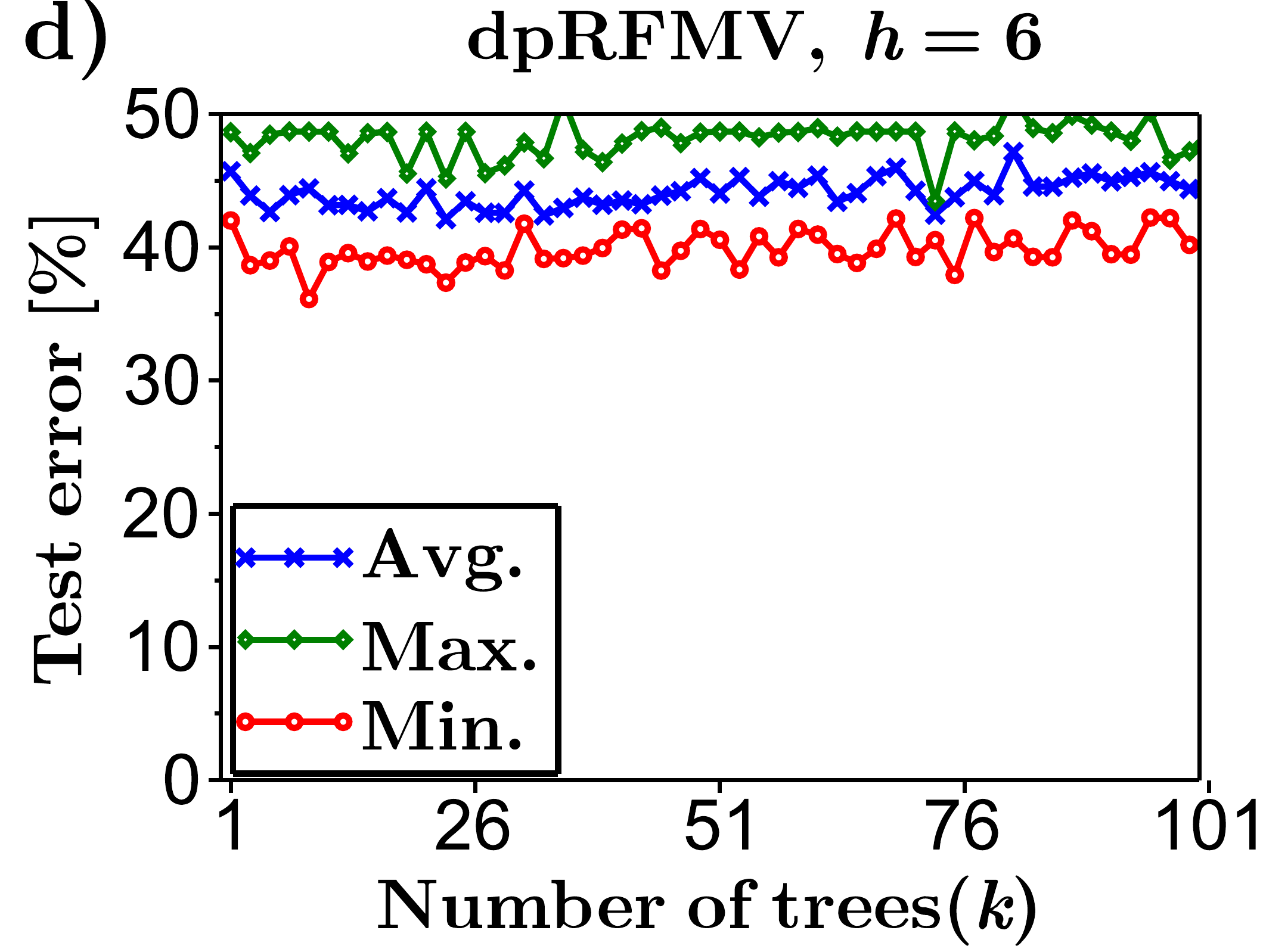} 
\hspace{-0.04in}\includegraphics[width = 1.54in]{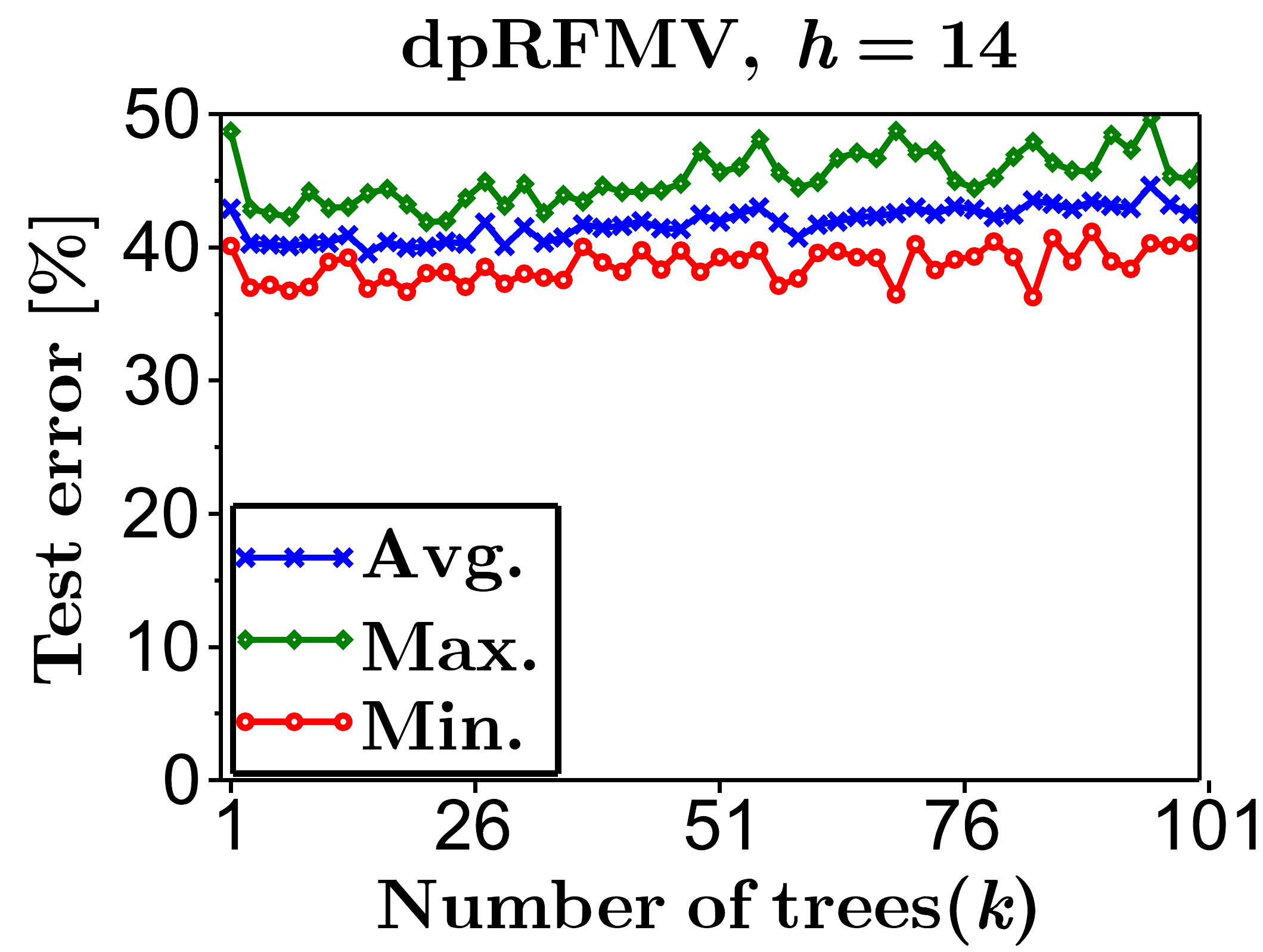}
\end{tabular}
\caption{\textit{Covertype} dataset. Comparison of dpRFMV, dpRFTA and dpRFPA. $\eta = 1000/n_{tr}$. Test error resp. vs. \textbf{a)} $h$ across various settings of $k$ and vs. \textbf{c)} $k$ across various settings of $h$; Minimal, average and maximal test error resp. vs. $h$ (\textbf{b)}) and vs. $k$ (\textbf{d)}) for dpRFMV.}
\label{fig:mam_mas_one}
\vspace{-0.04in}
\end{figure}

\begin{figure}[h]
  \center
\includegraphics[width = 2.8in,height = 1.25in]{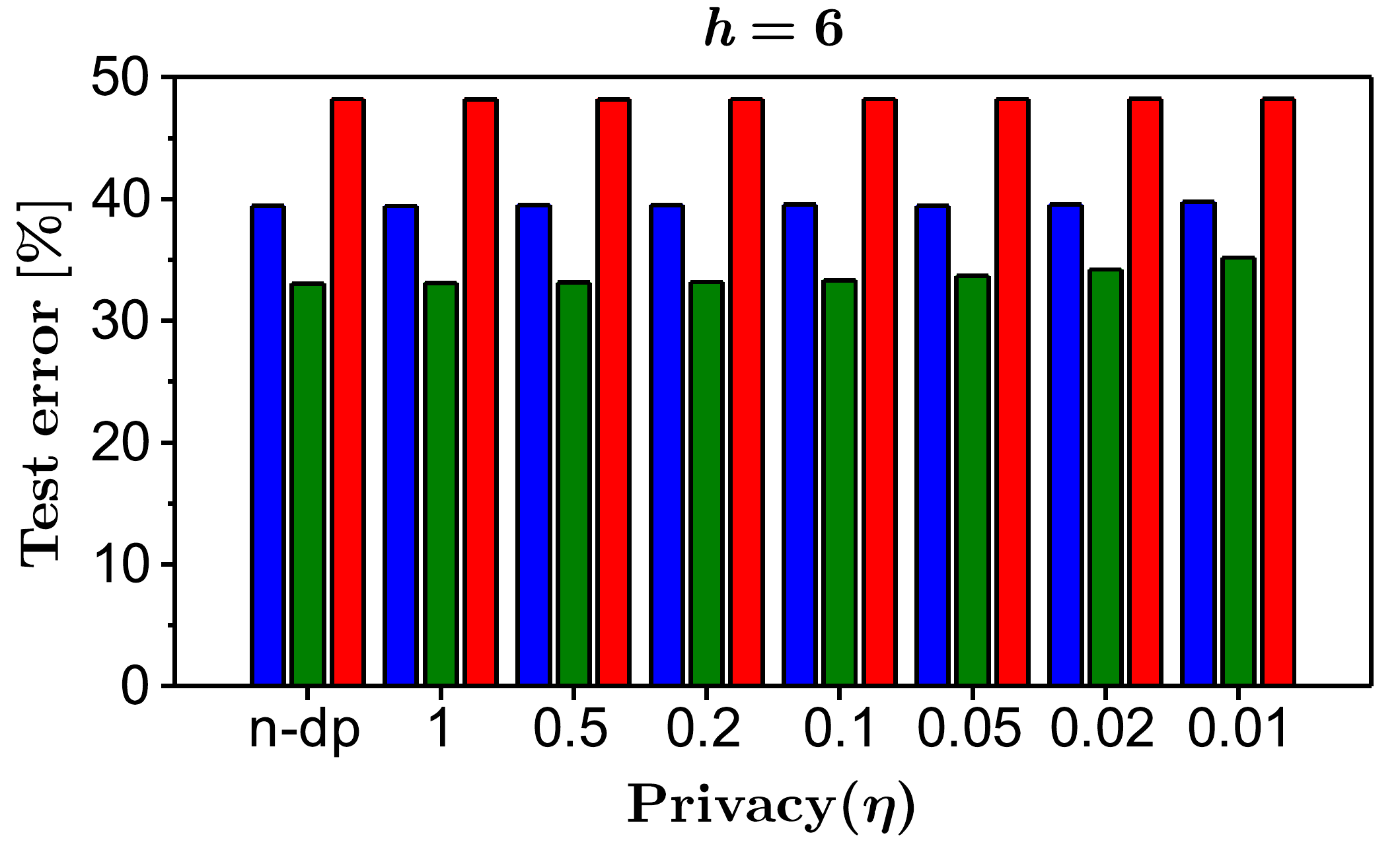} 
\includegraphics[width = 2.8in,height = 1.25in]{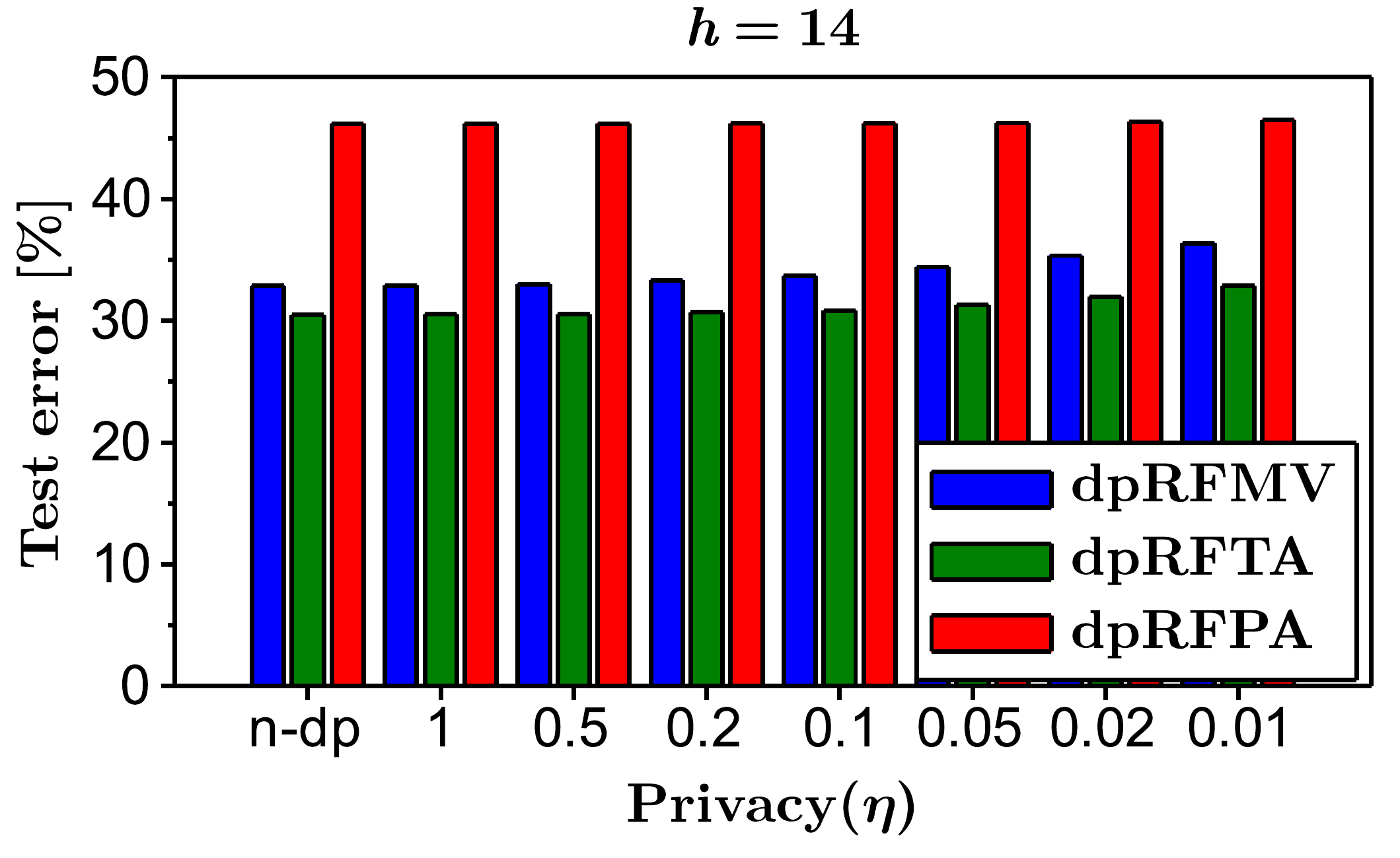}\\
\includegraphics[width = 1.6in]{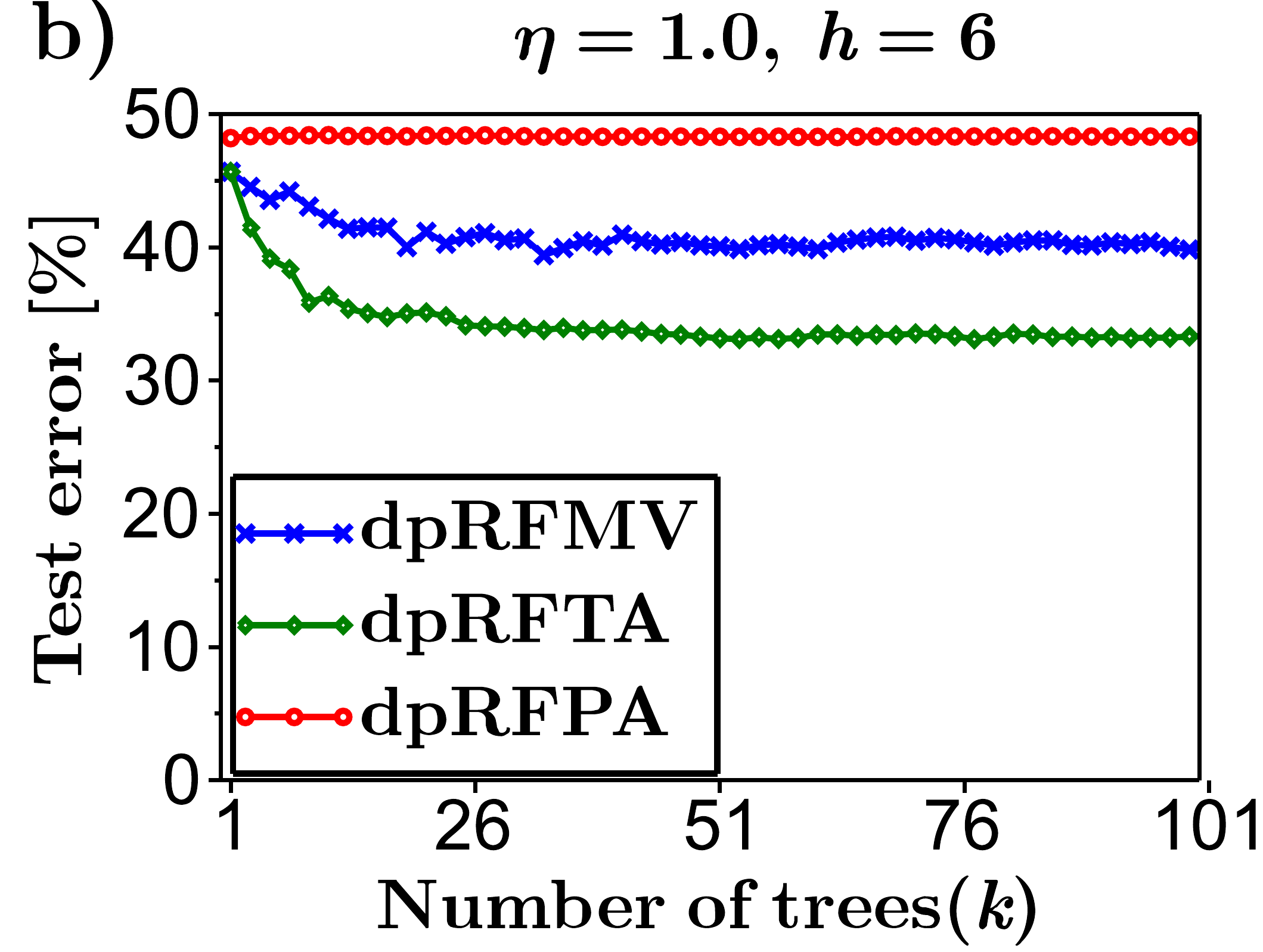} 
\hspace{-0.03in}\includegraphics[width = 1.6in]{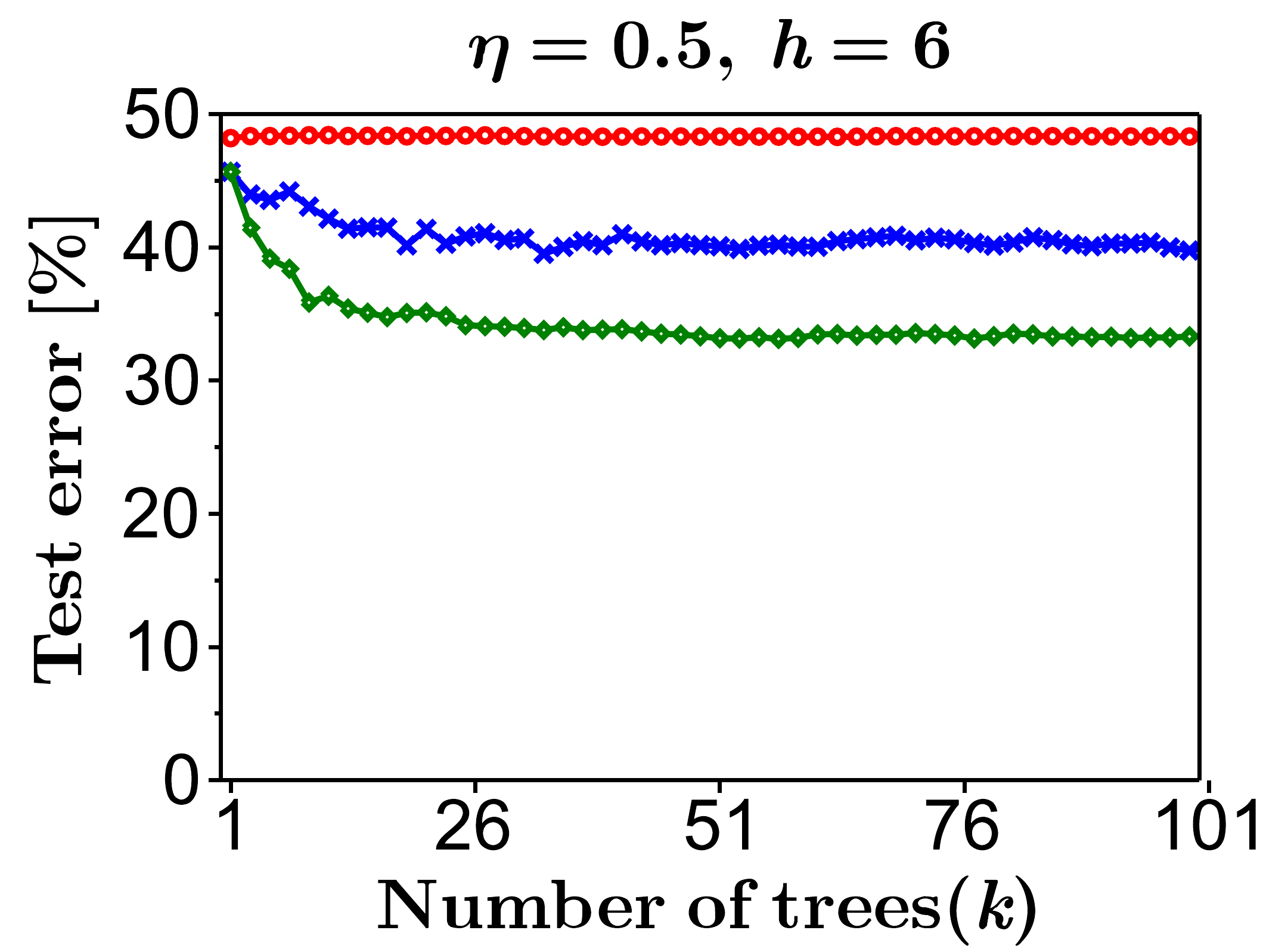} 
\hspace{-0.03in}\includegraphics[width = 1.6in]{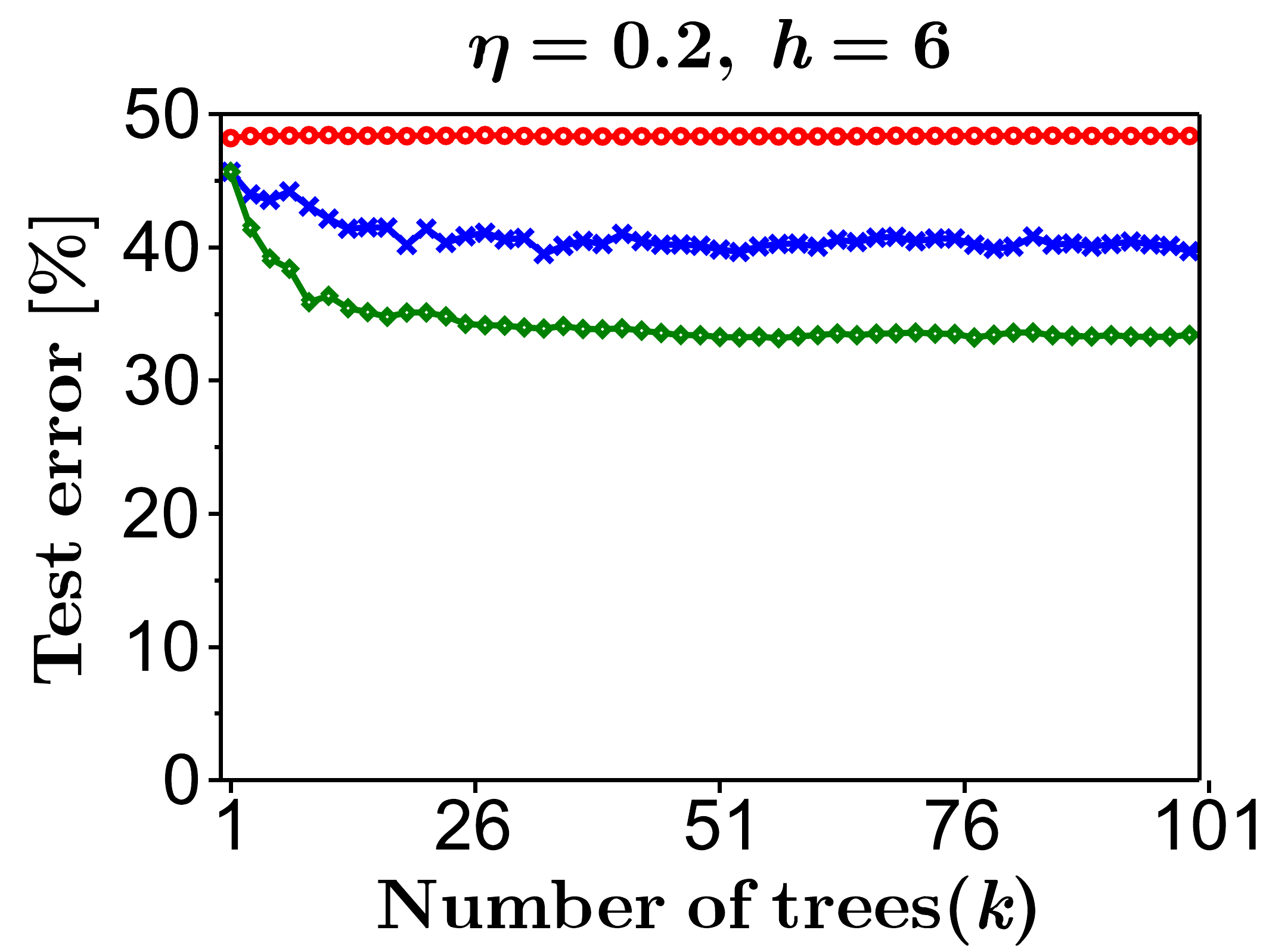} 
\hspace{-0.03in}\includegraphics[width = 1.6in]{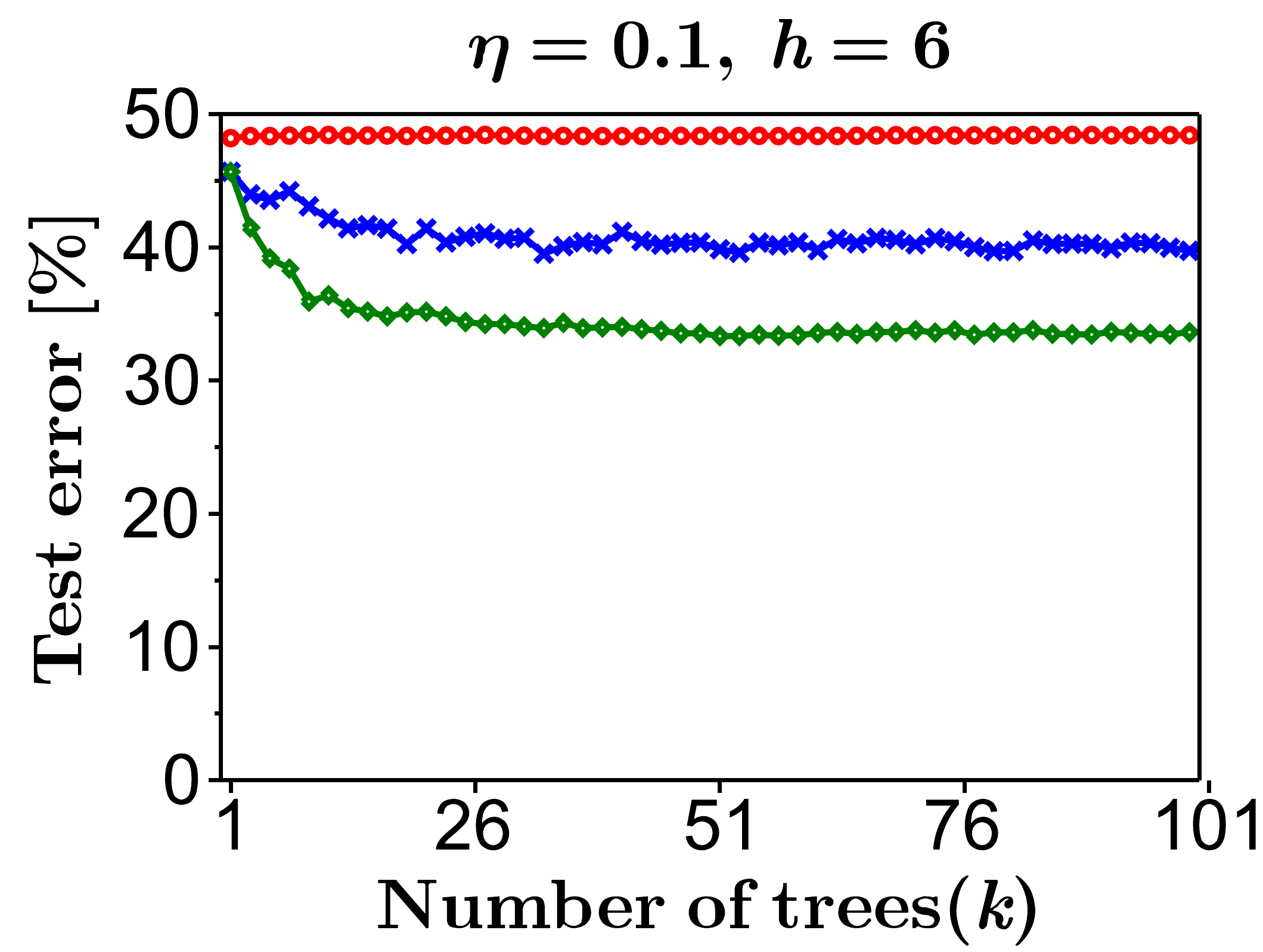}
\caption{\textit{Covertype} dataset. Comparison of dpRFMV, dpRFTA and dpRFPA. \textbf{a)} Test error vs. $\eta$ for two settings of $h$. \textbf{b)} Test error vs. $k$ for fixed $h$ and across different settings of $\eta$.}
\label{fig:mam_mas_two}
\end{figure}

\begin{figure}[h]
\center
\begin{tabular}{cc}
\hspace{-0.1in}\includegraphics[width = 1.54in]{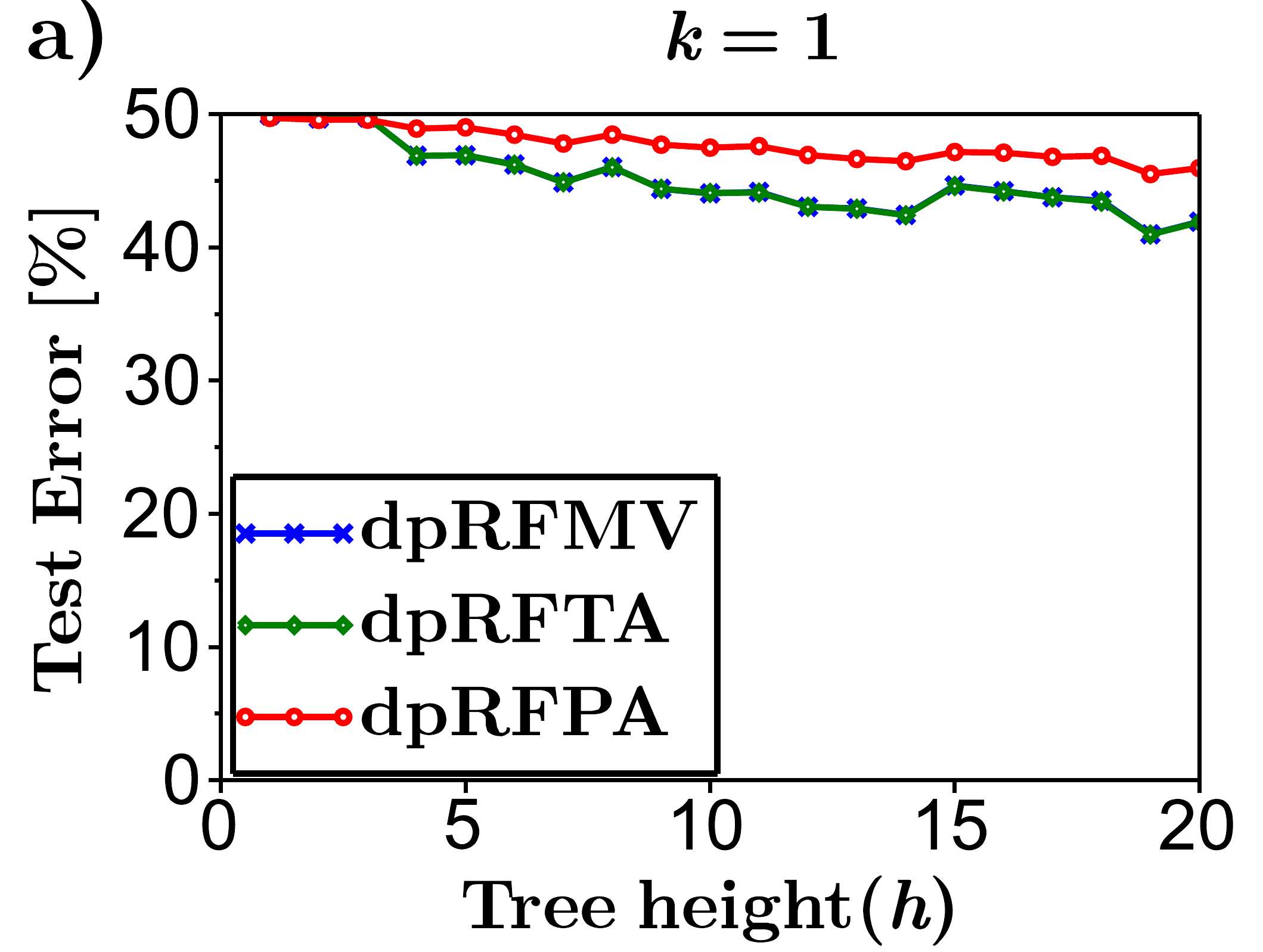} 
\hspace{-0.04in}\includegraphics[width = 1.54in]{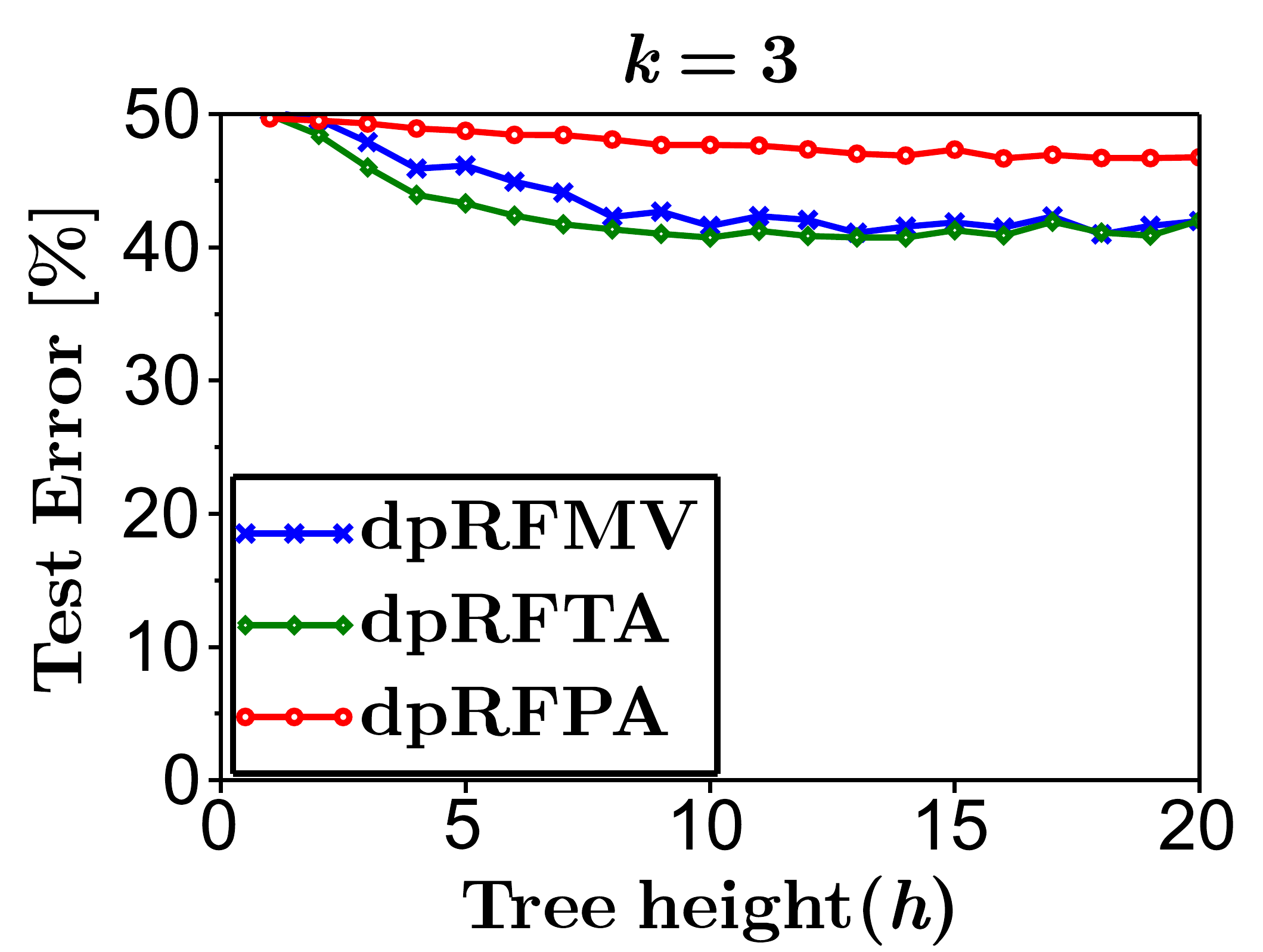} 
\hspace{-0.035in}\includegraphics[width = 1.54in]{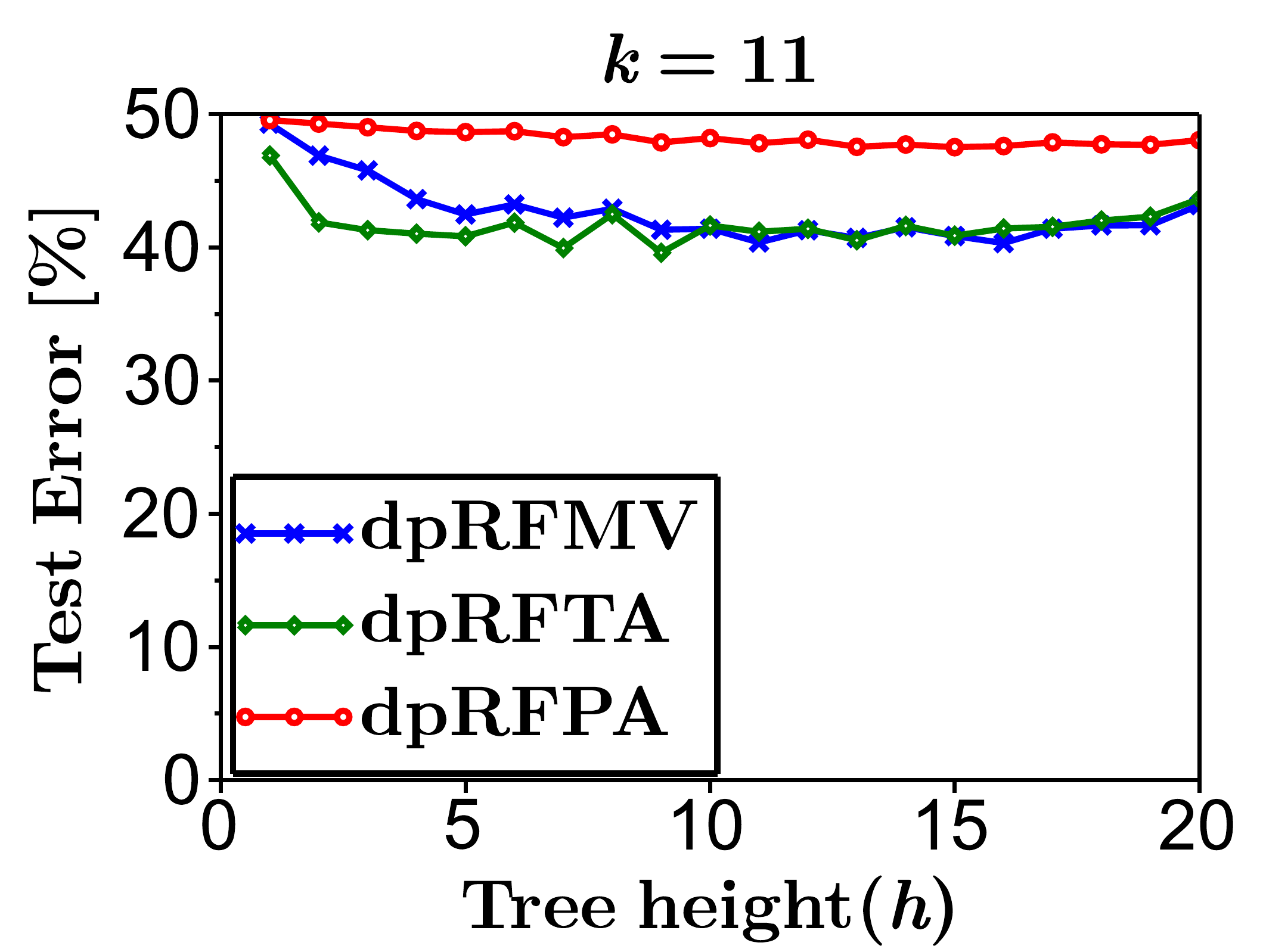} 
\hspace{-0.04in}\includegraphics[width = 1.54in]{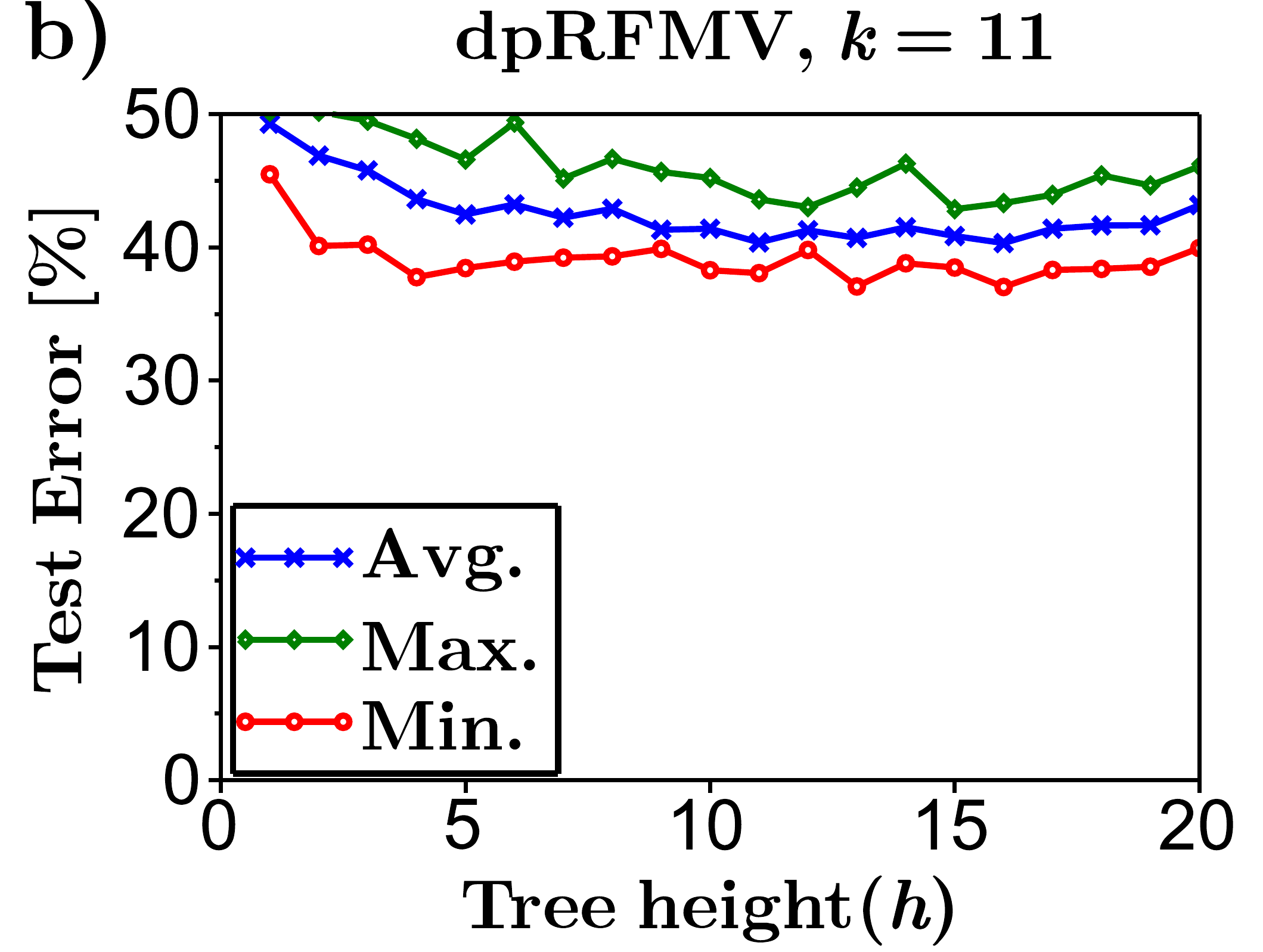}\\  
\hspace{-0.1in}\includegraphics[width = 1.54in]{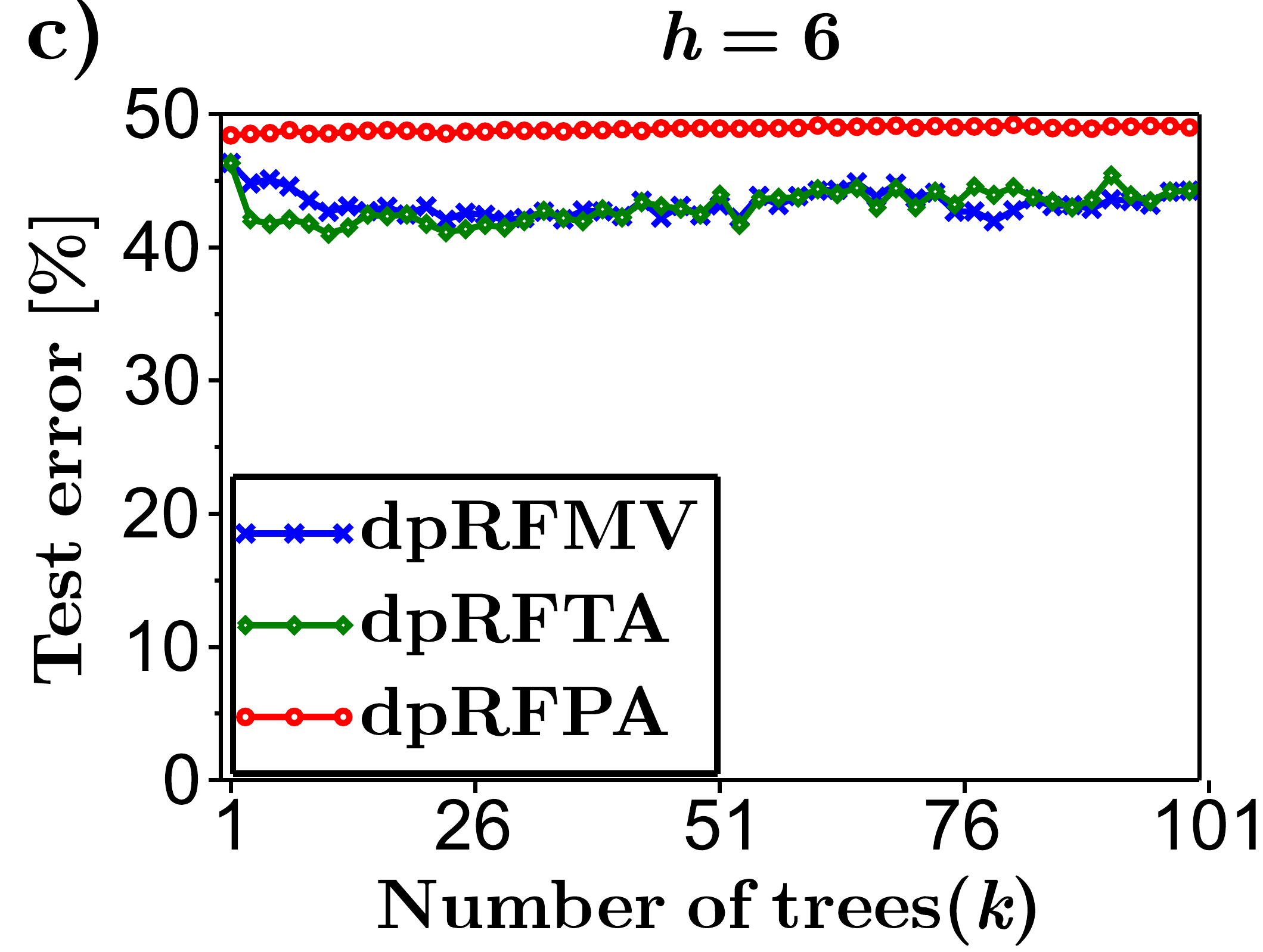} 
\hspace{-0.04in}\includegraphics[width = 1.54in]{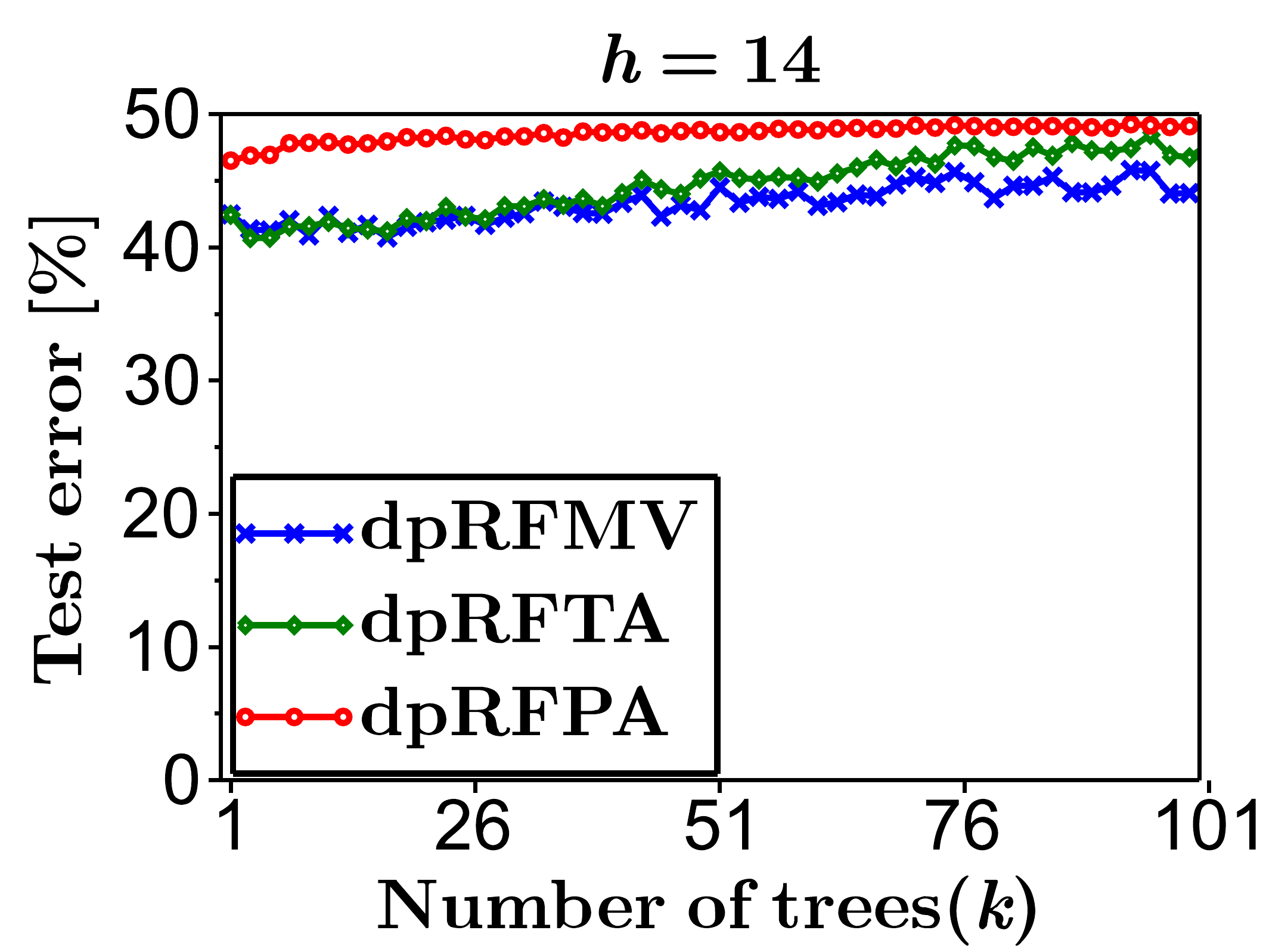} 
\hspace{-0.035in}\includegraphics[width = 1.54in]{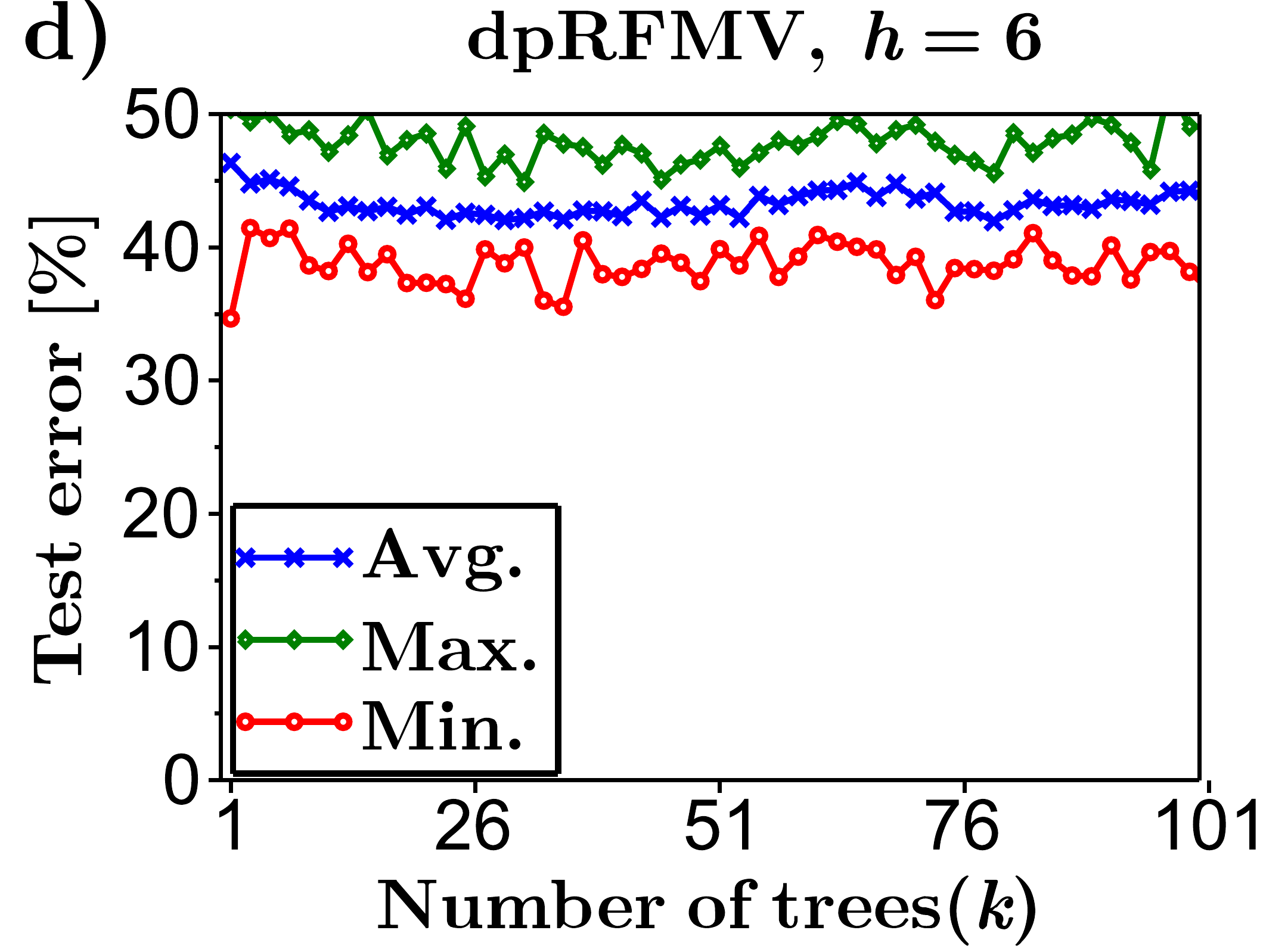} 
\hspace{-0.04in}\includegraphics[width = 1.54in]{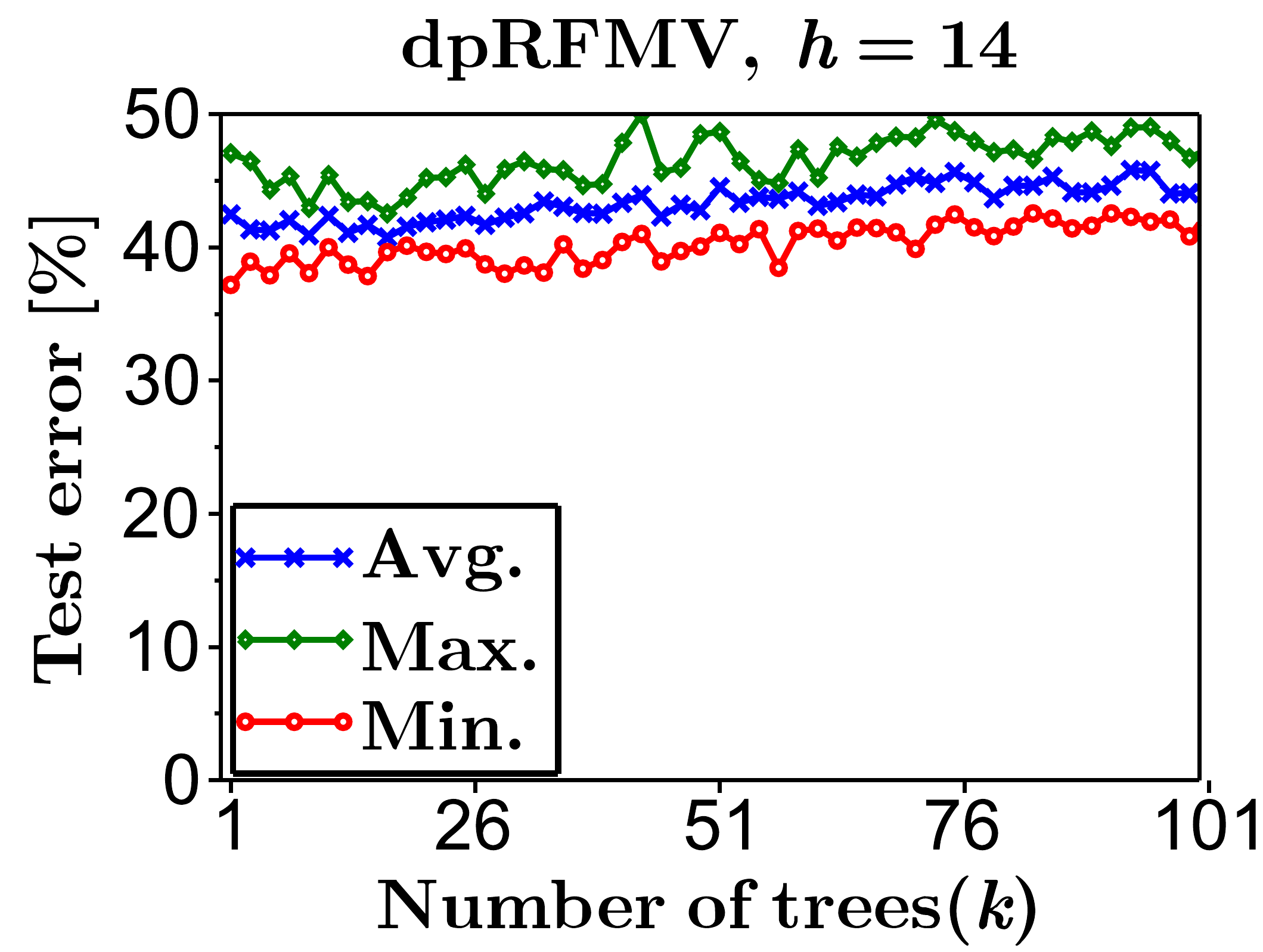}
\end{tabular}
\caption{\textit{Quantum} dataset. Comparison of dpRFMV, dpRFTA and dpRFPA. $\eta = 1000/n_{tr}$. Test error resp. vs. \textbf{a)} $h$ across various settings of $k$ and vs. \textbf{c)} $k$ across various settings of $h$; Minimal, average and maximal test error resp. vs. $h$ (\textbf{b)}) and vs. $k$ (\textbf{d)}) for dpRFMV.}
\label{fig:mam_mas_one}
\end{figure}

\makeatletter
\setlength{\@fptop}{0pt}
\makeatother

\begin{figure}[t!]
\center
\includegraphics[width = 2.8in,height = 1.25in]{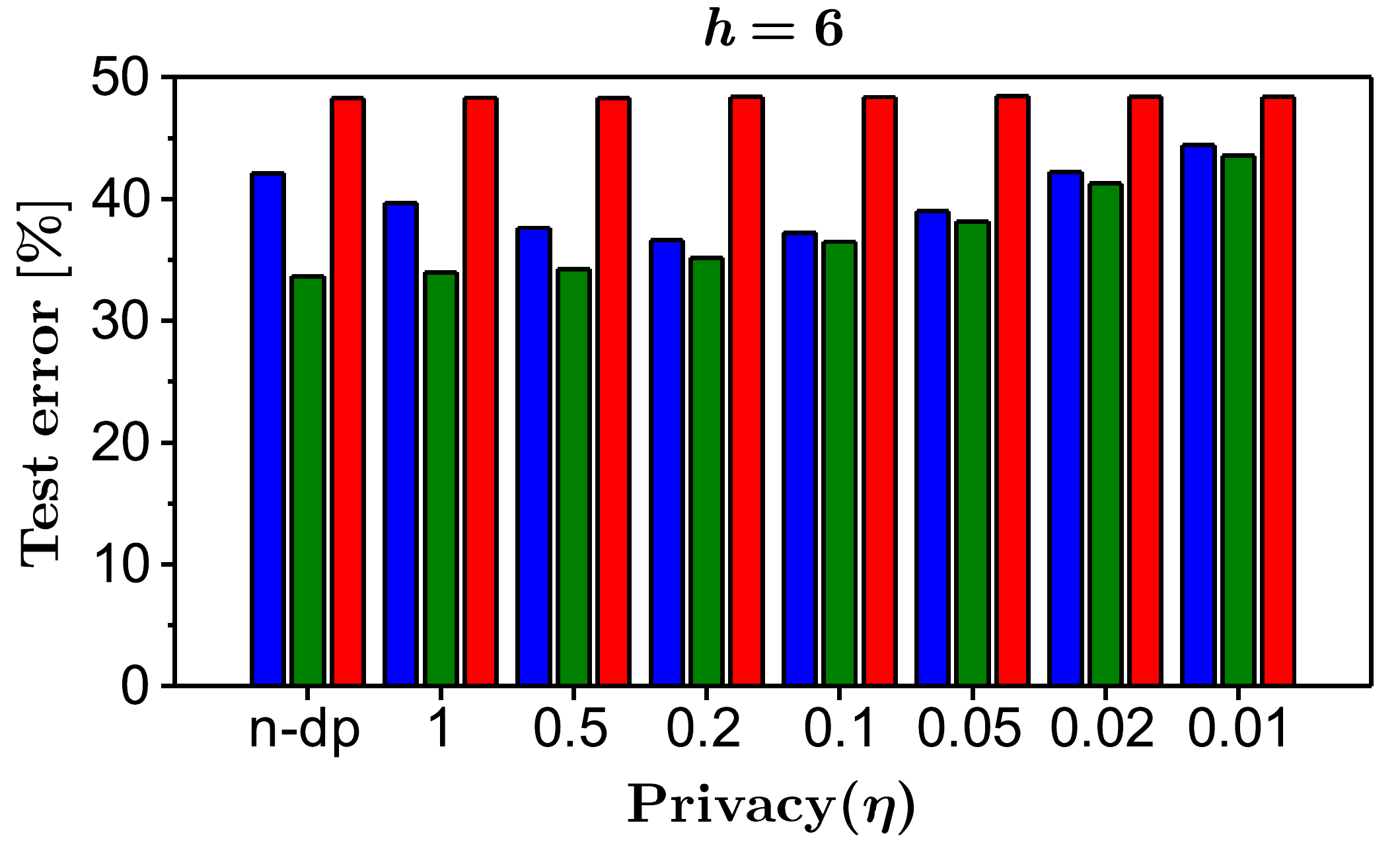} 
\includegraphics[width = 2.8in,height = 1.25in]{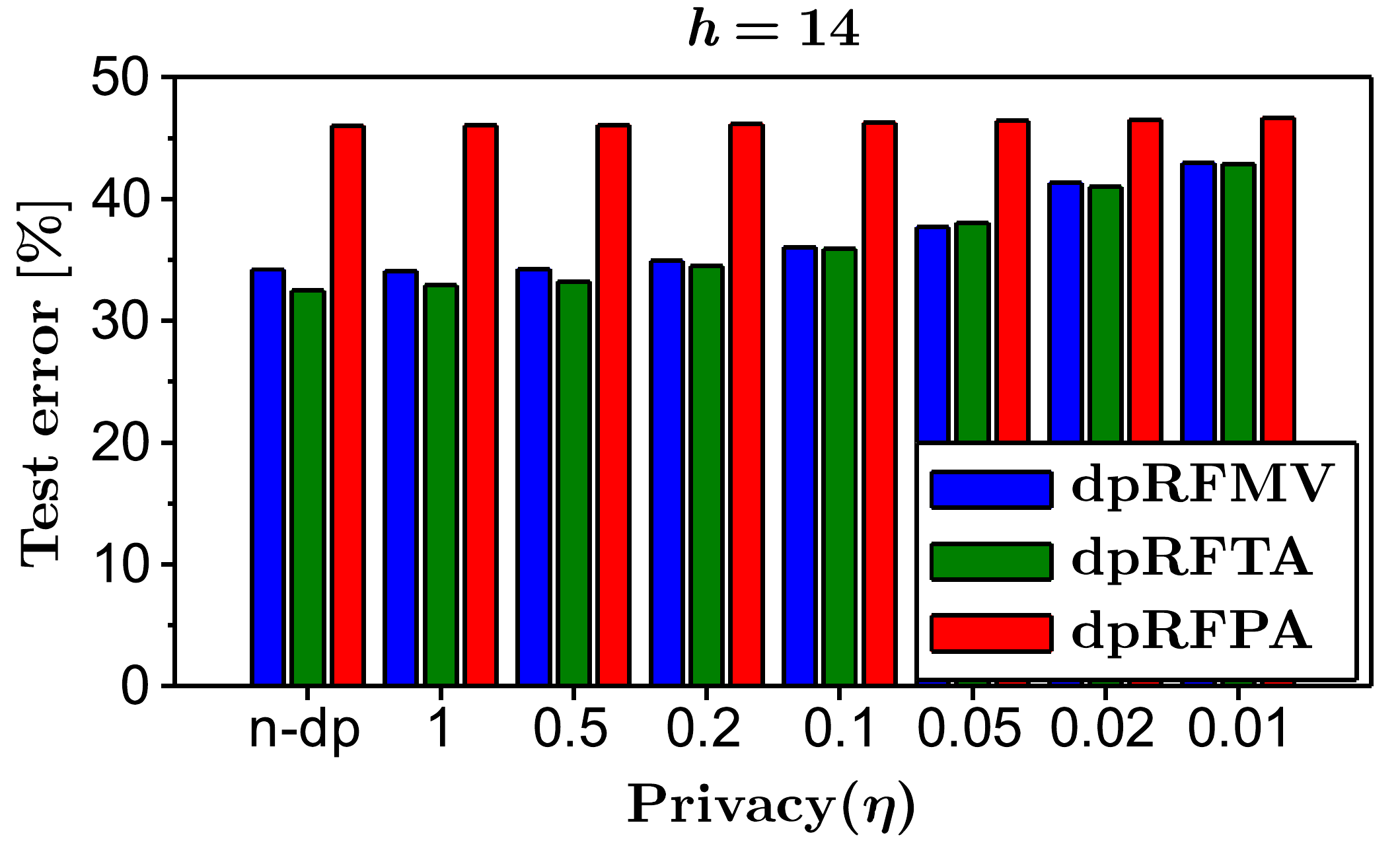}\\
\includegraphics[width = 1.6in]{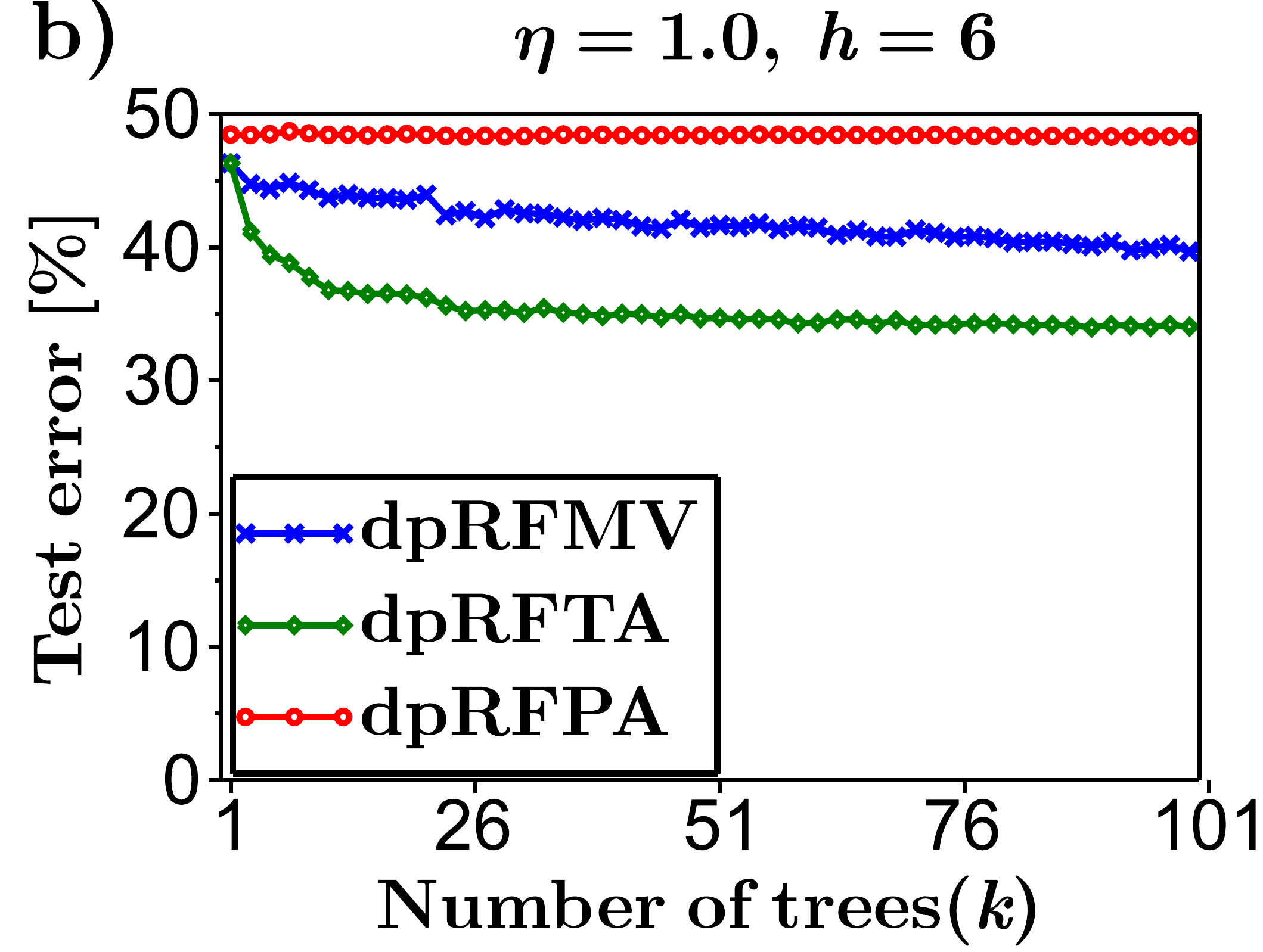} 
\hspace{-0.03in}\includegraphics[width = 1.6in]{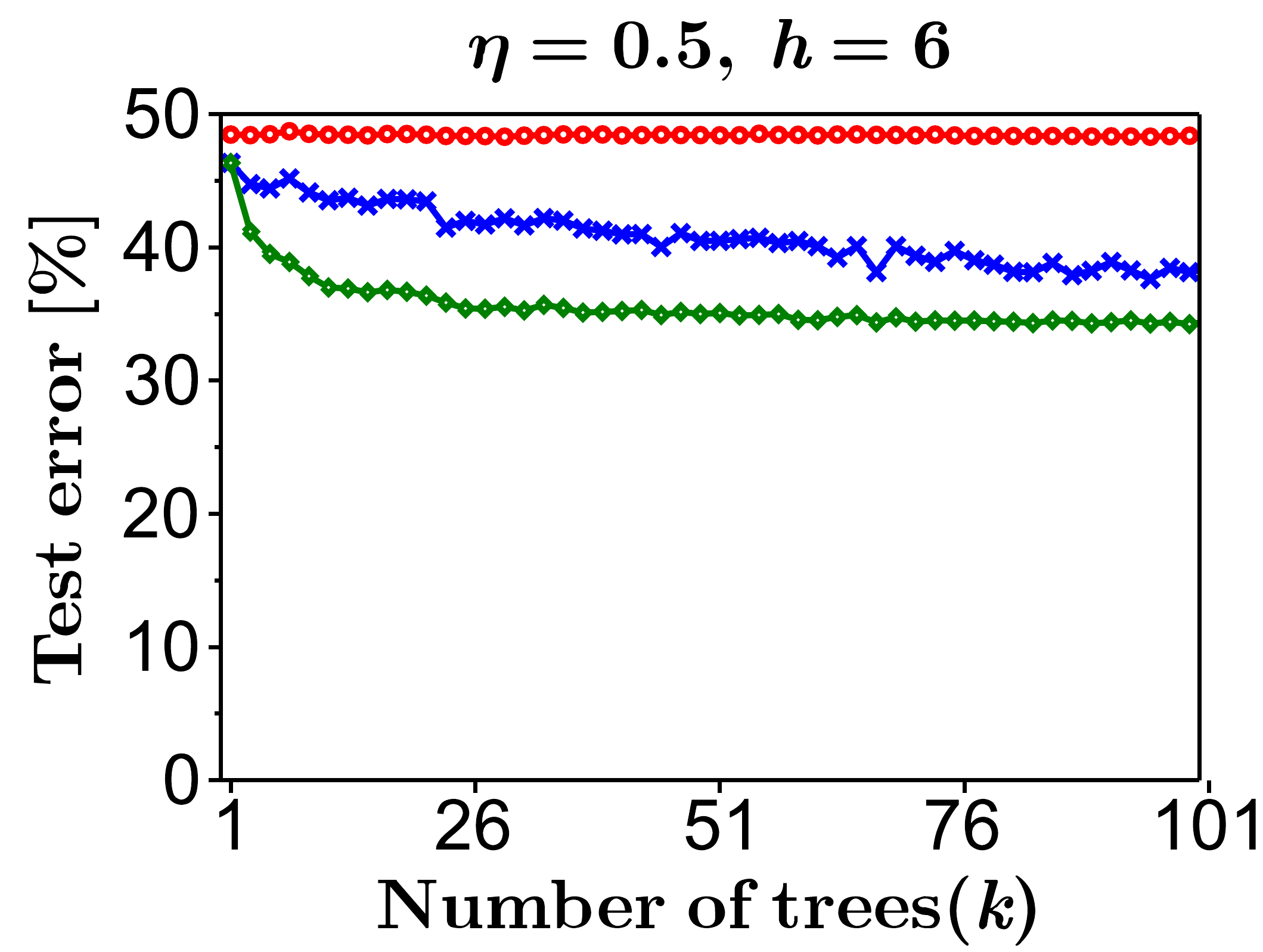} 
\hspace{-0.03in}\includegraphics[width = 1.6in]{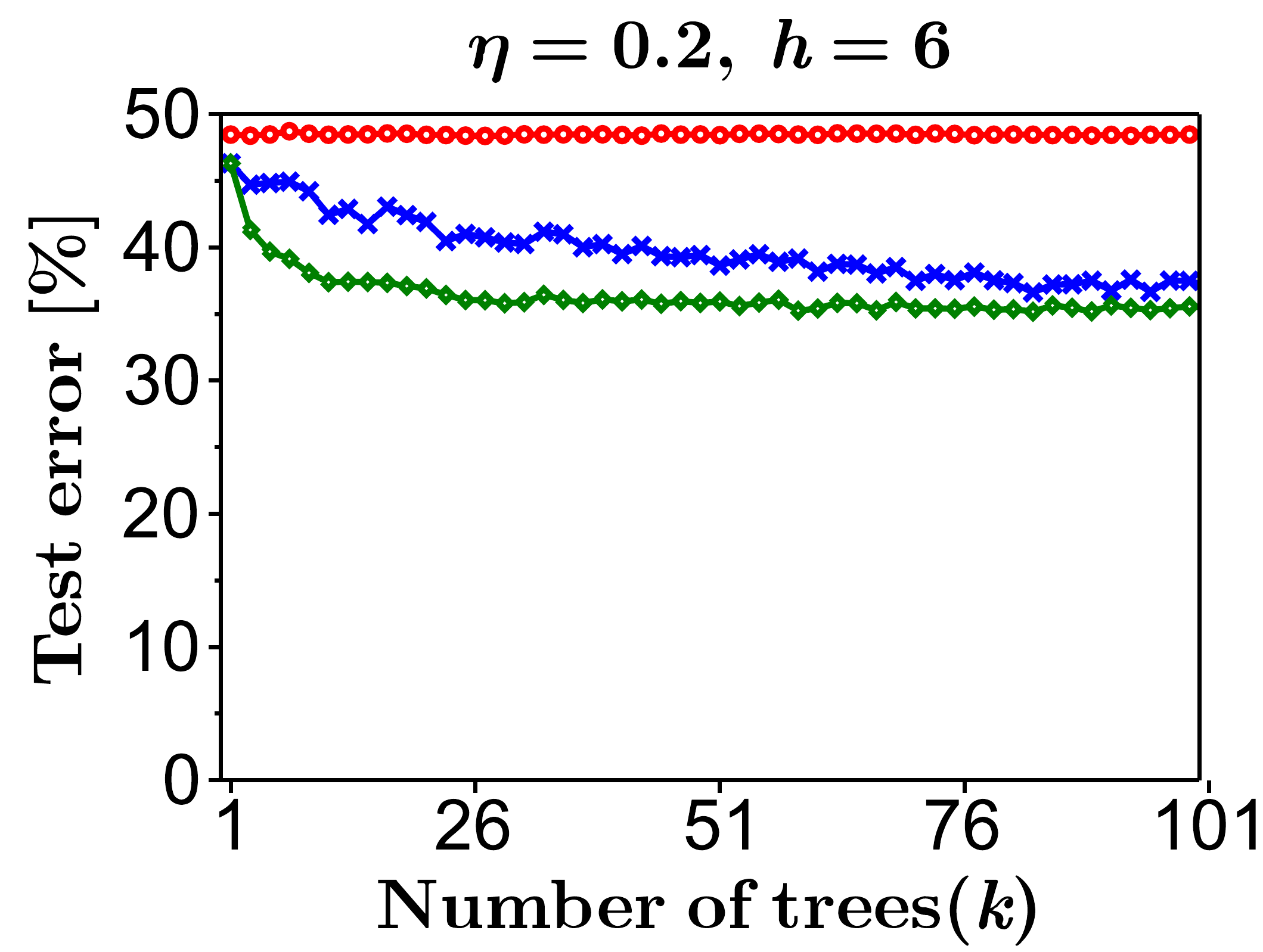} 
\hspace{-0.03in}\includegraphics[width = 1.6in]{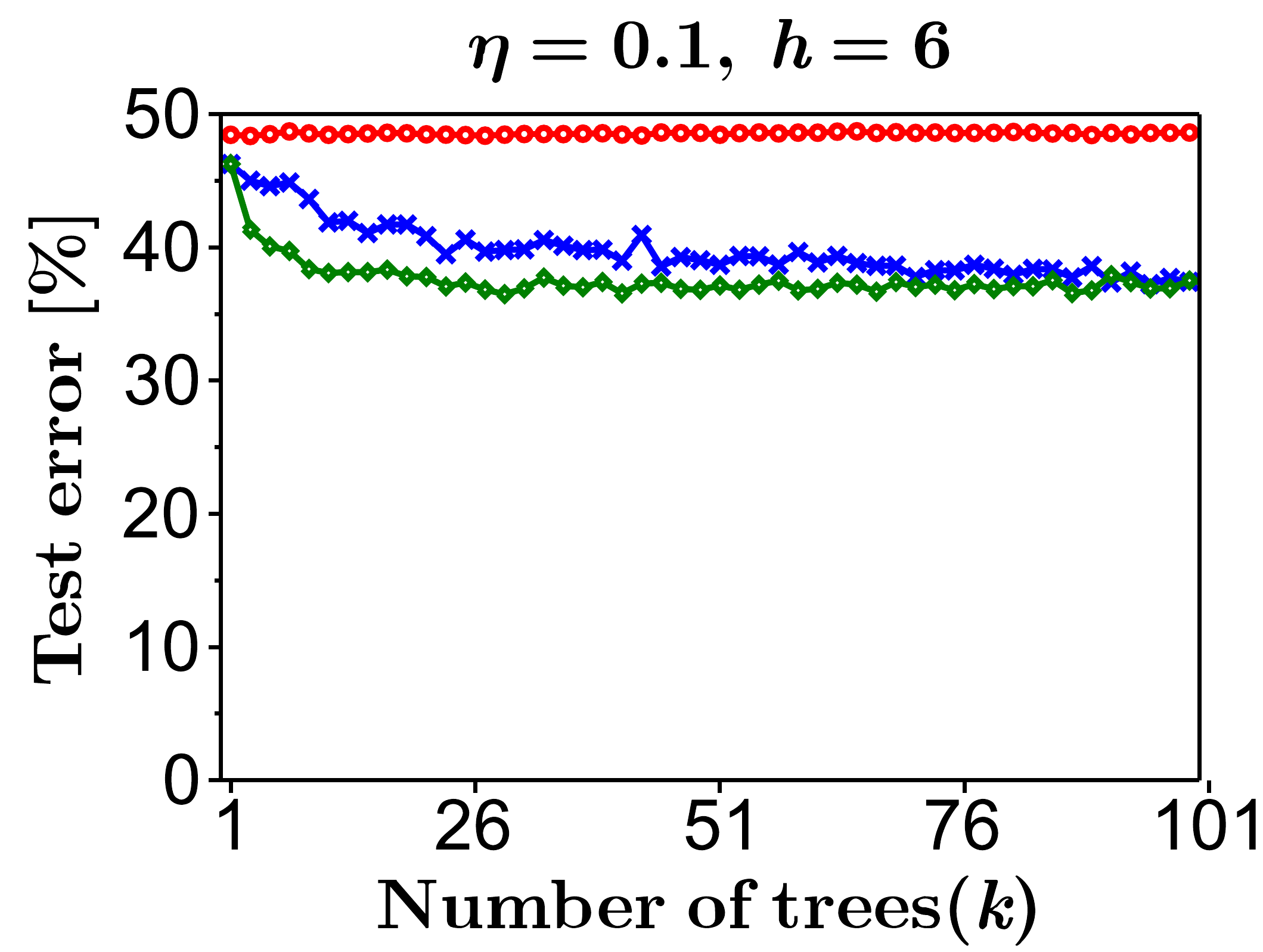}
\caption{\textit{Quantum} dataset. Comparison of dpRFMV, dpRFTA and dpRFPA. \textbf{a)} Test error vs. $\eta$ for two settings of $h$. \textbf{b)} Test error vs. $k$ for fixed $h$ and across different settings of $\eta$.}
\label{fig:mam_mas_two}
\end{figure}

\end{document}